\documentclass{article}

\PassOptionsToPackage{numbers,sort&compress}{natbib}
\usepackage[preprint]{neurips_2026}
\usepackage[T1]{fontenc}
\usepackage[utf8]{inputenc}
\usepackage{amsmath}
\usepackage{booktabs}
\usepackage{colortbl}
\usepackage{float}
\usepackage{graphicx}
\usepackage{hyperref}
\usepackage{microtype}
\usepackage{needspace}
\usepackage{tabularx}
\usepackage{xcolor}
\usepackage{url}
\usepackage{xspace}

\hypersetup{
  pdftitle={Trace: A Taxonomy-Guided Environment for Multidomain Visual Reasoning},
  pdfauthor={Md Tanvirul Alam},
  pdfsubject={A taxonomy-guided procedural environment for multidomain visual reasoning and RLVR},
  pdfkeywords={visual reasoning, multimodal reasoning, RLVR, procedural data generation, task taxonomy}
}

\newcommand{\trace}{\textsc{Trace}\xspace}

\definecolor{TraceChartsBand}{HTML}{E5EFF8}
\definecolor{TraceMathBand}{HTML}{E9EAF6}
\definecolor{TraceScienceBand}{HTML}{E3F0EE}
\definecolor{TraceSpatialBand}{HTML}{EAF2E6}
\definecolor{TracePerceptionBand}{HTML}{EEE9F3}
\definecolor{TracePuzzlesBand}{HTML}{EDF1F4}
\definecolor{TraceOverallBand}{HTML}{DCE5EB}
\definecolor{TraceGain}{HTML}{4E8A3A}
\definecolor{TraceLoss}{HTML}{C35A5A}

\makeatletter
\renewcommand{\@notice}{}
\makeatother

\title{%
  \begin{tabular}{@{}c@{\hspace{0.32em}}l@{}}
    \raisebox{-0.35\height}{\includegraphics[height=0.58in]{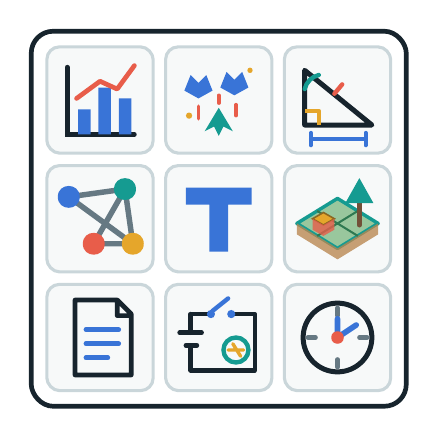}} &
    \begin{tabular}{@{}c@{}}
      Trace: A Taxonomy-Guided Environment\\
      for Multidomain Visual Reasoning
    \end{tabular}
  \end{tabular}
}

\author{%
  Md Tanvirul Alam \\
  Rochester Institute of Technology \\
  Rochester, NY, USA \\
  \texttt{ma8235@rit.edu} \\
  \url{https://maveryn.github.io/trace/}
}

\begin{document}
\raggedbottom

\maketitle

\begin{abstract}
\sloppy
Reinforcement learning with verifiable rewards (RLVR) has substantially improved
language-model reasoning, yet its extension to vision-language models remains
constrained by the lack of training data that are simultaneously broad, exactly
verifiable, and reproducible. We introduce \trace, a taxonomy-guided environment
for multidomain visual reasoning. \trace factorizes task construction into a
scene grammar and an executable task program, separating visual realization from
answer computation. A shared semantic state determines the rendered image,
prompt, typed answer, verifier state, and replayable instance trace. The resulting
environment comprises 1,000 tasks over 277 scene grammars and 11 visual domains,
with controlled semantic and visual variation. RLVR on 64,000 \trace instances
improves the macro-average across 24 external benchmarks by 3.51 percentage
points for Qwen2.5-VL-3B and 4.06 points for Qwen2.5-VL-7B, providing evidence
that broad procedural training can transfer beyond the generated task
distributions.
\textbf{Project page:} \url{https://maveryn.github.io/trace/}
\end{abstract}

\section{Introduction}
\label{sec:introduction}

Reinforcement learning with verifiable rewards (RLVR) has emerged as an effective
approach for improving language-model reasoning, particularly in mathematical and
programmatic domains where solutions can be evaluated with inexpensive, objective
feedback~\citep{shao2024deepseekmath,guo2025deepseek}. Extending this paradigm to
vision-language models requires training data whose questions are grounded in
visual content, whose answers can be verified exactly, and whose variation extends
beyond a narrow set of templates. Recent multimodal RL methods report gains in
visual mathematics, counting, spatial reasoning, and general visual
understanding~\citep{liu2025visualrft,huang2025visionr1,peng2025lmmr1,
meng2025mmeureka}, yet constructing visual training data that support broad and
transferable reasoning remains a central bottleneck.

Existing work has pursued two main strategies. Large training mixtures broaden
coverage by combining many datasets with task-specific rewards
~\citep{sarch2026vero}, but they inherit heterogeneous task granularity, sampling
units, and reasoning boundaries from their source collections. Procedural systems
instead provide exact supervision and controlled generation, but typically focus
on a single scene grammar or a small set of closely related objectives, such as
jigsaws, visual logic puzzles, games, or abstract perception tasks
~\citep{wang2025jigsawr1,feng2025visualsphinx,tong2025gamerl,
alam2026sphinx,yang2026tron}. As a result, breadth, controllability, and exact
verification are rarely combined within one coherent task organization.

We introduce \trace, a taxonomy-guided environment that addresses this gap by
separating visual construction from reasoning computation. A scene grammar defines
the objects, relations, and layout family that can be rendered; a task program
specifies the computation and answer type; and a reward contract defines exact
verification. Bounded query variation changes a meaningful task argument, such as
extremum direction, without creating a separate task. Building on structured scene
representations and executable question programs in CLEVR and Task Me Anything
~\citep{johnson2017clevr,zhang2024taskmeanything}, \trace uses this decomposition
not only to generate instances, but also to define the units used for sampling,
verification, and cross-domain analysis.

\begin{figure}[t]
  \centering
  \includegraphics[width=\textwidth]{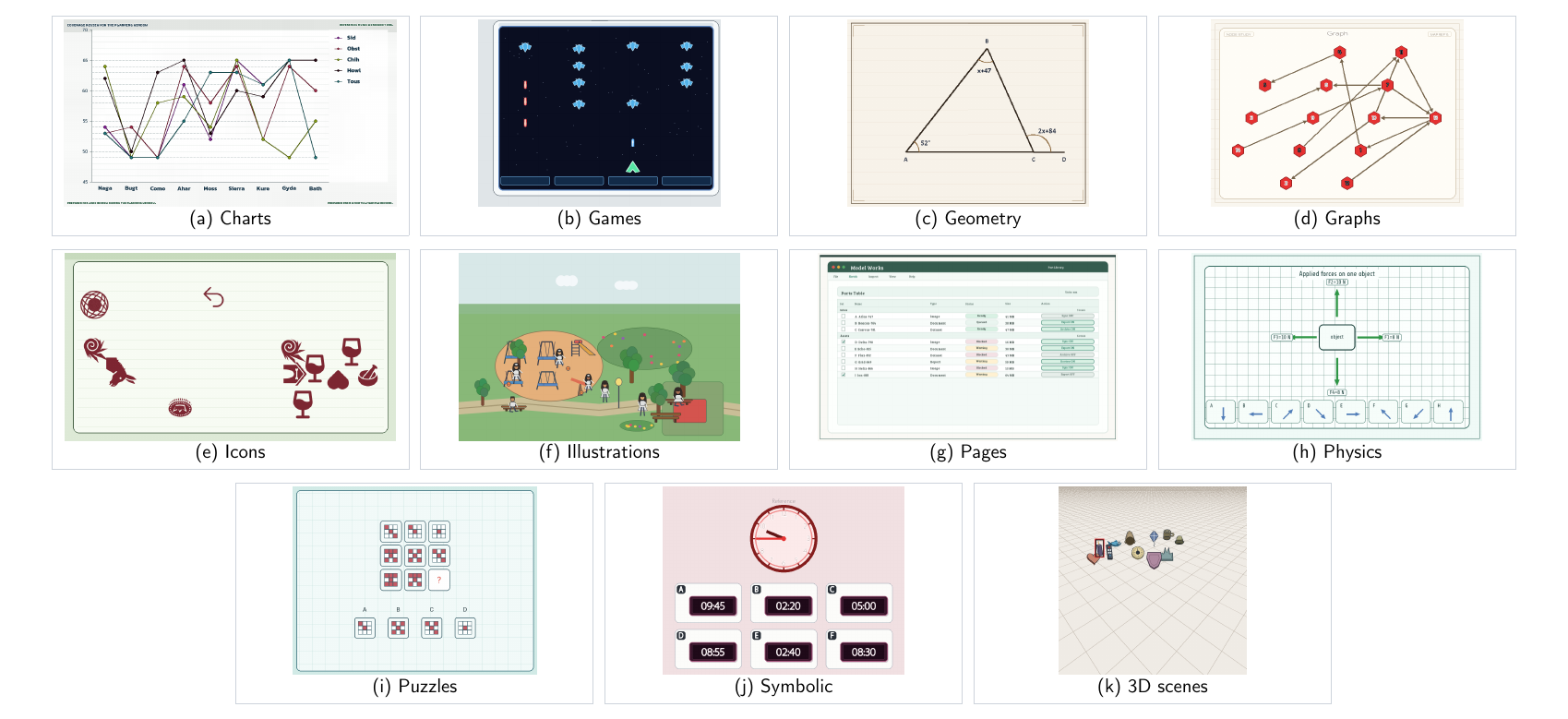}
  \caption{Representative instances from the 11 visual domains in \trace,
  spanning charts, games, geometry, graphs, icons, illustrations, pages,
  physics, puzzles, symbolic notation, and three-dimensional scenes.}
  \label{fig:domain-montage}
\end{figure}

This organization yields 1,000 tasks over 277 scene grammars spanning 11 visual
domains. Each instance is generated from a semantic state on which the task program
is executed; the same state determines the image, prompt, typed answer, and verifier
state. Supervision therefore follows directly from the task computation rather than
from labels assigned after rendering. The environment supports controlled variation
in both semantic content and visual realization while preserving stable task
identity and exact rewards. Figure~\ref{fig:domain-montage} illustrates the resulting
visual breadth.

We use \trace to test whether RLVR on procedural visual tasks transfers beyond
held-out instances from the same environment. Qwen2.5-VL-3B and
Qwen2.5-VL-7B are trained on the same 64,000 \trace instances and evaluated on
24 external visual-reasoning benchmarks. Training improves the benchmark
macro-average by 3.51 percentage points at 3B and 4.06 points at 7B, providing
evidence that broad procedural training can support transfer beyond the
generated task distributions.

Our contributions are:
\begin{itemize}
    \item A program-centered taxonomy that separates scene grammars, executable
    task programs, and bounded query variation, providing stable task units across
    1,000 tasks, 277 scene grammars, and 11 visual domains;

    \item An executable generator--renderer--verifier environment in which images,
    prompts, typed answers, verifier states, and replayable instance traces derive
    from shared semantic state, enabling exact supervision and controlled semantic
    and visual variation; and

    \item A two-scale RLVR study showing gains of 3.51 and 4.06 points on the
    24-benchmark macro-average at 3B and 7B, respectively, with positive mean
    changes on 21 of 24 benchmarks at 3B and all 24 benchmarks at 7B.
\end{itemize}
\section{Related Work}
\label{sec:related}

\paragraph{Controlled visual task generation.}
Structured representations have long been used to make visual-reasoning datasets
controllable and diagnostically useful. CLEVR generates synthetic scenes from
explicit scene graphs and associates questions with functional programs, enabling
controlled evaluation of compositional reasoning; GQA extends this paradigm to
compositional questions grounded in real-image scene graphs
\citep{johnson2017clevr,hudson2019gqa}. PuzzleVQA automatically constructs
abstract-pattern questions with reasoning explanations and uses them to distinguish
failures in visual perception, inductive reasoning, and deductive reasoning
\citep{chia2024puzzlevqa}. Task Me Anything generalizes programmatic benchmark
construction into an on-demand engine built around an extensible taxonomy of
visual assets, attributes, relations, and task plans
\citep{zhang2024taskmeanything}. WorldBench instead uses a broad taxonomy of
visual concepts to curate diverse real images and manually construct challenging
questions that require reasoning beyond object recognition
\citep{yin2026worldbench}. These works establish explicit structure as a basis for
controlling evaluation coverage. \trace builds on this lineage but makes the
executable task definition a stable unit not only of question generation, but also
of sampling, verification, and RL training.

\paragraph{Executable generator--verifier environments.}
Procedural reasoning environments replace fixed example collections with programs
that generate instances and evaluate their solutions. Reasoning Gym provides more
than one hundred generator--verifier environments across arithmetic, algebra,
logic, graphs, geometry, computation, and games, with instance difficulty controlled
through generation parameters \citep{stojanovski2025reasoninggym}. Enigmata
develops the same model for puzzle reasoning through 36 tasks in seven categories,
each equipped with a controllable generator and rule-based verifier, and studies
their use in multi-task RLVR \citep{chen2025enigmata}. These systems demonstrate
how inspectable generation and exact rewards can support effectively unbounded
training distributions. They primarily operate over textual or symbolic inputs,
whereas \trace extends the generator--verifier model to rendered visual states and
separates the visual scene grammar from the task program executed over that state.

\paragraph{Multimodal RL data and mixture design.}
A major line of multimodal RL research constructs broad training mixtures from
existing datasets and assigns rewards according to their answer formats.
Visual-RFT, Vision-R1, LMM-R1, MM-Eureka, R1-Onevision, Reason-RFT, and
VLM-R1 develop RL data and reward functions spanning classification, visual
mathematics, counting, structural perception, spatial reasoning, and referring
expressions
\citep{liu2025visualrft,huang2025visionr1,peng2025lmmr1,
meng2025mmeureka,yang2025r1onevision,tan2025reasonrft,shen2025vlmr1}.
Vero scales this approach to 600K examples drawn from 59 datasets across six
task categories, using task-routed rewards to accommodate heterogeneous output
spaces; its mixture ablations further indicate that broad category coverage is
important for transfer \citep{sarch2026vero}. More generally, OpenThoughts
emphasizes that data sources, filtering, teacher generation, and mixture proportions
should be treated as explicit and reproducible parts of a reasoning-data recipe
\citep{guha2025openthoughts}.

A complementary line asks which examples and source proportions are most useful.
SoTA with Less estimates sample difficulty through Monte Carlo tree search and
retains a compact set of challenging examples, while RAP identifies high-value
multimodal examples using discrepancies between multimodal and text-only behavior
together with attention-based confidence
\citep{wang2025sotawithless,li2025truth}. MoDoMoDo formulates multidomain
RLVR as a mixture-optimization problem and learns to predict downstream training
outcomes from the source distribution \citep{liang2025modomodo}. Vision-G1
combines data from 46 sources with influence-based selection, difficulty filtering,
and curriculum training, whereas WeThink constructs a 120K-example mixture from
18 sources and combines rule-based and model-based rewards
\citep{zha2026visiong1,yang2025wethink}. DeepVision-103K broadens verified
training data within visual mathematics through large-scale synthesis and curation
\citep{sun2026deepvision}. Across these approaches, source selection, mixture
weighting, and reward routing are central design variables. \trace instead defines
its sampling units through authored task programs before instances are generated,
rather than inheriting task granularity from source datasets.

\paragraph{Synthetic and verifiable visual RL data.}
The most closely related systems create new visual experience specifically for
reinforcement learning. Jigsaw-R1 uses jigsaw puzzles as a controlled testbed for
studying visual RL, including generalization across puzzle configurations and the
interaction between RL, supervised warm starts, and explicit reasoning
\citep{wang2025jigsawr1}. VisualSphinx expands a small collection of puzzle
rules through a rule-to-image pipeline that generates executable rendering code,
grounded questions, and verifiable answers \citep{feng2025visualsphinx}. Game-RL
adapts executable game code into visual states and verifiable tasks, thereby
separating game environments from the objectives defined over them
\citep{tong2025gamerl}. COGS takes a compositional approach: it decomposes seed
questions into primitive perception and reasoning factors, recombines those factors
with new images, and derives process-level rewards from the resulting intermediate
subquestions and answers \citep{gu2025cogs}.

Other systems construct verifiable objectives by modifying existing data.
SynthRL produces harder, answer-preserving variants of visual-mathematics
questions and applies an explicit verification stage, whereas ViCrit injects a
localized error into an otherwise valid image description and rewards the model
for identifying the corrupted span
\citep{wu2025synthrl,wang2025vicrit}. Sphinx procedurally generates 25 task
families over motifs, tiles, charts, icons, and geometric primitives, with
deterministic answers for both evaluation and RLVR training
\citep{alam2026sphinx}. Concurrent TRON provides 520 online
generator--verifier environments across five ability groups, generating fresh
instances during training and auditing generation reliability, diversity, and
difficulty \citep{yang2026tron}.

Game-RL distinguishes games from their associated tasks within one visual domain,
while TRON treats each generator--verifier program as an online environment.
\trace extends the structural separation across 11 visual domains: 277 scene
grammars define the states that can be rendered, while 1,000 stable task programs
define the computations and reward contracts applied to those states. This
factorization allows multiple reasoning objectives to share a scene grammar and
allows visual realization to change without redefining the task being sampled,
verified, or analyzed.
\section{A Program-Centered Task Taxonomy}
\label{sec:taxonomy}

A procedural environment requires a stable unit of generation, sampling,
verification, and analysis. Defining this unit is nontrivial: changes to the
reasoning objective should produce distinct tasks, whereas changes to prompt
arguments or visual realization should not artificially inflate the task
inventory. \trace addresses this problem through the hierarchy
\begin{equation}
  \text{domain} \longrightarrow \text{scene grammar} \longrightarrow \text{task}.
  \label{eq:public-taxonomy}
\end{equation}
A domain groups related visual inputs for reporting and mixture balancing. A
scene grammar defines a reusable family of semantic states and visual
realizations through its object vocabulary, relations, layout family, and
question-facing scaffold. A task couples that grammar to an executable task
program, answer schema, and reward contract. This factorization separates the
visual structure of an instance from the computation used to answer it.

\subsection{Task identity and equivalence}

We represent a task as
\begin{equation}
  T = (\mathcal{S}, P, \mathcal{Y}, \mathcal{C}),
\end{equation}
where $\mathcal{S}$ is the scene grammar, $P$ is the task program,
$\mathcal{Y}$ is the answer schema, and $\mathcal{C}$ is the reward contract.
The task program specifies how candidate objects are constructed, the roles of
its operands, any intermediate computations, the final operator, and the
binding between the computed result and the returned answer. Broad descriptions
such as \textit{count objects} or \textit{select an extremum} therefore do not
identify a task precisely enough for sampling or analysis.

Within \trace, two candidate task definitions are merged only when
\begin{equation}
  T_i \equiv T_j
  \quad\Longleftrightarrow\quad
  \mathcal{S}_i \simeq \mathcal{S}_j,\;
  P_i \simeq P_j,\;
  \mathcal{Y}_i = \mathcal{Y}_j,\;
  \mathcal{C}_i \simeq \mathcal{C}_j.
  \label{eq:task-equivalence}
\end{equation}
Here, $\mathcal{S}_i \simeq \mathcal{S}_j$ requires the same object
vocabulary, relations, layout family, and visible scaffold;
$P_i \simeq P_j$ requires the same candidate construction, operand roles,
intermediate operations, final operator, and output binding; and
$\mathcal{C}_i \simeq \mathcal{C}_j$ requires equivalent answer
normalization and scoring behavior. These relations permit the renaming of
literal operands, such as a color or category, but not changes to the
underlying visual or computational structure. A change to any of these
elements is represented as a separate task in \trace, even when the resulting
prompts appear similar or share the same answer type.

Table~\ref{tab:task-boundaries} summarizes the operational consequences of
this rule.

\begin{table}[t]
  \centering
  \small
  \caption{Operational examples of the \trace task-boundary rule. Changes to
  bounded program arguments remain queries, while semantic and render-only
  variation preserve task identity. Changes to the computation or scene
  grammar define a new task.}
  \label{tab:task-boundaries}
  \begin{tabular}{@{}p{0.30\columnwidth}p{0.43\columnwidth}p{0.21\columnwidth}@{}}
    \toprule
    Variation & Effect on task definition & Classification \\
    \midrule
    Palette or layout
      & Fixed semantic state, task program, and reward contract unchanged
      & Render-only generation \\
    Item count or sampled values
      & Semantic state changes within the same scene grammar and task program
      & Semantic generation \\
    Largest vs. smallest IQR
      & Only extremum direction changes
      & Query \\
    Shape vs. shape-and-color count
      & Predicate arity and selected set change
      & New task \\
    AND vs. inclusive OR count
      & Intermediate set operator changes
      & New task \\
    Direct vs. algebraic exterior angle
      & Derivation rule changes
      & New task \\
    Same extremum family in another scene grammar
      & Scene grammar changes
      & New task \\
    \bottomrule
  \end{tabular}
\end{table}

\subsection{Task, query, and generation boundaries}

A task may expose an internal query variable $q$ that selects a bounded
argument of $P$ while preserving the task program, answer schema, and reward
contract. For example, choosing between the largest and smallest interquartile
range changes the extremum direction but not the candidate set, metric, output
binding, or verification rule.

\begin{figure}[t]
  \centering
  \includegraphics[width=\textwidth]{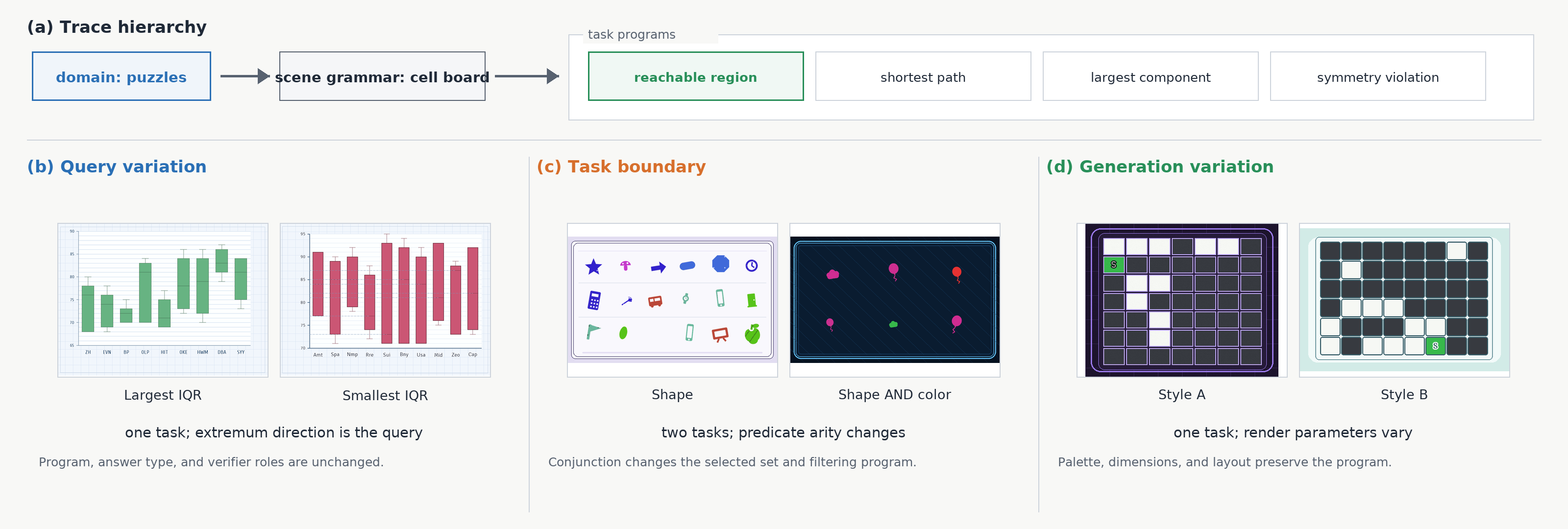}
  \caption{Examples of the task boundary in \trace. \textbf{Left:} Changing
  extremum direction modifies only a bounded query argument and therefore
  remains within one boxplot task. \textbf{Center:} Adding a conjunctive
  attribute changes the candidate predicate and selected set, producing a new
  icon-filtering task. \textbf{Right:} Changes to board dimensions, layout, and
  visual treatment remain generation parameters because the reachability
  program and reward contract are unchanged.}
  \label{fig:taxonomy-boundaries}
\end{figure}

Generation parameters $\theta$ instead determine the instance sampled from a
fixed task definition. Semantic parameters vary the underlying state---such as
object counts, values, thresholds, or board dimensions---and may therefore
change the answer. Render-only parameters vary the visual realization of a
fixed semantic state and query while preserving all answer-relevant relations
and the resulting reward. Figure~\ref{fig:taxonomy-boundaries} illustrates
these distinctions.

This separation prevents both forms of inventory distortion. Treating every
operand choice as a new task would overcount a single computation, whereas
merging distinct intermediate operations would obscure meaningful differences
in reasoning.

\subsection{Canonical task programs and extensibility}

Although task identity remains bound to a scene grammar, task-program operators
are canonicalized across grammars so that recurring computational structures
can be analyzed jointly. For example, selecting the boxplot with the extreme
IQR and selecting the grid line with the extreme shape count both instantiate
the canonical signature
\[
  \emph{filter candidates}
  \rightarrow
  \emph{compute metric}
  \rightarrow
  \emph{select extremum}
  \rightarrow
  \emph{return candidate label}.
\]
They nevertheless remain distinct tasks because they operate over different
scene grammars, construct different metrics, and bind the result to different
visual entities. Canonicalization therefore exposes shared reasoning structure
without collapsing scene-specific task identity.

The same distinction determines the training mixture. \trace samples tasks
uniformly by default and samples query values only after a task has been
selected. Splitting a task into operand-specific variants would consequently
overweight one program, whereas merging programs with different computations
would hide heterogeneous reasoning behind a single sampling unit and aggregate
score.

The factorization also provides a controlled path for extension. A new scene
grammar introduces a new family of semantic states and visual realizations,
while a new task adds an executable program and reward contract to an existing
grammar. Renderers can therefore evolve without redefining the tasks applied
to their semantic states, and additional reasoning objectives can reuse an
existing scene grammar without duplicating its visual construction.
\section{The \trace Environment}
\label{sec:design}

The taxonomy defines what constitutes a task; the environment instantiates that
definition as a deterministic generator--renderer--verifier pipeline. For each
sampled task, \trace constructs a semantic state, executes the corresponding
task program, renders the image and prompt, and binds the resulting answer and
verifier state to an exact scorer. This design keeps generation, supervision,
and replay grounded in the same underlying state.

\subsection{Executable instance generation}

Each generated instance is represented as
\begin{equation}
  \mathcal{D}_i =
  \bigl(I_i, p_i, y_i, c_i, \tau_i\bigr),
  \label{eq:instance-contract}
\end{equation}
where $I_i$ is the rendered image, $p_i$ is the prompt, $y_i$ is the typed
answer, $c_i$ is the instance-bound scorer, and $\tau_i$ is the corresponding
instance trace. The trace records the semantic state, sampled query, task
execution, rendering decisions, and verifier state, making each instance
replayable and supporting failure analysis.

Let $s$ denote a scene grammar, $t$ a task, $q$ an internal query, $z$ an
instance seed, and $\theta$ the sampled generation parameters. Instance
construction proceeds as
\begin{align}
  x &= G_s(z, \theta_s), &
  (y,v) &= P_t(x;q), \label{eq:semantic-execution}\\
  I &= R_s(x;z,\theta_r), &
  p &= H_{s,t}(x,q;z,\theta_p), \label{eq:render-prompt}\\
  c &= \operatorname{bind}\!\left(\mathcal{C}_t;y,v\right), &
  \tau &= \Gamma(s,t,q,z,\theta,x,v).
  \label{eq:projection-contract}
\end{align}
The scene generator $G_s$ first constructs semantic state $x$. The task
program $P_t$ then executes over that state and returns both the typed answer
$y$ and verifier state $v$. The renderer $R_s$ and prompt function $H_{s,t}$
independently realize the image and question from the same semantic state.
Finally, the task-level reward contract $\mathcal{C}_t$ is bound to the
expected answer and verifier state to produce the instance scorer $c$, while
$\Gamma$ records the complete trace.

Semantic, prompt, and rendering choices are governed by independent seeded
streams. The resulting instance can therefore be reconstructed from its
recorded state without coupling superficial prompt or rendering variation to
the task computation.

\trace supports four answer interfaces: integers, canonical numeric values,
strings, and option letters. Each reward contract specifies the corresponding
type-aware normalization and comparison rule, which the bound scorer applies
to the submitted answer.

\begin{figure}[t]
  \centering
  \includegraphics[width=\textwidth]{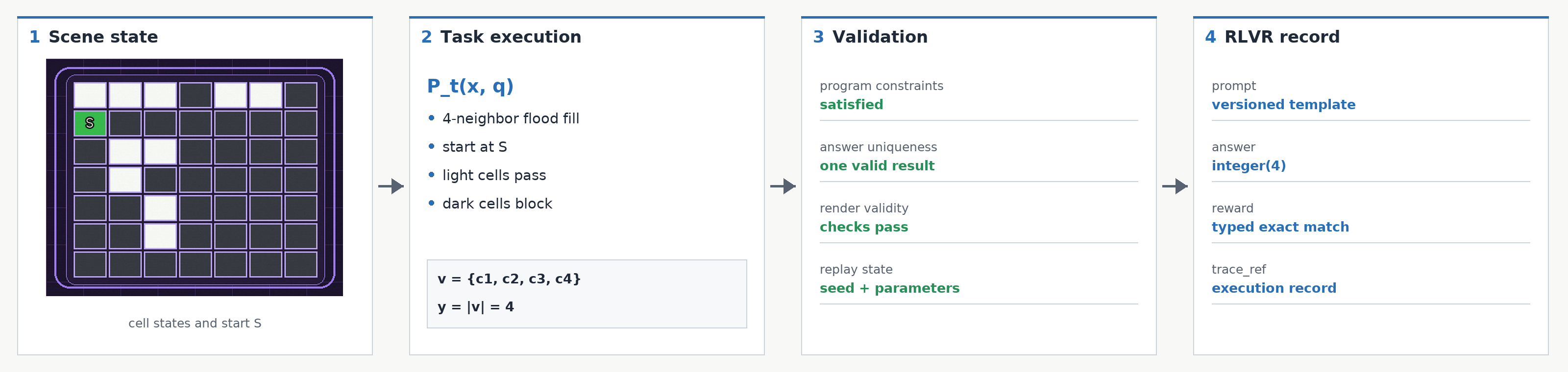}
  \caption{End-to-end construction of a \trace instance. Four-neighbor
  reachability is computed over the semantic cell board, and the resulting
  state determines the typed answer, rendered image, prompt, scorer, and
  replayable instance trace.}
  \label{fig:reachable-pipeline}
\end{figure}

Figure~\ref{fig:reachable-pipeline} illustrates the full pipeline for a
cell-board task. A four-neighbor flood fill from the marked start cell produces
the set of reachable passable cells, and the cardinality of that set gives
$y=4$. Passable cells outside the connected region remain visual distractors
but do not enter the task computation.

\subsection{Environment composition}

\begin{figure}[t]
  \centering
  \includegraphics[width=\textwidth]{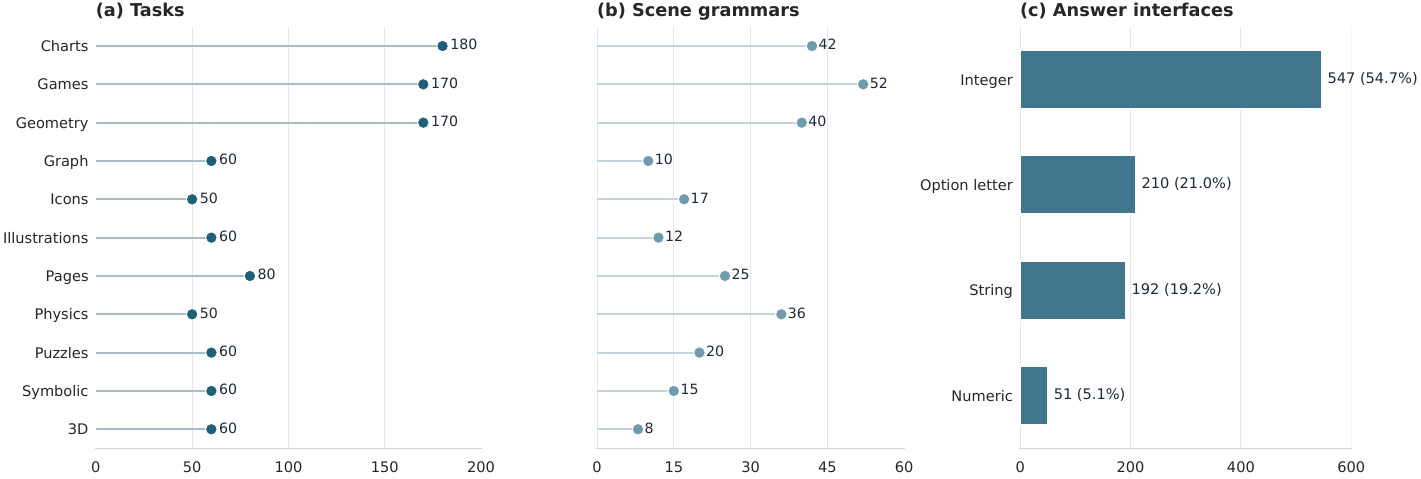}
  \caption{Composition of \trace across 11 visual domains. The panels report
  the number of task programs and scene grammars in each domain, together with
  the overall distribution of answer interfaces.}
  \label{fig:environment-composition}
\end{figure}

\trace comprises 1,000 tasks defined over 277 scene grammars spanning 11
visual domains. The domains organize related visual structures, rendering
conventions, and object vocabularies for reporting and balancing; they are not
themselves reasoning categories or sampling units.
Figure~\ref{fig:environment-composition} summarizes the distribution of tasks,
scene grammars, and answer interfaces across domains.
Figure~\ref{fig:domain-montage} provides representative renderings, and
Appendix~\ref{app:task-atlas} presents a domain-stratified task atlas.

The 277 scene grammars support a median of three tasks each
(mean 3.61, range 1--26), indicating that visual construction is reused across
multiple reasoning objectives rather than duplicated for each task. The task
inventory includes integer, canonical numeric, string, and option-letter
outputs; 790 tasks use open-form integer, numeric, or string answers.

Of the 1,000 tasks, 317 expose more than one internal query, producing 1,475
query variants in total. Because task selection precedes query selection,
adding bounded query variants does not increase the sampling probability of the
underlying task. The sampling distribution therefore reflects the authored task
programs rather than the number of prompt-level variants available to each one.

\begin{figure}[t]
  \centering
  \includegraphics[width=\textwidth]{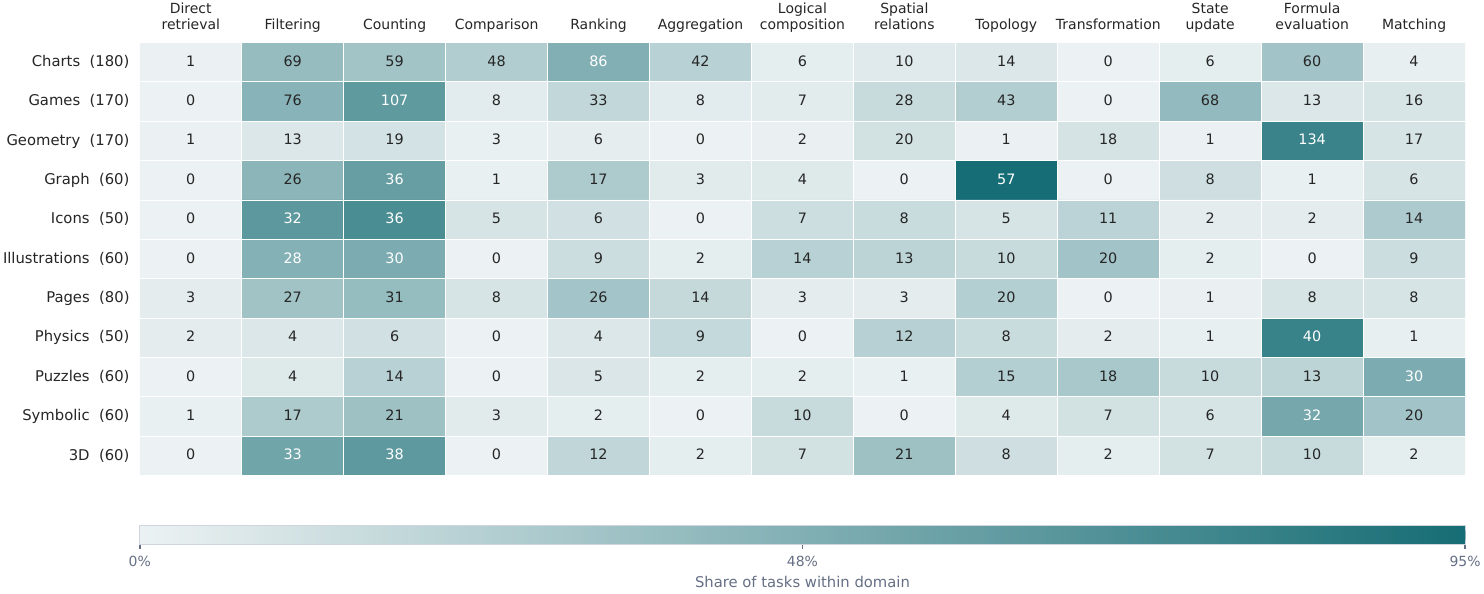}
  \caption{Distribution of task-program operations across visual domains.
  Cell labels report the number of tasks assigned to each operation family,
  while color indicates the corresponding within-domain share. Assignments are
  exhaustive and multi-label.}
  \label{fig:operation-coverage}
\end{figure}

Visual diversity alone does not establish diversity of computation. We
therefore project the task inventory onto 13 compositional operation families,
whose distribution across domains is shown in
Figure~\ref{fig:operation-coverage}. These families are an analytical view of
the task programs rather than an additional taxonomy level. Assignments are
multi-label because a composed program may require several operations; every
task belongs to at least one family, and \emph{direct retrieval} is assigned
only when no other operation materially determines the answer.
Appendix~\ref{app:operation-families} defines the complete set of families.

\subsection{Controlled visual realization}

Instance variation occurs at two distinct levels. Semantic controls alter the
state on which the task program operates, including dimensions, cardinalities,
values, thresholds, sequence lengths, and candidate sets. Prompt templates draw
their dynamic labels and values from that same state, ensuring that the
question and answer remain synchronized with task execution. All sampled
choices are retained in the instance trace.

Once semantic execution fixes the state, query, and answer, scene-compatible
realization controls vary how that state is presented. These controls include
placement, panel arrangement, camera, palette, typography, materials,
non-answer context, and bounded raster effects. A realization control is
admitted only when it preserves the relations queried by the task program and
therefore leaves the typed answer unchanged. Appendix~\ref{app:generation}
formalizes these variation layers and their required invariants.

For information-rich scenes, \trace may additionally introduce curated
context outside answer-bearing regions. Such context is constrained by
scene-specific fit, collision, and contrast checks so that visual complexity
can increase without modifying the underlying task state.

\begin{figure}[t]
  \centering
  \includegraphics[width=0.72\textwidth]{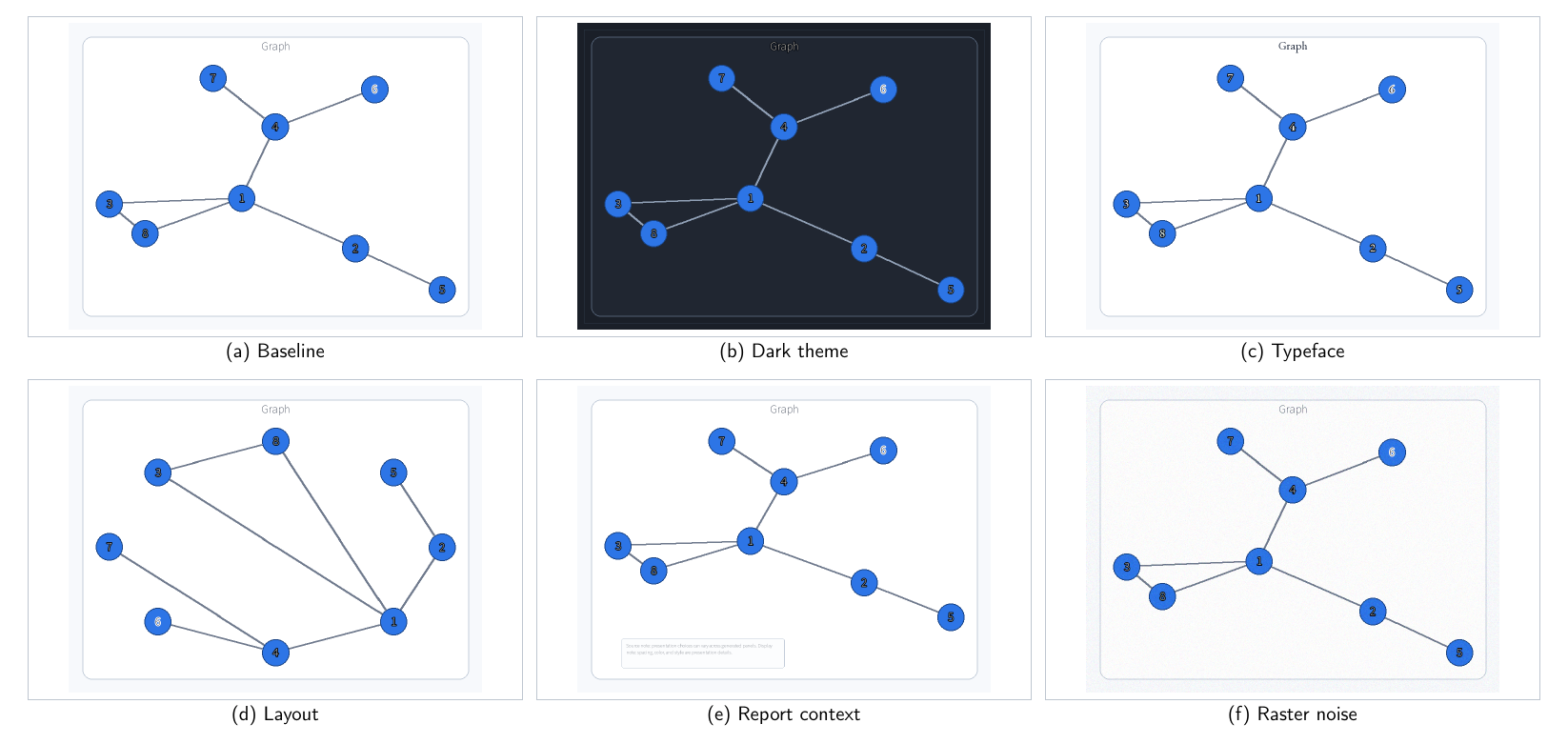}
  \caption{Answer-preserving realizations of a fixed semantic graph. Theme,
  typeface, layout, non-answer context, and raster treatment vary while the
  prompt, topology, query, and typed answer remain unchanged.}
  \label{fig:rendering-variation}
\end{figure}

\subsection{Instance validation}

An instance is retained only if task execution produces a unique typed answer,
the prompt and rendered image remain consistent with the semantic state, and
deterministic replay reconstructs the same instance. Scene-specific rendering
checks additionally test visibility, contrast, fit, and collisions.

These structural checks cannot guarantee that every generated question is
semantically clear or that every valid rendering is equally legible. We
therefore also inspect samples from every task for semantic clarity,
prompt--image consistency, and visual legibility, complementing automatic
validation with direct review of the rendered instances.

\section{Experimental Setup}
\label{sec:experiments}

Our experiments test whether RLVR on procedurally generated \trace instances
improves visual reasoning beyond the distributions represented by the training
tasks. We initialize from Qwen2.5-VL-3B-Instruct and
Qwen2.5-VL-7B-Instruct~\citep{bai2025qwen25vl} and apply the same training
procedure at both model scales. Each resulting checkpoint is compared with its
corresponding base checkpoint under a common external-evaluation protocol.

\subsection{Training data and reward}

Both model scales are trained on the same fixed set of 64,000 instances,
comprising 64 independently generated examples from each of the 1,000 \trace
tasks. A separate held-out \trace validation set contains two previously unseen
instances per task, for 2,000 examples in total. All reported results use the
final checkpoint after 500 updates.

The model may produce an unrestricted reasoning trace, but its response must
terminate with a JSON object of the form
\texttt{\{"answer": ...\}}. Let $R_a \in \{0,1\}$ indicate whether the
submitted answer exactly matches the target after type-aware canonicalization,
and let $R_f \in \{0,1\}$ indicate whether the response ends with a valid JSON
object containing exactly the requested key. The reward is
\begin{equation}
  R = 0.95 R_a + 0.05 R_f.
\end{equation}
Answer correctness therefore determines the reward, while the smaller format
term encourages a consistently parseable output interface.

\subsection{Optimization}

We optimize both models with group relative policy optimization (GRPO). Each
update samples 128 prompts and generates eight responses per prompt. Rewards
are normalized within each eight-response group and optimized using a clipped
token-level policy objective. The 500 updates correspond to one shuffled pass
over the 64,000 training prompts and produce 512,000 sampled responses in
total.

We use full-parameter training with BF16 parameters, a learning rate of
$10^{-6}$, and one policy epoch per update. Rollout sampling uses temperature
1.0 and top-$p$ 1.0. Training is performed on eight H100 80GB GPUs and takes
12.2 hours for the 3B model and 13.9 hours for the 7B model. The two scales
share the same data, reward, sampling schedule, and optimization
hyperparameters; Appendix~\ref{app:experiments} provides the complete
configuration.

\subsection{External evaluation}

To measure transfer beyond the \trace task distributions, we evaluate each
base and trained checkpoint on 24 external benchmarks spanning six analysis
groups: charts and tables, visual mathematics, science and general reasoning,
spatial reasoning, perception and counting, and puzzles and logic. Each group
contains four benchmarks. Table~\ref{tab:main-results} reports the complete
per-benchmark results, while Appendix Table~\ref{tab:evaluation-suite} lists
the evaluated splits, sample counts, and benchmark-specific references.

We evaluate each designated split in full, yielding 32,805 examples per model
and decoding seed. Each benchmark retains its native evaluation metric.
Category scores are computed as unweighted means of the four benchmarks in
that group, and the overall score is the unweighted mean across all 24
benchmarks. This macro-averaging prevents benchmarks with larger evaluation
sets from dominating the aggregate result.

For each fixed checkpoint, we repeat decoding with three seeds and report the
mean and sample standard deviation. The seeds affect only response generation;
the model weights remain unchanged. All checkpoints are evaluated with the same
benchmark prompts, image-preprocessing bounds, parsers, scorers, and
VLMEvalKit version~\citep{duan2024vlmevalkit}. Exact inference settings are provided in
Appendix~\ref{app:experiments}.

At the 7B scale, we additionally evaluate publicly available synthetic-data
RLVR checkpoints from Game-RL, Sphinx, and
PC-GRPO~\citep{tong2025gamerl,alam2026sphinx,jeddi2025pcgrpo}. We report
Vero~\citep{sarch2026vero} separately as a real-image reference because its
600K-example training mixture draws from 59 datasets and includes real-image
data, whereas our runs use 64K procedurally generated instances. All
checkpoints are evaluated under the same external protocol. Because their
training data, optimization, and compute are not matched, these comparisons
characterize checkpoint outcomes rather than isolate the effect of a specific
training method or data source.

\section{Results}
\label{sec:results}

\subsection{\trace validation}

RLVR substantially improves performance on unseen instances generated from the
same task programs. Accuracy increases from 24.45 to 41.05 at 3B and from
34.25 to 51.55 at 7B, corresponding to gains of 16.60 and 17.30 percentage
points, respectively. These results measure generalization to new semantic and
visual realizations within the \trace task distributions; the external
benchmarks in Section~\ref{sec:external-transfer} test whether the gains extend beyond
those distributions.

\begin{figure}[t]
  \centering
  \includegraphics[width=0.80\linewidth]{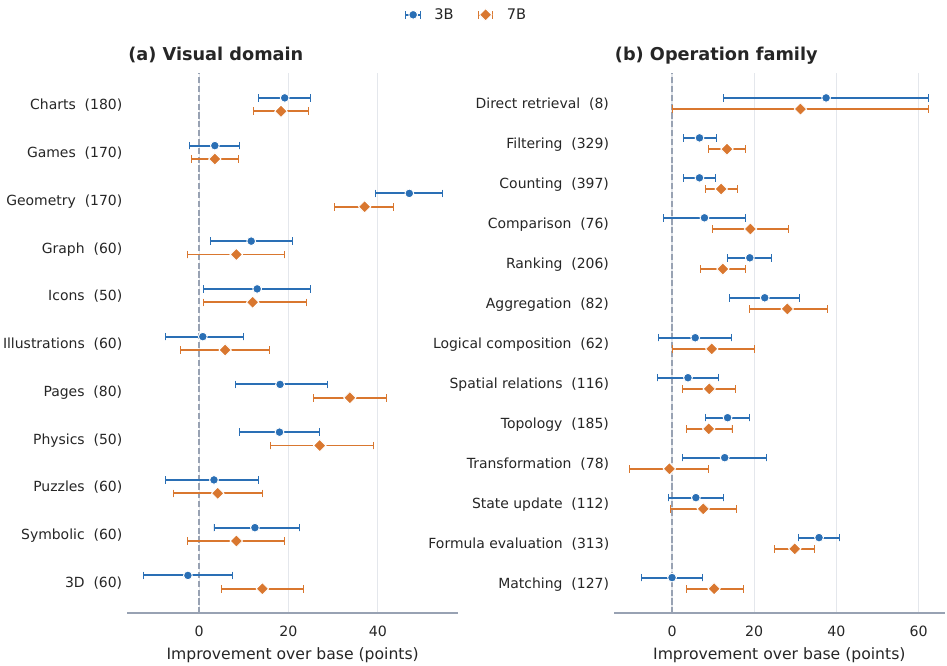}
  \caption{Held-out \trace accuracy gains by visual domain (left) and operation
  family (right). Whiskers show 95\% task-cluster bootstrap intervals;
  parentheses give task counts, and operation-family assignments overlap.}
  \label{fig:iid-taxonomy-gains}
\end{figure}

\begin{figure}[t]
  \centering
  \includegraphics[width=0.50\linewidth]{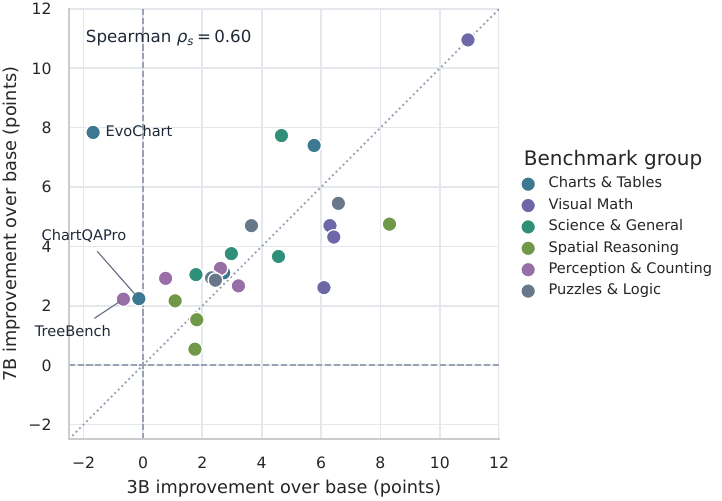}
  \caption{Benchmark gains at 3B versus 7B. Colors denote benchmark groups;
  dashed axes mark zero, the dotted diagonal marks equal gains, and
  $\rho_s=0.60$.}
  \label{fig:scale-gain-correlation}
\end{figure}

The improvements are distributed across domains and computational structures
rather than being driven by a single task family. Formula evaluation and
aggregation show the largest operation-family gains at both model scales,
whereas the 3D domain, transformation, and matching exhibit stronger
scale-dependent behavior. Performance also improves across all four answer
interfaces and for both single-query and multi-query tasks. Numeric-answer
tasks show the largest increase, although this interface contains only 51
tasks, while option-letter tasks show the smallest. Appendix
Table~\ref{tab:iid-slice-results} reports these overlapping slices using two
instances per task and one decoding seed.

\newpage
\begin{table}[t]
  \centering
  \fontsize{7.5}{8.5}\selectfont
  \setlength{\tabcolsep}{0.5pt}
  \renewcommand{\arraystretch}{1.04}
  \caption{Per-benchmark transfer on the 24-benchmark external evaluation suite. Each score reports the mean and sample standard deviation over three decoding seeds, in percent. Smaller green/red subscripts in the \trace columns give the change from the corresponding base model. Bold marks the best score among the base model and synthetic-data RLVR checkpoints at each model scale. Category and overall averages are unweighted across benchmarks.}
  \label{tab:main-results}
  \begin{tabularx}{\textwidth}{@{}>{\raggedright\arraybackslash}p{0.16\textwidth} *{5}{>{\centering\arraybackslash}X}|>{\centering\arraybackslash}X|*{2}{>{\centering\arraybackslash}X}@{}}
    \toprule
     & \multicolumn{5}{c|}{7B base + synthetic RLVR} & \multicolumn{1}{c|}{Real-image} & \multicolumn{2}{c}{3B base + synthetic RLVR} \\
    \cmidrule(lr){2-6}\cmidrule(lr){7-7}\cmidrule(l){8-9}
    Benchmark & Base & \trace & Game-RL & Sphinx & PC-GRPO & Vero & Base & \trace \\
    \midrule
    \rowcolor{TraceChartsBand} \multicolumn{9}{@{}l@{}}{\textbf{Charts \& Tables}} \\
    \rowcolor{TraceChartsBand} ChartQAPro & 45.8\ensuremath{\mathord{\pm}}0.2 & \textbf{48.1}\ensuremath{\mathord{\boldsymbol{\pm}}}\textbf{0.4}\raisebox{-0.45ex}{\fontsize{4}{4}\selectfont\color{TraceGain}{+2.2}} & 45.4\ensuremath{\mathord{\pm}}0.4 & 43.9\ensuremath{\mathord{\pm}}2.3 & 39.3\ensuremath{\mathord{\pm}}1.2 & 40.9\ensuremath{\mathord{\pm}}0.4 & \textbf{31.6}\ensuremath{\mathord{\boldsymbol{\pm}}}\textbf{0.7} & 31.4\ensuremath{\mathord{\pm}}1.3\raisebox{-0.45ex}{\fontsize{4}{4}\selectfont\color{TraceLoss}{-0.1}} \\
    \rowcolor{TraceChartsBand} CharXivReason & 39.7\ensuremath{\mathord{\pm}}1.2 & \textbf{47.1}\ensuremath{\mathord{\boldsymbol{\pm}}}\textbf{0.4}\raisebox{-0.45ex}{\fontsize{4}{4}\selectfont\color{TraceGain}{+7.4}} & 39.7\ensuremath{\mathord{\pm}}1.0 & 40.7\ensuremath{\mathord{\pm}}0.4 & 41.2\ensuremath{\mathord{\pm}}0.7 & 45.8\ensuremath{\mathord{\pm}}0.8 & 28.9\ensuremath{\mathord{\pm}}1.1 & \textbf{34.7}\ensuremath{\mathord{\boldsymbol{\pm}}}\textbf{1.5}\raisebox{-0.45ex}{\fontsize{4}{4}\selectfont\color{TraceGain}{+5.8}} \\
    \rowcolor{TraceChartsBand} TableVQABench & 75.2\ensuremath{\mathord{\pm}}0.9 & \textbf{78.3}\ensuremath{\mathord{\boldsymbol{\pm}}}\textbf{0.2}\raisebox{-0.45ex}{\fontsize{4}{4}\selectfont\color{TraceGain}{+3.1}} & 75.6\ensuremath{\mathord{\pm}}0.8 & 74.9\ensuremath{\mathord{\pm}}1.7 & 72.3\ensuremath{\mathord{\pm}}2.5 & 78.5\ensuremath{\mathord{\pm}}0.5 & 69.3\ensuremath{\mathord{\pm}}0.9 & \textbf{72.0}\ensuremath{\mathord{\boldsymbol{\pm}}}\textbf{0.3}\raisebox{-0.45ex}{\fontsize{4}{4}\selectfont\color{TraceGain}{+2.7}} \\
    \rowcolor{TraceChartsBand} EvoChart & 57.1\ensuremath{\mathord{\pm}}0.7 & \textbf{64.9}\ensuremath{\mathord{\boldsymbol{\pm}}}\textbf{0.0}\raisebox{-0.45ex}{\fontsize{4}{4}\selectfont\color{TraceGain}{+7.8}} & 58.6\ensuremath{\mathord{\pm}}0.8 & 58.4\ensuremath{\mathord{\pm}}0.2 & 58.8\ensuremath{\mathord{\pm}}0.6 & 63.5\ensuremath{\mathord{\pm}}0.2 & \textbf{48.5}\ensuremath{\mathord{\boldsymbol{\pm}}}\textbf{0.5} & 46.8\ensuremath{\mathord{\pm}}1.4\raisebox{-0.45ex}{\fontsize{4}{4}\selectfont\color{TraceLoss}{-1.7}} \\
    \rowcolor{TraceChartsBand} \textit{Category average} & 54.5\ensuremath{\mathord{\pm}}0.3 & \textbf{59.6}\ensuremath{\mathord{\boldsymbol{\pm}}}\textbf{0.2}\raisebox{-0.45ex}{\fontsize{4}{4}\selectfont\color{TraceGain}{+5.1}} & 54.8\ensuremath{\mathord{\pm}}0.4 & 54.5\ensuremath{\mathord{\pm}}0.9 & 52.9\ensuremath{\mathord{\pm}}1.0 & 57.2\ensuremath{\mathord{\pm}}0.3 & 44.6\ensuremath{\mathord{\pm}}0.1 & \textbf{46.2}\ensuremath{\mathord{\boldsymbol{\pm}}}\textbf{0.9}\raisebox{-0.45ex}{\fontsize{4}{4}\selectfont\color{TraceGain}{+1.7}} \\
    \specialrule{0.35pt}{0pt}{0pt}
    \rowcolor{TraceMathBand} \multicolumn{9}{@{}l@{}}{\textbf{Visual Math}} \\
    \rowcolor{TraceMathBand} MathVision & 24.8\ensuremath{\mathord{\pm}}0.7 & \textbf{27.4}\ensuremath{\mathord{\boldsymbol{\pm}}}\textbf{0.6}\raisebox{-0.45ex}{\fontsize{4}{4}\selectfont\color{TraceGain}{+2.6}} & 25.3\ensuremath{\mathord{\pm}}0.3 & 26.2\ensuremath{\mathord{\pm}}0.5 & 25.3\ensuremath{\mathord{\pm}}1.0 & 28.1\ensuremath{\mathord{\pm}}0.1 & 19.3\ensuremath{\mathord{\pm}}0.7 & \textbf{25.4}\ensuremath{\mathord{\boldsymbol{\pm}}}\textbf{0.9}\raisebox{-0.45ex}{\fontsize{4}{4}\selectfont\color{TraceGain}{+6.1}} \\
    \rowcolor{TraceMathBand} MathVista & 68.7\ensuremath{\mathord{\pm}}0.4 & \textbf{73.4}\ensuremath{\mathord{\boldsymbol{\pm}}}\textbf{0.5}\raisebox{-0.45ex}{\fontsize{4}{4}\selectfont\color{TraceGain}{+4.7}} & 69.9\ensuremath{\mathord{\pm}}1.0 & 71.0\ensuremath{\mathord{\pm}}0.7 & 70.8\ensuremath{\mathord{\pm}}0.6 & 76.8\ensuremath{\mathord{\pm}}0.4 & 58.1\ensuremath{\mathord{\pm}}3.7 & \textbf{64.4}\ensuremath{\mathord{\boldsymbol{\pm}}}\textbf{2.1}\raisebox{-0.45ex}{\fontsize{4}{4}\selectfont\color{TraceGain}{+6.3}} \\
    \rowcolor{TraceMathBand} MathVerse & 43.4\ensuremath{\mathord{\pm}}0.9 & \textbf{47.8}\ensuremath{\mathord{\boldsymbol{\pm}}}\textbf{0.5}\raisebox{-0.45ex}{\fontsize{4}{4}\selectfont\color{TraceGain}{+4.3}} & 44.1\ensuremath{\mathord{\pm}}1.5 & 45.3\ensuremath{\mathord{\pm}}2.4 & 44.9\ensuremath{\mathord{\pm}}1.0 & 50.9\ensuremath{\mathord{\pm}}1.1 & 33.6\ensuremath{\mathord{\pm}}2.0 & \textbf{40.0}\ensuremath{\mathord{\boldsymbol{\pm}}}\textbf{1.5}\raisebox{-0.45ex}{\fontsize{4}{4}\selectfont\color{TraceGain}{+6.4}} \\
    \rowcolor{TraceMathBand} WeMath & 35.2\ensuremath{\mathord{\pm}}2.9 & \textbf{46.2}\ensuremath{\mathord{\boldsymbol{\pm}}}\textbf{1.3}\raisebox{-0.45ex}{\fontsize{4}{4}\selectfont\color{TraceGain}{+11.0}} & 38.2\ensuremath{\mathord{\pm}}1.0 & 36.3\ensuremath{\mathord{\pm}}0.5 & 39.0\ensuremath{\mathord{\pm}}1.4 & 46.9\ensuremath{\mathord{\pm}}0.4 & 17.9\ensuremath{\mathord{\pm}}1.2 & \textbf{28.8}\ensuremath{\mathord{\boldsymbol{\pm}}}\textbf{0.4}\raisebox{-0.45ex}{\fontsize{4}{4}\selectfont\color{TraceGain}{+10.9}} \\
    \rowcolor{TraceMathBand} \textit{Category average} & 43.0\ensuremath{\mathord{\pm}}0.8 & \textbf{48.7}\ensuremath{\mathord{\boldsymbol{\pm}}}\textbf{0.1}\raisebox{-0.45ex}{\fontsize{4}{4}\selectfont\color{TraceGain}{+5.6}} & 44.4\ensuremath{\mathord{\pm}}0.3 & 44.7\ensuremath{\mathord{\pm}}0.3 & 45.0\ensuremath{\mathord{\pm}}0.4 & 50.7\ensuremath{\mathord{\pm}}0.2 & 32.2\ensuremath{\mathord{\pm}}1.7 & \textbf{39.7}\ensuremath{\mathord{\boldsymbol{\pm}}}\textbf{0.8}\raisebox{-0.45ex}{\fontsize{4}{4}\selectfont\color{TraceGain}{+7.4}} \\
    \specialrule{0.35pt}{0pt}{0pt}
    \rowcolor{TraceScienceBand} \multicolumn{9}{@{}l@{}}{\textbf{Science \& General}} \\
    \rowcolor{TraceScienceBand} PhyX mini MC & 41.0\ensuremath{\mathord{\pm}}3.6 & \textbf{48.7}\ensuremath{\mathord{\boldsymbol{\pm}}}\textbf{0.8}\raisebox{-0.45ex}{\fontsize{4}{4}\selectfont\color{TraceGain}{+7.7}} & 41.4\ensuremath{\mathord{\pm}}4.5 & 42.4\ensuremath{\mathord{\pm}}3.3 & 42.0\ensuremath{\mathord{\pm}}2.9 & 46.6\ensuremath{\mathord{\pm}}0.2 & 32.8\ensuremath{\mathord{\pm}}10.0 & \textbf{37.5}\ensuremath{\mathord{\boldsymbol{\pm}}}\textbf{5.6}\raisebox{-0.45ex}{\fontsize{4}{4}\selectfont\color{TraceGain}{+4.7}} \\
    \rowcolor{TraceScienceBand} MMMU-ProVis & 35.7\ensuremath{\mathord{\pm}}0.4 & \textbf{39.4}\ensuremath{\mathord{\boldsymbol{\pm}}}\textbf{0.7}\raisebox{-0.45ex}{\fontsize{4}{4}\selectfont\color{TraceGain}{+3.7}} & 36.1\ensuremath{\mathord{\pm}}0.6 & 37.6\ensuremath{\mathord{\pm}}0.2 & 36.9\ensuremath{\mathord{\pm}}1.2 & 39.7\ensuremath{\mathord{\pm}}0.5 & 26.6\ensuremath{\mathord{\pm}}0.3 & \textbf{31.2}\ensuremath{\mathord{\boldsymbol{\pm}}}\textbf{1.2}\raisebox{-0.45ex}{\fontsize{4}{4}\selectfont\color{TraceGain}{+4.6}} \\
    \rowcolor{TraceScienceBand} RealWorldQA & 65.4\ensuremath{\mathord{\pm}}0.9 & \textbf{68.5}\ensuremath{\mathord{\boldsymbol{\pm}}}\textbf{0.7}\raisebox{-0.45ex}{\fontsize{4}{4}\selectfont\color{TraceGain}{+3.1}} & 66.0\ensuremath{\mathord{\pm}}0.3 & 67.3\ensuremath{\mathord{\pm}}0.2 & 67.8\ensuremath{\mathord{\pm}}1.0 & 69.8\ensuremath{\mathord{\pm}}0.7 & 60.3\ensuremath{\mathord{\pm}}0.4 & \textbf{62.1}\ensuremath{\mathord{\boldsymbol{\pm}}}\textbf{1.2}\raisebox{-0.45ex}{\fontsize{4}{4}\selectfont\color{TraceGain}{+1.8}} \\
    \rowcolor{TraceScienceBand} MMStar & 61.9\ensuremath{\mathord{\pm}}0.9 & \textbf{65.6}\ensuremath{\mathord{\boldsymbol{\pm}}}\textbf{0.3}\raisebox{-0.45ex}{\fontsize{4}{4}\selectfont\color{TraceGain}{+3.8}} & 62.6\ensuremath{\mathord{\pm}}1.5 & 61.8\ensuremath{\mathord{\pm}}0.5 & 62.5\ensuremath{\mathord{\pm}}0.8 & 65.8\ensuremath{\mathord{\pm}}0.3 & 52.3\ensuremath{\mathord{\pm}}0.5 & \textbf{55.2}\ensuremath{\mathord{\boldsymbol{\pm}}}\textbf{1.0}\raisebox{-0.45ex}{\fontsize{4}{4}\selectfont\color{TraceGain}{+3.0}} \\
    \rowcolor{TraceScienceBand} \textit{Category average} & 51.0\ensuremath{\mathord{\pm}}0.8 & \textbf{55.6}\ensuremath{\mathord{\boldsymbol{\pm}}}\textbf{0.4}\raisebox{-0.45ex}{\fontsize{4}{4}\selectfont\color{TraceGain}{+4.5}} & 51.5\ensuremath{\mathord{\pm}}1.5 & 52.2\ensuremath{\mathord{\pm}}0.9 & 52.3\ensuremath{\mathord{\pm}}0.9 & 55.5\ensuremath{\mathord{\pm}}0.3 & 43.0\ensuremath{\mathord{\pm}}2.5 & \textbf{46.5}\ensuremath{\mathord{\boldsymbol{\pm}}}\textbf{1.8}\raisebox{-0.45ex}{\fontsize{4}{4}\selectfont\color{TraceGain}{+3.5}} \\
    \specialrule{0.35pt}{0pt}{0pt}
    \rowcolor{TraceSpatialBand} \multicolumn{9}{@{}l@{}}{\textbf{Spatial Reasoning}} \\
    \rowcolor{TraceSpatialBand} EmbSpatial & 69.4\ensuremath{\mathord{\pm}}0.9 & 71.0\ensuremath{\mathord{\pm}}0.7\raisebox{-0.45ex}{\fontsize{4}{4}\selectfont\color{TraceGain}{+1.5}} & 69.5\ensuremath{\mathord{\pm}}0.9 & 70.9\ensuremath{\mathord{\pm}}0.7 & \textbf{72.7}\ensuremath{\mathord{\boldsymbol{\pm}}}\textbf{0.2} & 70.6\ensuremath{\mathord{\pm}}0.1 & 59.1\ensuremath{\mathord{\pm}}1.0 & \textbf{60.9}\ensuremath{\mathord{\boldsymbol{\pm}}}\textbf{1.1}\raisebox{-0.45ex}{\fontsize{4}{4}\selectfont\color{TraceGain}{+1.8}} \\
    \rowcolor{TraceSpatialBand} SpatialVizBench & 35.1\ensuremath{\mathord{\pm}}0.1 & \textbf{35.7}\ensuremath{\mathord{\boldsymbol{\pm}}}\textbf{0.3}\raisebox{-0.45ex}{\fontsize{4}{4}\selectfont\color{TraceGain}{+0.5}} & 32.7\ensuremath{\mathord{\pm}}2.3 & 33.8\ensuremath{\mathord{\pm}}1.5 & 30.8\ensuremath{\mathord{\pm}}1.8 & 32.1\ensuremath{\mathord{\pm}}0.5 & 30.1\ensuremath{\mathord{\pm}}1.2 & \textbf{31.8}\ensuremath{\mathord{\boldsymbol{\pm}}}\textbf{1.5}\raisebox{-0.45ex}{\fontsize{4}{4}\selectfont\color{TraceGain}{+1.8}} \\
    \rowcolor{TraceSpatialBand} CV-Bench 3D & 76.2\ensuremath{\mathord{\pm}}2.0 & \textbf{81.0}\ensuremath{\mathord{\boldsymbol{\pm}}}\textbf{0.4}\raisebox{-0.45ex}{\fontsize{4}{4}\selectfont\color{TraceGain}{+4.8}} & 76.2\ensuremath{\mathord{\pm}}0.8 & 79.8\ensuremath{\mathord{\pm}}1.0 & 80.2\ensuremath{\mathord{\pm}}0.5 & 83.1\ensuremath{\mathord{\pm}}0.6 & 58.7\ensuremath{\mathord{\pm}}8.3 & \textbf{67.0}\ensuremath{\mathord{\boldsymbol{\pm}}}\textbf{3.7}\raisebox{-0.45ex}{\fontsize{4}{4}\selectfont\color{TraceGain}{+8.3}} \\
    \rowcolor{TraceSpatialBand} ERQA & 39.0\ensuremath{\mathord{\pm}}1.6 & \textbf{41.2}\ensuremath{\mathord{\boldsymbol{\pm}}}\textbf{1.3}\raisebox{-0.45ex}{\fontsize{4}{4}\selectfont\color{TraceGain}{+2.2}} & 38.8\ensuremath{\mathord{\pm}}1.7 & 40.3\ensuremath{\mathord{\pm}}2.0 & 40.4\ensuremath{\mathord{\pm}}0.8 & 44.6\ensuremath{\mathord{\pm}}0.8 & 35.3\ensuremath{\mathord{\pm}}0.8 & \textbf{36.4}\ensuremath{\mathord{\boldsymbol{\pm}}}\textbf{0.5}\raisebox{-0.45ex}{\fontsize{4}{4}\selectfont\color{TraceGain}{+1.1}} \\
    \rowcolor{TraceSpatialBand} \textit{Category average} & 55.0\ensuremath{\mathord{\pm}}0.8 & \textbf{57.2}\ensuremath{\mathord{\boldsymbol{\pm}}}\textbf{0.3}\raisebox{-0.45ex}{\fontsize{4}{4}\selectfont\color{TraceGain}{+2.2}} & 54.3\ensuremath{\mathord{\pm}}1.0 & 56.2\ensuremath{\mathord{\pm}}1.0 & 56.0\ensuremath{\mathord{\pm}}0.5 & 57.6\ensuremath{\mathord{\pm}}0.3 & 45.8\ensuremath{\mathord{\pm}}2.4 & \textbf{49.0}\ensuremath{\mathord{\boldsymbol{\pm}}}\textbf{0.9}\raisebox{-0.45ex}{\fontsize{4}{4}\selectfont\color{TraceGain}{+3.2}} \\
    \specialrule{0.35pt}{0pt}{0pt}
    \rowcolor{TracePerceptionBand} \multicolumn{9}{@{}l@{}}{\textbf{Perception \& Counting}} \\
    \rowcolor{TracePerceptionBand} BLINK & 53.3\ensuremath{\mathord{\pm}}1.2 & 56.6\ensuremath{\mathord{\pm}}0.7\raisebox{-0.45ex}{\fontsize{4}{4}\selectfont\color{TraceGain}{+3.3}} & 53.8\ensuremath{\mathord{\pm}}1.2 & 55.7\ensuremath{\mathord{\pm}}1.1 & \textbf{56.6}\ensuremath{\mathord{\boldsymbol{\pm}}}\textbf{0.5} & 57.5\ensuremath{\mathord{\pm}}0.7 & 44.5\ensuremath{\mathord{\pm}}2.1 & \textbf{47.1}\ensuremath{\mathord{\boldsymbol{\pm}}}\textbf{0.7}\raisebox{-0.45ex}{\fontsize{4}{4}\selectfont\color{TraceGain}{+2.6}} \\
    \rowcolor{TracePerceptionBand} CountBenchQA & 82.1\ensuremath{\mathord{\pm}}1.5 & \textbf{84.8}\ensuremath{\mathord{\boldsymbol{\pm}}}\textbf{1.4}\raisebox{-0.45ex}{\fontsize{4}{4}\selectfont\color{TraceGain}{+2.7}} & 82.6\ensuremath{\mathord{\pm}}1.4 & 83.3\ensuremath{\mathord{\pm}}1.3 & 80.6\ensuremath{\mathord{\pm}}1.2 & 83.0\ensuremath{\mathord{\pm}}0.2 & 65.4\ensuremath{\mathord{\pm}}1.6 & \textbf{68.7}\ensuremath{\mathord{\boldsymbol{\pm}}}\textbf{0.8}\raisebox{-0.45ex}{\fontsize{4}{4}\selectfont\color{TraceGain}{+3.2}} \\
    \rowcolor{TracePerceptionBand} CountQA & 19.6\ensuremath{\mathord{\pm}}1.2 & \textbf{22.5}\ensuremath{\mathord{\boldsymbol{\pm}}}\textbf{0.9}\raisebox{-0.45ex}{\fontsize{4}{4}\selectfont\color{TraceGain}{+2.9}} & 19.9\ensuremath{\mathord{\pm}}1.0 & 21.4\ensuremath{\mathord{\pm}}0.2 & 17.2\ensuremath{\mathord{\pm}}0.4 & 23.2\ensuremath{\mathord{\pm}}0.3 & 14.9\ensuremath{\mathord{\pm}}1.8 & \textbf{15.6}\ensuremath{\mathord{\boldsymbol{\pm}}}\textbf{0.6}\raisebox{-0.45ex}{\fontsize{4}{4}\selectfont\color{TraceGain}{+0.8}} \\
    \rowcolor{TracePerceptionBand} TreeBench & 38.0\ensuremath{\mathord{\pm}}1.4 & 40.2\ensuremath{\mathord{\pm}}2.1\raisebox{-0.45ex}{\fontsize{4}{4}\selectfont\color{TraceGain}{+2.2}} & 37.9\ensuremath{\mathord{\pm}}1.0 & 39.8\ensuremath{\mathord{\pm}}1.8 & \textbf{40.4}\ensuremath{\mathord{\boldsymbol{\pm}}}\textbf{2.0} & 41.5\ensuremath{\mathord{\pm}}0.9 & \textbf{39.3}\ensuremath{\mathord{\boldsymbol{\pm}}}\textbf{2.0} & 38.6\ensuremath{\mathord{\pm}}1.2\raisebox{-0.45ex}{\fontsize{4}{4}\selectfont\color{TraceLoss}{-0.7}} \\
    \rowcolor{TracePerceptionBand} \textit{Category average} & 48.3\ensuremath{\mathord{\pm}}0.3 & \textbf{51.0}\ensuremath{\mathord{\boldsymbol{\pm}}}\textbf{0.4}\raisebox{-0.45ex}{\fontsize{4}{4}\selectfont\color{TraceGain}{+2.8}} & 48.5\ensuremath{\mathord{\pm}}0.2 & 50.0\ensuremath{\mathord{\pm}}0.2 & 48.7\ensuremath{\mathord{\pm}}0.2 & 51.3\ensuremath{\mathord{\pm}}0.0 & 41.0\ensuremath{\mathord{\pm}}0.5 & \textbf{42.5}\ensuremath{\mathord{\boldsymbol{\pm}}}\textbf{0.3}\raisebox{-0.45ex}{\fontsize{4}{4}\selectfont\color{TraceGain}{+1.5}} \\
    \specialrule{0.35pt}{0pt}{0pt}
    \rowcolor{TracePuzzlesBand} \multicolumn{9}{@{}l@{}}{\textbf{Puzzles \& Logic}} \\
    \rowcolor{TracePuzzlesBand} PuzzleVQA & 44.6\ensuremath{\mathord{\pm}}0.6 & \textbf{50.0}\ensuremath{\mathord{\boldsymbol{\pm}}}\textbf{0.5}\raisebox{-0.45ex}{\fontsize{4}{4}\selectfont\color{TraceGain}{+5.5}} & 40.1\ensuremath{\mathord{\pm}}1.5 & 48.6\ensuremath{\mathord{\pm}}1.1 & 47.9\ensuremath{\mathord{\pm}}1.1 & 49.2\ensuremath{\mathord{\pm}}0.2 & 32.6\ensuremath{\mathord{\pm}}1.2 & \textbf{39.1}\ensuremath{\mathord{\boldsymbol{\pm}}}\textbf{1.1}\raisebox{-0.45ex}{\fontsize{4}{4}\selectfont\color{TraceGain}{+6.6}} \\
    \rowcolor{TracePuzzlesBand} VisualPuzzles & 31.1\ensuremath{\mathord{\pm}}0.1 & \textbf{34.0}\ensuremath{\mathord{\boldsymbol{\pm}}}\textbf{0.8}\raisebox{-0.45ex}{\fontsize{4}{4}\selectfont\color{TraceGain}{+2.9}} & 29.4\ensuremath{\mathord{\pm}}0.7 & 33.2\ensuremath{\mathord{\pm}}0.9 & 33.1\ensuremath{\mathord{\pm}}1.7 & 37.0\ensuremath{\mathord{\pm}}0.4 & 26.2\ensuremath{\mathord{\pm}}2.7 & \textbf{28.5}\ensuremath{\mathord{\boldsymbol{\pm}}}\textbf{0.7}\raisebox{-0.45ex}{\fontsize{4}{4}\selectfont\color{TraceGain}{+2.3}} \\
    \rowcolor{TracePuzzlesBand} LogicVista & 42.7\ensuremath{\mathord{\pm}}2.2 & \textbf{47.4}\ensuremath{\mathord{\boldsymbol{\pm}}}\textbf{0.7}\raisebox{-0.45ex}{\fontsize{4}{4}\selectfont\color{TraceGain}{+4.7}} & 43.3\ensuremath{\mathord{\pm}}0.9 & 44.7\ensuremath{\mathord{\pm}}1.2 & 43.0\ensuremath{\mathord{\pm}}0.9 & 47.7\ensuremath{\mathord{\pm}}0.7 & 36.5\ensuremath{\mathord{\pm}}1.2 & \textbf{40.1}\ensuremath{\mathord{\boldsymbol{\pm}}}\textbf{1.7}\raisebox{-0.45ex}{\fontsize{4}{4}\selectfont\color{TraceGain}{+3.7}} \\
    \rowcolor{TracePuzzlesBand} MME-Reasoning & 25.2\ensuremath{\mathord{\pm}}1.2 & \textbf{28.0}\ensuremath{\mathord{\boldsymbol{\pm}}}\textbf{0.9}\raisebox{-0.45ex}{\fontsize{4}{4}\selectfont\color{TraceGain}{+2.9}} & 26.1\ensuremath{\mathord{\pm}}1.1 & 27.1\ensuremath{\mathord{\pm}}0.3 & 27.5\ensuremath{\mathord{\pm}}1.0 & 30.6\ensuremath{\mathord{\pm}}0.7 & 22.6\ensuremath{\mathord{\pm}}1.8 & \textbf{25.1}\ensuremath{\mathord{\boldsymbol{\pm}}}\textbf{1.5}\raisebox{-0.45ex}{\fontsize{4}{4}\selectfont\color{TraceGain}{+2.4}} \\
    \rowcolor{TracePuzzlesBand} \textit{Category average} & 35.9\ensuremath{\mathord{\pm}}0.7 & \textbf{39.9}\ensuremath{\mathord{\boldsymbol{\pm}}}\textbf{0.5}\raisebox{-0.45ex}{\fontsize{4}{4}\selectfont\color{TraceGain}{+4.0}} & 34.7\ensuremath{\mathord{\pm}}0.7 & 38.4\ensuremath{\mathord{\pm}}0.1 & 37.9\ensuremath{\mathord{\pm}}0.8 & 41.1\ensuremath{\mathord{\pm}}0.1 & 29.5\ensuremath{\mathord{\pm}}0.6 & \textbf{33.2}\ensuremath{\mathord{\boldsymbol{\pm}}}\textbf{0.9}\raisebox{-0.45ex}{\fontsize{4}{4}\selectfont\color{TraceGain}{+3.7}} \\
    \specialrule{0.7pt}{1pt}{0pt}
    \rowcolor{TraceOverallBand} \textbf{Overall average} & 47.9\ensuremath{\mathord{\pm}}0.3 & \textbf{52.0}\ensuremath{\mathord{\boldsymbol{\pm}}}\textbf{0.2}\raisebox{-0.45ex}{\fontsize{4}{4}\selectfont\color{TraceGain}{+4.1}} & 48.0\ensuremath{\mathord{\pm}}0.4 & 49.3\ensuremath{\mathord{\pm}}0.2 & 48.8\ensuremath{\mathord{\pm}}0.3 & 52.2\ensuremath{\mathord{\pm}}0.1 & 39.3\ensuremath{\mathord{\pm}}0.6 & \textbf{42.9}\ensuremath{\mathord{\boldsymbol{\pm}}}\textbf{0.4}\raisebox{-0.45ex}{\fontsize{4}{4}\selectfont\color{TraceGain}{+3.5}} \\
    \bottomrule
  \end{tabularx}
\end{table}

\subsection{External benchmark transfer}
\label{sec:external-transfer}

Table~\ref{tab:main-results} reports transfer across the complete
24-benchmark evaluation suite.

Training on \trace improves the benchmark-macro average at both model scales.
At 3B, the score increases from $39.34\pm0.63$ to $42.85\pm0.39$, a gain of
3.51 percentage points. At 7B, it increases from $47.93\pm0.30$ to
$51.99\pm0.17$, a gain of 4.06 points. The improvements are broad: all six
category averages increase at both scales, mean performance improves on 21 of
24 benchmarks at 3B and on all 24 benchmarks at 7B, and paired item-bootstrap
intervals exclude zero on 18 and 21 benchmarks, respectively. EvoChart at 3B
is the only benchmark whose interval lies entirely below zero; the remaining
intervals either exclude zero in the positive direction or include it.
Appendix Table~\ref{tab:external-bootstrap-cis} reports the complete
benchmark-level intervals.

Visual mathematics shows the largest category-level improvement at both scales,
with gains of 7.44 points at 3B and 5.65 points at 7B. WeMath shows the largest
individual improvement, increasing by 10.95 points at 3B and 10.96 points at
7B. The gains therefore do not arise solely from improved performance on
held-out procedural images, but extend across external benchmarks with
different visual formats, task structures, and evaluation metrics.

Under the common evaluation protocol, the \trace-trained 7B checkpoint obtains
a macro-average 3.94 points higher than Game-RL-7B, 2.64 points higher than
Sphinx-7B, and 3.18 points higher than PC-GRPO-7B. Among these synthetic-data
RLVR checkpoints, it also has the highest reported mean on 21 of 24 benchmarks
and the highest category average in all six groups. These comparisons describe
checkpoint outcomes rather than isolate the effect of a particular method or
training distribution, because the systems are not matched in data,
optimization, or compute. Table~\ref{tab:main-results} reports Vero separately
because its substantially larger training mixture includes real images.

Figure~\ref{fig:scale-gain-correlation} shows moderate consistency in the
relative transfer patterns across model scales. The negative mean changes at
3B on ChartQAPro ($-0.14$), TreeBench ($-0.66$), and EvoChart ($-1.68$)
become gains of 2.24, 2.22, and 7.84 points at 7B, respectively. EvoChart is
the largest departure from the equal-gain diagonal. Thus, although the overall
transfer pattern is positively associated across scales, the benefit of a
fixed procedural training mixture remains dependent on model capacity and
benchmark demands.

\subsection{Training dynamics}
\label{sec:training-dynamics}

\begin{figure}[t]
  \centering
  \includegraphics[width=\linewidth]{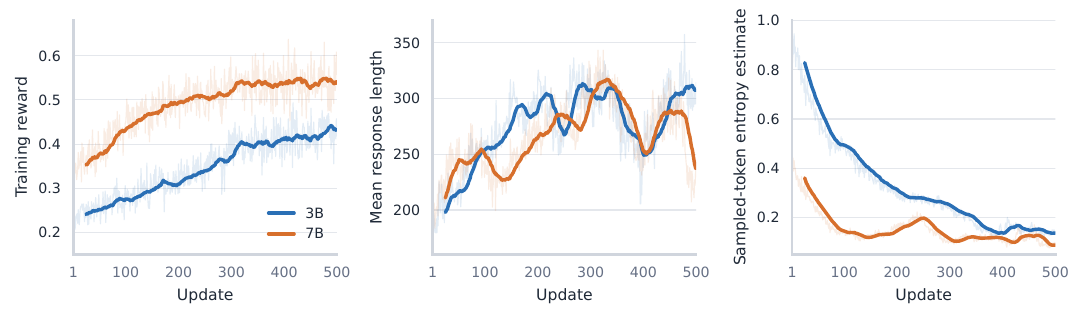}
  \caption{Optimization trajectories over 500 RLVR updates. Faint lines show
  per-update values, and solid lines show 25-update moving averages. Training
  reward is the bounded answer-and-format reward defined in
  Section~\ref{sec:experiments}. Response length is measured over sampled
  completions, and entropy is the sampled-token estimate recorded by the policy
  optimizer.}
  \label{fig:training-dynamics}
\end{figure}

Training reward increases at both model scales despite substantial
batch-to-batch variation. Averaged over the first and final 25 updates, mean
reward rises from 0.240 to 0.432 at 3B and from 0.353 to 0.542 at 7B. The 7B
model therefore begins and ends with higher reward, while both models exhibit
continued improvement throughout the single training pass.

The scales differ more clearly in their response-length trajectories. Mean
completion length increases from 198 to 308 tokens at 3B, compared with a
smaller increase from 211 to 237 tokens at 7B. Over the same intervals, the
sampled-token entropy estimate decreases from 0.829 to 0.137 at 3B and from
0.361 to 0.088 at 7B. Optimization therefore raises reward while making the
sampled token distributions more concentrated, without reducing the models to
shorter responses; both retain multi-hundred-token completions at the end of
training.

\section{Limitations and Future Work}
\label{sec:limitations}

The \trace taxonomy is explicit but human-designed: task boundaries can still
involve judgment, and the 1,000 authored tasks do not exhaust visual reasoning.
The current operation families characterize computation types but do not define
a common notion of difficulty across domains. Future work could model complexity
through program structure, semantic-state size, visual density, and empirical
solve rates.

Our experiments show transfer from the complete \trace mixture but do not
attribute gains to particular domains, operation families, sampling choices, or
rendering controls. Such attribution would require mixture ablations. The
current study also uses uniform task sampling over one fixed training pass;
the explicit task identities and generation parameters could instead support
difficulty-aware curricula, adaptive sampling, and mixture optimization based on
model competence or learning progress.

Results are based on one training run per model scale, so variation across
decoding seeds reflects inference variability rather than independent training
variation. Comparisons with released RLVR checkpoints are descriptive because
their data, optimization, and compute are unmatched. The external suite is broad
but cannot establish transfer to every natural-image distribution, and some
benchmarks use model-based graders. Finally, procedural consistency does not
guarantee equal perceptual difficulty: synthetic renderers may introduce
regularities or occasional ambiguities not found in natural images. Richer
rendering sources and replay-based failure analysis are natural directions for
future work.
\section{Conclusion}
\label{sec:conclusion}

\trace formulates procedural visual data as an executable environment in which
scene grammars, task programs, bounded queries, and reward contracts are explicit
and separable. This structure provides stable units for generation, task-balanced
sampling, verification, and analysis while keeping every instance exactly
verifiable and deterministically replayable across semantic and visual variation.
RLVR on 64,000 \trace instances improves the macro-average across 24 external
benchmarks by 3.51 points at 3B and 4.06 points at 7B, with positive mean changes
on 21 of 24 and all 24 benchmarks, respectively. These results provide evidence
that broad, structured procedural experience can support transferable visual
reasoning rather than only improving performance on regenerated instances from
the same task distributions.

\section*{Broader Impact}

\trace may support more reproducible multimodal-reasoning research by making
task definitions, supervision, and reward computation inspectable, while reducing
reliance on sensitive or copyrighted source imagery in settings where procedural
data are suitable. At the same time, stronger visual-reasoning systems may be
used in surveillance, automated assessment, or other consequential decision
pipelines, and large-scale RLVR training can require substantial computational
resources. Releases built on \trace should therefore document intended uses,
licensing, resource requirements, and downstream-use constraints.

\bibliographystyle{plainnat}
\bibliography{references}

\clearpage
\appendix
\section{Environment Construction and Validation}
\label{app:generation}

\subsection{Domain construction profiles}

Prompts are assembled from scene-, task-, and query-level templates. Dynamic
labels, values, and answers are derived from the same semantic state and task
execution recorded in the instance trace. The profiles below therefore
summarize the domain-specific state representations, rendering conventions,
and visible vocabularies; the common generation and reward pipeline is
described in Section~\ref{sec:design}.

\paragraph{Charts.}
Chart scenes are constructed from procedural marks, axes, scales, legends, and
panels spanning Cartesian, radial, geographic, flow, tabular, and compositional
layouts. Rendering controls vary chart treatment, palette, typography, tick
density, mark style, and compatible report context without modifying the
underlying data. Curated entity, temporal, scientific, and compact label pools
provide distinct series, category, panel, and region names within each figure.

\paragraph{Games.}
Game scenes combine procedural boards, tracks, hands, scores, pieces, enemies,
and action states with curated cards, dice, tokens, and sprite-like assets.
Each renderer follows the layout and interaction rules of its scene grammar
rather than a shared generic board template. Coordinates, score displays,
status labels, and action cues are derived from the simulated state, while
optional interface overlays and surface treatments provide visual variation.

\paragraph{Geometry.}
Geometry scenes are generated from analytic constructions of points, lines,
angles, polygons, circles, solids, coordinate systems, and measurement
instruments under solved numeric constraints. Textbook, whiteboard, and
technical-diagram profiles vary stroke, shading, notation placement, and
typography. Point labels, measurements, equations, and answer options are bound
after construction so that the visible diagram and symbolic problem statement
remain consistent with the same geometric state.

\paragraph{Graphs.}
Graph scenes begin with explicit nodes, edges, weights, paths, and state
transitions, which are then realized through circular, layered, tree, route, or
force-directed layouts. Node shape, edge style, palette, and framing vary
independently of topology. Numeric, alphabetic, and named node vocabularies are
sampled without collisions, and labels and edge values remain attached to
their semantic elements across layout changes.

\paragraph{Icons.}
Icon scenes populate fields, grids, paths, rings, paired canvases, and rendered
option sets with curated icon assets. Procedural controls vary icon identity,
color, scale, rotation, spacing, ordering, and transformation while preserving
legibility and answer uniqueness. Marker and option labels are placed as part
of the scene geometry, allowing counting, adjacency, transformation, and
matching tasks to share a consistent visual vocabulary.

\paragraph{Illustrations.}
Illustration scenes compose semantic object, part, and tile catalogs into
rooms, construction sites, libraries, tactical maps, and isometric
environments. Scene-specific layout rules determine occupancy, overlap, depth,
and permitted object relations, after which palette, asset, framing, and
surface treatments vary the appearance. Explicit tile, object, and option
labels are introduced only when required by the task program.

\paragraph{Pages.}
Page scenes represent structured records through forms, tables, calendars,
schedules, application controls, documents, and infographic components.
Templates govern visual hierarchy, alignment, interface framing, spacing, and
typography, while information-density and context profiles vary the
presentation. Field names, section headings, dates, values, and control labels
remain linked to the underlying document state.

\paragraph{Physics.}
Physics scenes assemble apparatus, forces, wires, rays, fields, graphs, and
measurement instruments from explicit physical quantities and component
states. Technical-diagram profiles vary line work, component treatment,
background, and label placement while preserving the solved configuration.
Quantities, units, directions, and component names are rendered from the same
state used by formula-evaluation, comparison, and state-update programs.

\paragraph{Puzzles.}
Puzzle scenes include rule-governed grids, mazes, pipes, matrices, nets, voxel
boards, and candidate sets sampled under validity and uniqueness constraints.
Symbols, board skins, spacing, missing-cell treatments, and option scaffolds
vary the presentation without revealing the underlying rule. Coordinates,
candidate labels, and compact clues are added only when required by the puzzle
definition or answer interface.

\paragraph{Symbolic.}
Symbolic scenes use notation-specific generators for clocks, abaci, dice,
spinners, musical staves, logic circuits, formal tables, tapes, and encoded
glyph systems. Symbols and transitions are represented as executable state
rather than decorative text. Fonts, notation cards, palettes, spacing, and
layout may vary, while tokens, values, note names, and state labels remain
bound to the formal structure being queried.

\paragraph{Three-dimensional scenes.}
Three-dimensional scenes arrange procedural meshes and object assets under
explicit camera, projection, depth, occlusion, and surface constraints.
Viewpoint, material, lighting, ground or wall treatment, and framing vary
without changing the queried spatial relation. Object, lane, wall, point, and
candidate labels are placed after projection and checked against the final
image geometry for visibility and separation.

\subsection{Operation-family definitions}
\label{app:operation-families}

Figure~\ref{fig:operation-coverage} classifies tasks according to the
operations that materially contribute to their outputs. Assignments are
multi-label when a task program composes several operations.

\textit{Direct retrieval} denotes a localized visible lookup for which no
other substantive operation determines the answer. \textit{Filtering}
constructs a subset using a predicate, relation, or named scope, and
\textit{counting} returns the cardinality of such a set. \textit{Comparison}
evaluates equality, order, thresholds, or intervals, whereas \textit{ranking}
orders candidates or selects an extremum or ordinal item.
\textit{Aggregation} reduces a collection using a sum, mean, median,
cumulative total, share, mass, or related statistic.

\textit{Logical composition} combines predicates or sets through conjunction,
disjunction, negation, union, intersection, or difference.
\textit{Spatial relations} covers positional, directional, metric, overlap,
containment, and occlusion relations. \textit{Topology} covers connectivity,
paths, components, cycles, traversal, and reachability.
\textit{Matching} evaluates equivalence, correspondence, consistency,
completion, or candidate-to-reference agreement.

\textit{Transformation} applies or infers a geometric, symbolic,
representational, folding, projection, overlay, or reconstruction operation.
\textit{State update} applies a move, action, counterfactual edit, or discrete
transition. \textit{Formula evaluation} uses arithmetic, algebra, a
domain-specific formula, a derived metric, or a numeric difference, ratio, or
rate.

\subsection{Invariant enforcement}

Semantic generation first enforces task-specific constraints and answer
uniqueness. Structural realization is completed before geometry-dependent
verifier state is projected into image coordinates, and post-render effects
preserve the final canvas coordinate system. Render-only controls may not
modify task-program operands, answer values, or the reward contract.

Independent seeded streams and the recorded instance trace allow the semantic
state, prompt, rendering choices, and verifier state to be replayed together.
The precise ordering of optional finishing layers remains renderer-specific.
Table~\ref{tab:variation-layers} summarizes the variation levels and their
required invariants.

\begin{table}[t]
  \centering
  \scriptsize
  \setlength{\tabcolsep}{3.5pt}
  \renewcommand{\arraystretch}{1.02}
  \caption{Variation layers and the invariants they preserve.}
  \label{tab:variation-layers}
  \begin{tabular}{@{}>{\raggedright\arraybackslash}p{0.21\linewidth}
                       >{\raggedright\arraybackslash}p{0.44\linewidth}
                       >{\raggedright\arraybackslash}p{0.25\linewidth}@{}}
    \toprule
    Layer & Representative controls & Required invariant \\
    \midrule
    Semantic instance
      & Dimensions, cardinalities, values, operation length, candidate set
      & Scene grammar and task program \\
    Structural realization
      & Placement, spacing, panel arrangement, framing, answer-preserving camera
      & Semantic relations and answer \\
    Surface and context
      & Palette, typography, skins, materials, labels, non-answer text
      & Semantic state and answer \\
    Post-render
      & Blur, resampling, compression, tone, noise, texture
      & Canvas geometry and answer \\
    \bottomrule
  \end{tabular}
\end{table}

\begin{figure}[t]
  \centering
  \includegraphics[width=0.9\textwidth]{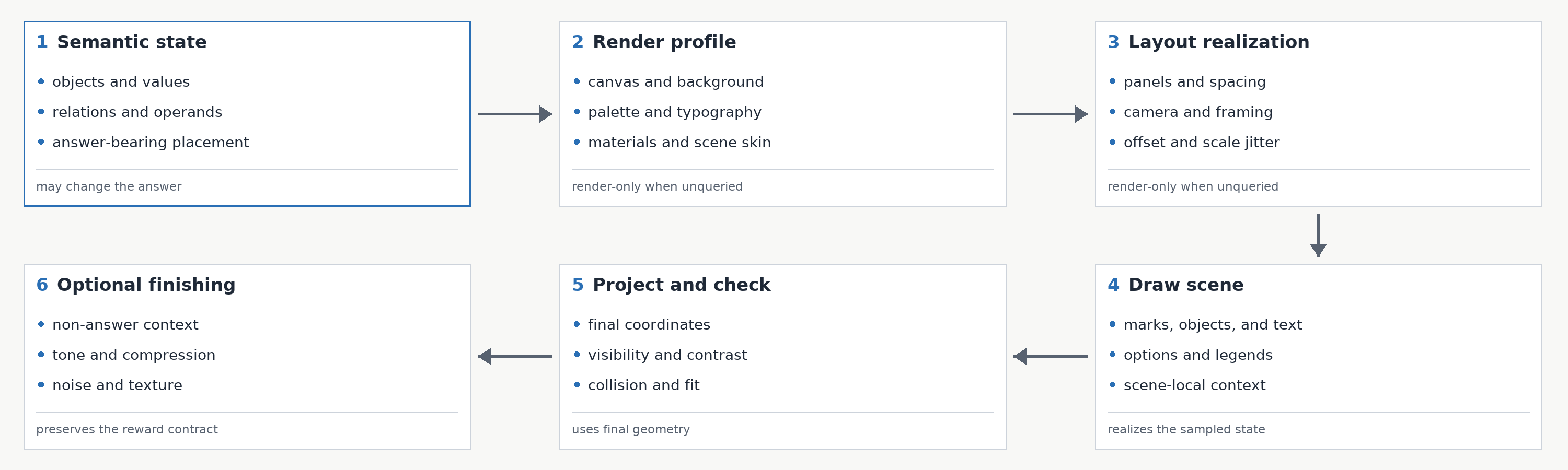}
  \caption{Control flow for visual realization. Structural and surface
  controls are treated as render-only when they preserve the queried semantic
  relations and typed answer. Answer-bearing geometry is projected before
  validation, while optional context and raster finishing remain
  scene-specific and are recorded in the instance trace.}
  \label{fig:rendering-pipeline}
\end{figure}

Figure~\ref{fig:rendering-pipeline} illustrates how semantic execution,
structural realization, projection, and surface treatment interact while
preserving the task output.

Table~\ref{tab:rendering-variation-profiles} specifies the controlled
realizations used in Figure~\ref{fig:rendering-variation}.

\begin{table}[t]
  \centering
  \scriptsize
  \setlength{\tabcolsep}{4pt}
  \renewcommand{\arraystretch}{1.08}
  \caption{Controlled profiles in the fixed-instance rendering sweep.}
  \label{tab:rendering-variation-profiles}
  \begin{tabular}{@{}p{0.18\linewidth}p{0.18\linewidth}p{0.53\linewidth}@{}}
    \toprule
    Profile & Controlled axis & Realization \\
    \midrule
    Baseline & Reference & Light neutral theme, sans-serif type, fixed layout, clean background \\
    Dark theme & Theme & Dark publication treatment and analytics palette; topology and layout retained \\
    Typeface & Typography & EB Garamond node-label typeface; baseline palette and layout retained \\
    Layout & Structure & Circular node placement replaces the baseline spring layout \\
    Report context & Non-answer context & Curated paragraph context in reserved non-answer panel space \\
    Raster noise & Post-render & Gaussian noise with fixed sigma; coordinate geometry retained \\
    \bottomrule
  \end{tabular}
\end{table}

\section{Experimental Details}
\label{app:experiments}

Table~\ref{tab:training-configuration} reports the configuration shared by the
3B and 7B training runs.

\begin{table}[t]
  \centering
  \small
  \setlength{\tabcolsep}{5pt}
  \caption{RLVR configuration shared across the 3B and 7B runs.}
  \label{tab:training-configuration}
  \begin{tabular}{@{}l l@{}}
    \toprule
    Setting & Value \\
    \midrule
    Training data & 64,000 prompts; 1,000 tasks \\
    Image pixel cap & 1,280,000 per image \\
    Updates & 500; one shuffled data pass \\
    Validation & 2,000 instances every 100 updates \\
    Prompt batch / rollouts & 128 / 8 \\
    Sampled responses & 512,000 \\
    Prompt / response caps & 2,048 / 2,048 tokens \\
    Optimization scope & Full model; vision encoder trainable; BF16 \\
    Actor optimizer & AdamW; LR $10^{-6}$; weight decay 0.01 \\
    Policy epochs & 1 \\
    Policy clipping & 0.2 lower; 0.3 upper; dual 3.0 \\
    Rollout sampling & temperature 1.0; top-$p$ 1.0 \\
    Reward & 0.95 exact answer $+$ 0.05 JSON format \\
    Hardware & 8 $\times$ H100 80GB \\
    Runtime & 12.2 h (3B); 13.9 h (7B) \\
    \bottomrule
  \end{tabular}
\end{table}

External inference uses temperature 0.6, top-$p$ 1.0, a 4,096-token response
limit, and decoding seeds 42, 43, and 44. The seeds affect response generation
only; model weights remain fixed across decoding runs.

\subsection{\trace validation slices}

Table~\ref{tab:iid-slice-results} reports held-out \trace accuracy by answer
interface and query structure. Each task contributes two previously unseen
instances, and all slice scores use one decoding seed. The rows are overlapping
analytical views of the same task inventory rather than separate evaluation
sets.

\begin{table}[t]
  \centering
  \fontsize{7.5}{8.6}\selectfont
  \setlength{\tabcolsep}{4.8pt}
  \caption{Accuracy by answer interface and query structure on the 2,000-instance \trace validation set. Each task contributes two unseen instances and scores use one decoding seed. Rows are analytical slices of the same task inventory.}
  \label{tab:iid-slice-results}
  \begin{tabular}{@{}l c c c c c c c@{}}
    \toprule
    & & \multicolumn{3}{c}{3B} & \multicolumn{3}{c}{7B} \\
    \cmidrule(lr){3-5}\cmidrule(l){6-8}
    Slice & Tasks & Base & After RLVR & $\Delta$ & Base & After RLVR & $\Delta$ \\
    \midrule
    \multicolumn{8}{@{}l}{\textit{Answer interface}} \\
    Integer & 547 & 22.39 & 37.84 & +15.45 & 30.35 & 49.36 & +19.01 \\
    Numeric & 51 & 24.51 & 95.10 & +70.59 & 49.02 & 94.12 & +45.10 \\
    Option letter & 210 & 26.43 & 30.95 & +4.52 & 33.33 & 40.48 & +7.14 \\
    String & 192 & 28.12 & 46.88 & +18.75 & 42.45 & 58.59 & +16.15 \\
    \midrule
    \multicolumn{8}{@{}l}{\textit{Query structure}} \\
    Single-query tasks & 683 & 24.52 & 39.82 & +15.30 & 33.53 & 50.88 & +17.35 \\
    Multi-query tasks & 317 & 24.29 & 43.69 & +19.40 & 35.80 & 53.00 & +17.19 \\
    \bottomrule
  \end{tabular}
\end{table}

\subsection{External evaluation suite}

Each benchmark is evaluated with its standard scorer when available. A small
number of open-answer benchmarks retain their fixed model-based grading
procedures; the same grader, prompt, and parser are used for every checkpoint.
Table~\ref{tab:evaluation-suite} lists the complete benchmark suite, references,
and evaluated sample counts.

\begin{table}[t]
  \centering
  \scriptsize
  \setlength{\tabcolsep}{4pt}
  \caption{External evaluation suite. Each model and decoding seed is scored on 32,805 examples.}
  \label{tab:evaluation-suite}
  \begin{tabular}{@{}p{0.30\textwidth} p{0.38\textwidth} c@{}}
    \toprule
    Category & Benchmark & Rows \\
    \midrule
    Charts \& Tables & ChartQAPro~\citep{masry2025chartqapro} & 1,948 \\
     & CharXivReason~\citep{wang2024charxiv} & 1,000 \\
     & TableVQABench~\citep{kim2024tablevqabench} & 1,500 \\
     & EvoChart~\citep{huang2025evochart} & 1,250 \\
    \midrule
    Visual Math & MathVision~\citep{wang2024mathvision} & 3,040 \\
     & MathVista~\citep{lu2024mathvista} & 1,000 \\
     & MathVerse~\citep{zhang2024mathverse} & 788 \\
     & WeMath~\citep{qiao2025wemath} & 1,740 \\
    \midrule
    Science \& General & PhyX mini MC~\citep{shen2025phyx} & 1,000 \\
     & MMMU-ProVis~\citep{yue2025mmmupro} & 1,730 \\
     & RealWorldQA~\citep{xai2024realworldqa} & 765 \\
     & MMStar~\citep{chen2024mmstar} & 1,500 \\
    \midrule
    Spatial Reasoning & EmbSpatial~\citep{du2024embspatial} & 3,640 \\
     & SpatialVizBench~\citep{wang2026spatialviz} & 1,180 \\
     & CV-Bench 3D~\citep{tong2024cambrian} & 1,200 \\
     & ERQA~\citep{deepmind2025erqa} & 400 \\
    \midrule
    Perception \& Counting & BLINK~\citep{fu2024blink} & 1,901 \\
     & CountBenchQA~\citep{paiss2023countbench} & 487 \\
     & CountQA~\citep{tamarapalli2025countqa} & 1,528 \\
     & TreeBench~\citep{wang2026treebench} & 405 \\
    \midrule
    Puzzles \& Logic & PuzzleVQA~\citep{chia2024puzzlevqa} & 2,000 \\
     & VisualPuzzles~\citep{song2025visualpuzzles} & 1,168 \\
     & LogicVista~\citep{xiao2024logicvista} & 447 \\
     & MME-Reasoning~\citep{yuan2025mmereasoning} & 1,188 \\
    \bottomrule
  \end{tabular}
\end{table}

\subsection{External benchmark uncertainty}

Table~\ref{tab:external-bootstrap-cis} reports paired confidence intervals for
the improvement from each base checkpoint to its \trace-trained counterpart.
For each bootstrap replicate, evaluation items are resampled jointly with the
predictions from all three decoding seeds. Benchmarks with grouped metrics are
resampled at their official scoring unit: problem families for WeMath and
items within official splits for TableVQABench.

\begin{table}[t]
  \centering
  \fontsize{8.5}{9.5}\selectfont
  \setlength{\tabcolsep}{5pt}
  \renewcommand{\arraystretch}{1.10}
  \caption{Paired Base-to-\trace improvements on the external evaluation suite, in percentage points. Intervals are 95\% bootstrap percentile intervals, computed from the 2.5th and 97.5th percentiles of 10,000 paired item-bootstrap replicates; each sampled item carries results from all three decoding seeds. Bold intervals exclude zero. These intervals quantify variation over evaluation items rather than independently trained checkpoints.}
  \label{tab:external-bootstrap-cis}
  \begin{tabularx}{\textwidth}{@{}>{\raggedright\arraybackslash}p{0.20\textwidth} *{4}{>{\centering\arraybackslash}X}@{}}
    \toprule
    & \multicolumn{2}{c}{3B} & \multicolumn{2}{c}{7B} \\
    \cmidrule(lr){2-3}\cmidrule(l){4-5}
    Benchmark & $\Delta$ & 95\% CI & $\Delta$ & 95\% CI \\
    \midrule
    \rowcolor{TraceChartsBand} \multicolumn{5}{@{}l@{}}{\textbf{Charts \& Tables}} \\
    \rowcolor{TraceChartsBand} ChartQAPro & $-0.14$ & $[-1.30,\ +1.06]$ & $+2.24$ & \textbf{$[+0.76,\ +3.66]$} \\
    \rowcolor{TraceChartsBand} CharXivReason & $+5.77$ & \textbf{$[+4.13,\ +7.40]$} & $+7.40$ & \textbf{$[+5.33,\ +9.50]$} \\
    \rowcolor{TraceChartsBand} TableVQABench & $+2.72$ & \textbf{$[+1.78,\ +3.69]$} & $+3.11$ & \textbf{$[+2.01,\ +4.21]$} \\
    \rowcolor{TraceChartsBand} EvoChart & $-1.68$ & \textbf{$[-3.20,\ -0.16]$} & $+7.84$ & \textbf{$[+5.89,\ +9.79]$} \\
    \specialrule{0.35pt}{0pt}{0pt}
    \rowcolor{TraceMathBand} \multicolumn{5}{@{}l@{}}{\textbf{Visual Math}} \\
    \rowcolor{TraceMathBand} MathVision & $+6.10$ & \textbf{$[+5.02,\ +7.18]$} & $+2.61$ & \textbf{$[+1.54,\ +3.66]$} \\
    \rowcolor{TraceMathBand} MathVista & $+6.30$ & \textbf{$[+4.53,\ +8.07]$} & $+4.70$ & \textbf{$[+2.80,\ +6.53]$} \\
    \rowcolor{TraceMathBand} MathVerse & $+6.43$ & \textbf{$[+4.27,\ +8.54]$} & $+4.31$ & \textbf{$[+2.12,\ +6.56]$} \\
    \rowcolor{TraceMathBand} WeMath & $+10.95$ & \textbf{$[+8.73,\ +13.24]$} & $+10.96$ & \textbf{$[+8.39,\ +13.56]$} \\
    \specialrule{0.35pt}{0pt}{0pt}
    \rowcolor{TraceScienceBand} \multicolumn{5}{@{}l@{}}{\textbf{Science \& General}} \\
    \rowcolor{TraceScienceBand} PhyX mini MC & $+4.67$ & \textbf{$[+3.40,\ +5.90]$} & $+7.73$ & \textbf{$[+5.10,\ +10.37]$} \\
    \rowcolor{TraceScienceBand} MMMU-ProVis & $+4.57$ & \textbf{$[+3.12,\ +6.05]$} & $+3.66$ & \textbf{$[+2.22,\ +5.13]$} \\
    \rowcolor{TraceScienceBand} RealWorldQA & $+1.79$ & \textbf{$[+0.92,\ +2.66]$} & $+3.05$ & \textbf{$[+1.44,\ +4.71]$} \\
    \rowcolor{TraceScienceBand} MMStar & $+2.98$ & \textbf{$[+1.93,\ +4.04]$} & $+3.76$ & \textbf{$[+2.44,\ +5.16]$} \\
    \specialrule{0.35pt}{0pt}{0pt}
    \rowcolor{TraceSpatialBand} \multicolumn{5}{@{}l@{}}{\textbf{Spatial Reasoning}} \\
    \rowcolor{TraceSpatialBand} EmbSpatial & $+1.81$ & \textbf{$[+1.34,\ +2.28]$} & $+1.53$ & \textbf{$[+0.94,\ +2.09]$} \\
    \rowcolor{TraceSpatialBand} SpatialVizBench & $+1.75$ & $[-0.28,\ +3.84]$ & $+0.54$ & $[-1.64,\ +2.85]$ \\
    \rowcolor{TraceSpatialBand} CV-Bench 3D & $+8.31$ & \textbf{$[+6.50,\ +10.11]$} & $+4.75$ & \textbf{$[+3.11,\ +6.39]$} \\
    \rowcolor{TraceSpatialBand} ERQA & $+1.08$ & $[-0.50,\ +2.75]$ & $+2.17$ & $[-0.17,\ +4.50]$ \\
    \specialrule{0.35pt}{0pt}{0pt}
    \rowcolor{TracePerceptionBand} \multicolumn{5}{@{}l@{}}{\textbf{Perception \& Counting}} \\
    \rowcolor{TracePerceptionBand} BLINK & $+2.61$ & \textbf{$[+1.75,\ +3.49]$} & $+3.26$ & \textbf{$[+2.03,\ +4.54]$} \\
    \rowcolor{TracePerceptionBand} CountBenchQA & $+3.22$ & \textbf{$[+1.85,\ +4.65]$} & $+2.67$ & \textbf{$[+1.03,\ +4.31]$} \\
    \rowcolor{TracePerceptionBand} CountQA & $+0.76$ & $[-0.07,\ +1.59]$ & $+2.92$ & \textbf{$[+1.44,\ +4.43]$} \\
    \rowcolor{TracePerceptionBand} TreeBench & $-0.66$ & $[-2.47,\ +1.15]$ & $+2.22$ & $[+0.00,\ +4.44]$ \\
    \specialrule{0.35pt}{0pt}{0pt}
    \rowcolor{TracePuzzlesBand} \multicolumn{5}{@{}l@{}}{\textbf{Puzzles \& Logic}} \\
    \rowcolor{TracePuzzlesBand} PuzzleVQA & $+6.58$ & \textbf{$[+4.90,\ +8.23]$} & $+5.45$ & \textbf{$[+3.85,\ +7.08]$} \\
    \rowcolor{TracePuzzlesBand} VisualPuzzles & $+2.31$ & \textbf{$[+0.37,\ +4.22]$} & $+2.94$ & \textbf{$[+0.88,\ +4.97]$} \\
    \rowcolor{TracePuzzlesBand} LogicVista & $+3.65$ & \textbf{$[+0.45,\ +6.79]$} & $+4.70$ & \textbf{$[+1.42,\ +7.98]$} \\
    \rowcolor{TracePuzzlesBand} MME-Reasoning & $+2.44$ & \textbf{$[+0.76,\ +4.07]$} & $+2.86$ & \textbf{$[+1.04,\ +4.69]$} \\
    \bottomrule
  \end{tabularx}
\end{table}

\section{Representative Task Atlas}
\label{app:task-atlas}

Each page presents twelve seeded examples from one visual domain. Panels pair each rendered instance with its question and typed answer. Rule-heavy questions are condensed only when required for legibility.

\begingroup
\setlength{\fboxsep}{2pt}
\setlength{\fboxrule}{0.3pt}
\definecolor{traceatlasborder}{HTML}{D1D5DB}
\definecolor{traceatlasanswer}{HTML}{2A6FB5}
\newcommand{\traceatlaspromptnormal}{\fontsize{5.5}{6.0}\selectfont}
\newcommand{\traceatlaspromptdense}{\fontsize{4.75}{5.2}\selectfont}
\newcommand{\traceatlascard}[5]{%
  \begingroup
  \color{traceatlasborder}%
  \fbox{%
    \color{black}%
    \begin{minipage}[t][0.176\textheight][t]{0.298\textwidth}
      \raggedright\setlength{\parindent}{0pt}%
      {\fontsize{5.1}{5.6}\selectfont\sffamily\color{gray}\textbf{#1}\par}%
      \vspace{1pt}%
      \centering\includegraphics[width=\linewidth,height=0.088\textheight,keepaspectratio]{#2}\par
      \vspace{1pt}%
      \raggedright{\sffamily #5 #3\par}%
      \vfill
      {\fontsize{5.6}{6.1}\selectfont\sffamily\color{traceatlasanswer}\textbf{Answer: #4}\par}%
    \end{minipage}%
  }%
  \endgroup
}

\begin{figure}[t]
\centering
\traceatlascard{Error Interval: Reference Containment}{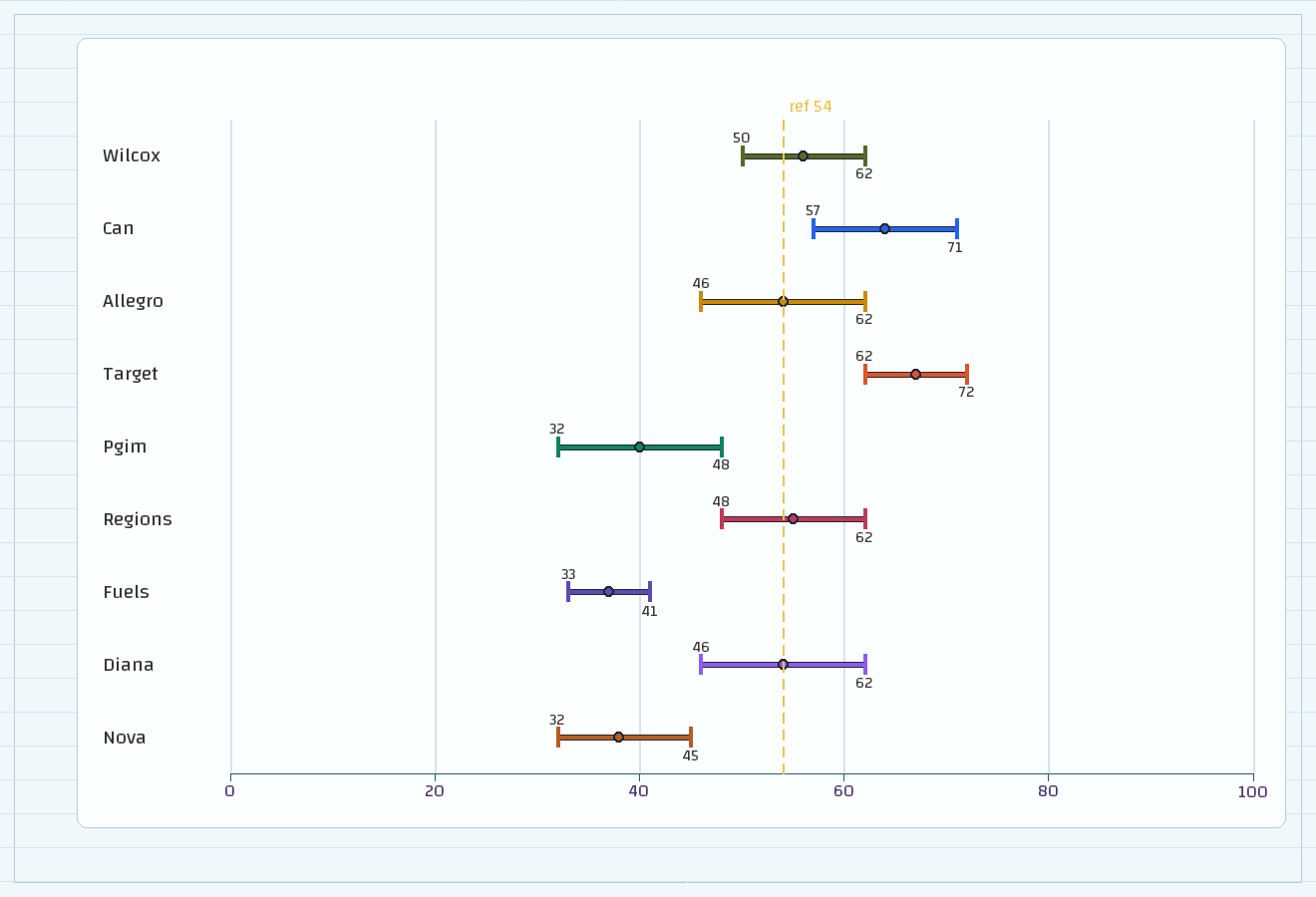}{The image shows a horizontal interval chart with labeled categories, point estimates, and lower-to-upper interval whiskers. How many intervals contain the reference value 54?}{4}{\traceatlaspromptnormal}
\hfill
\traceatlascard{Region Map: Numeric Interval Region}{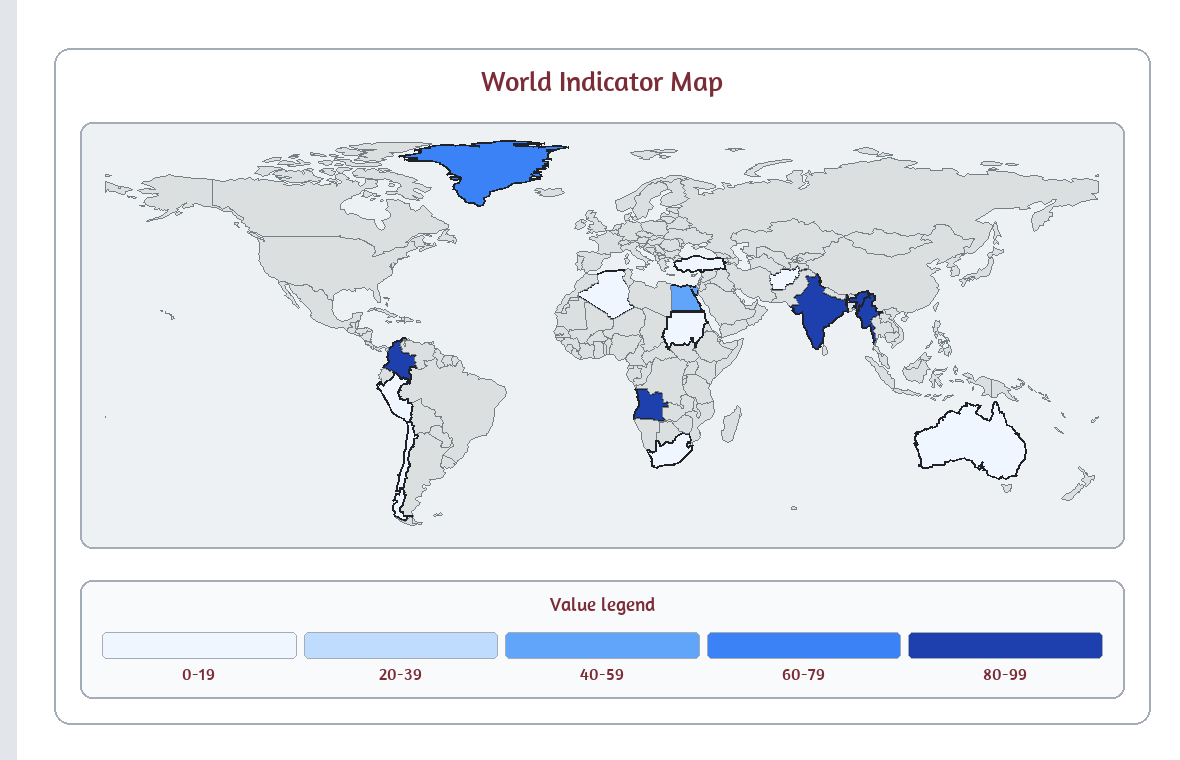}{This image shows a world map with selected countries colored by value and a color legend. How many countries fall in value bins between 20 and 99, inclusive?}{6}{\traceatlaspromptnormal}
\hfill
\traceatlascard{Scatter Cluster: Centroid Option Selection}{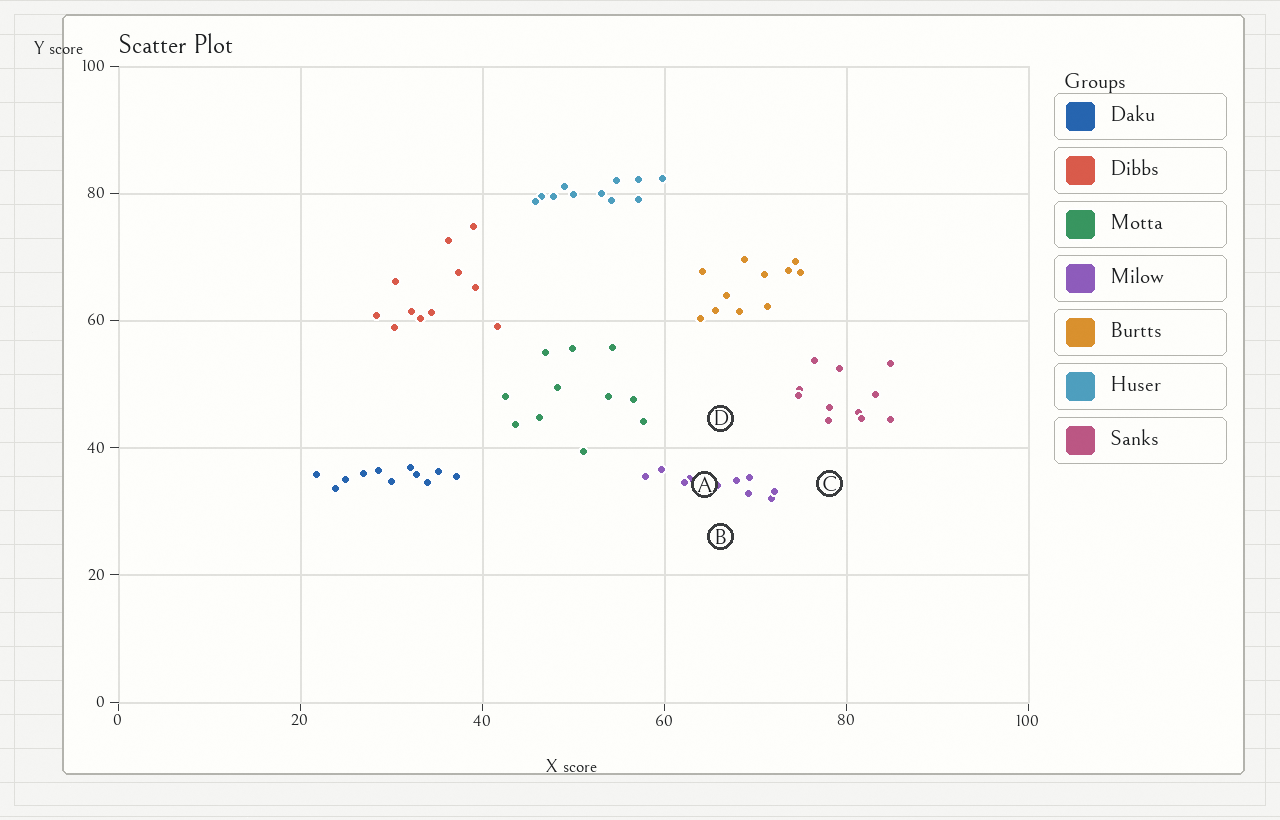}{The visual shows a scatter plot with several colored point clusters and a matching legend. Find the centroid of cluster "Milow" by eye. Which option marker is closest?}{A}{\traceatlaspromptnormal}
\par\vspace{2pt}
\noindent
\traceatlascard{Size Encoding: Category Extremum Across Panels}{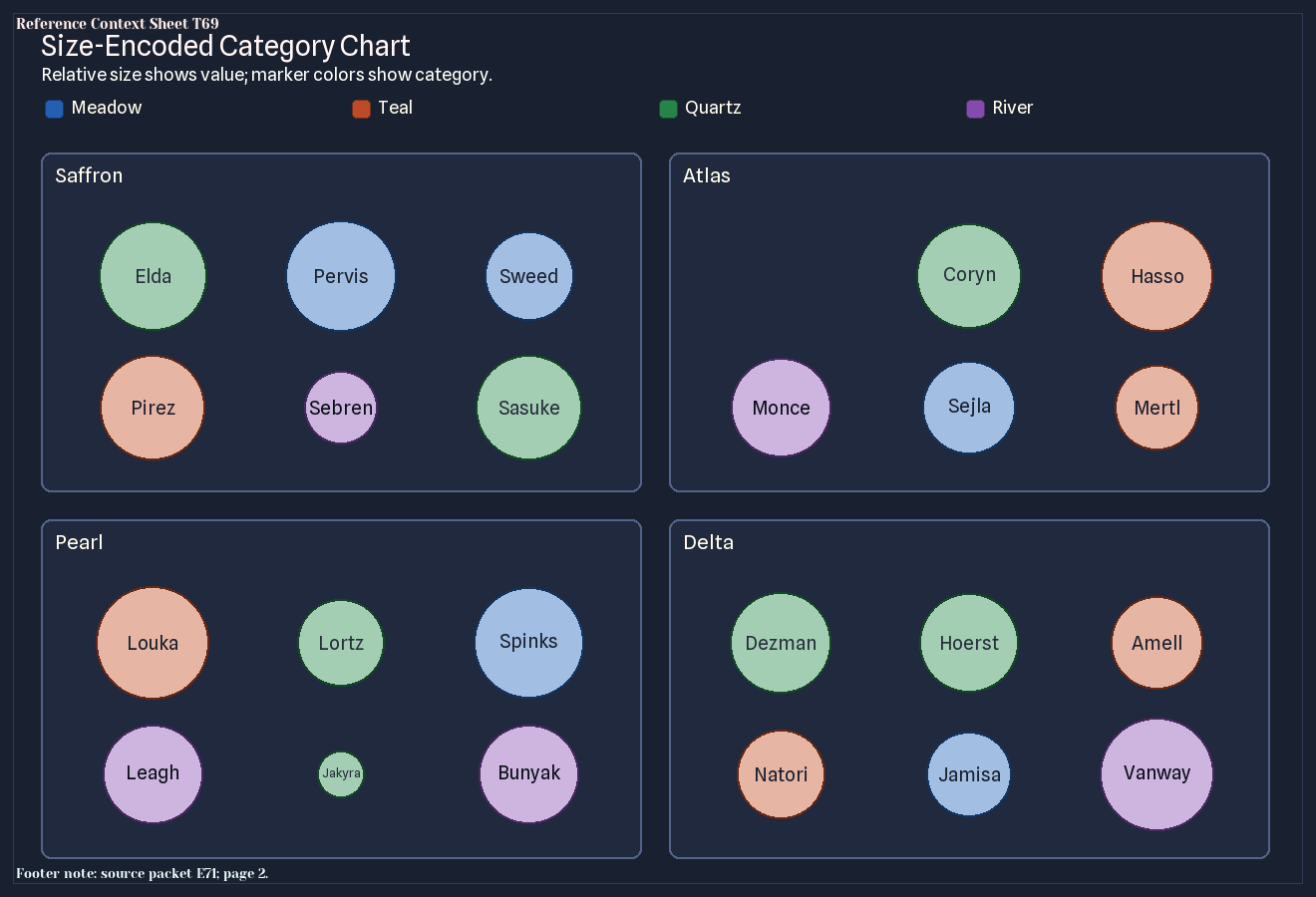}{The image shows multiple packed bubble-chart panels where bubble size indicates value and color indicates category. For category "River", which panel contains the item with the largest size?}{Delta}{\traceatlaspromptnormal}
\hfill
\traceatlascard{Scatter Points: Category Threshold Point}{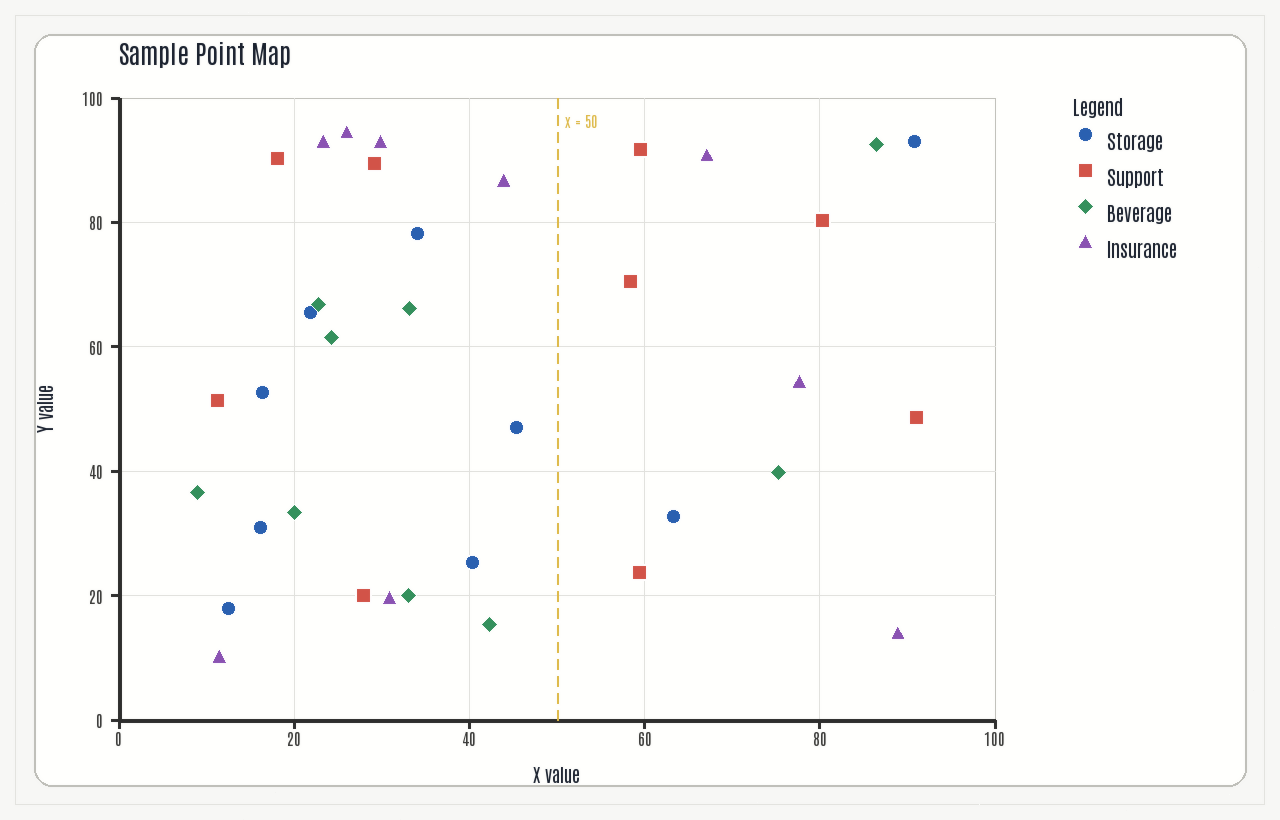}{The chart shows a categorized scatter plot with individual data points, numeric x- and y-axes, and a legend. How many plotted points from category "Storage" satisfy x greater than 50?}{2}{\traceatlaspromptnormal}
\hfill
\traceatlascard{Single Series: Threshold Count}{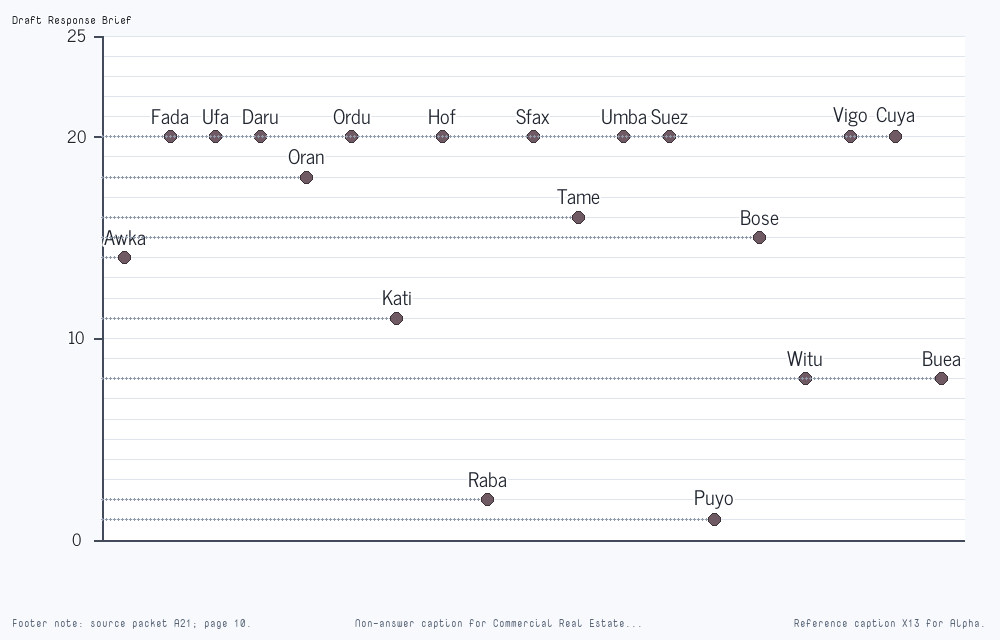}{The image shows an ordered labeled dot plot where point labels identify y-values from left to right. Determine how many marks show values strictly greater than 19.}{10}{\traceatlaspromptnormal}
\par\vspace{2pt}
\noindent
\traceatlascard{Population Pyramid: Age Group Threshold}{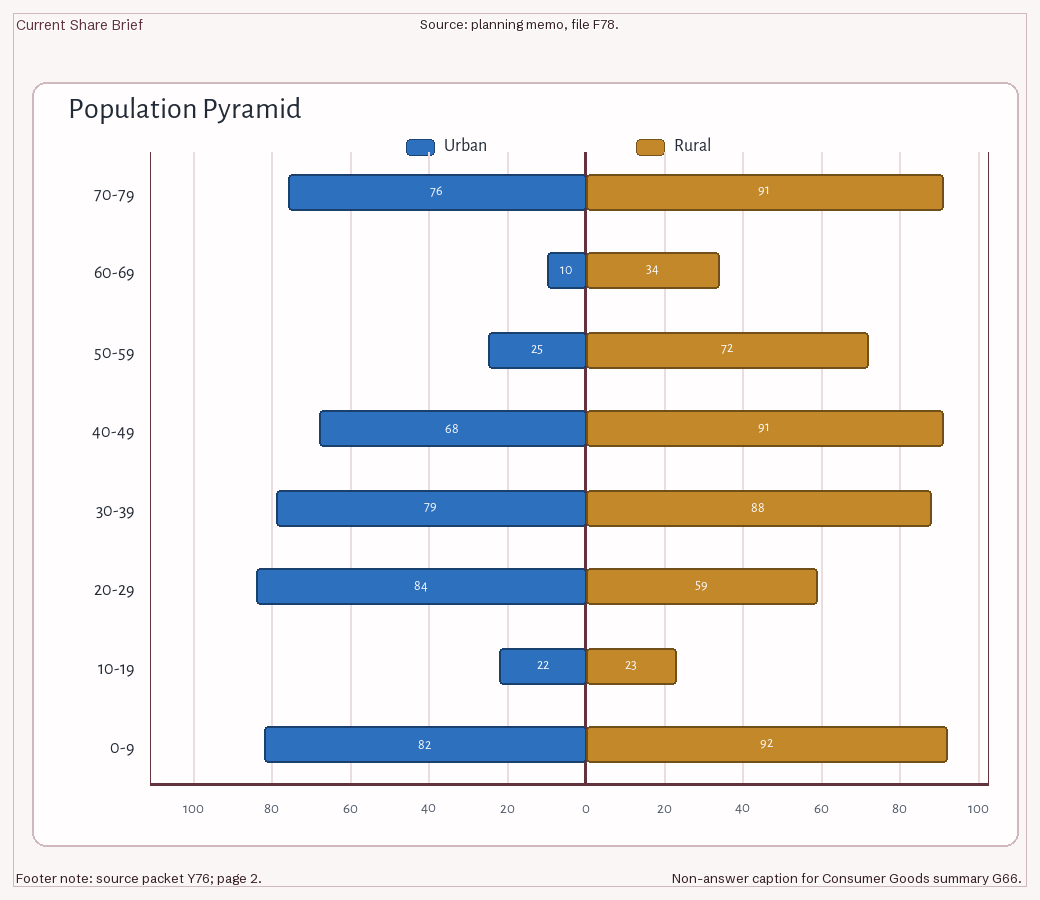}{The image shows a mirrored horizontal bar chart with one row per age group. The left and right bars show the two legend series on the same positive scale. In how many age groups is the sum of Urban and Rural at most 140?}{3}{\traceatlaspromptnormal}
\hfill
\traceatlascard{Area: Stacked Band Dominance}{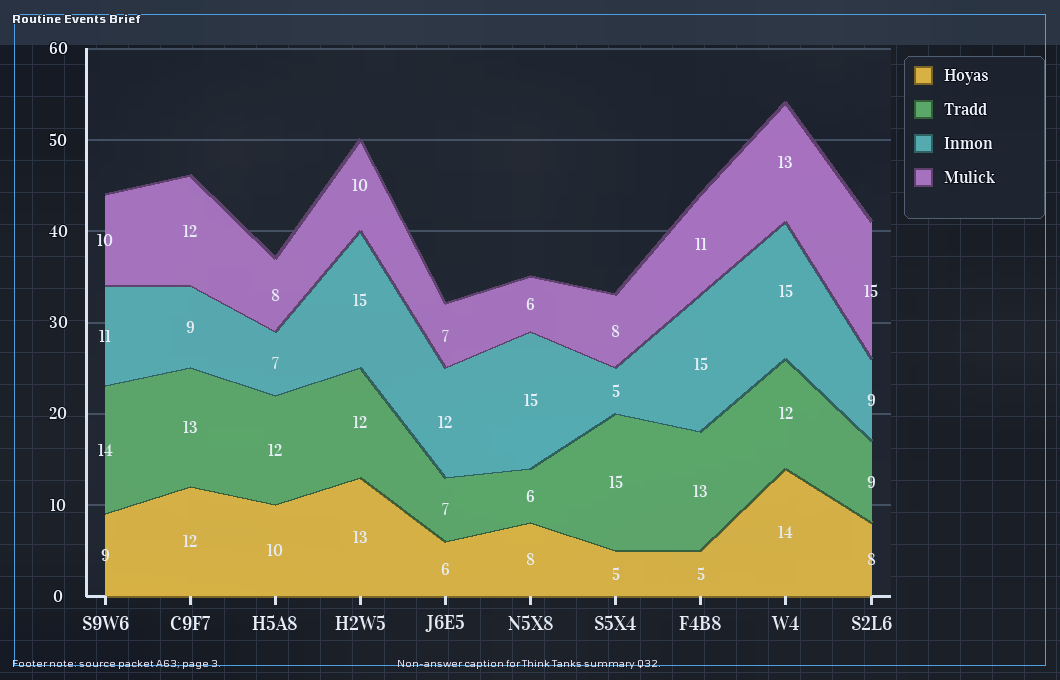}{The visual shows one stacked area chart with colored category bands, numeric band values, and a legend. Sum each category's band values over "S9W6" through "H2W5" in displayed x-axis order; which category is largest?}{Tradd}{\traceatlaspromptnormal}
\hfill
\traceatlascard{Annotated Series: Callout Endpoint Change}{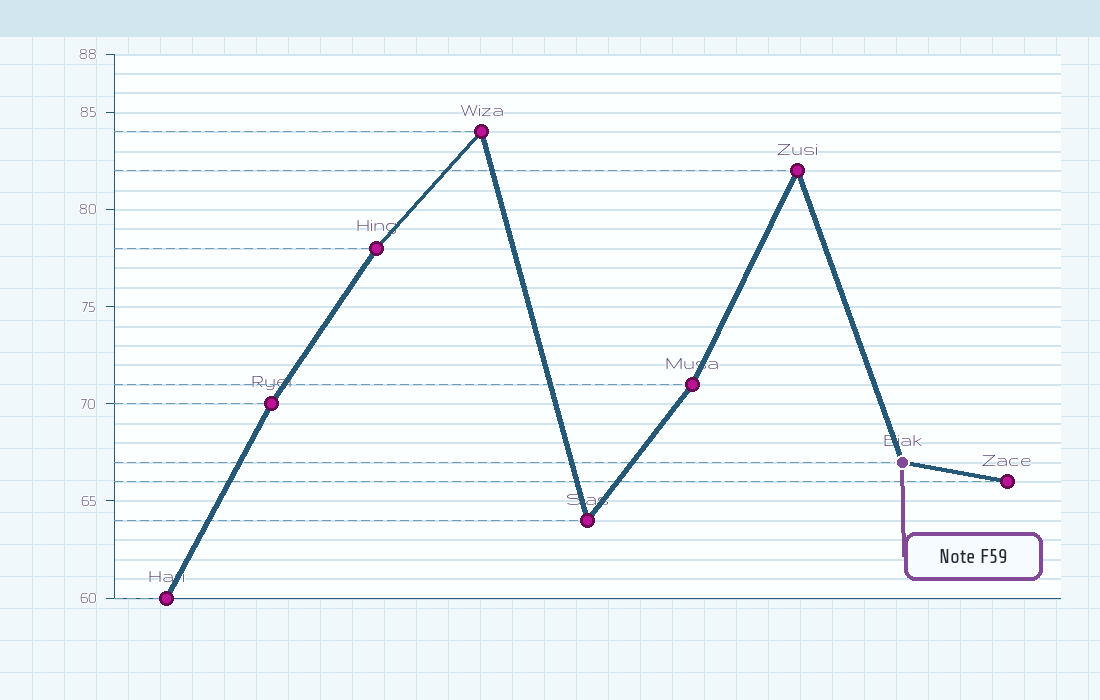}{The figure shows a labeled line chart with one visible annotation. Compare the callout mark with "Hari". What is the nonnegative difference between their values?}{7}{\traceatlaspromptnormal}
\par\vspace{2pt}
\noindent
\traceatlascard{Combo Mark: Conditioned Line Extremum}{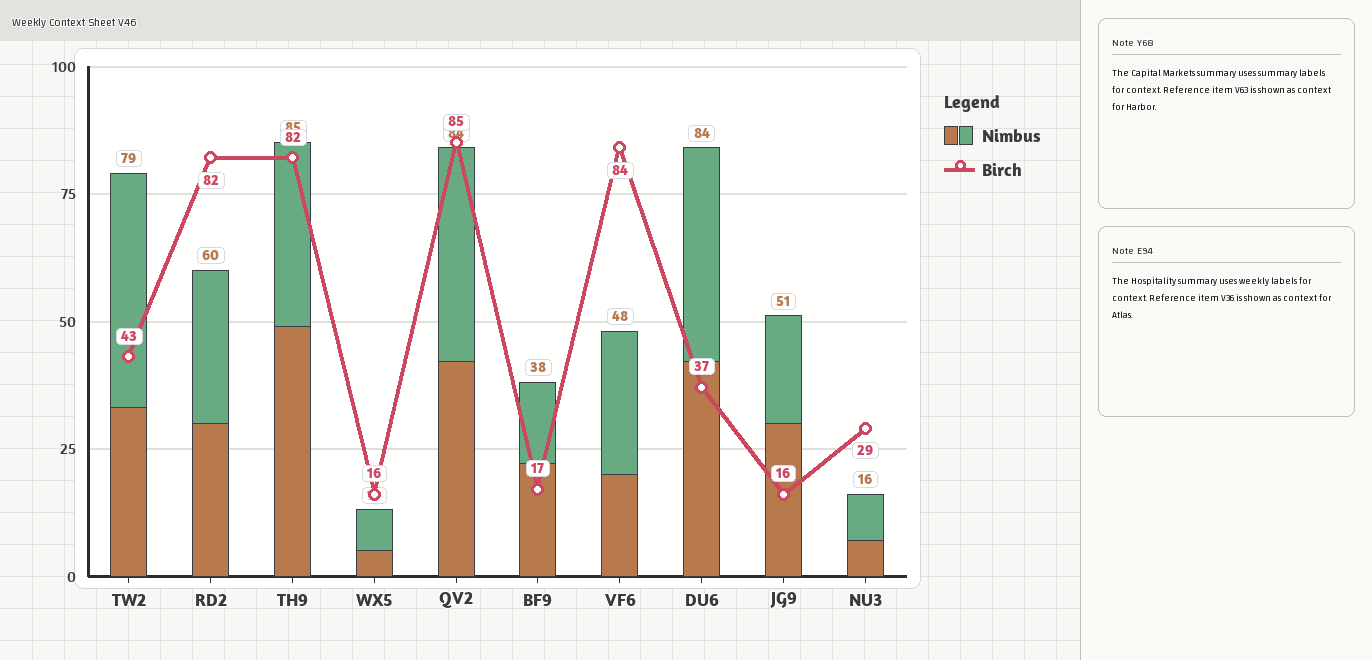}{The figure shows stacked bars for "Nimbus" with an overlaid line for "Birch"; each category displays the relevant printed values. First keep categories with "Nimbus" greater than 55; which remaining category has the maximum "Birch" value?}{QV2}{\traceatlaspromptdense}
\hfill
\traceatlascard{Uncertainty Band: Band Overlap}{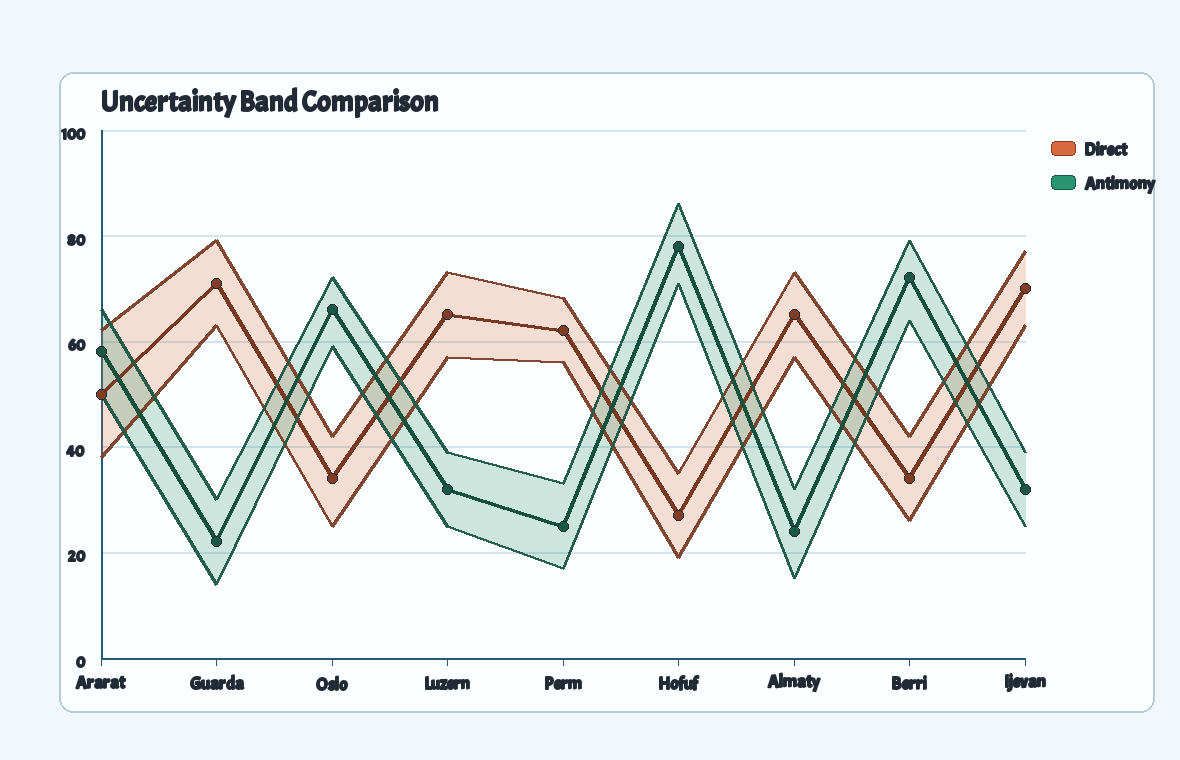}{The image shows a line chart with two labeled series, each shown with a central line and a shaded upper-to-lower uncertainty band. At how many x-axis labels do the shaded bands for "Direct" and "Antimony" overlap?}{1}{\traceatlaspromptnormal}
\hfill
\traceatlascard{Matrix: Off-Diagonal Confusion}{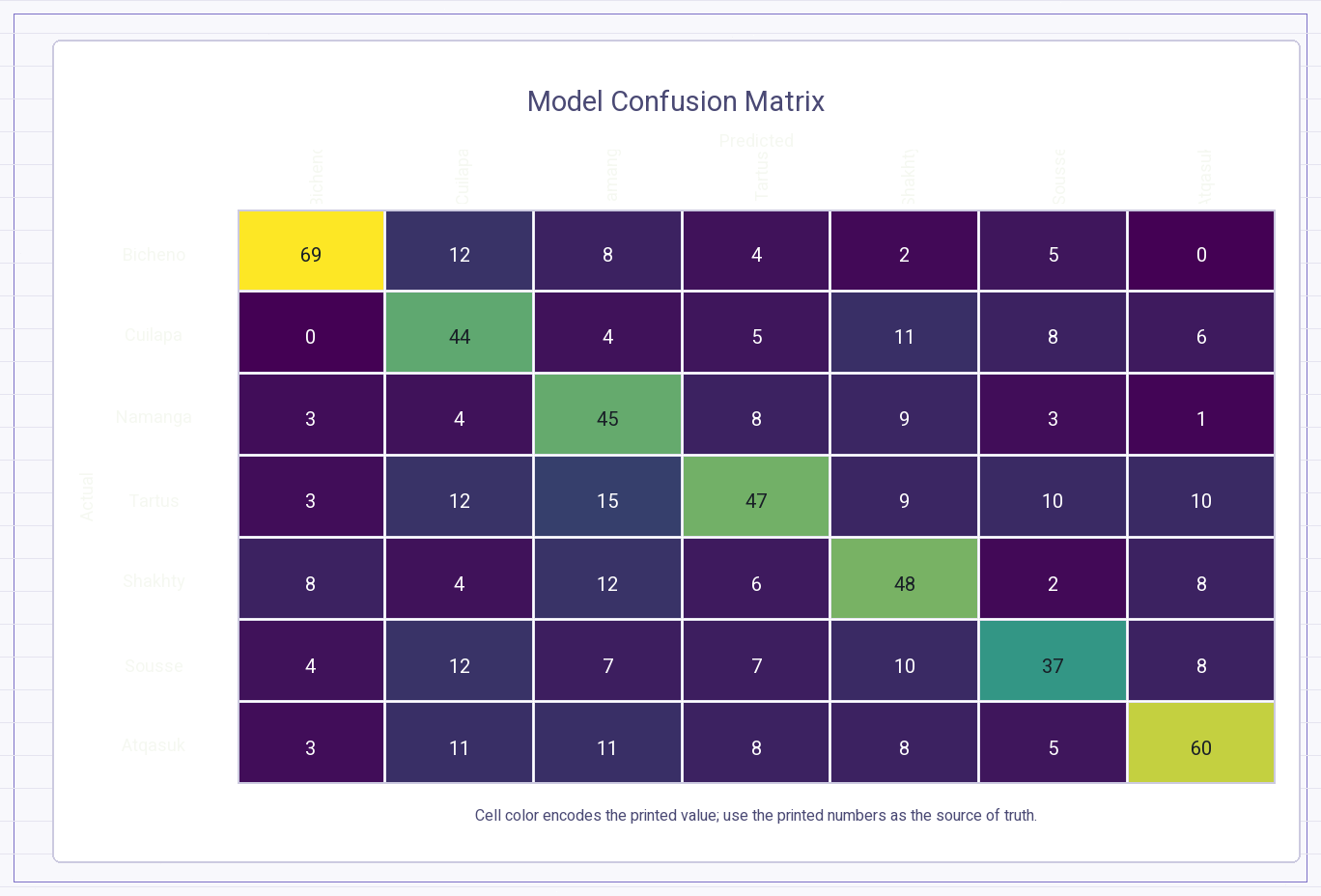}{The figure shows a labeled confusion matrix with printed integer counts in each active cell. In the row for actual class "Tartus", exclude the matching diagonal cell. What predicted column label has the largest count?}{Namanga}{\traceatlaspromptnormal}
\caption{Representative chart tasks.}
\label{fig:task-atlas-charts}
\end{figure}
\clearpage

\begin{figure}[p]
\centering
\traceatlascard{Sokoban: Box Goal Status}{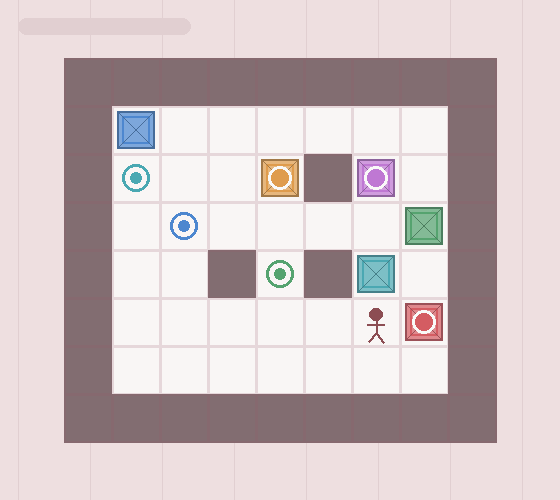}{The figure shows a Sokoban board with walls, a player, colored boxes, and colored goal dots; a box on its matching goal has the goal dot drawn on top of the box. How many boxes are not on their matching colored goal dots?}{3}{\traceatlaspromptdense}
\hfill
\traceatlascard{Checkers: Max Capture Chain Length}{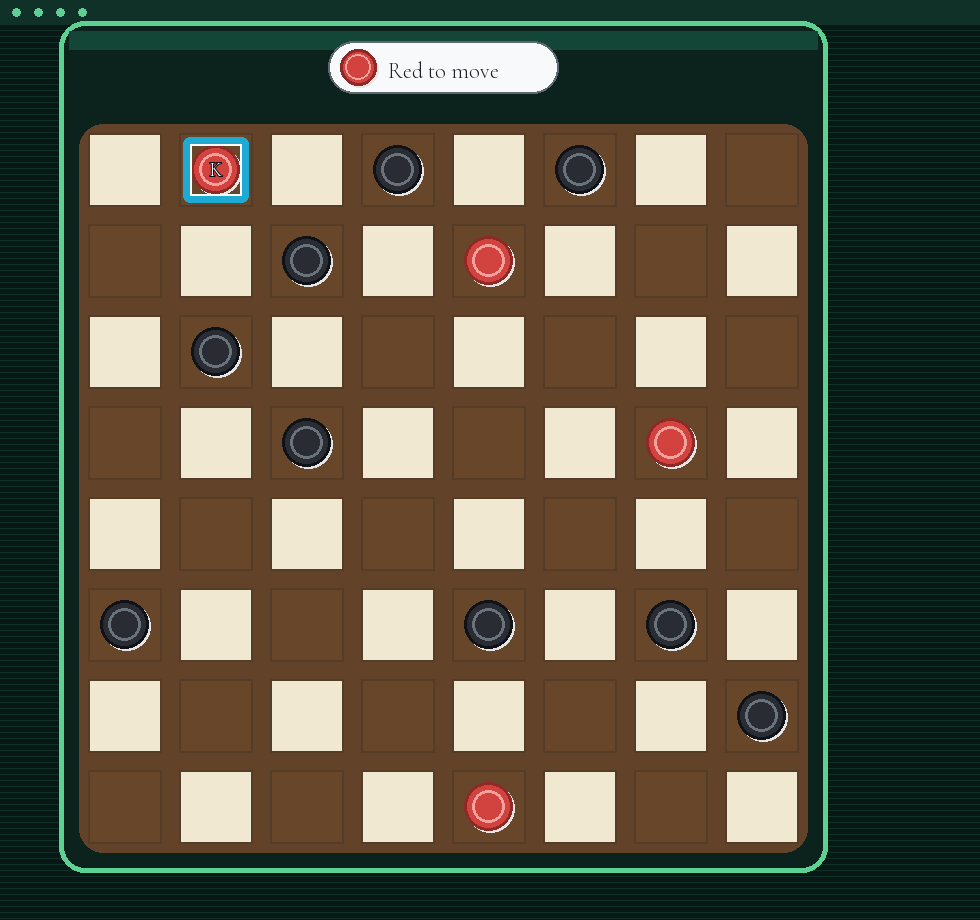}{Red moves on the shown 8 by 8 checkers board. A jump captures an adjacent opponent and lands on the empty square beyond. The marked piece is a king, so it may jump diagonally in either direction and continue while another capture is available. What is its maximum capture-chain length?}{2}{\traceatlaspromptdense}
\hfill
\traceatlascard{Darts: Dart Score}{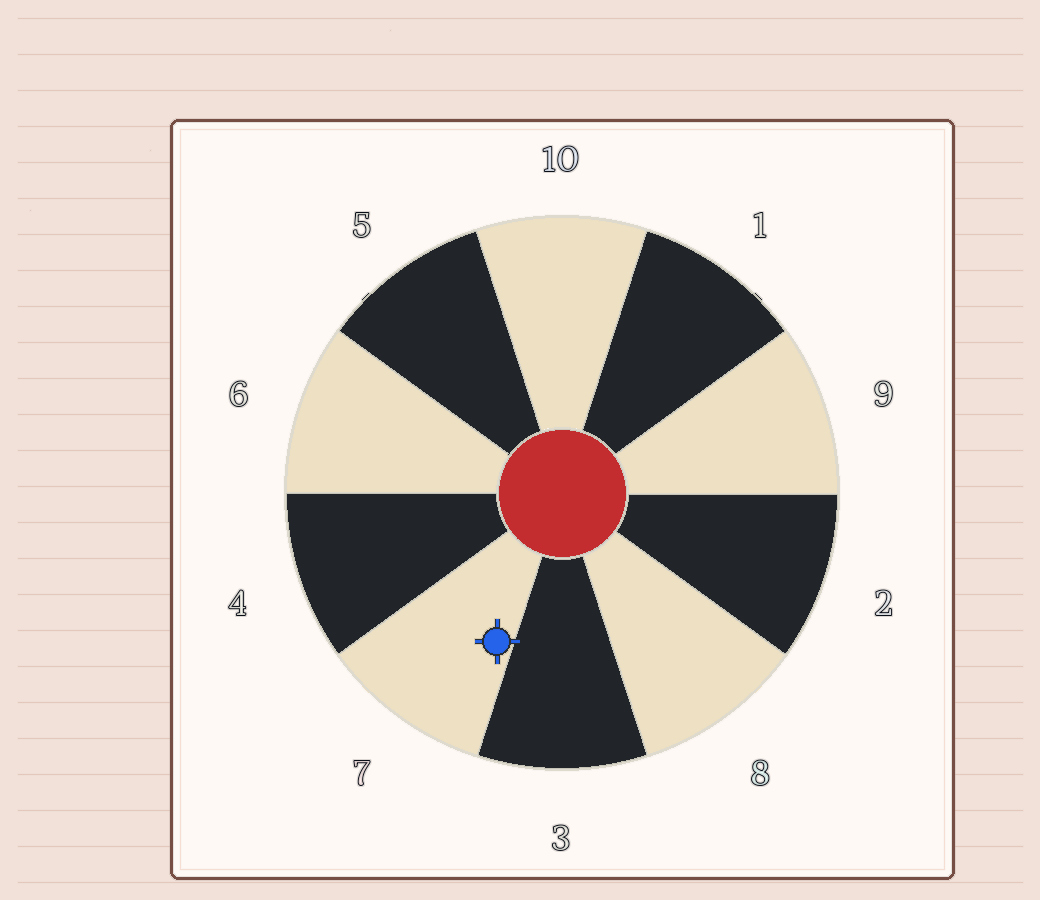}{The image shows a simplified labeled dartboard with visible dart markers. Using the simplified dartboard scoring rule, what is the dart's score? A dart in a numbered sector scores that number; a dart in the center bullseye scores 50.}{7}{\traceatlaspromptdense}
\par\vspace{2pt}
\noindent
\traceatlascard{Pac-Man: Next Item}{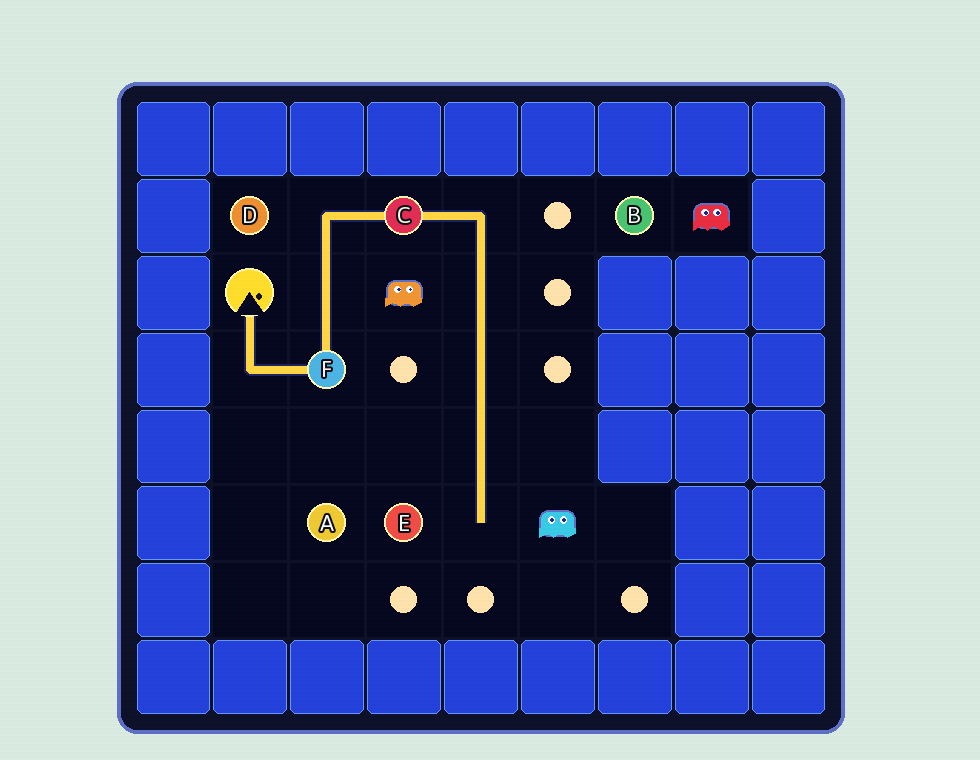}{This scene shows a Pac-Man-style maze with a visible Pac-Man marker, ghosts, pellets, bonus items, and a highlighted route. Which labeled bonus item is encountered first along the highlighted route?}{F}{\traceatlaspromptnormal}
\hfill
\traceatlascard{Minesweeper: Forced Cell}{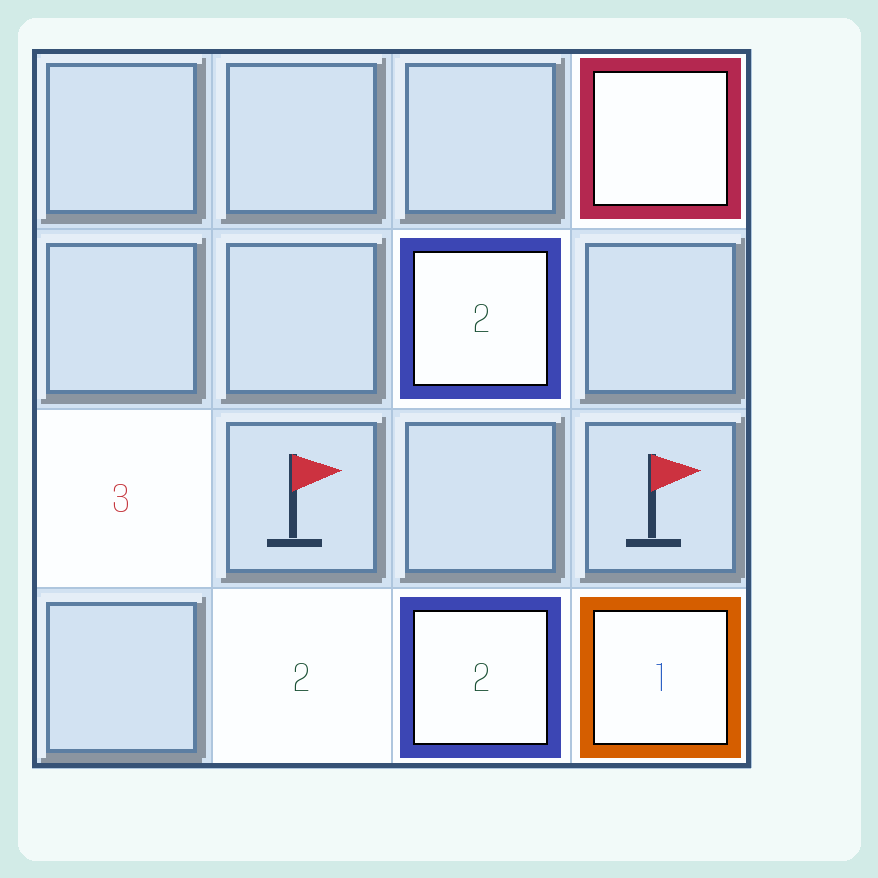}{In the Minesweeper grid, each number gives the mine count among its eight neighbors and each flag is a known mine. If a clue already has enough flags, its other hidden neighbors are safe; if every remaining hidden neighbor is needed, they are mines. From the outlined clue cells, how many hidden cells must be safe?}{5}{\traceatlaspromptdense}
\hfill
\traceatlascard{Chess: Target Square Attacker}{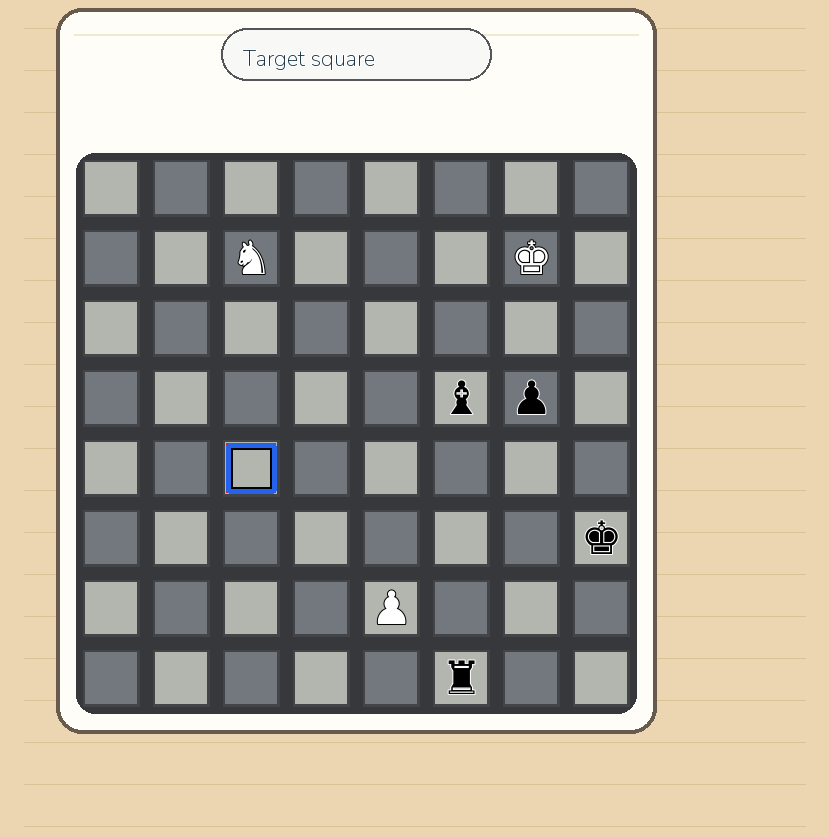}{The scene shows a partial chess board with white and black pieces. How many black pieces currently attack the red outlined target square?}{0}{\traceatlaspromptnormal}
\par\vspace{2pt}
\noindent
\traceatlascard{Reversi: Marked Move Flip}{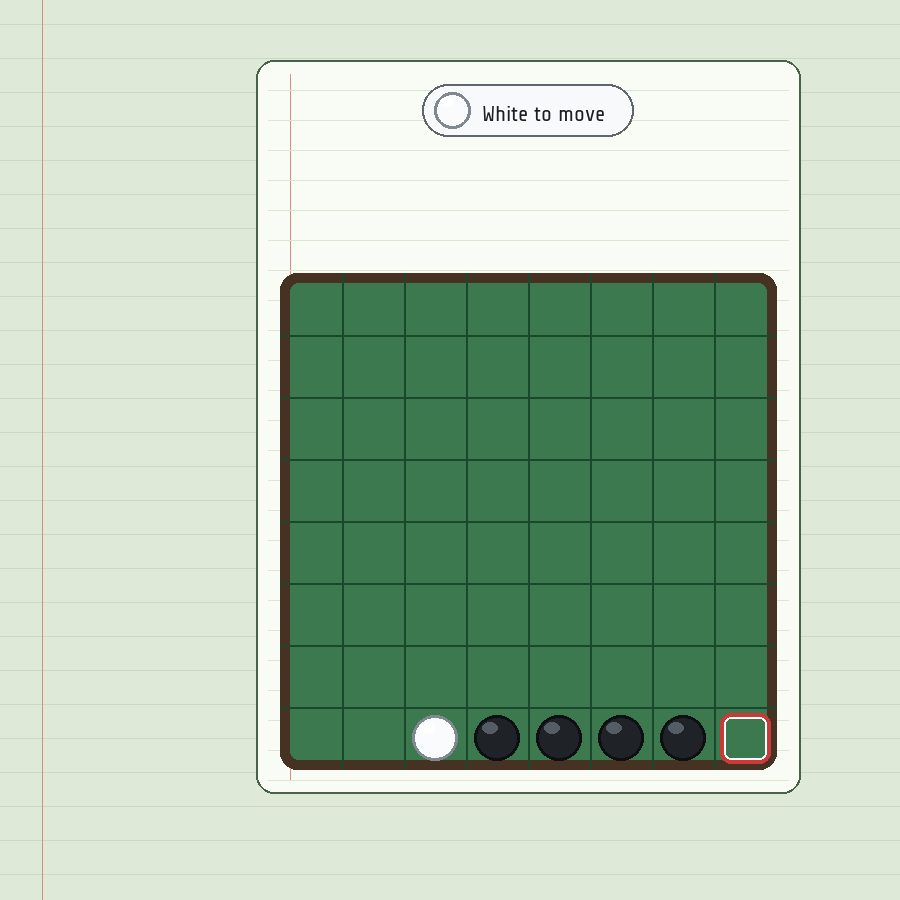}{White moves on the shown 8 by 8 Reversi board. A legal move brackets opponent discs along one or more straight lines, and all bracketed discs flip. How many discs flip after playing at the red marked square?}{4}{\traceatlaspromptdense}
\hfill
\traceatlascard{Bubble Shooter: Pop Color}{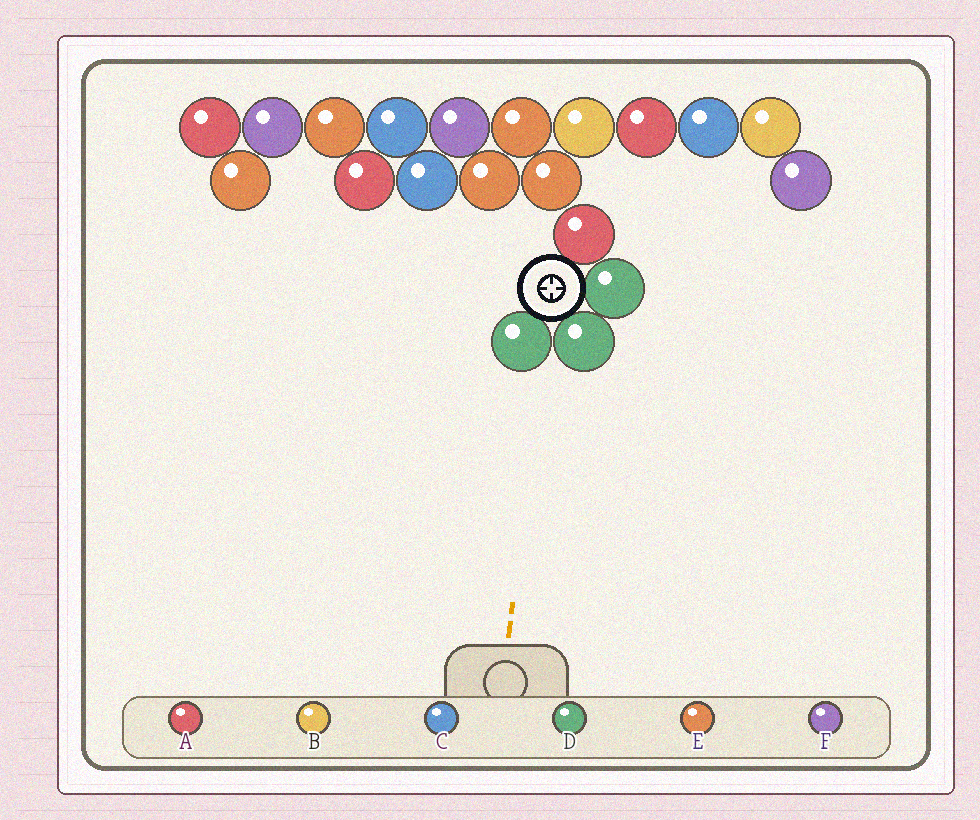}{A shot bubble attaches at the marked target. A connected same-color group pops when the placement makes its size at least three. Which option labels the bubble color that would pop a group there?}{D}{\traceatlaspromptdense}
\hfill
\traceatlascard{Snakes and Ladders: Remaining Distance to Finish}{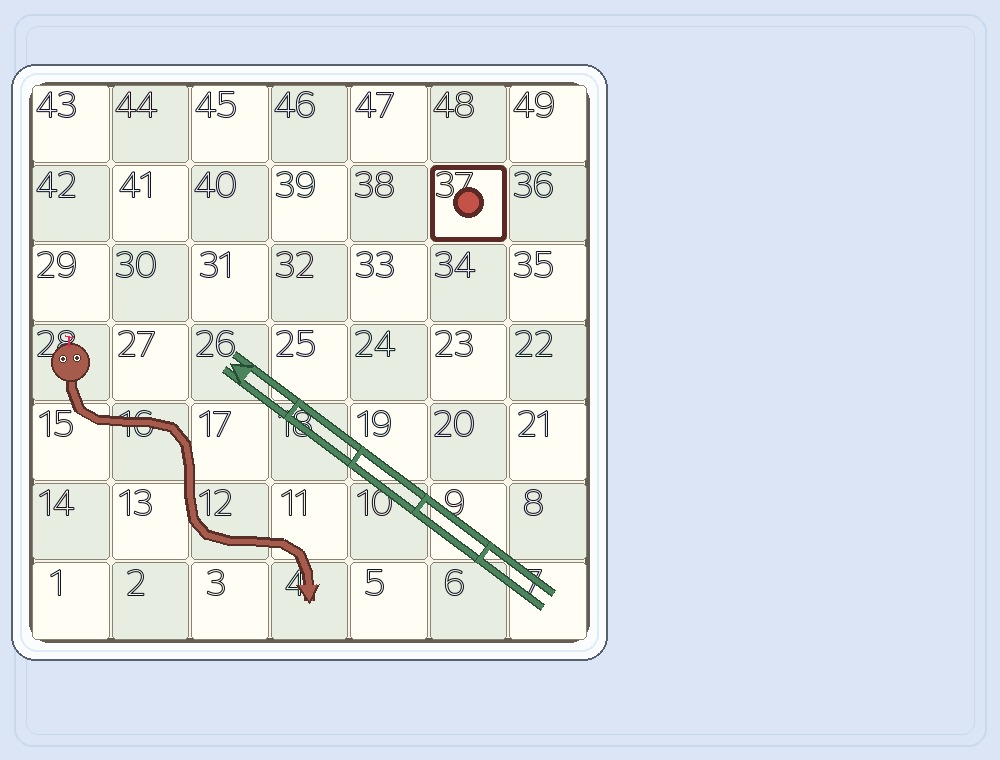}{The visual shows a numbered Snakes and Ladders board with one token and visible snakes and ladders. How far is the token from the final square by square number?}{12}{\traceatlaspromptnormal}
\par\vspace{2pt}
\noindent
\traceatlascard{Racing Track: Ahead Object}{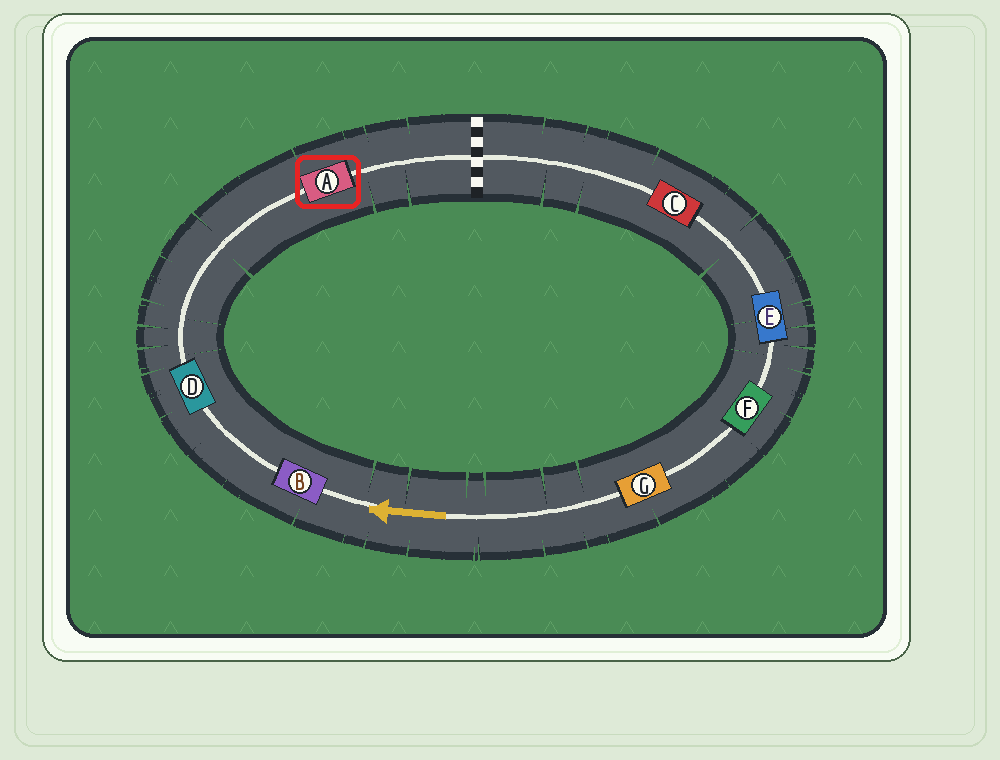}{The scene shows a top-down loop racing track with a checkered finish line, a direction arrow, a marked car, and other labeled cars. Follow the arrow direction from the marked car to the finish line. Do not count the marked car. Count only the cars ahead of the marked car before the finish line.}{0}{\traceatlaspromptdense}
\hfill
\traceatlascard{Circular Chess: Target Cell Reacher}{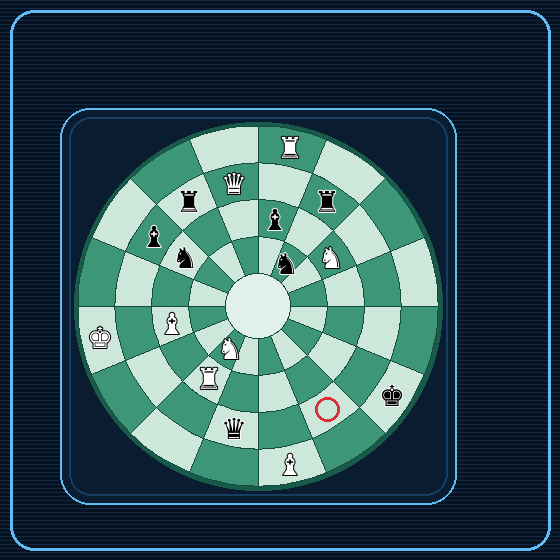}{On this four-ring, sixteen-sector circular chess board, sectors wrap but rings do not. The red cell is the target. Rooks move along or across rings, bishops diagonally, queens as both, knights jump, and kings one step; sliding pieces stop at the first occupied cell. How many Black pieces can reach the target?}{3}{\traceatlaspromptdense}
\hfill
\traceatlascard{Dots and Boxes: Owned Box}{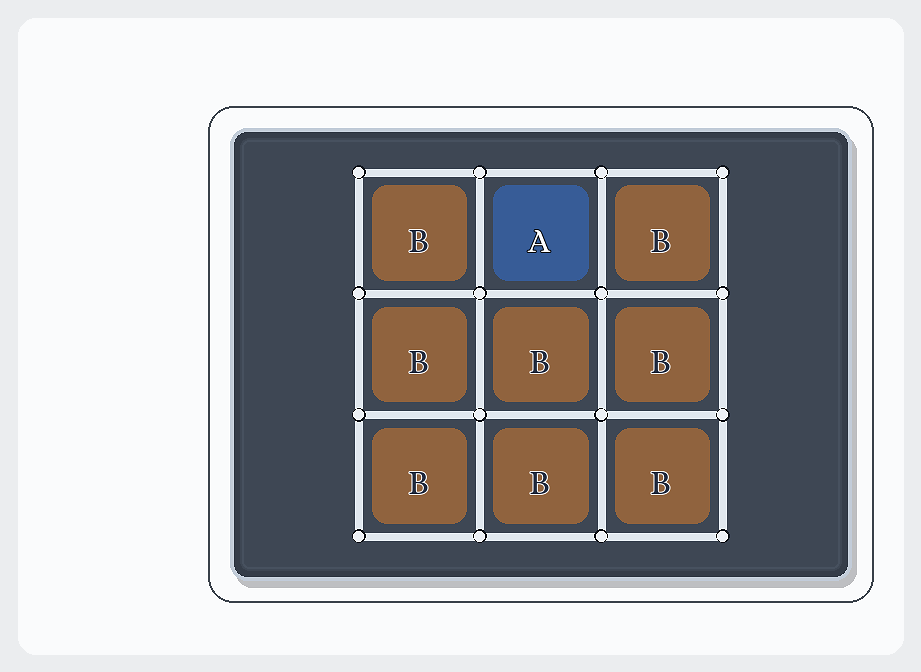}{The scene shows a dots-and-boxes grid with some edges already drawn. How many boxes are marked with player B?}{8}{\traceatlaspromptnormal}
\caption{Representative game tasks.}
\label{fig:task-atlas-games}
\end{figure}
\clearpage

\begin{figure}[p]
\centering
\traceatlascard{Circle Pair Tangents: Center Distance}{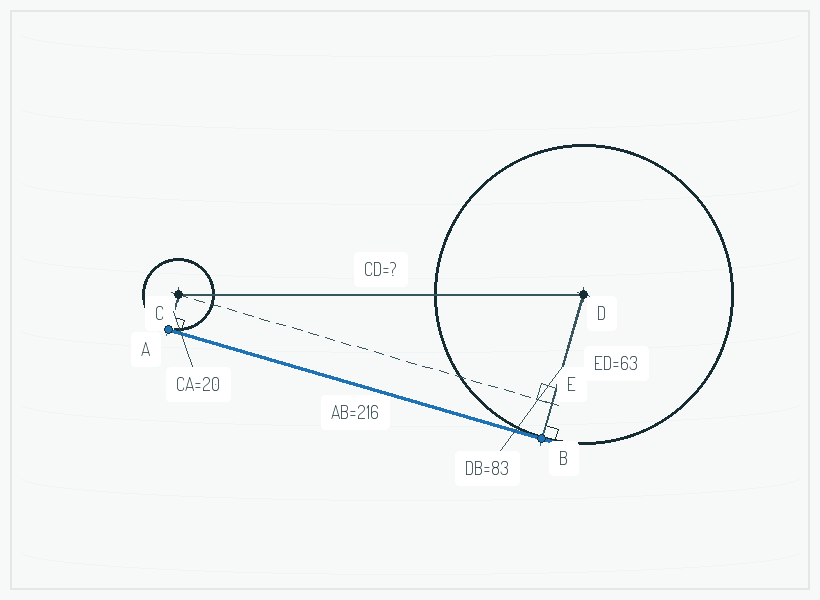}{This geometry diagram shows two separated non-concentric circles with centers C and D, a common tangent touching them at A and B, radii to the tangent points, and an auxiliary right triangle CED whose leg ED shows the radius difference. Find the center-to-center distance CD.}{225}{\traceatlaspromptdense}
\hfill
\traceatlascard{Solid Cross Section: Square Pyramid Parallel Slice Area}{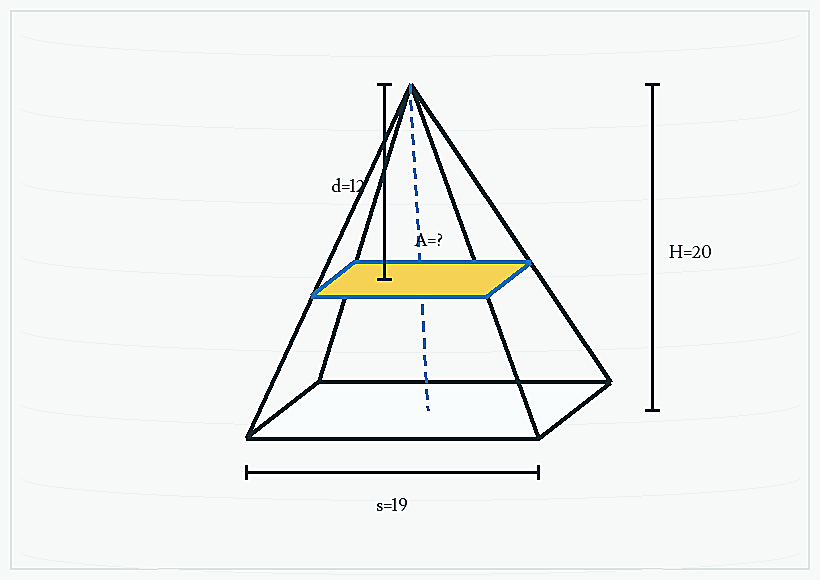}{The labeled diagram shows a square pyramid cut by a plane parallel to its base, with the cross-section marked and dimensions labeled. The marked slice is parallel to the base. Use similar solids with scale d/H to find the value of A for the cross-section.}{130}{\traceatlaspromptdense}
\hfill
\traceatlascard{Measuring Tools: Protractor Angle}{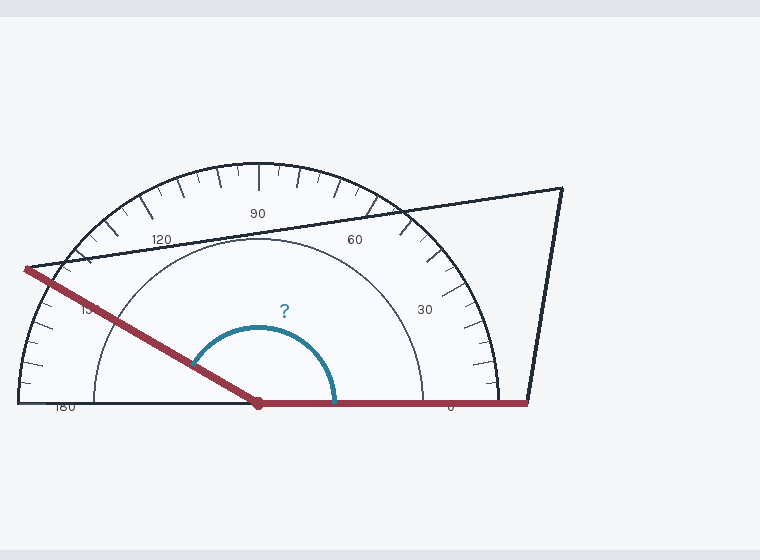}{The figure shows a geometric figure with a protractor placed at the marked angle. Read the protractor scale at the marked angle. What is the angle measure in degrees?}{150}{\traceatlaspromptnormal}
\par\vspace{2pt}
\noindent
\traceatlascard{Sector: Central Angle from Sector Measure}{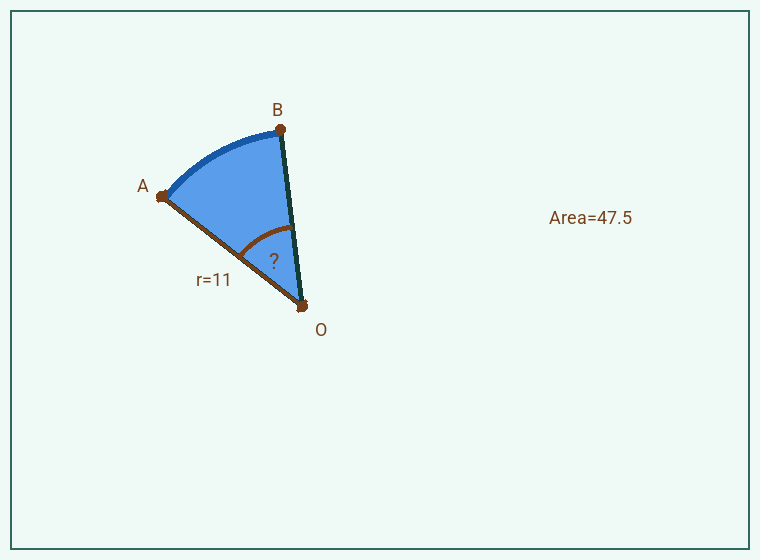}{The sector figure shows a circular sector with labeled points. Compute the measure of angle AOB. Use the labeled radius and shaded sector area.}{45}{\traceatlaspromptnormal}
\hfill
\traceatlascard{Special Quadrilateral: Segment Length}{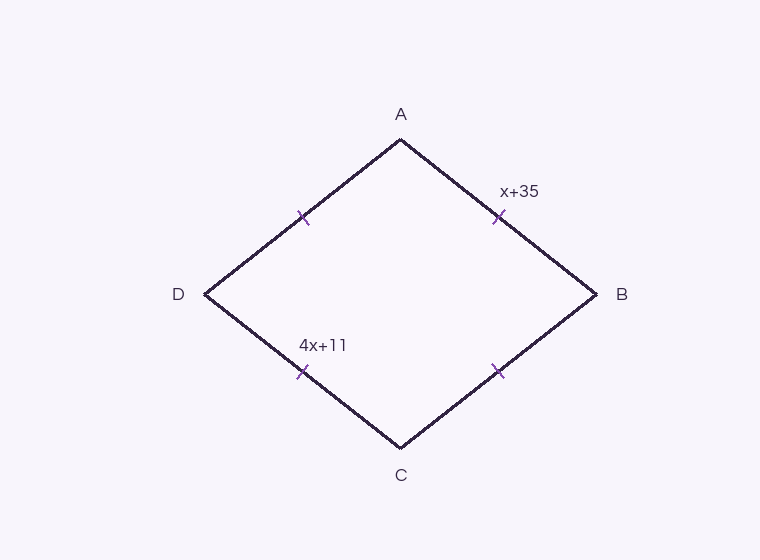}{The figure shows a quadrilateral with construction marks and measurement labels. Solve the visible rhombus side expressions. What is the length of "segment CD"?}{43}{\traceatlaspromptnormal}
\hfill
\traceatlascard{Paper Fold: Folded Segment Length}{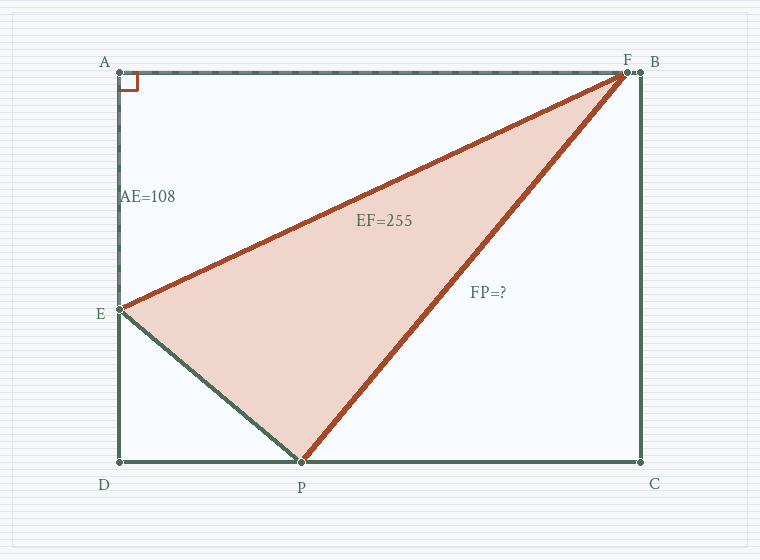}{This folded-paper diagram shows a folded paper corner with dashed original edges, a crease, a folded flap, and side-length labels. The crease EF folds A onto P. What is the value of the marked segment FP?}{231}{\traceatlaspromptnormal}
\par\vspace{2pt}
\noindent
\traceatlascard{Circle Theorem: Inscribed Angle from Arc}{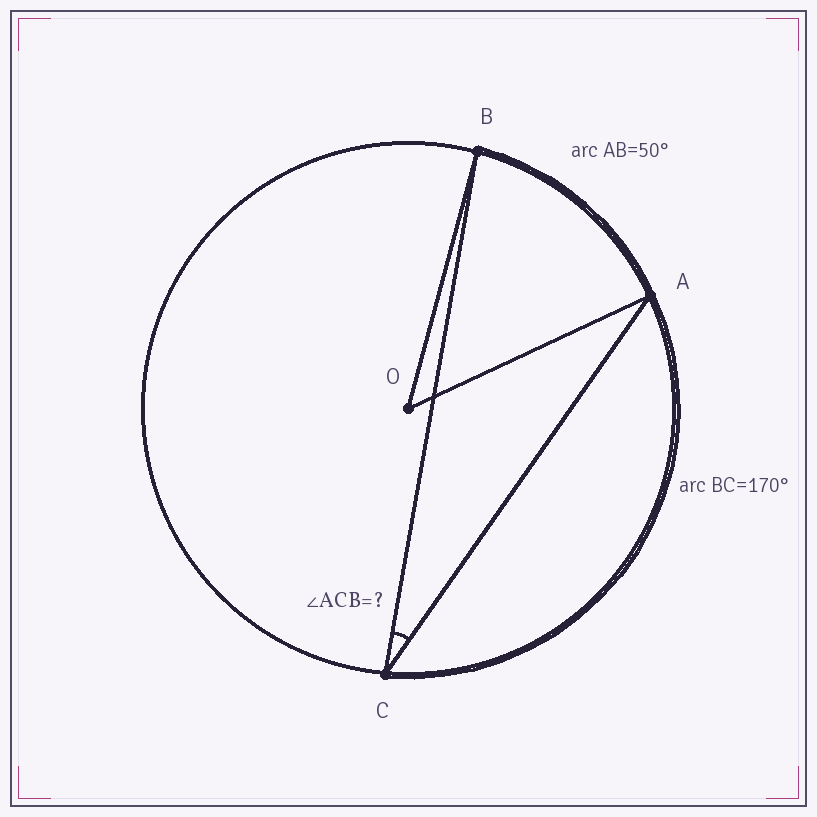}{The geometry diagram shows a circle with labeled points and shown measurements. Find the missing inscribed angle ACB.}{25}{\traceatlaspromptnormal}
\hfill
\traceatlascard{Rectangular Solid: Open Box Net Dimension}{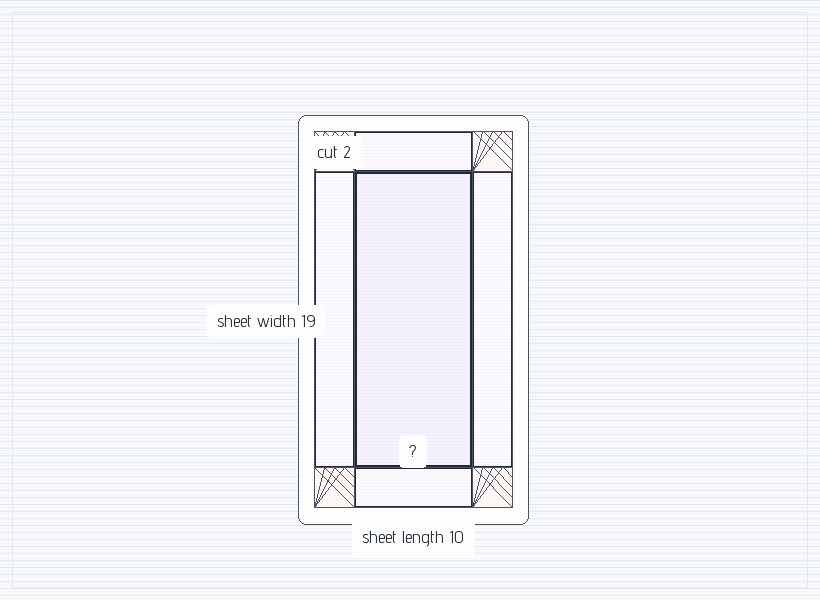}{The diagram shows an open-box net with sheet dimensions, equal corner cutouts, and one missing base dimension. Using the sheet dimensions and cut size, find the marked base dimension.}{6}{\traceatlaspromptnormal}
\hfill
\traceatlascard{Circle Centerline Overlap: Segment Length}{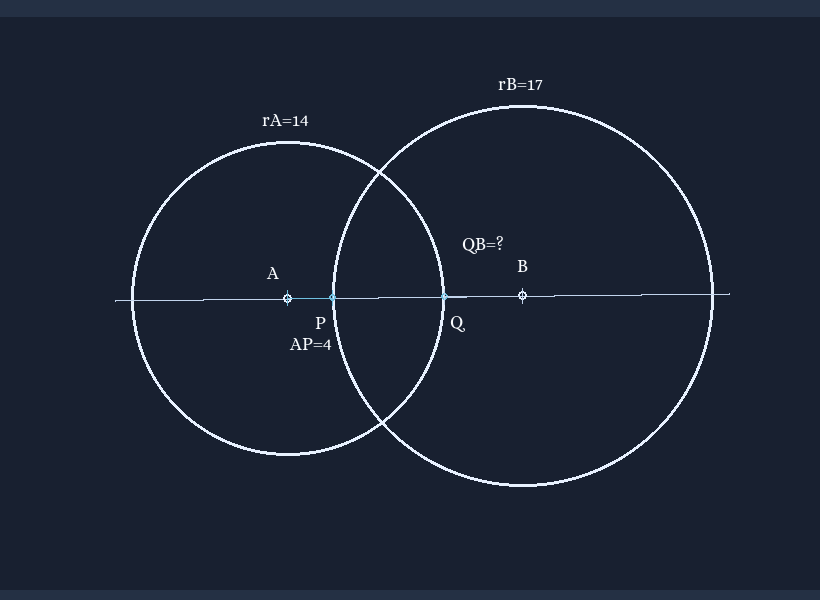}{The diagram shows two overlapping circles whose centers lie on one line. Boundary labels mark where circles cross the shared centerline; radius labels and centerline segment labels are visible. Determine QB.}{7}{\traceatlaspromptnormal}
\par\vspace{2pt}
\noindent
\traceatlascard{Coordinate Panels: Point Set Transform Match}{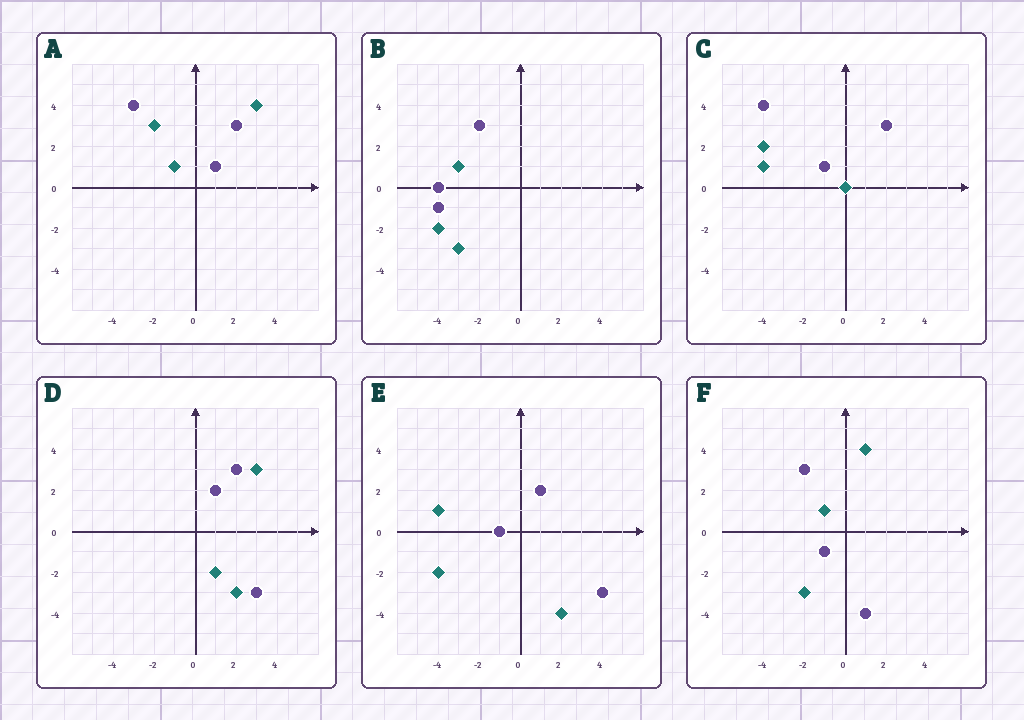}{The visual shows labeled coordinate panels, each containing a source point set and a candidate point set. Which labeled panel contains a source point set and its y-axis reflection?}{A}{\traceatlaspromptnormal}
\hfill
\traceatlascard{Coordinate Plane: Locus Point}{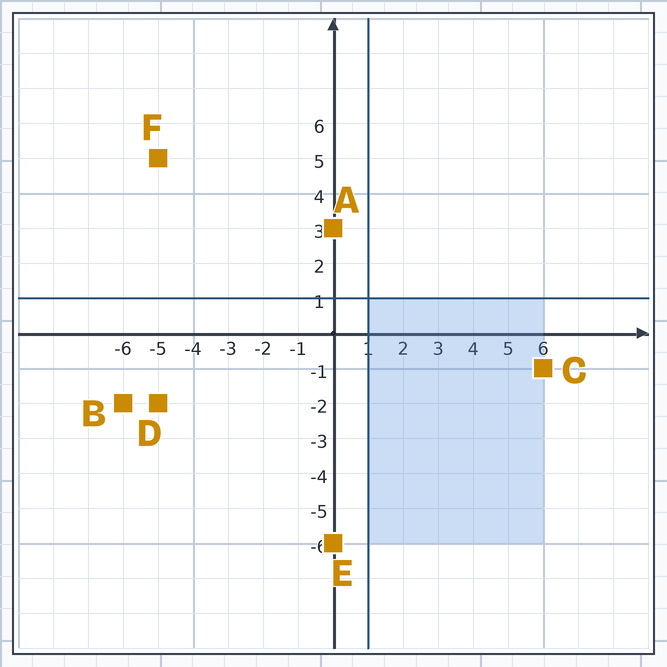}{The diagram shows a coordinate grid with one shaded locus region and lettered candidate points. Choose the lettered point inside the shaded region.}{C}{\traceatlaspromptnormal}
\hfill
\traceatlascard{Bearing Route: Endpoint Position}{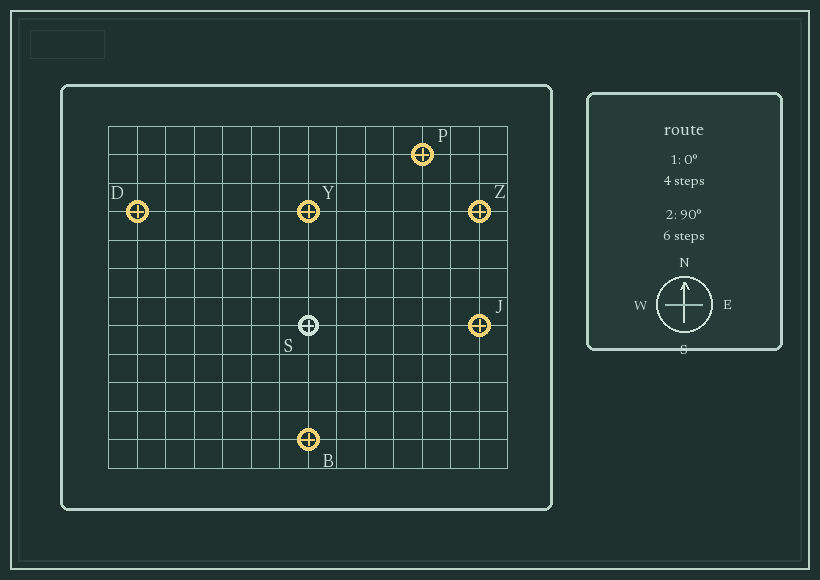}{The figure shows a compass-bearing route instruction panel, a graph-paper candidate grid where each square is one step, the start point marked S, and labeled candidate endpoint points, with bearings measured clockwise from north. Follow the two bearing instructions from S. Which candidate endpoint label is reached?}{Z}{\traceatlaspromptdense}
\caption{Representative geometry tasks.}
\label{fig:task-atlas-geometry}
\end{figure}
\clearpage

\begin{figure}[p]
\centering
\traceatlascard{Binary Tree: Local Relative Node}{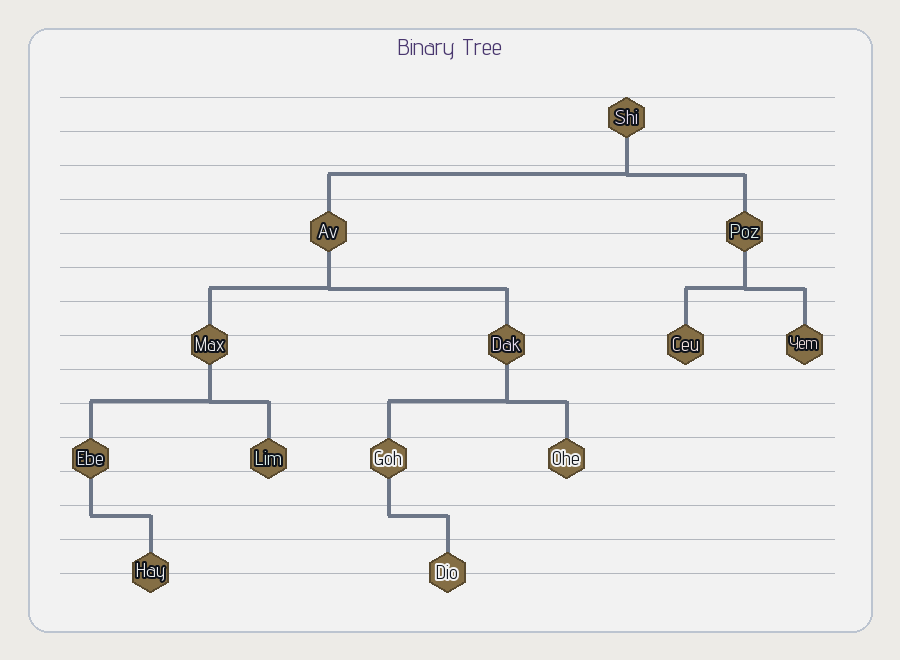}{The image shows a labeled binary tree with the root at the top and left/right children shown by position. Which node shares the same parent as "Max"?}{Dak}{\traceatlaspromptnormal}
\hfill
\traceatlascard{Adjacency: Undirected Component}{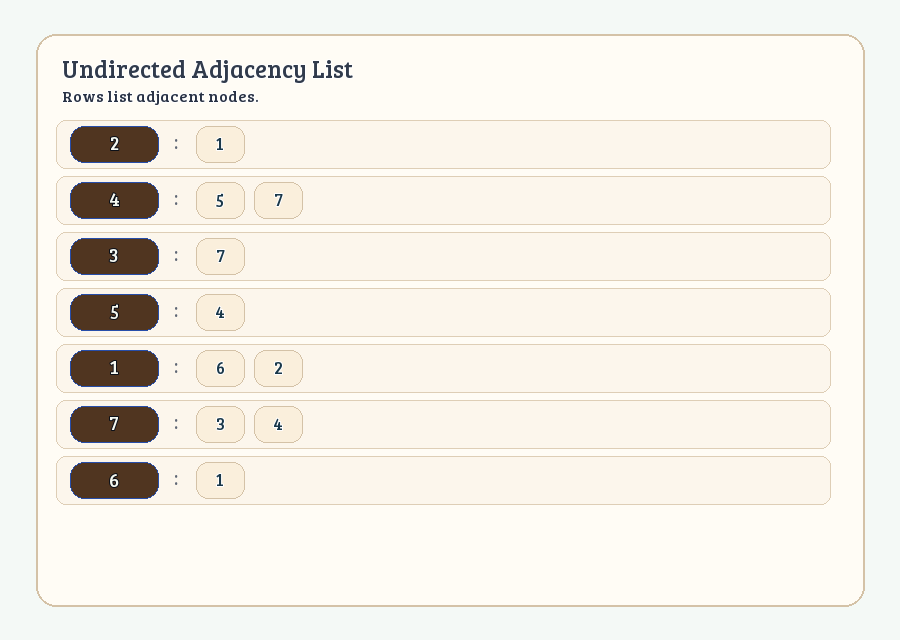}{You are shown an undirected graph as an adjacency list. Count the connected components in the undirected graph.}{2}{\traceatlaspromptnormal}
\hfill
\traceatlascard{Node Link: Bridge}{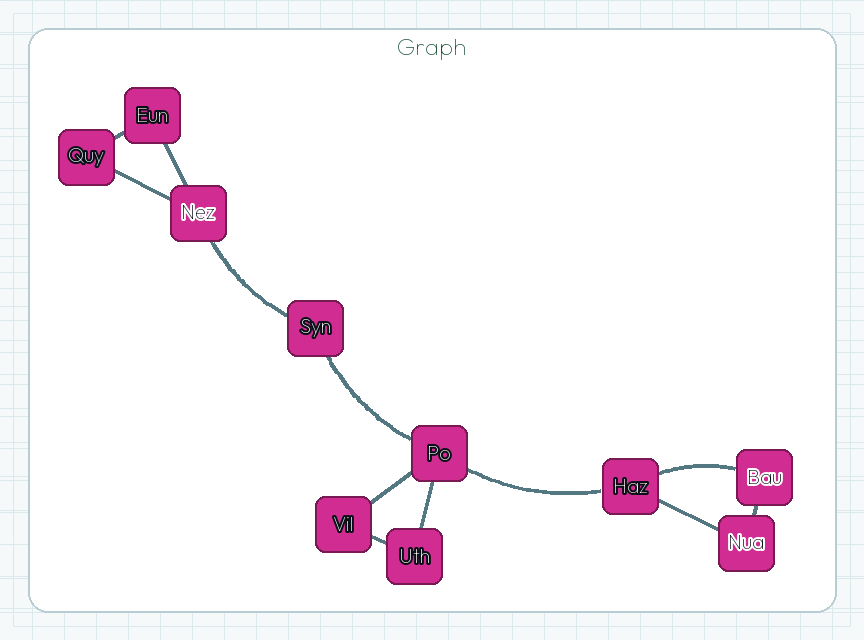}{The figure shows a labeled undirected graph. What is the number of bridges in the graph?}{3}{\traceatlaspromptnormal}
\par\vspace{2pt}
\noindent
\traceatlascard{Phylogeny Tree: MRCA Clade Membership}{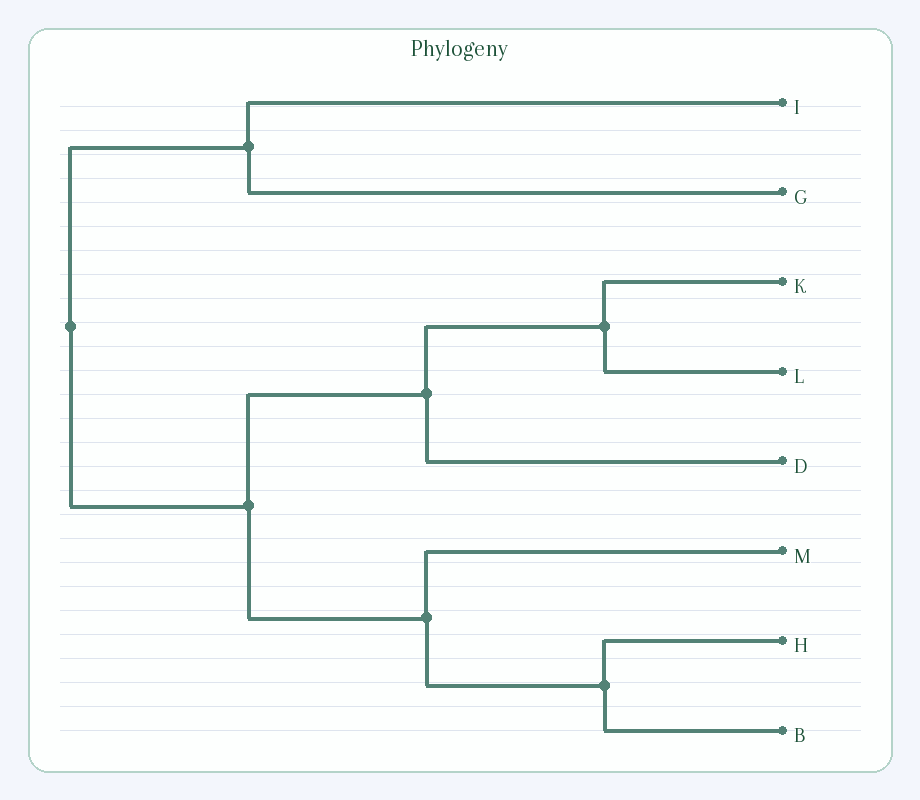}{The figure shows a rooted phylogeny cladogram with labeled taxa. Count the terminal taxa in the MRCA clade for "B" and "L".}{6}{\traceatlaspromptnormal}
\hfill
\traceatlascard{Graph Options: Same Structure}{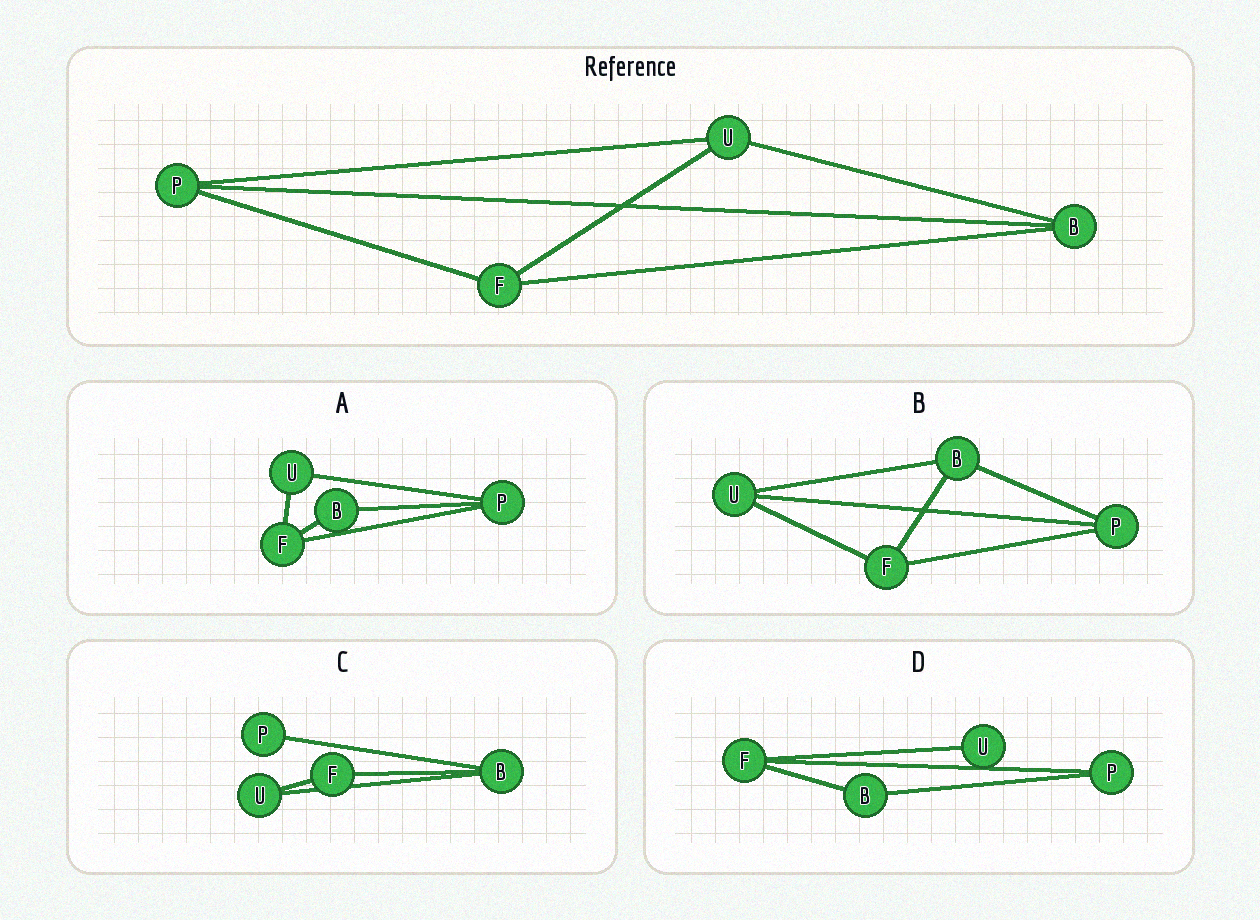}{You are shown a Reference labeled graph above four labeled graph options. Which option has the same labeled graph structure as the Reference, ignoring layout?}{B}{\traceatlaspromptnormal}
\hfill
\traceatlascard{Node Link: Reachable Count After Edge Edit}{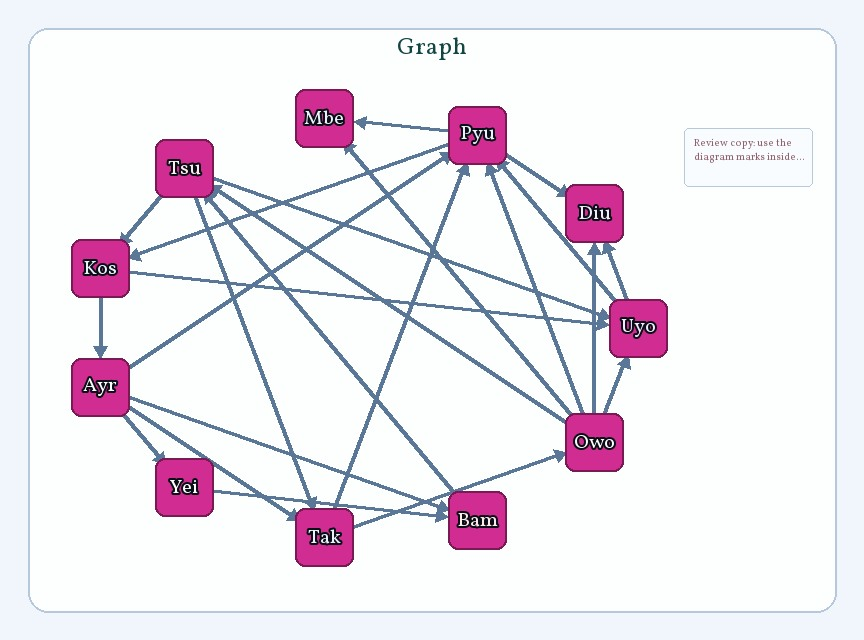}{The diagram shows a labeled directed graph. Remove the arrow from node "Uyo" to node "Pyu". Following arrow directions from "Uyo", how many nodes would be reachable?}{2}{\traceatlaspromptnormal}
\par\vspace{2pt}
\noindent
\traceatlascard{Automaton: State After Input}{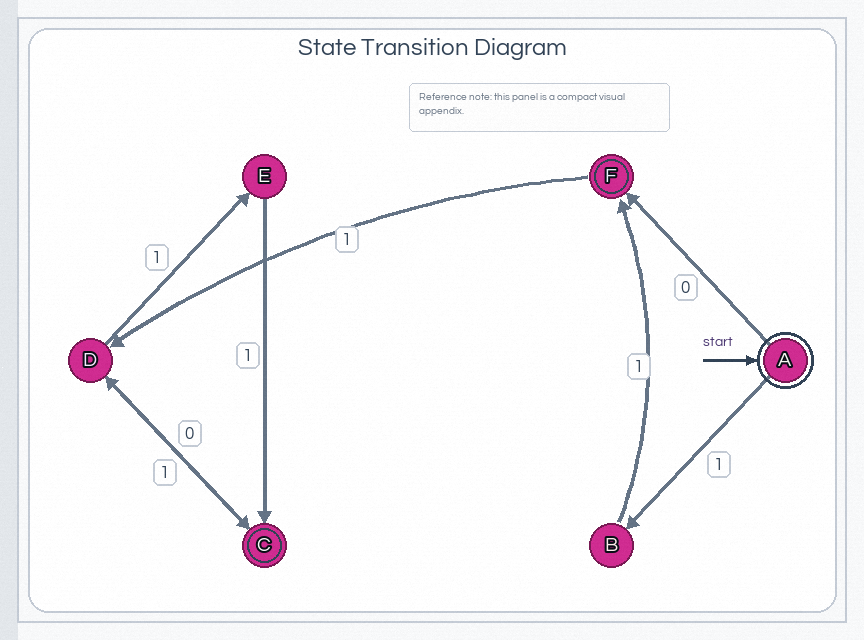}{The figure shows a deterministic state-transition diagram with a start arrow, double-ring accepting states, and transition labels. Read input "1110" from the start state. What state label is reached after 2 symbols?}{F}{\traceatlaspromptnormal}
\hfill
\traceatlascard{Pedigree Chart: Relationship}{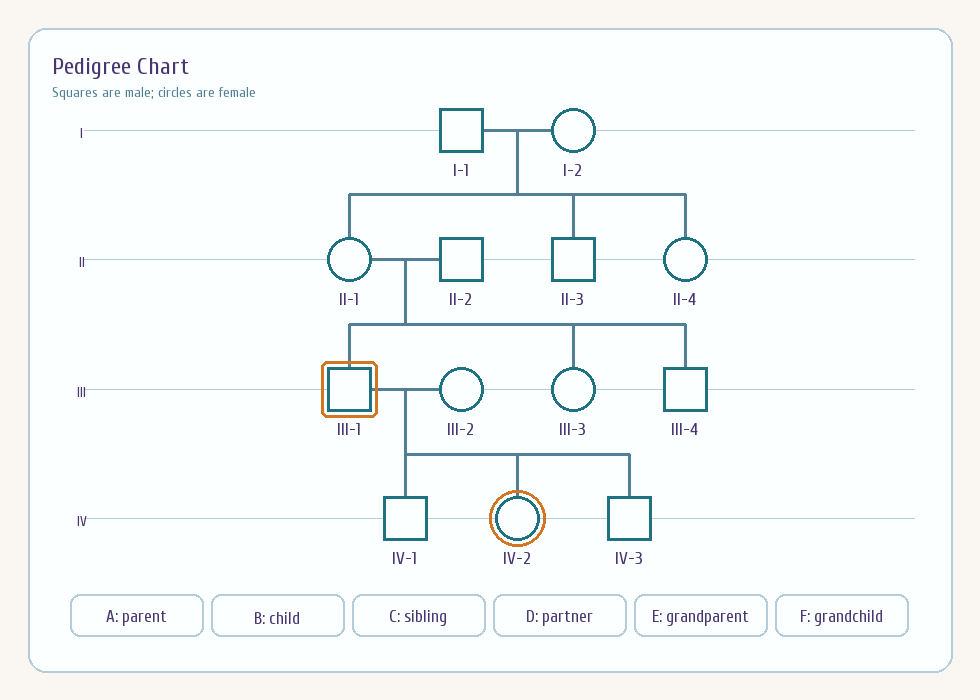}{The diagram shows a labeled pedigree chart with family connections and answer options. Read the pedigree chart. Which option is III-1 relative to IV-2?}{A}{\traceatlaspromptnormal}
\hfill
\traceatlascard{Flow Network: Max Flow}{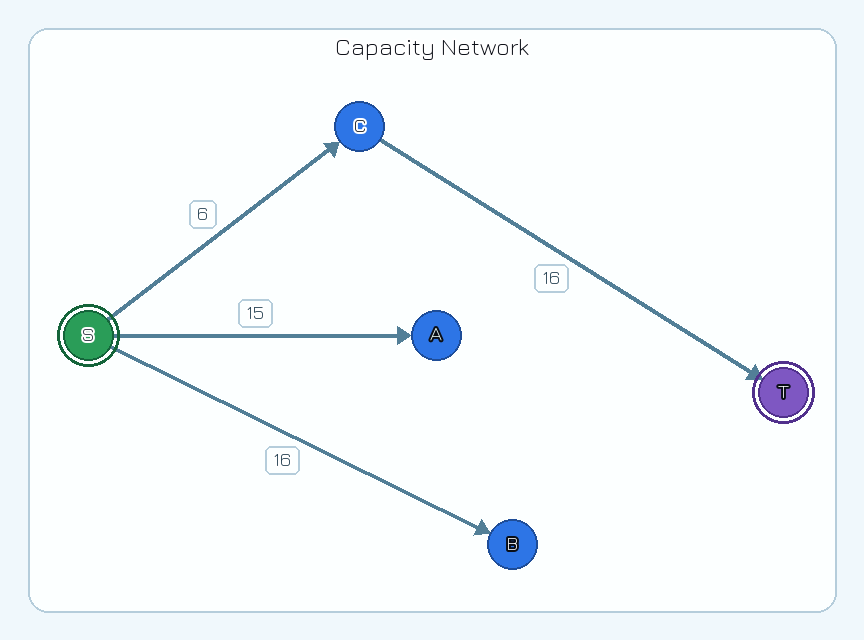}{You are shown a directed capacity network with source S and sink T. Find the max-flow value from source S to sink T in this capacity network.}{6}{\traceatlaspromptnormal}
\par\vspace{2pt}
\noindent
\traceatlascard{Pipe Network: Pipe Reachable Junction}{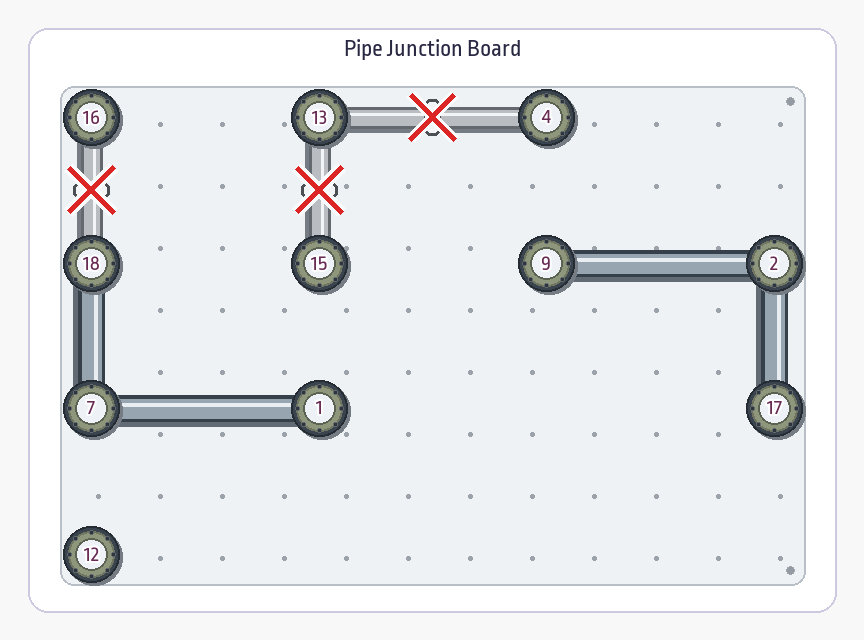}{The visual shows a labeled pipe-junction network with open pipes and blocked pipes (blocked pipes are marked with a red X). From 2, how many junctions are in the same open-pipe connected region, including the start junction?}{3}{\traceatlaspromptdense}
\hfill
\traceatlascard{Binary Tree: Lowest Common Ancestor}{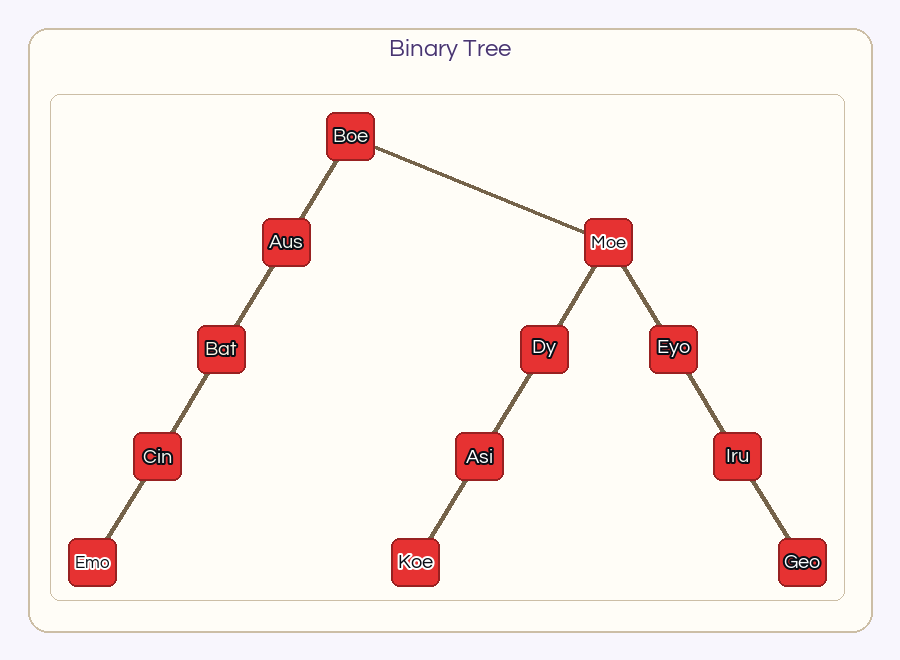}{This diagram shows a labeled binary tree with the root at the top and left/right children shown by position. Trace upward from "Eyo" and "Koe". What is the label of their lowest common ancestor?}{Moe}{\traceatlaspromptnormal}
\hfill
\traceatlascard{Metro: Shortest Path Length}{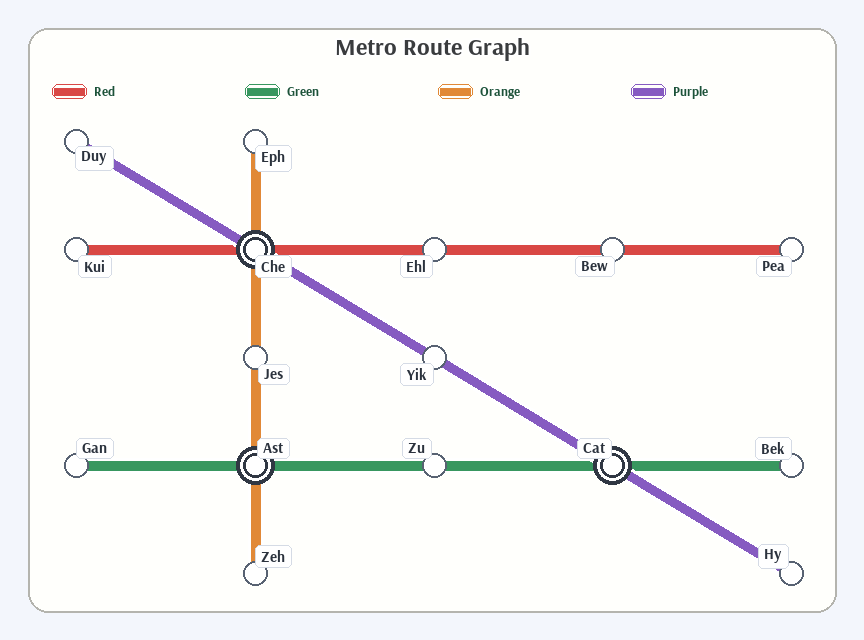}{The figure shows a labeled metro route map with colored routes and stations. What is the shortest-path length in route segments from station "Pea" to station "Bek"?}{6}{\traceatlaspromptnormal}
\caption{Representative graph tasks.}
\label{fig:task-atlas-graph}
\end{figure}
\clearpage

\begin{figure}[p]
\centering
\traceatlascard{Named Path: Immediate Successor}{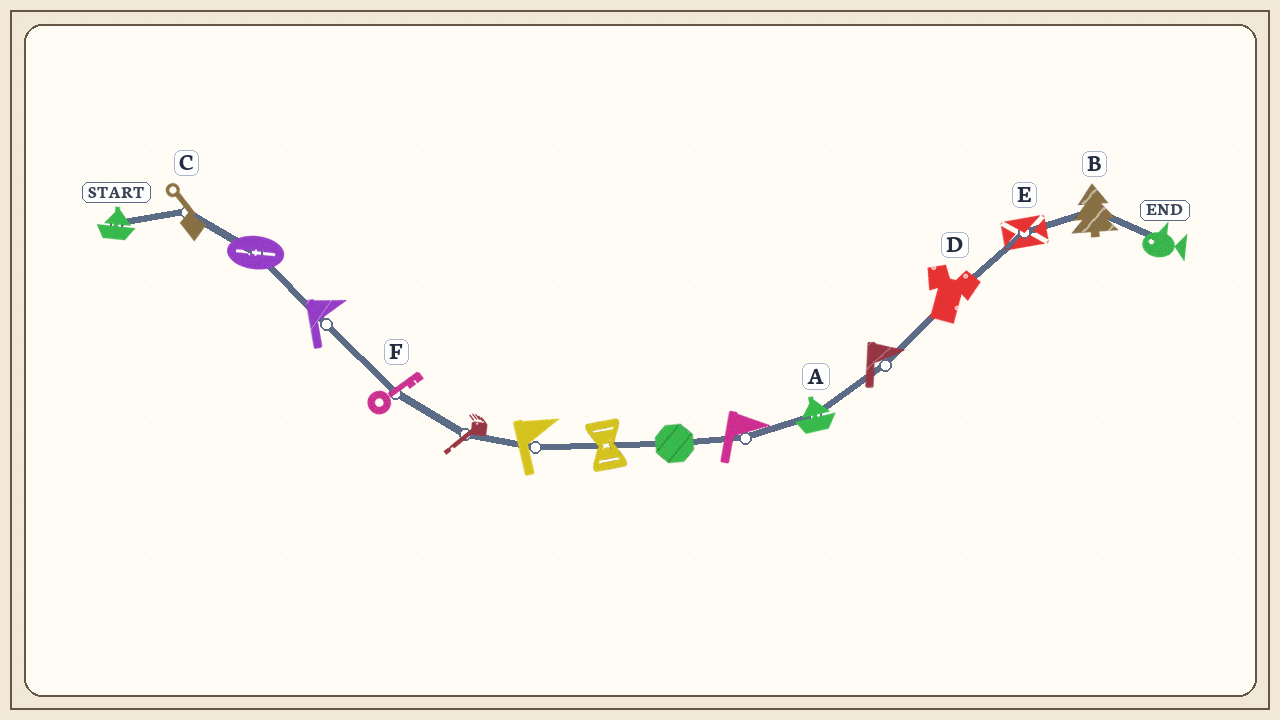}{The picture shows a START-to-END icon path with six option icons labeled A-F and other icons. Which labeled icon comes immediately after the first "flag" icon along the path from START to END?}{F}{\traceatlaspromptnormal}
\hfill
\traceatlascard{Wallpaper Panels: Same Pattern as Reference}{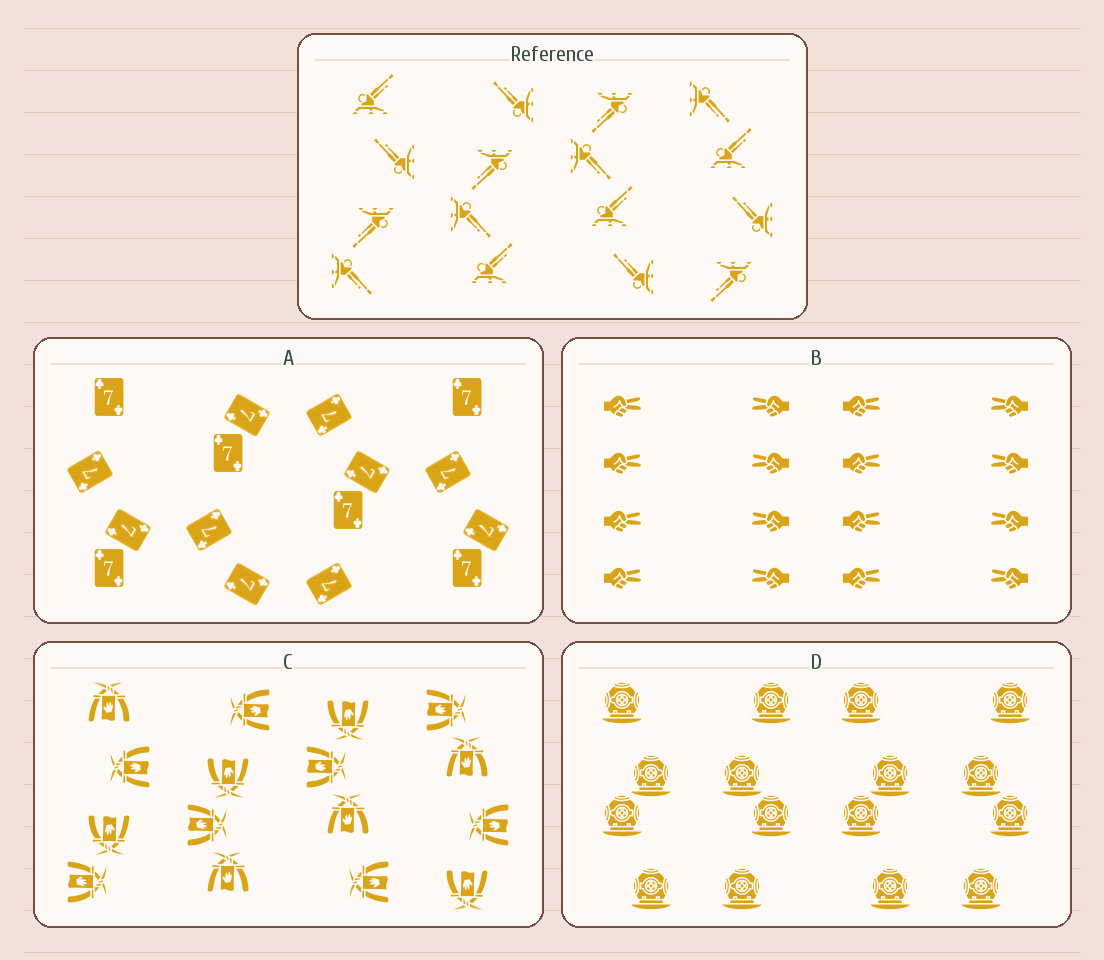}{The image shows one Reference wallpaper panel above four labeled wallpaper panels filled with repeated icon motifs. Select the labeled panel whose wallpaper pattern matches the Reference panel.}{C}{\traceatlaspromptnormal}
\hfill
\traceatlascard{Sequence Strip: Size Progression Completion}{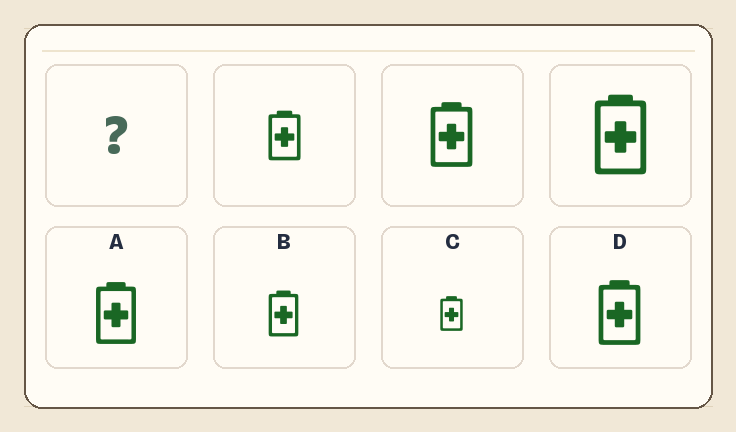}{The image shows a top row of four boxed sequence cells with one question-mark box and a bottom row of four labeled option boxes. Which labeled option should replace the question mark in the size sequence?}{C}{\traceatlaspromptnormal}
\par\vspace{2pt}
\noindent
\traceatlascard{Icon Cutout: Partial Match}{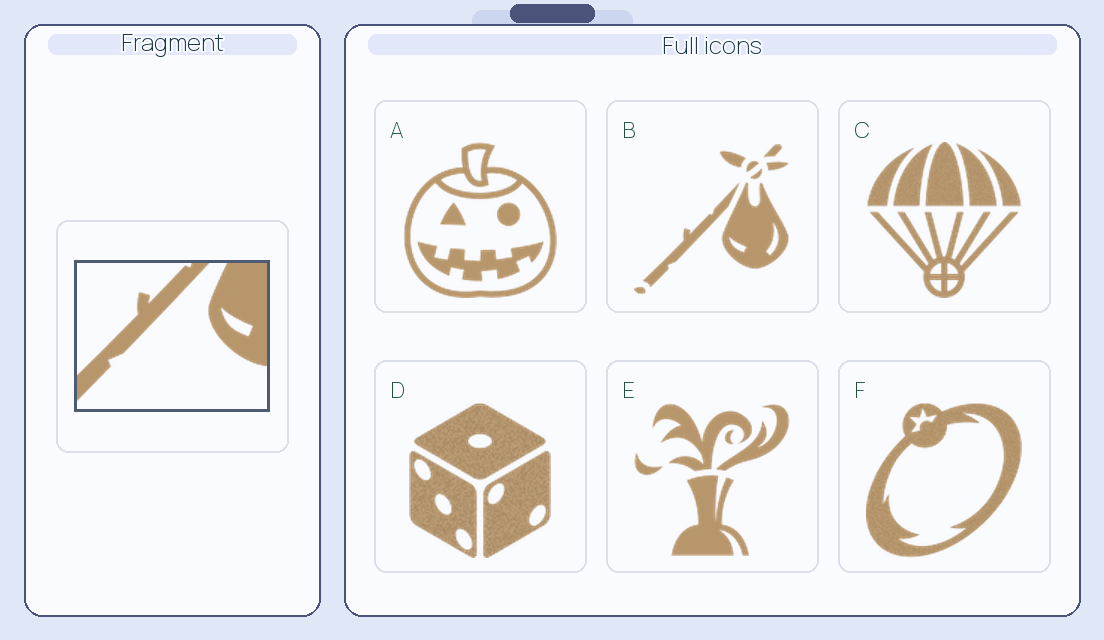}{The image shows a partial icon fragment beside six labeled full-icon options. Which labeled full icon does the partial fragment come from?}{B}{\traceatlaspromptnormal}
\hfill
\traceatlascard{Single Transform Options: Inverse Geometric Transform Source}{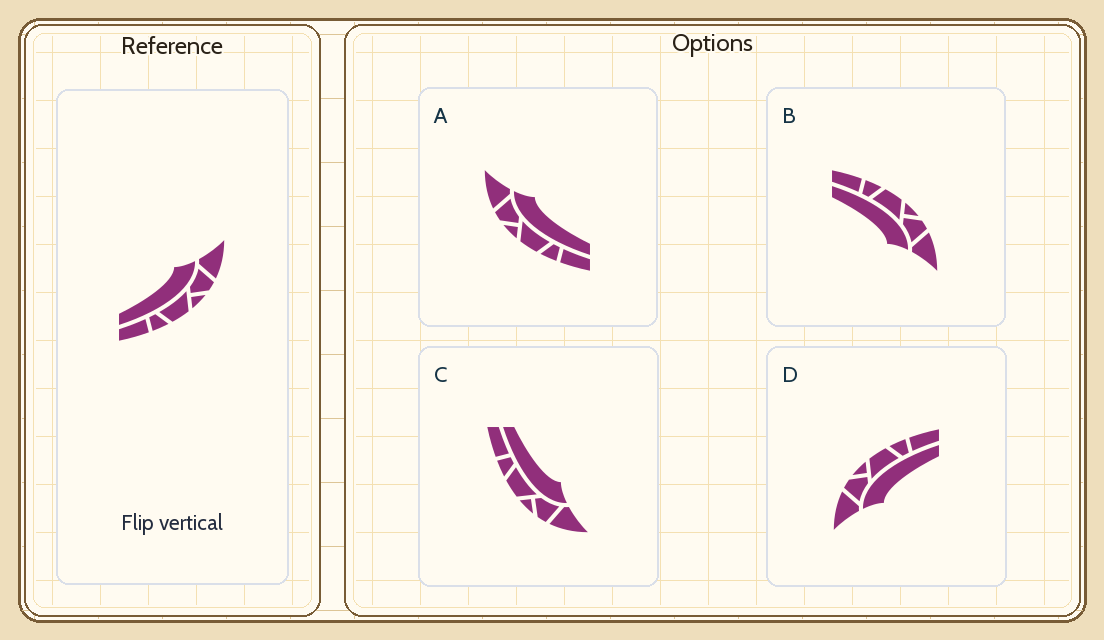}{The scene shows a transformed Reference icon with a transform cue and four labeled source-option icons. Select the option that would become the Reference icon after a vertical flip.}{B}{\traceatlaspromptnormal}
\hfill
\traceatlascard{Named Field: Reference Distance Rank}{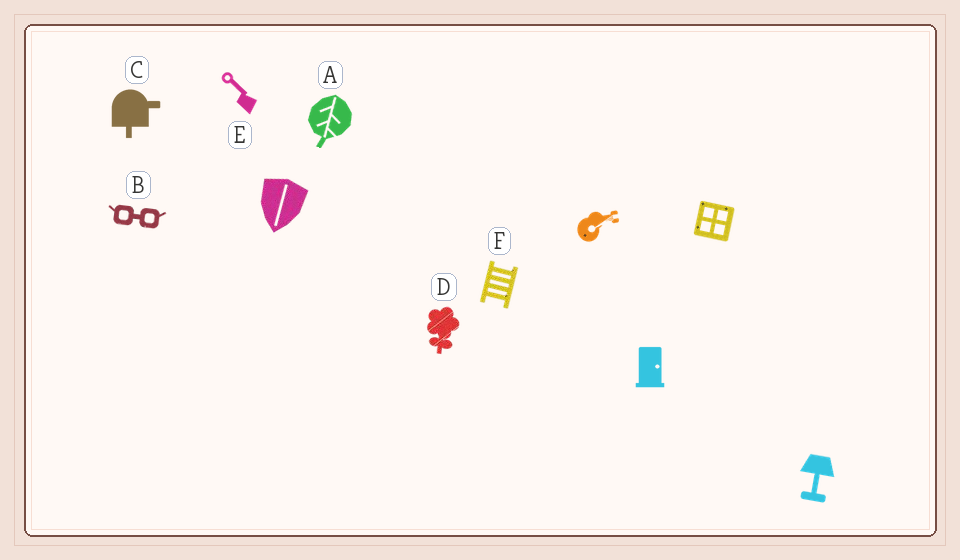}{The picture shows a panel with one reference icon, six option icons labeled A-F, and other icons. Which labeled icon is second closest to the magenta shield icon?}{E}{\traceatlaspromptnormal}
\par\vspace{2pt}
\noindent
\traceatlascard{Icon Field: Frequency Extreme Type}{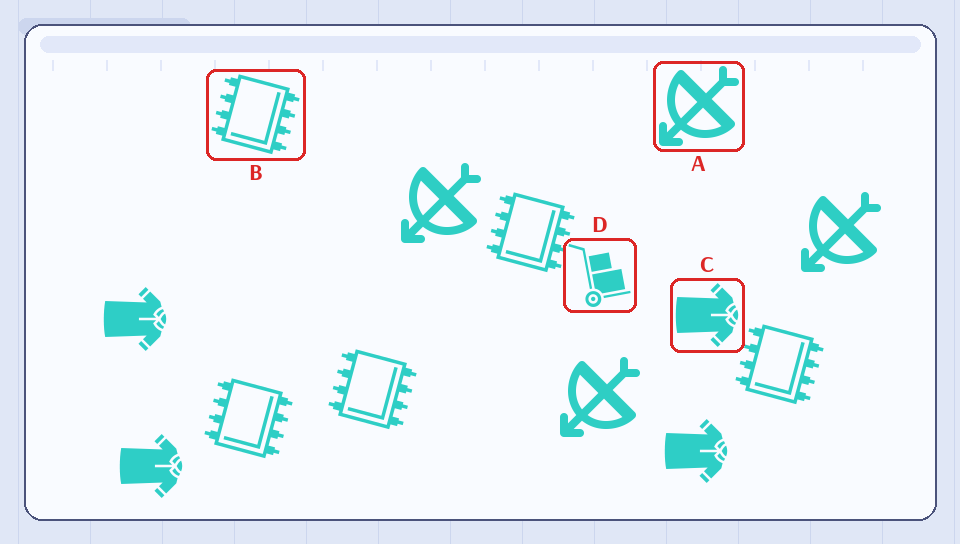}{The picture shows one panel of assorted icons. Which marked icon type appears most often in the image?}{B}{\traceatlaspromptnormal}
\hfill
\traceatlascard{Icon Grid: Distinct Color}{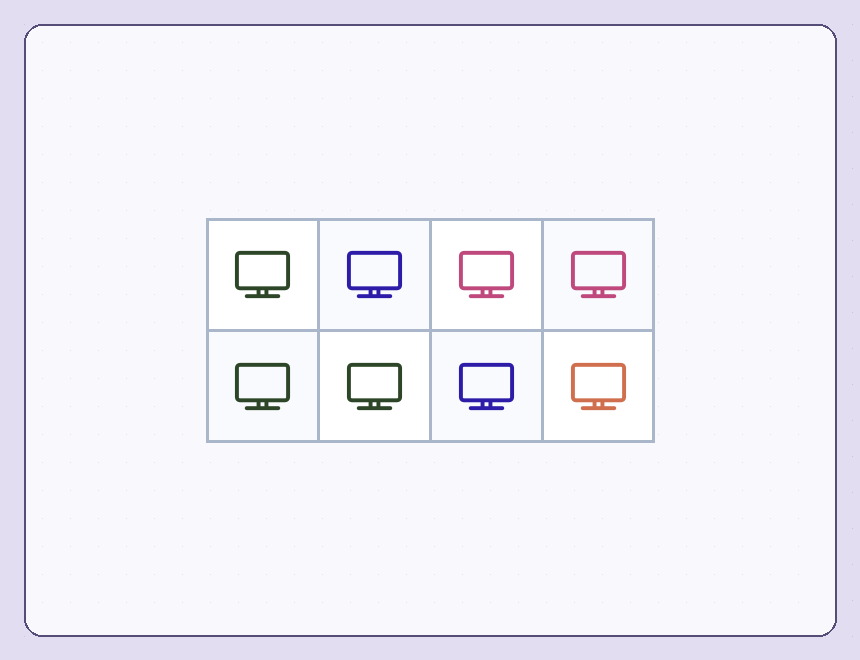}{The scene shows a visible grid of icon cells. How many unique icon colors appear in the grid?}{4}{\traceatlaspromptnormal}
\hfill
\traceatlascard{Paired Canvas: Rotation Change}{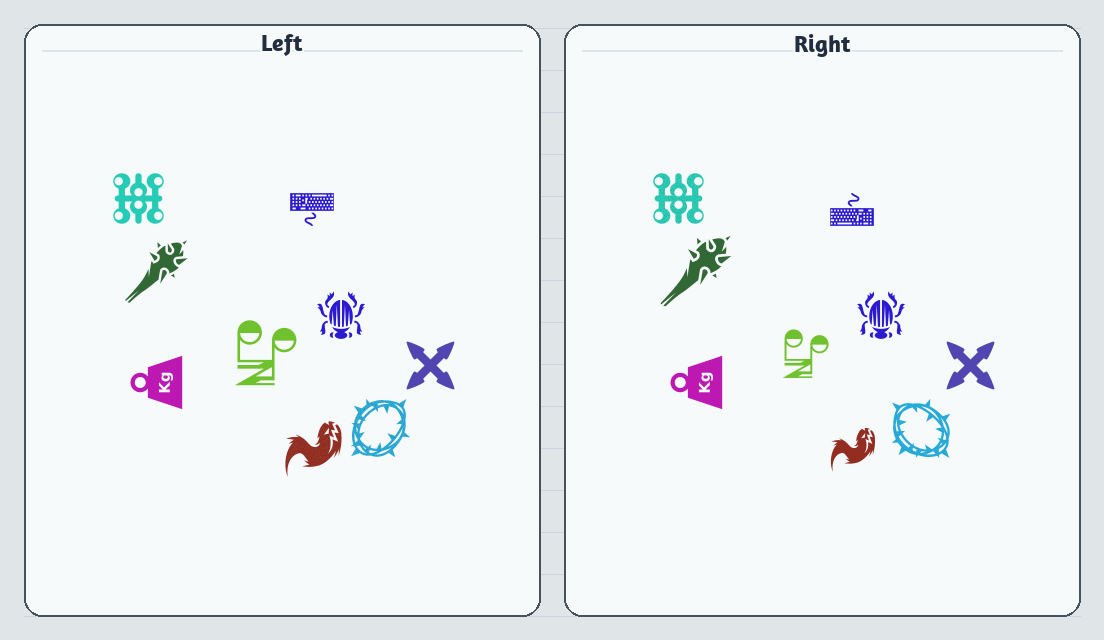}{The picture shows two icon panels labeled Left and Right with corresponding icons at matching positions. How many Right-panel icons changed rotation compared with the corresponding icon in the Left panel?}{4}{\traceatlaspromptnormal}
\par\vspace{2pt}
\noindent
\traceatlascard{Venn Field: Same Region as Reference}{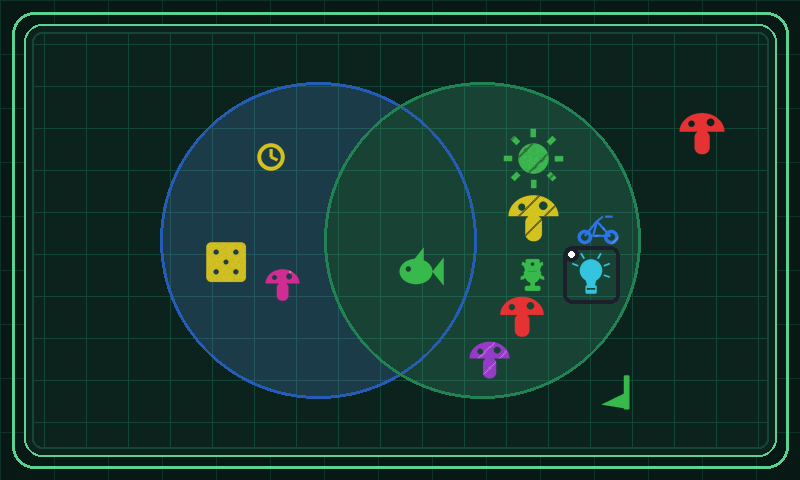}{The scene shows icons and two overlapping marked circles. By icon center, the possible regions are left-only, right-only, both-circle overlap, and outside both circles. Find the number of "mushroom" icons in the marked reference icon's region.}{3}{\traceatlaspromptdense}
\hfill
\traceatlascard{Overlap Grid: Occlusion Order}{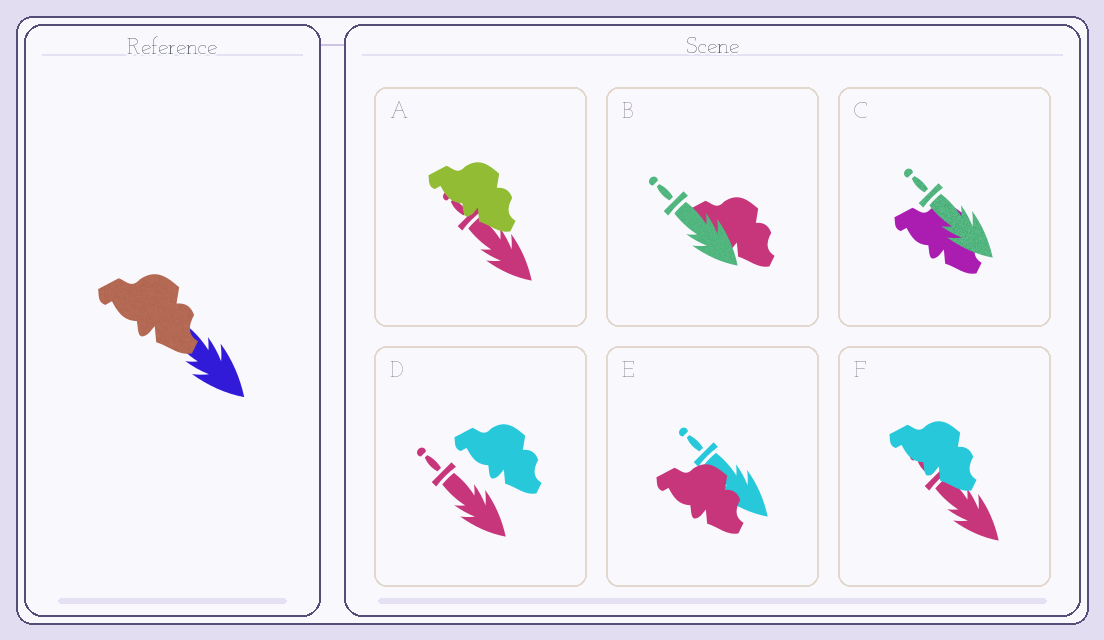}{The image shows a Reference cell beside a labeled Scene grid. How many labeled Scene cells match the Reference front-to-back order? Front-to-back order means which icon is on top.}{3}{\traceatlaspromptnormal}
\hfill
\traceatlascard{Named Grid: Scoped Attribute}{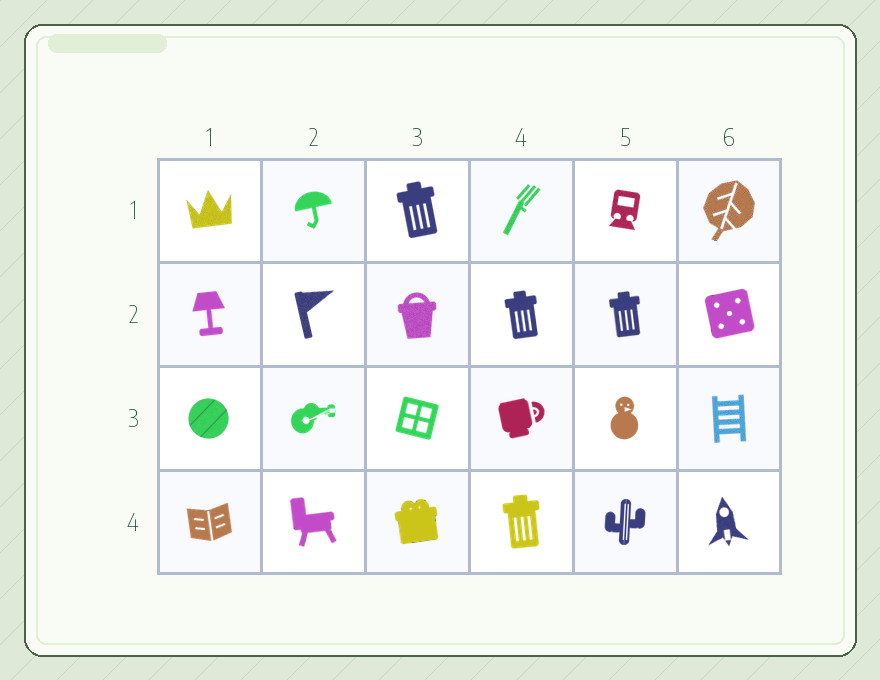}{The image shows a numbered grid with one icon in each cell. How many "trash can" icons are in row 4?}{1}{\traceatlaspromptnormal}
\caption{Representative icon tasks.}
\label{fig:task-atlas-icons}
\end{figure}
\clearpage

\begin{figure}[p]
\centering
\traceatlascard{Library: Filtered Book in Section}{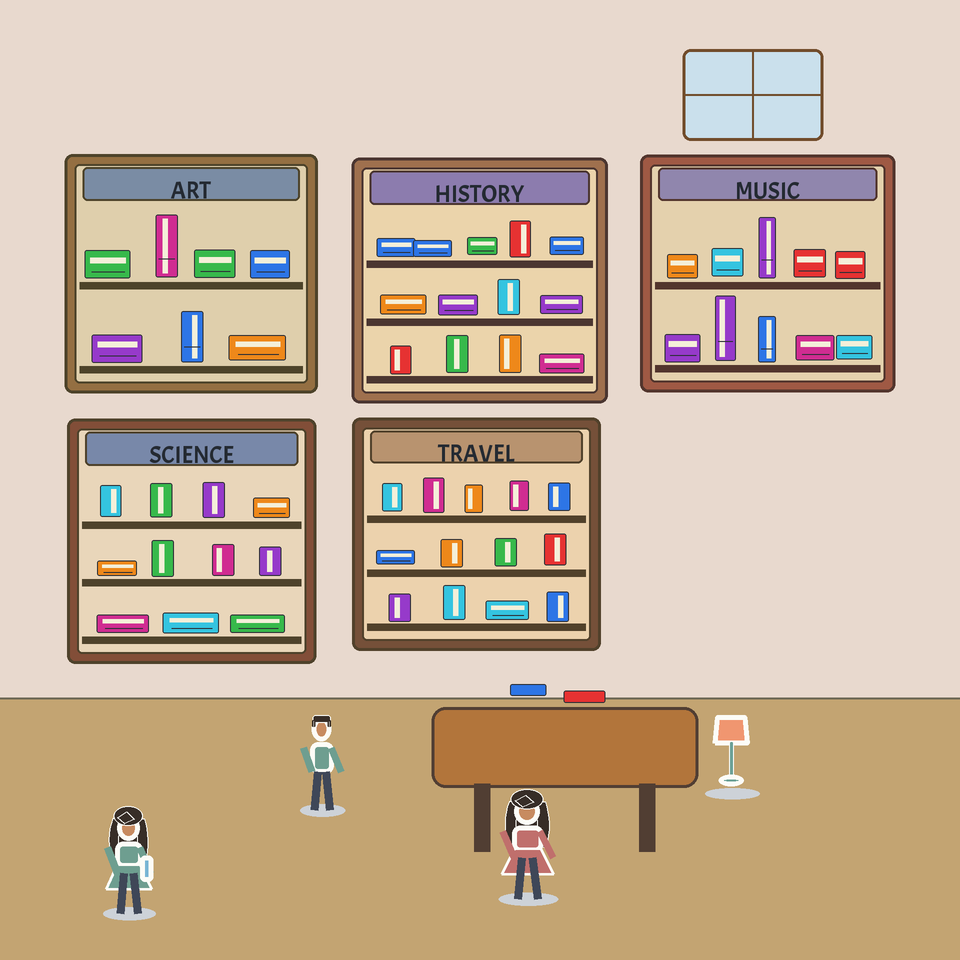}{The scene shows a library room containing 5 labeled book sections. How many upright books are in the Music section?}{3}{\traceatlaspromptnormal}
\hfill
\traceatlascard{Indoor Room: Swapped Tile Pair}{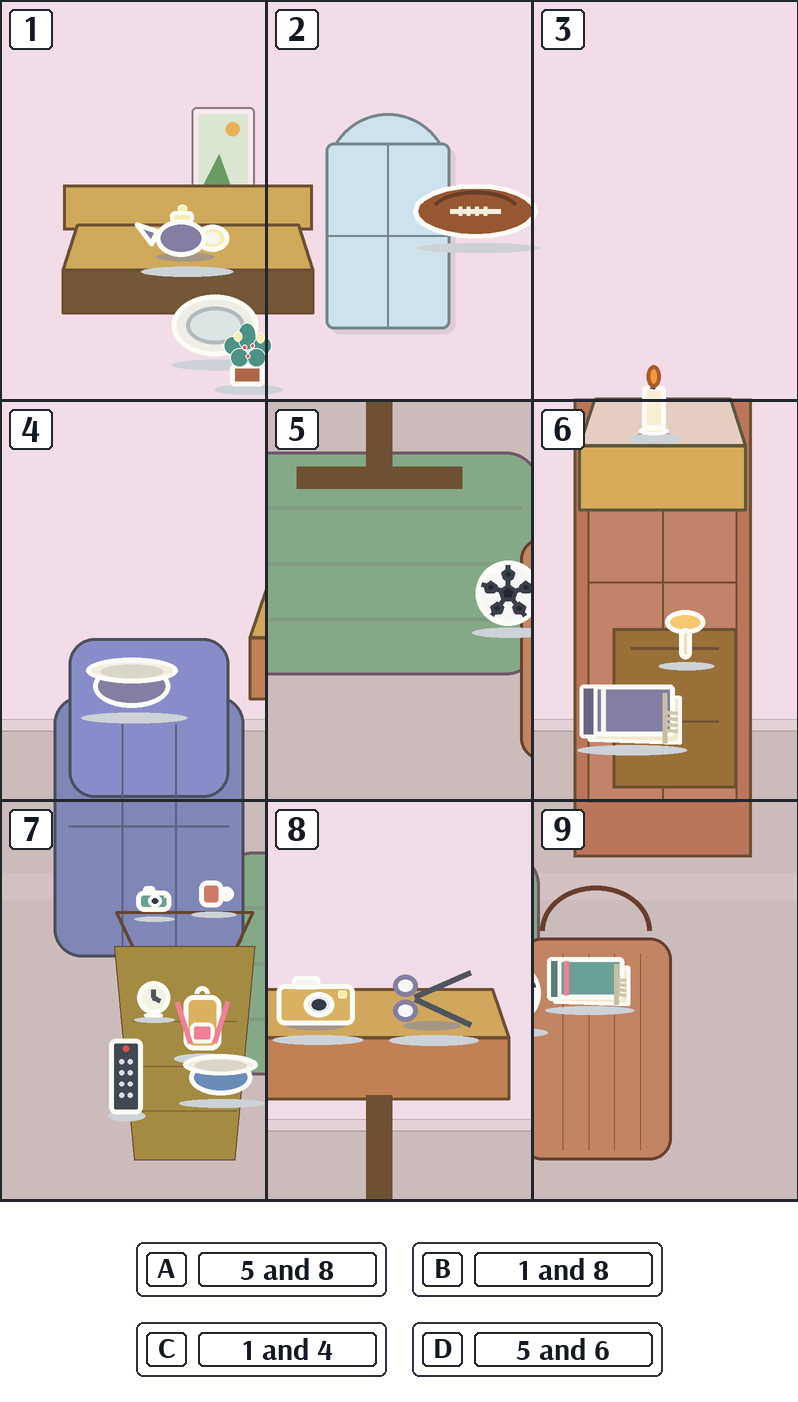}{The scene shows a bedroom with small objects. Find the pair of numbered cells that were swapped in the room image.}{A}{\traceatlaspromptnormal}
\hfill
\traceatlascard{RPG Dungeon: Safe Reachable Chest}{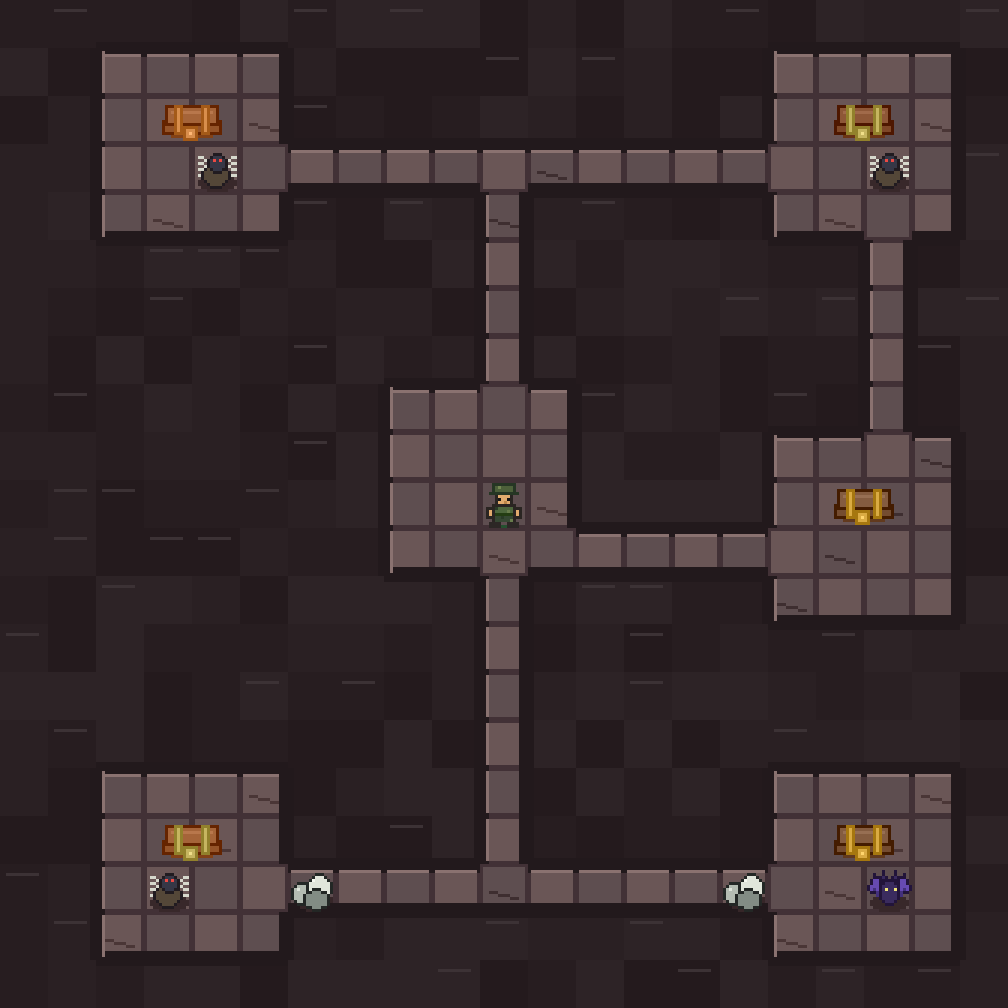}{The scene shows a top-down RPG-style dungeon map with walkable floor, blocked passages, treasure chests, and monsters inside some chambers. Starting from the player, how many treasure chests can be reached without crossing boulders, excluding any chest in a chamber that contains a monster?}{1}{\traceatlaspromptdense}
\par\vspace{2pt}
\noindent
\traceatlascard{Isometric Farmstead: Farmer Same Level Tile}{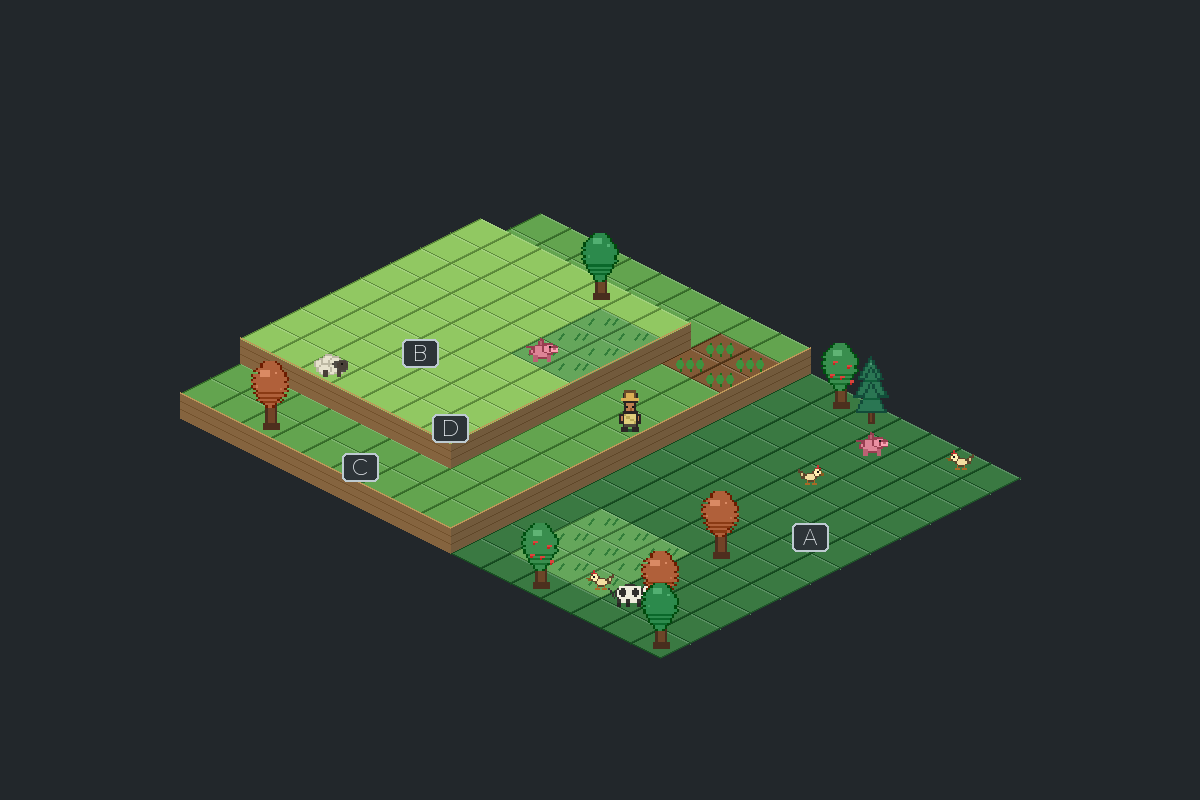}{This illustration shows a farmstead drawn on isometric tiles, with a farmer and lettered candidate ground tiles. Select the lettered farm tile whose ground level matches the farmer's tile.}{C}{\traceatlaspromptnormal}
\hfill
\traceatlascard{Park Playground: Missing Patch}{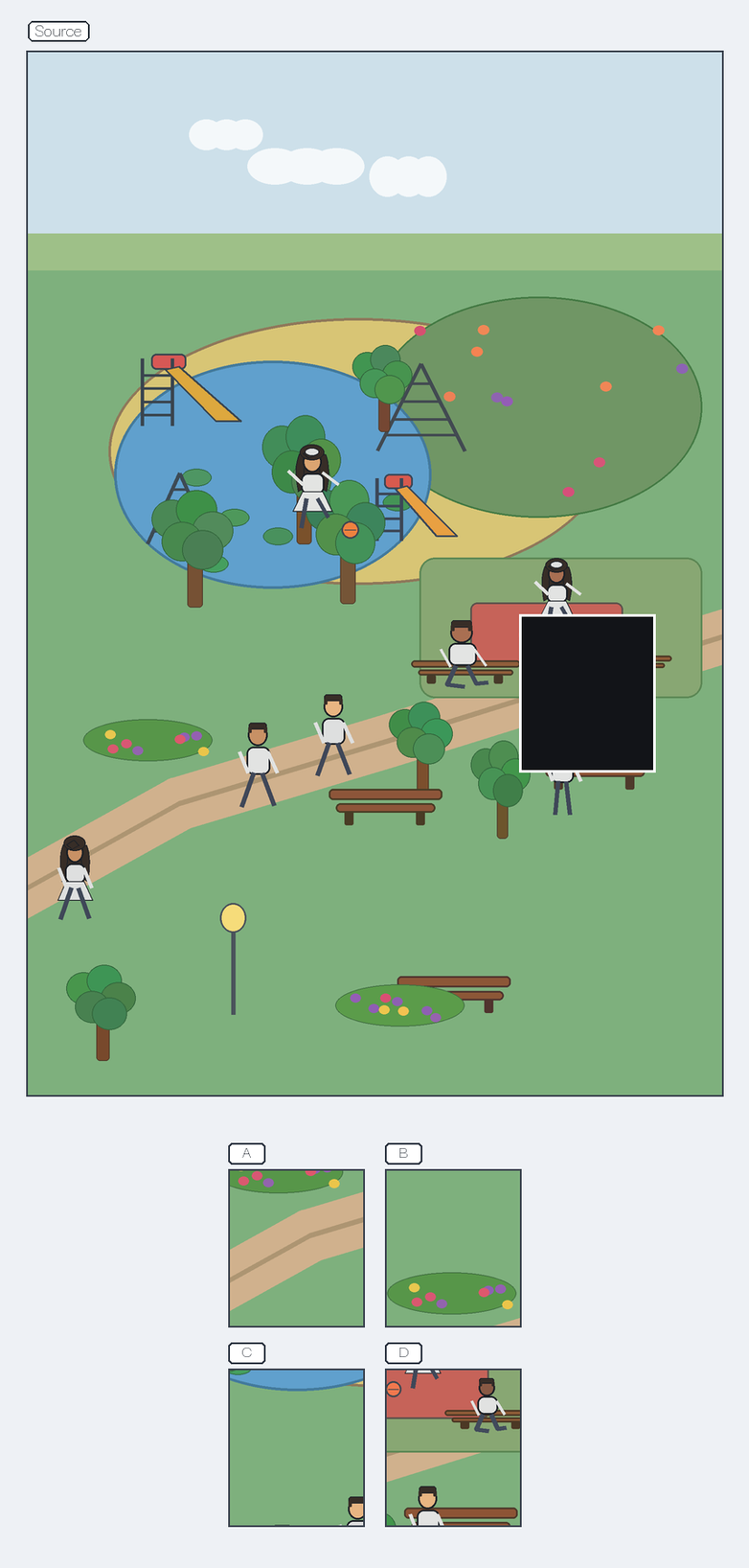}{The picture shows a synthetic park playground. Select the option that fills the missing patch exactly.}{D}{\traceatlaspromptnormal}
\hfill
\traceatlascard{Pixel Village: Rotated Tile}{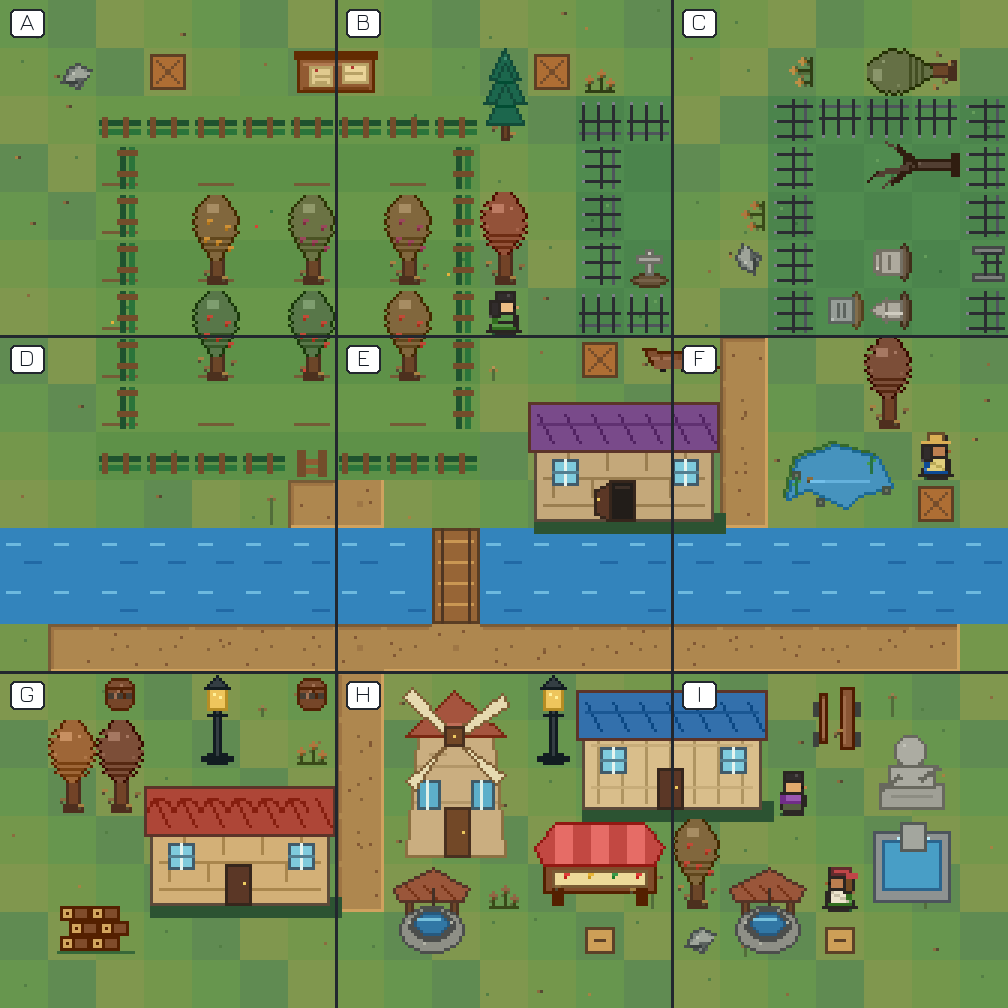}{The canvas contains a top-down pixel village scene. Which letter marks the tile with the incorrect orientation?}{C}{\traceatlaspromptnormal}
\par\vspace{2pt}
\noindent
\traceatlascard{RPG Tactical Map: Water-Barrier Unreachable Tile}{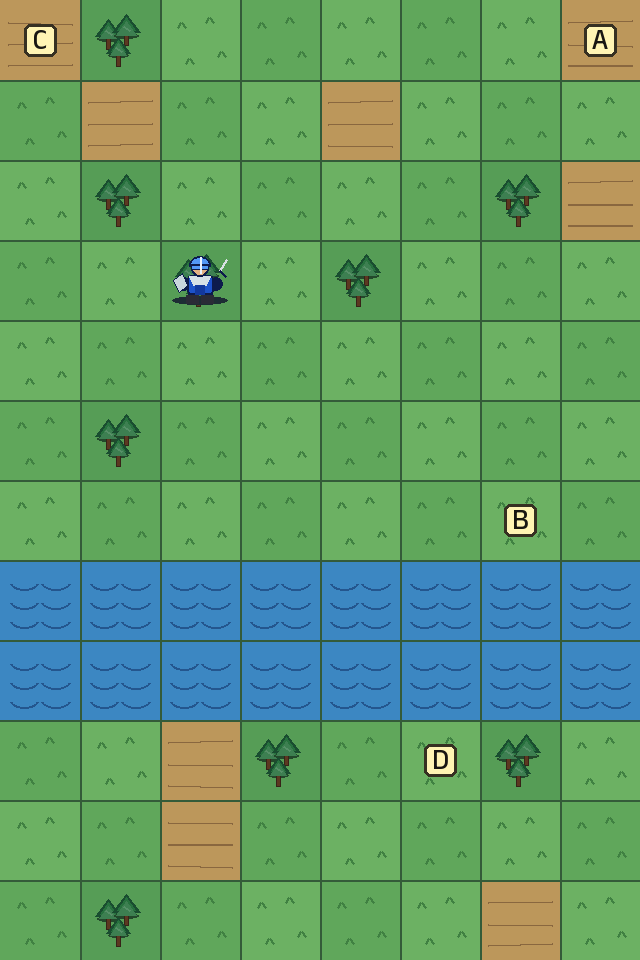}{Water tiles form a barrier and cannot be crossed; every other tile can. The blue unit moves only up, down, left, or right. Which candidate letter marks a tile the blue unit cannot reach?}{D}{\traceatlaspromptdense}
\hfill
\traceatlascard{Isometric Quarry: Highest Terrain Tile}{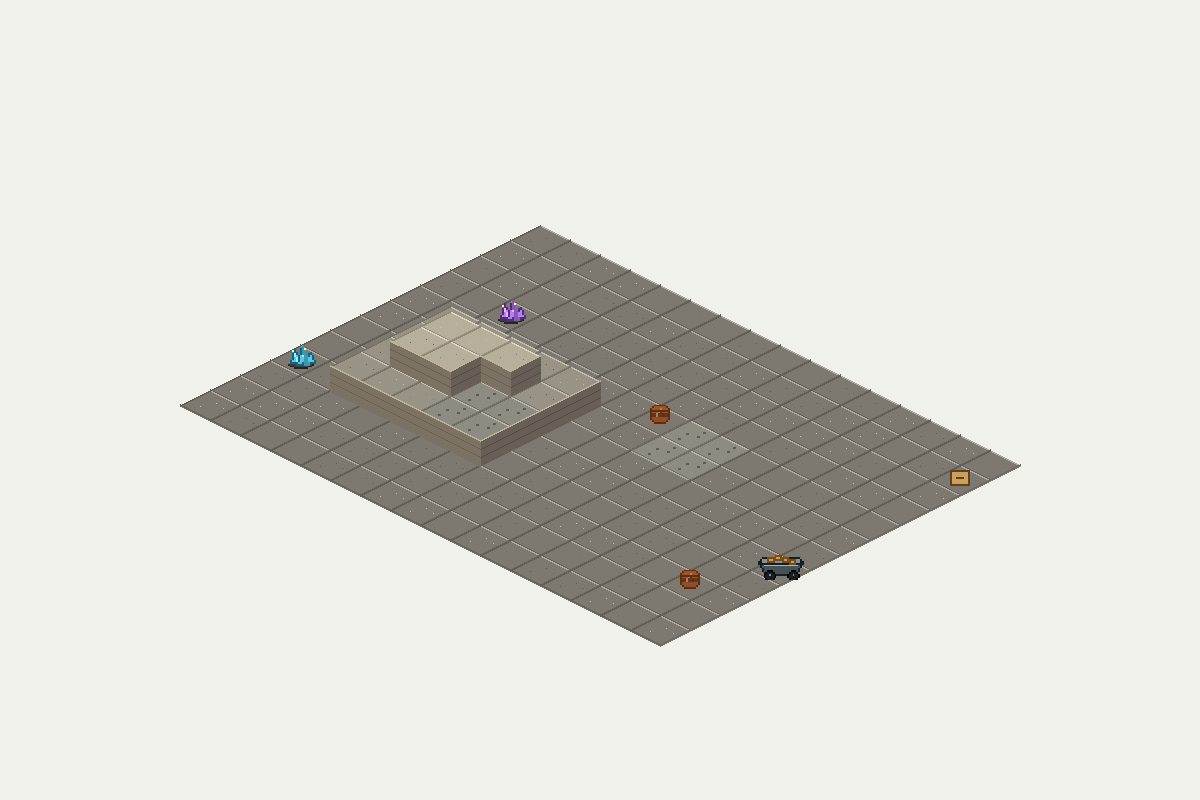}{The scene shows a pixel-art isometric quarry with a clean highest terrace and lower surrounding rock terrain. Count only the visible top-surface tiles on the highest elevated quarry layer.}{5}{\traceatlaspromptnormal}
\hfill
\traceatlascard{Environment: Rotated Tile}{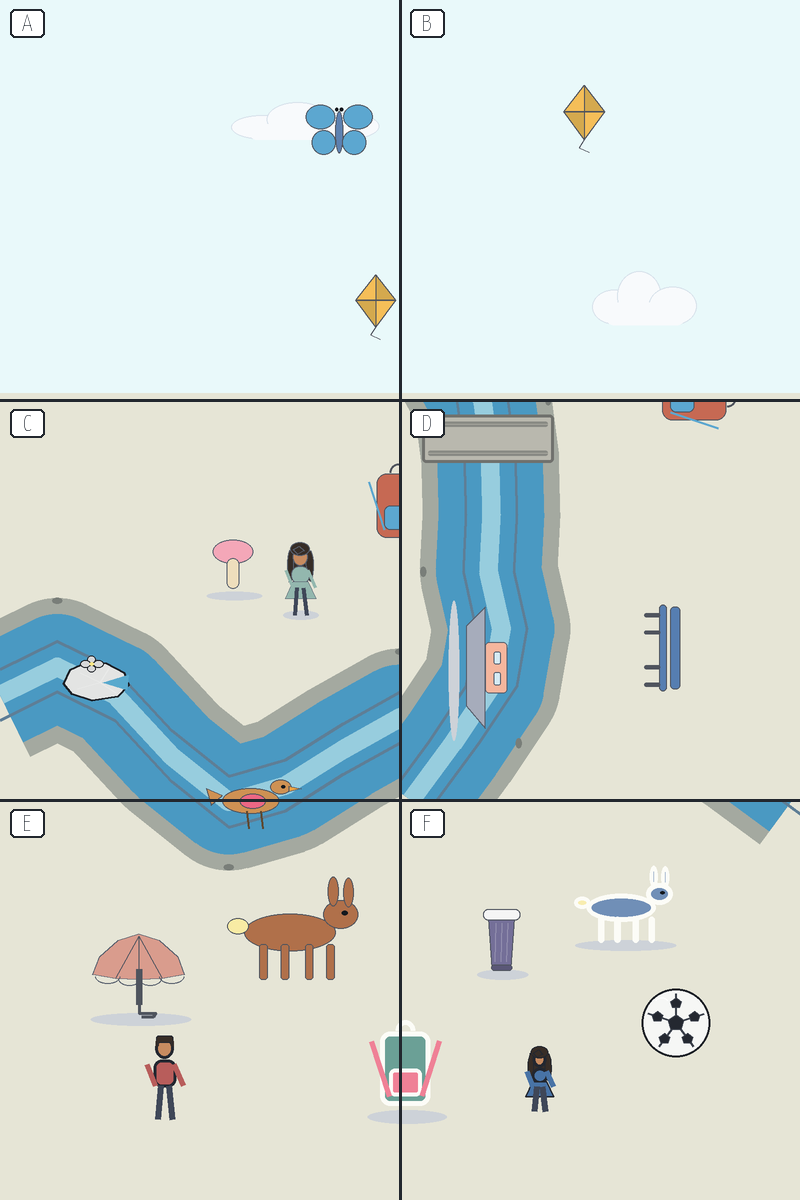}{The image shows a meadow river setting with illustrated foreground objects and visible environmental features. One of the lettered tiles is rotated relative to the rest of the environment image. Which tile is it?}{D}{\traceatlaspromptnormal}
\par\vspace{2pt}
\noindent
\traceatlascard{RPG House: Adjacent Room}{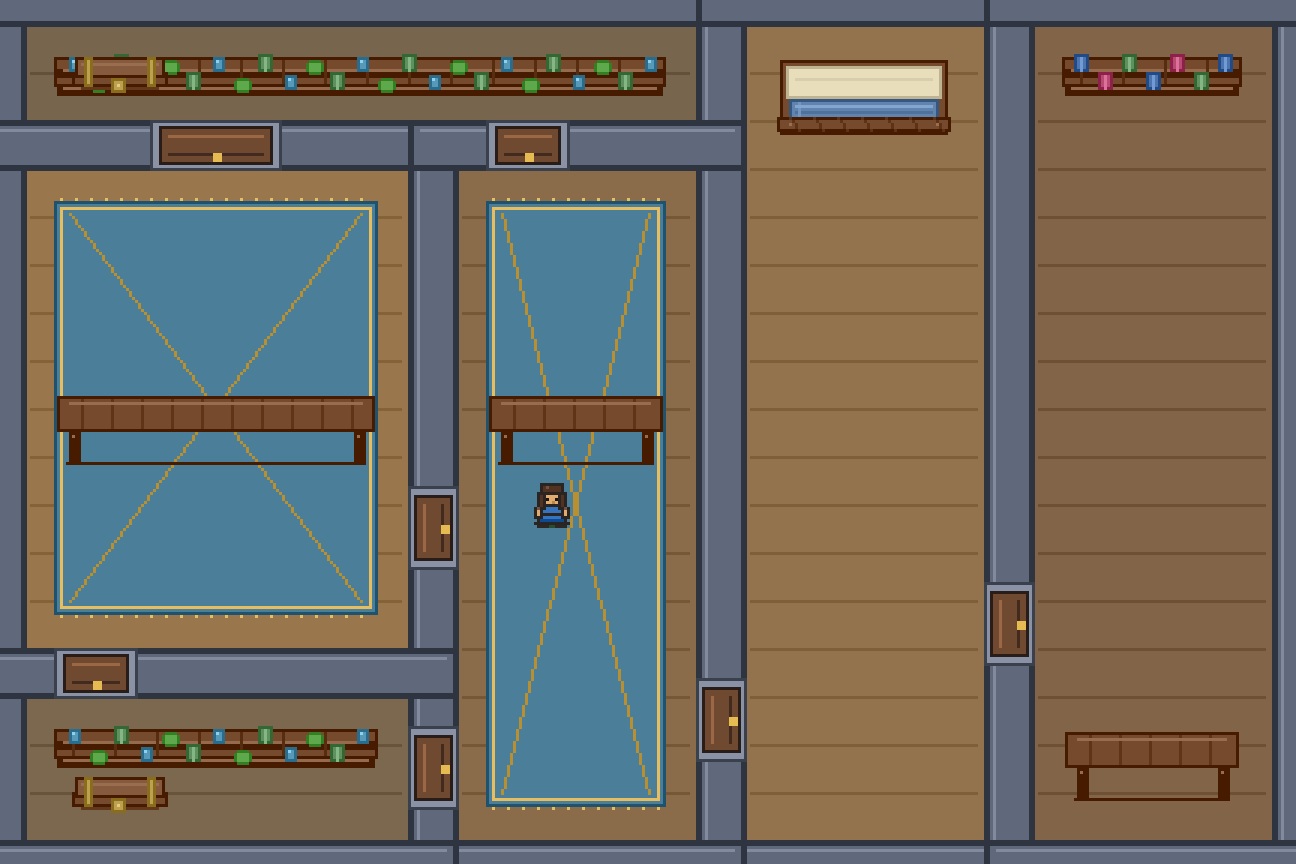}{The canvas contains a top-down RPG house map with enclosed rooms, walls, and doorways. How many rooms are immediately connected to the room with the player by a doorway?}{4}{\traceatlaspromptnormal}
\hfill
\traceatlascard{Isometric Harbor: Boat Side}{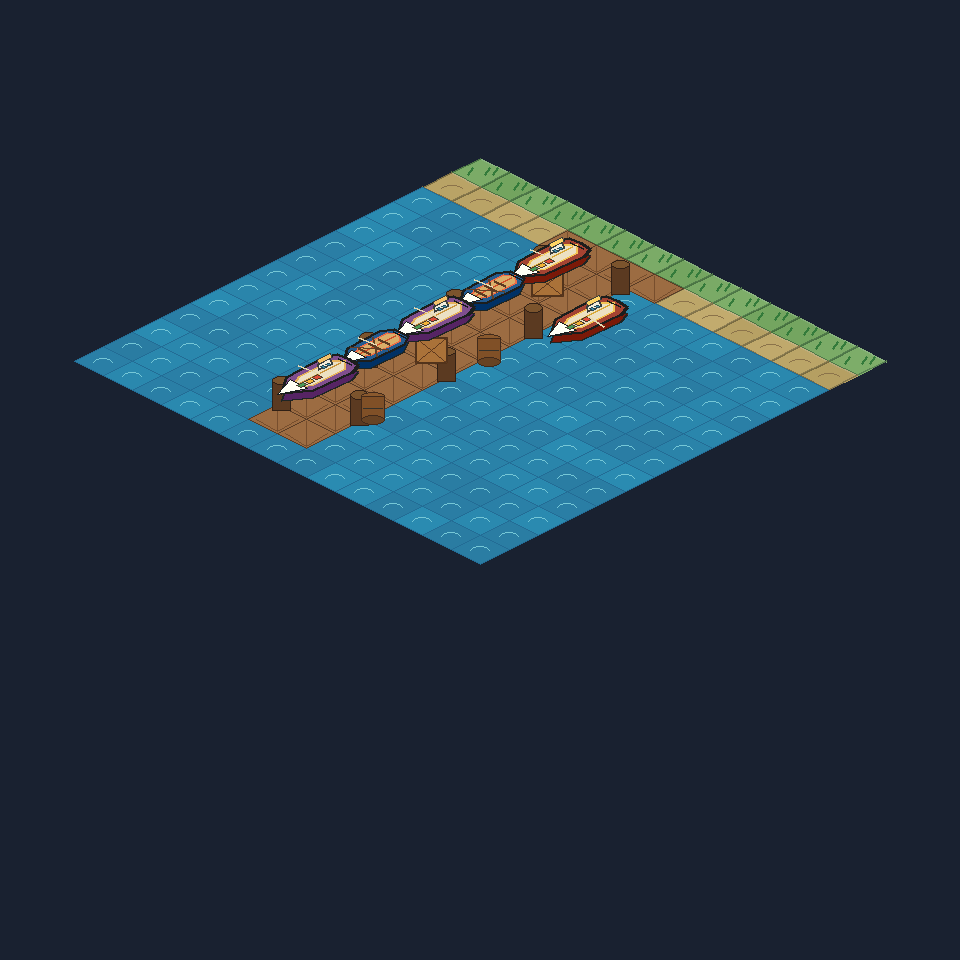}{The picture shows an isometric dockyard with shoreline tiles, water tiles, a wooden pier, dock objects, and boats tied along the dock. Count only the boats on the image-left side of the wooden main dock.}{5}{\traceatlaspromptnormal}
\hfill
\traceatlascard{Construction Site: Equipment Zone}{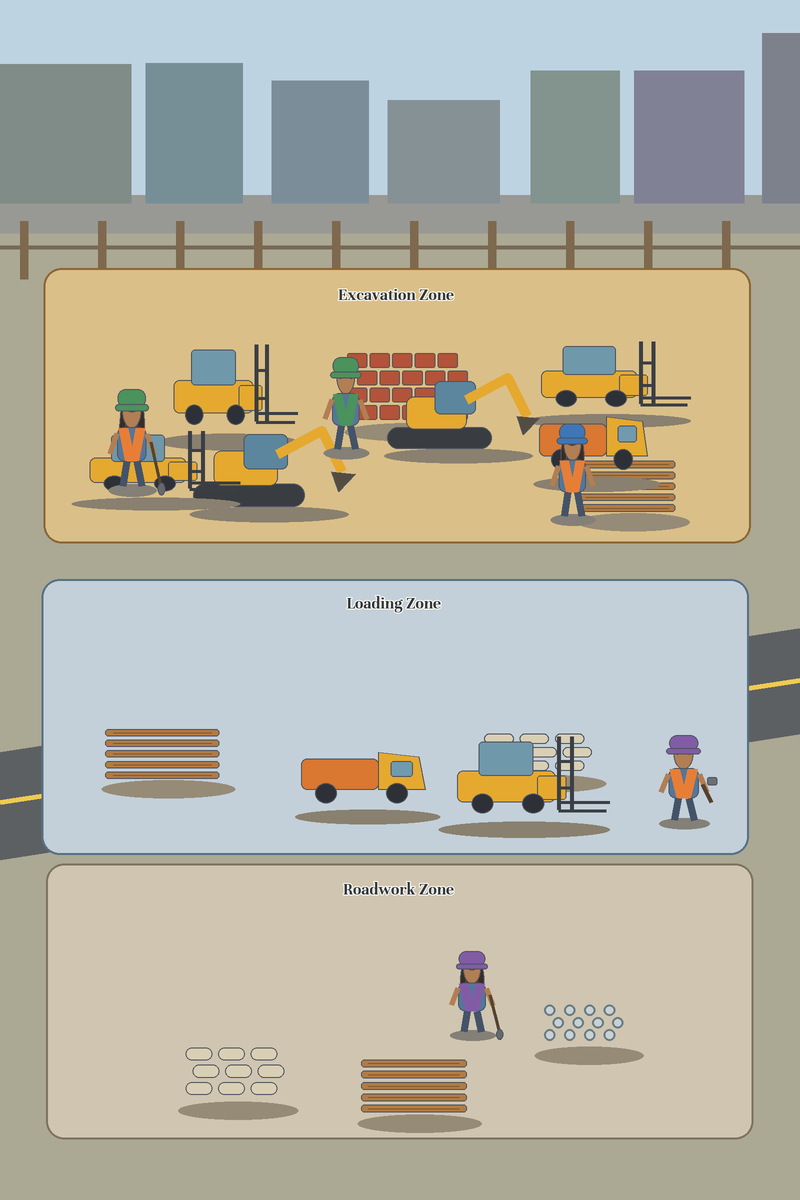}{The canvas contains a construction site with 5 workers, labeled work zones, materials, and equipment. How many construction vehicles are in the Roadwork Zone? If none are there, answer 0.}{0}{\traceatlaspromptnormal}
\caption{Representative illustration tasks.}
\label{fig:task-atlas-illustrations}
\end{figure}
\clearpage

\begin{figure}[p]
\centering
\traceatlascard{Web Action: Action Target}{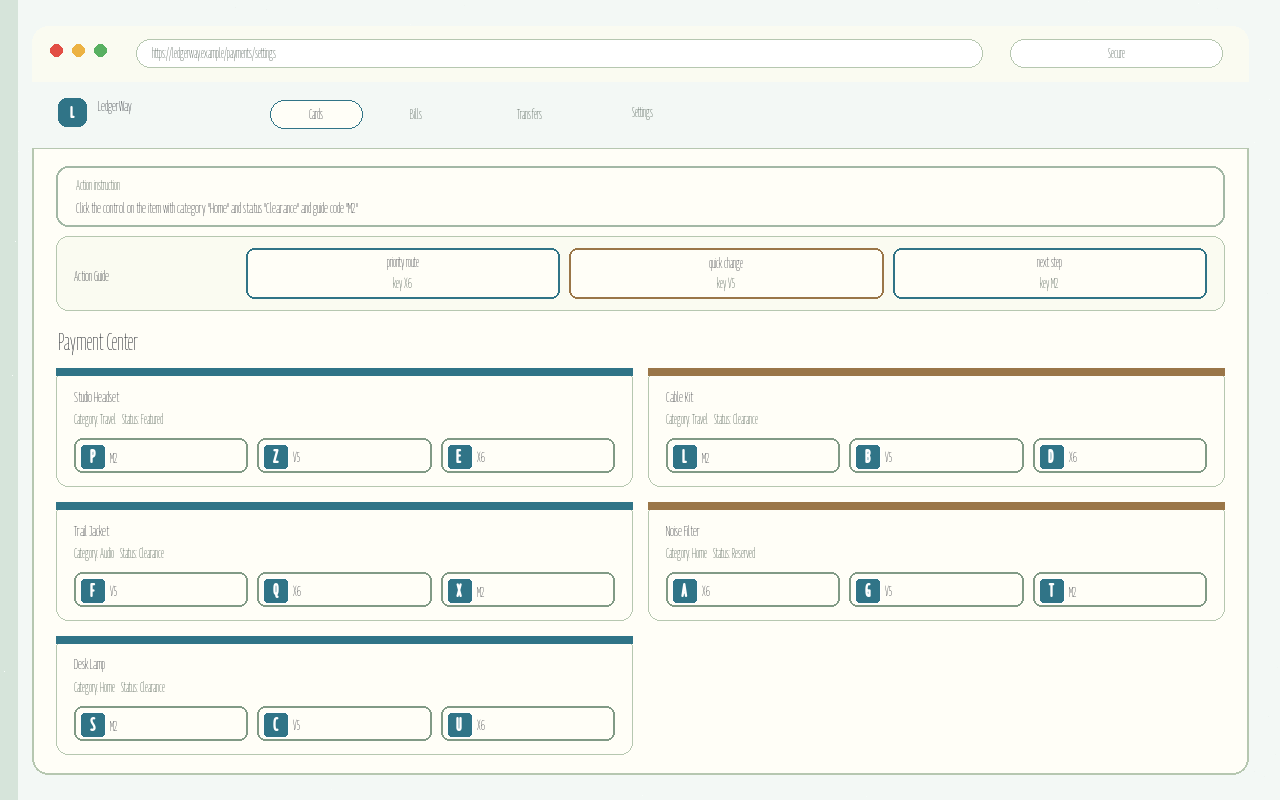}{The browser page includes a guide-code table and labeled candidate controls. Click the control on the item with category "Home", status "Clearance", and guide code "M2". Which candidate label marks that control?}{S}{\traceatlaspromptdense}
\hfill
\traceatlascard{Mixed Infographic Page: Field-Total Comparison}{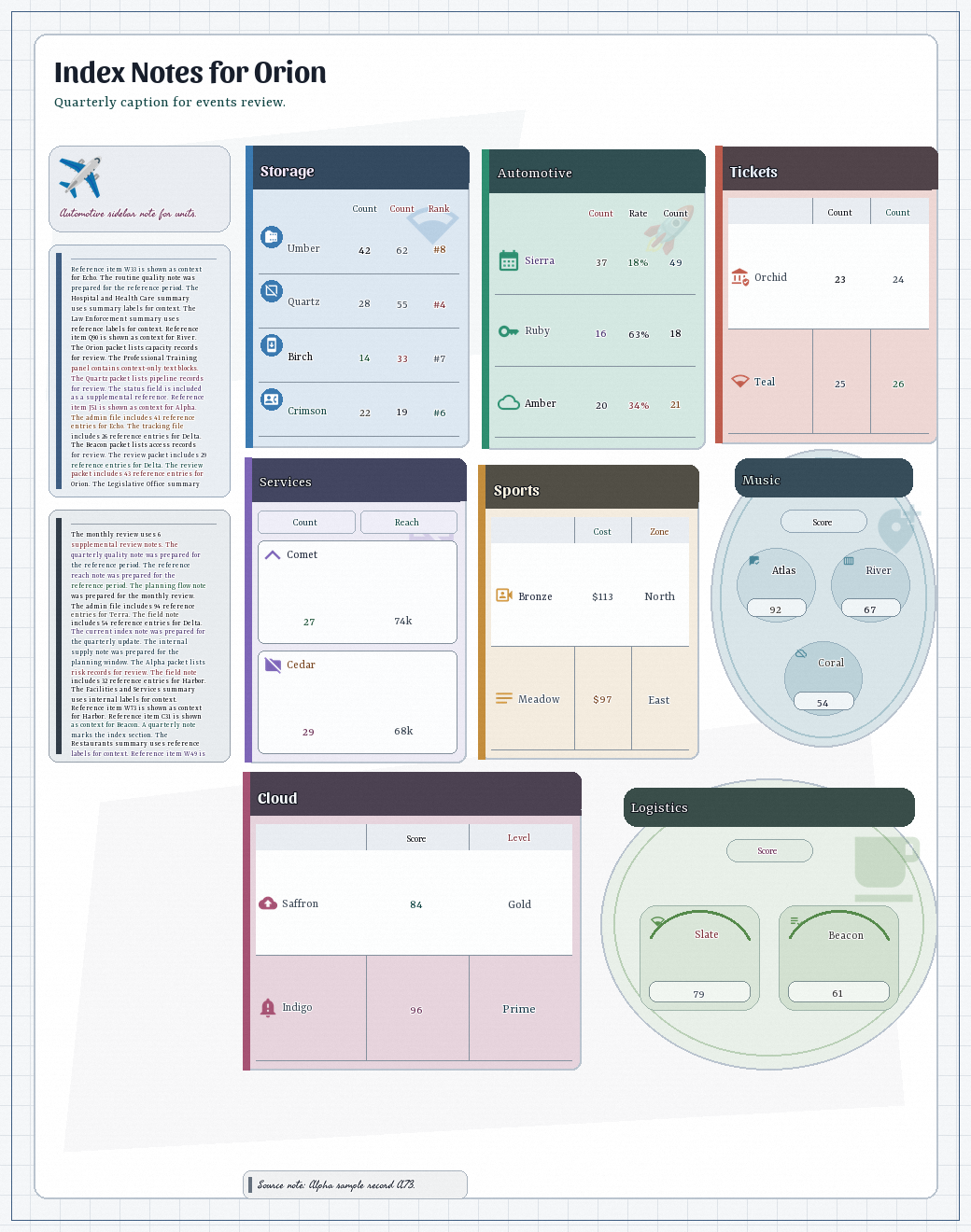}{The image shows a dense mixed infographic page with titled modules, item labels, field labels, printed values, and native context text. Between "Music" and "Logistics", which module has the greater sum of "Score" values?}{Music}{\traceatlaspromptnormal}
\hfill
\traceatlascard{Infographic: Section Icon Total Difference}{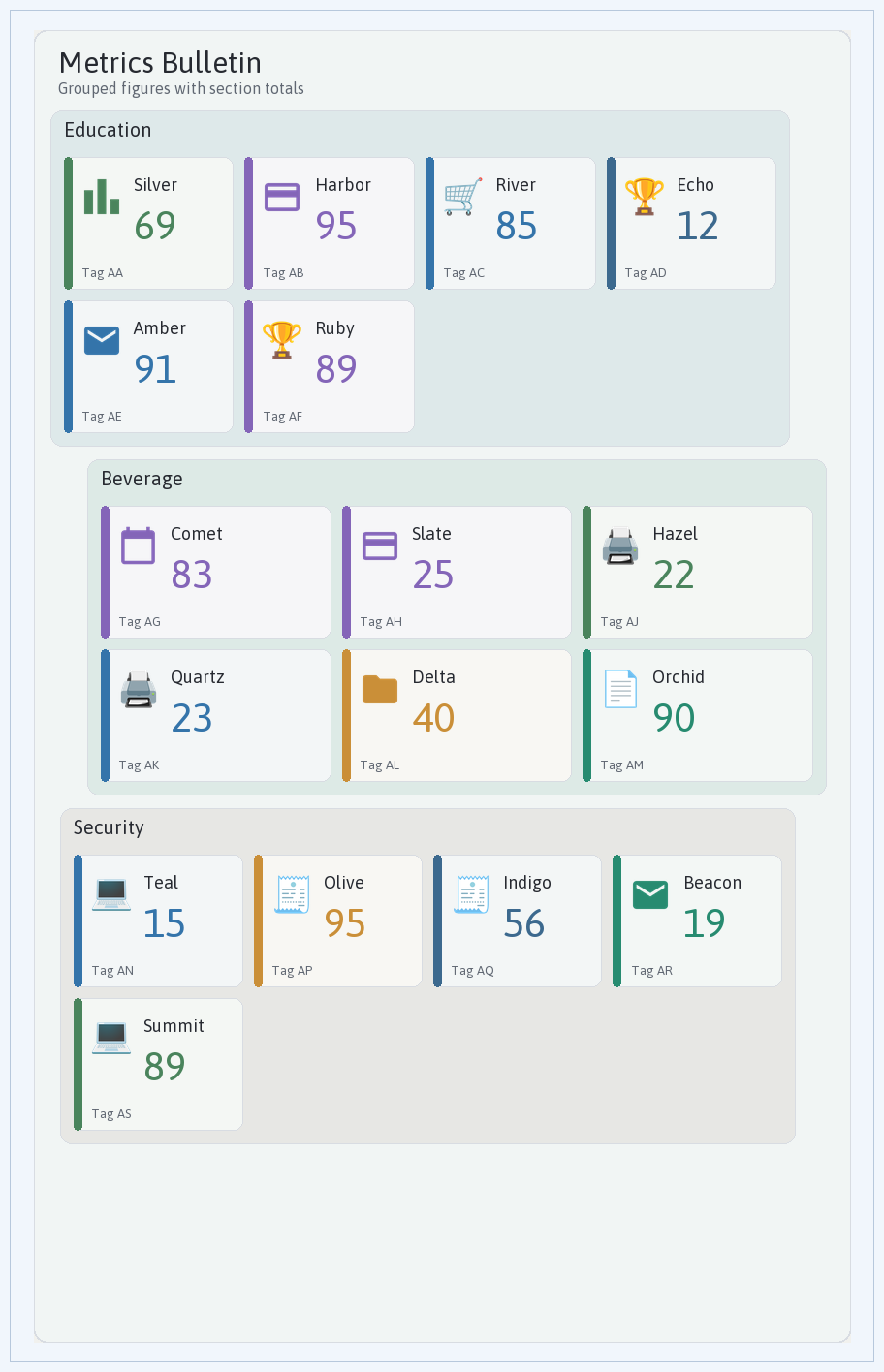}{The image shows a multi-section infographic with labeled metric cards and printed values. Add the credit card-icon cards in section "Education" and in section "Beverage". What is the nonnegative difference?}{70}{\traceatlaspromptnormal}
\par\vspace{2pt}
\noindent
\traceatlascard{Schedule: Longer Than Reference}{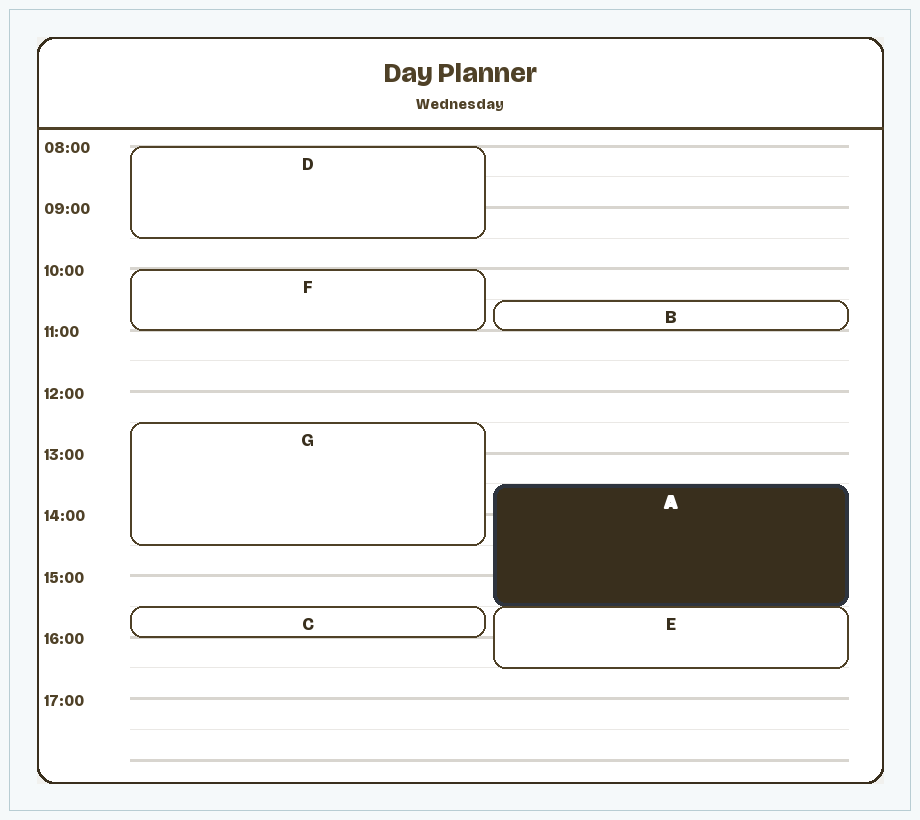}{This schedule shows one single-day schedule with time labels and scheduled event blocks. How many other scheduled events are longer than the highlighted reference event?}{0}{\traceatlaspromptnormal}
\hfill
\traceatlascard{Concept Map: Branch Child}{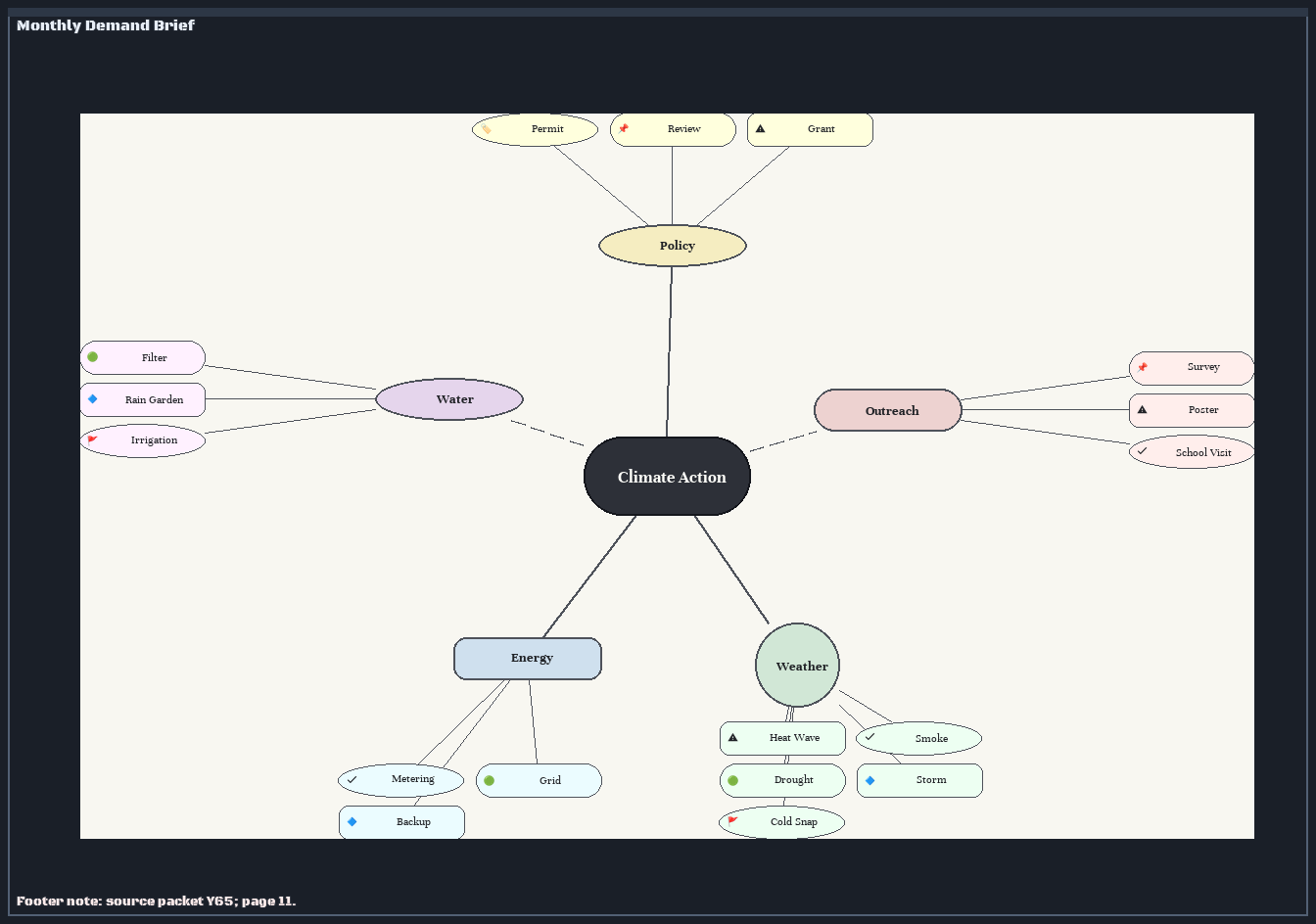}{The concept map shows a central topic, labeled branches, visible child-item nodes, and connector lines. How many child items are listed under the "Weather" branch?}{5}{\traceatlaspromptnormal}
\hfill
\traceatlascard{Hero Callout Infographic: Callout Composite Metric Extremum}{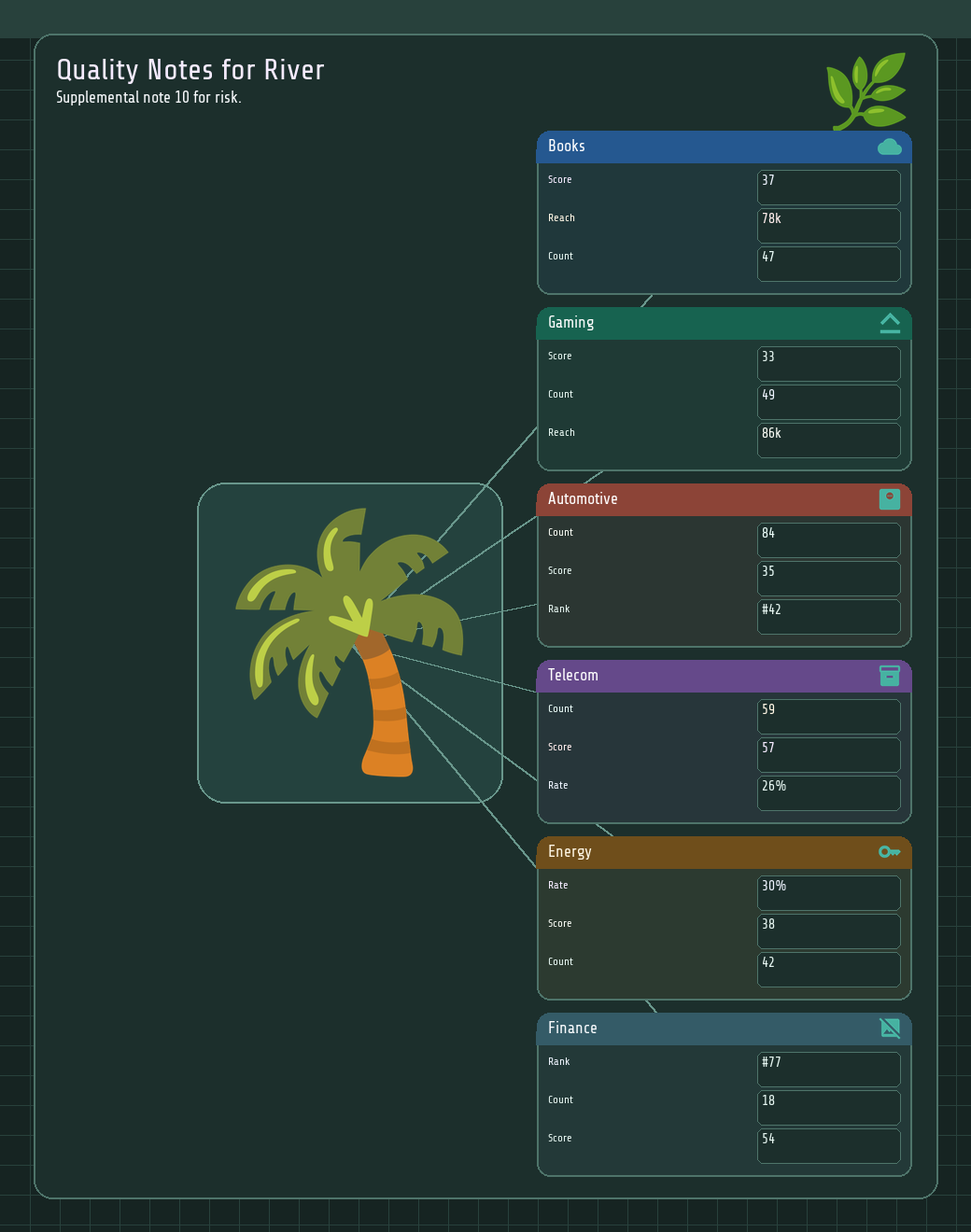}{The figure shows titled callout cards with field labels, visible values, badges, and connector lines. Order the callouts by "Score" plus "Count", highest to lowest. Which callout title is first?}{Automotive}{\traceatlaspromptnormal}
\par\vspace{2pt}
\noindent
\traceatlascard{Cycle: Offset Stage}{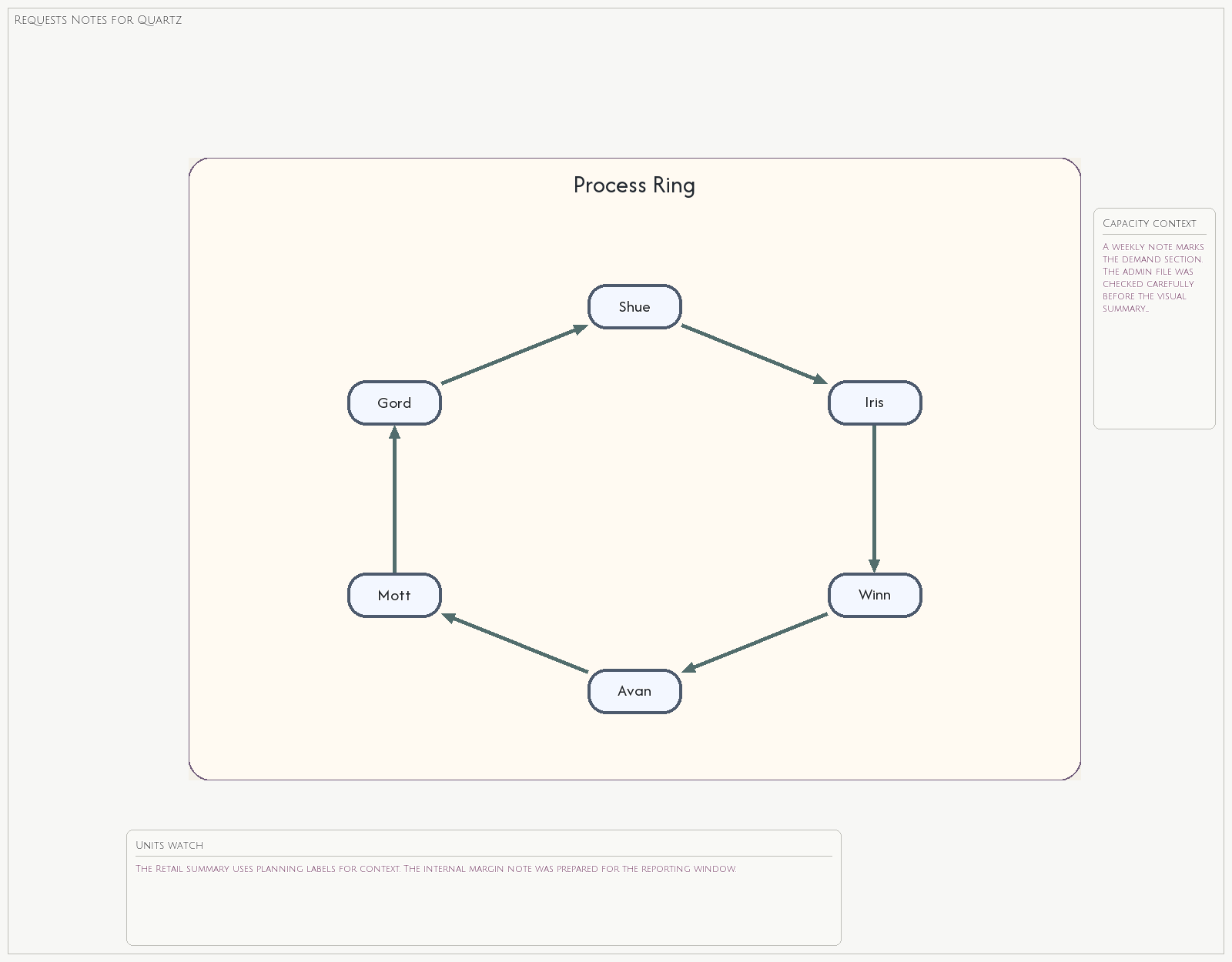}{This diagram shows a directed cycle with labeled stages connected by arrows. Use the arrows and count 2 steps after "Gord". What stage do you reach?}{Iris}{\traceatlaspromptnormal}
\hfill
\traceatlascard{Paired Forms: Total Amount Delta}{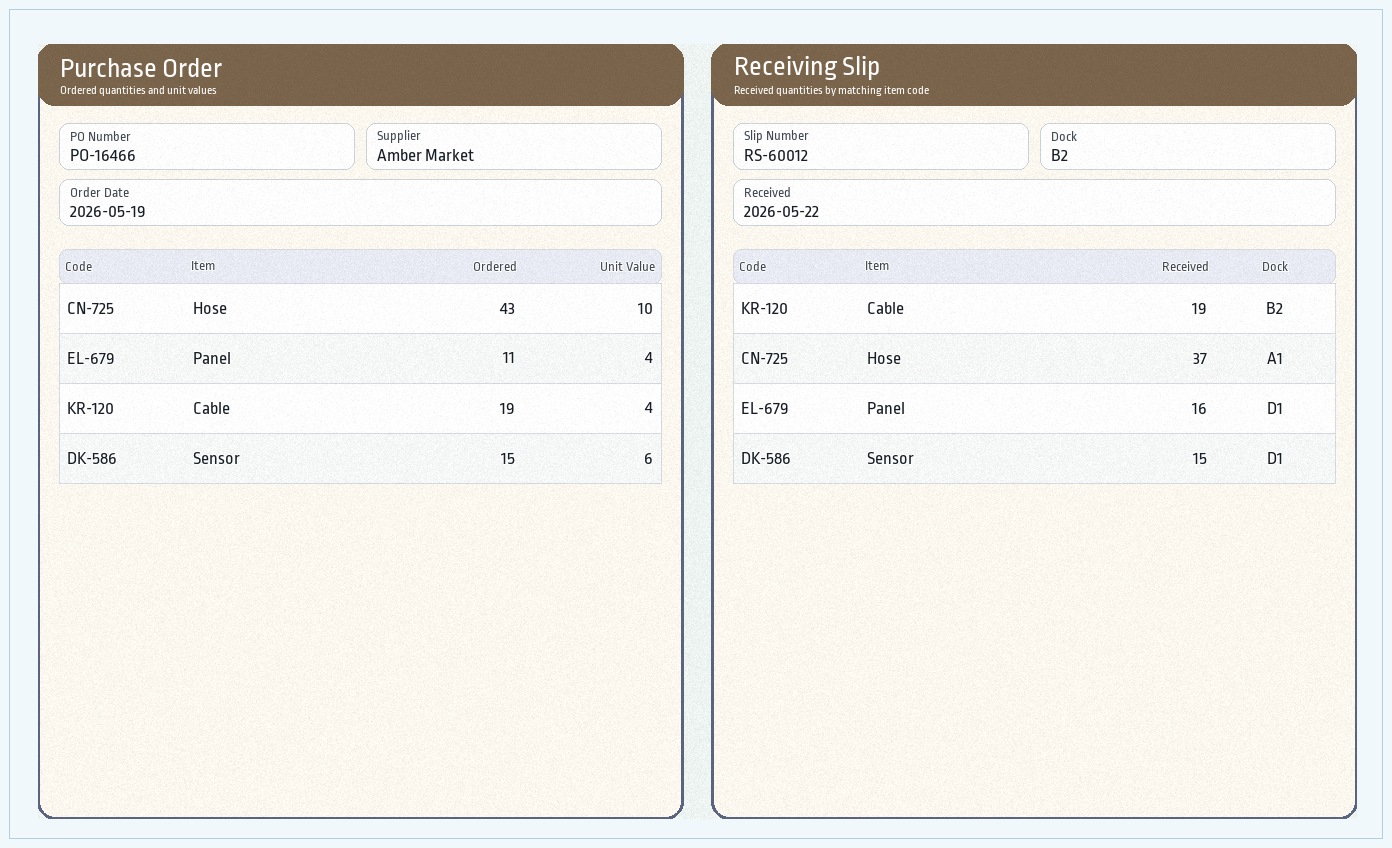}{The purchase order and receiving slip show matching item codes, quantities, and purchase-order unit values. For each item with differing quantities, multiply the absolute quantity difference by its unit value and sum the products. What is the total amount delta?}{80}{\traceatlaspromptdense}
\hfill
\traceatlascard{Instruction Panel: Step for Control Pair}{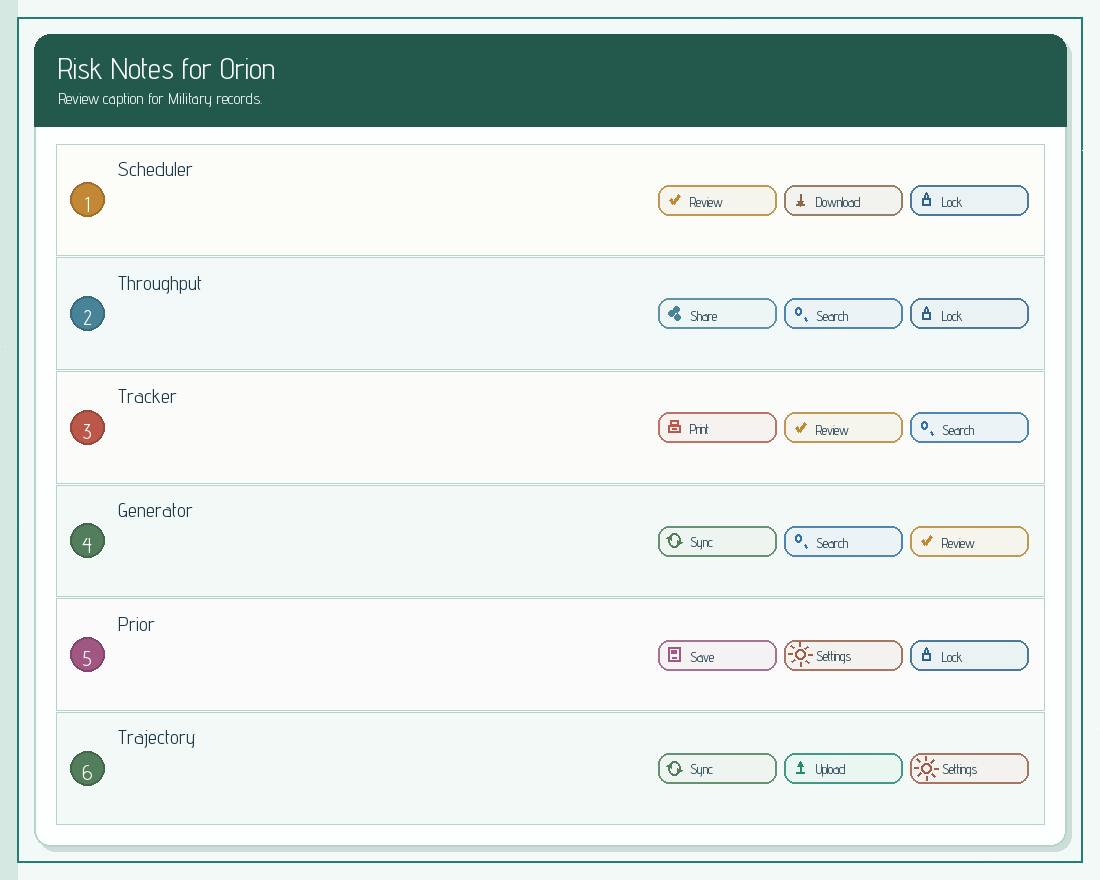}{The visual shows an instruction panel with numbered steps and visible control chips on each step. Find the step with control labels "Search" and "Share". What number is on that step?}{2}{\traceatlaspromptnormal}
\par\vspace{2pt}
\noindent
\traceatlascard{Step List: Relative Offset Step}{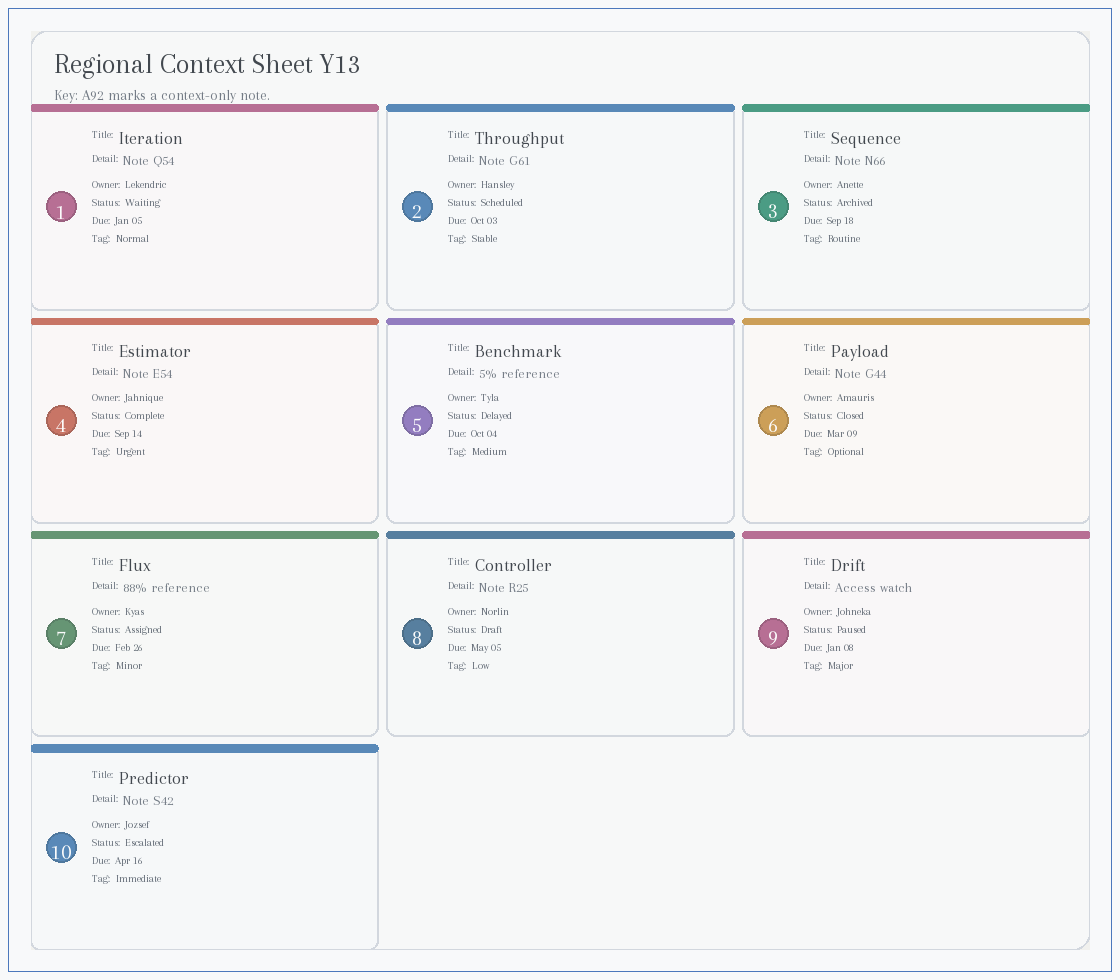}{This page shows a numbered workflow step list where each step card has labeled Title and Detail fields plus other small fields. What step title is 3 steps after the step titled "Payload"?}{Drift}{\traceatlaspromptnormal}
\hfill
\traceatlascard{Map: Landmark After Route Step}{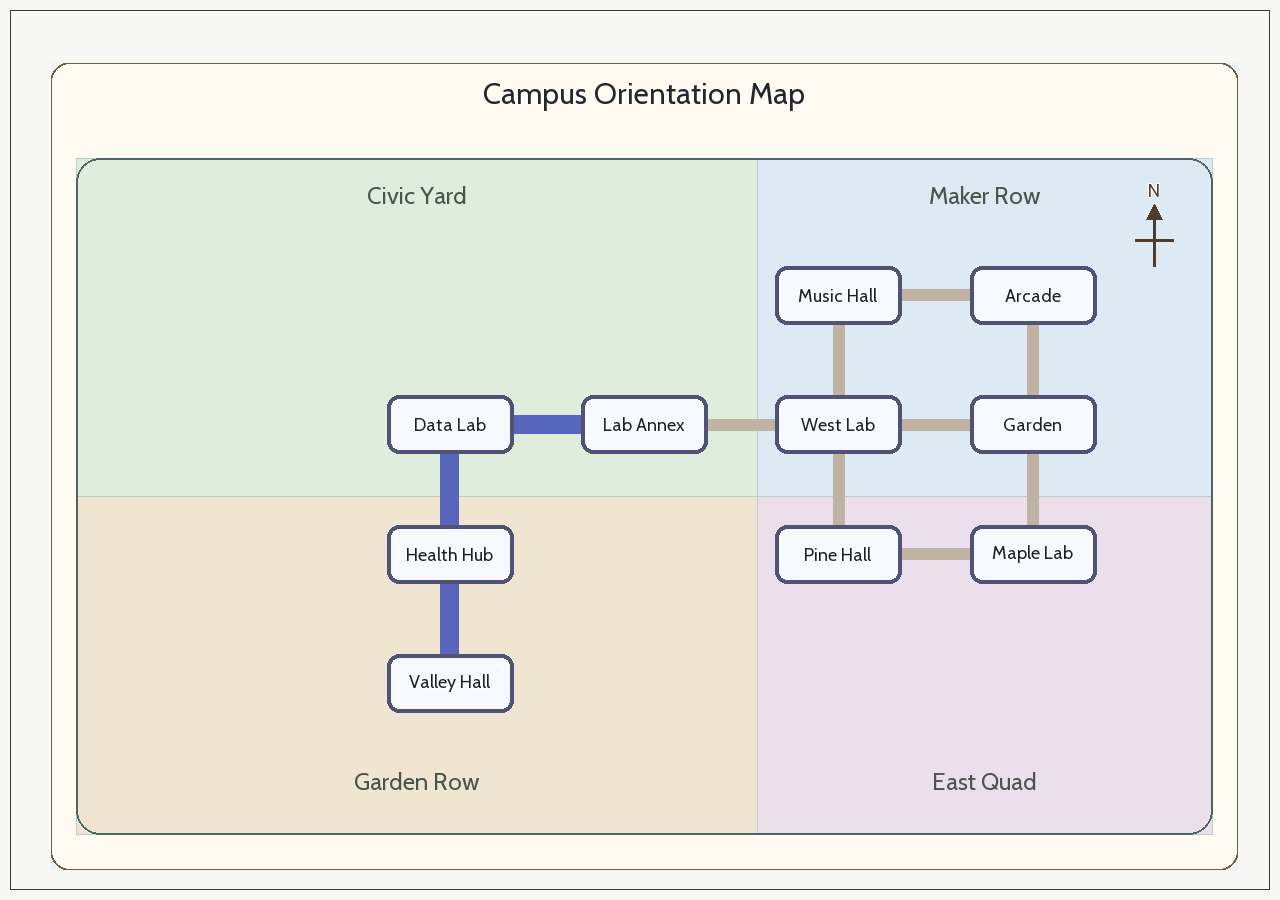}{The figure shows a printed campus layout with labeled landmarks, named zones, visible walking paths, and a highlighted orange route. Trace the highlighted path from "Valley Hall" to "Lab Annex". Which label is at the 3rd reached landmark?}{Lab Annex}{\traceatlaspromptdense}
\hfill
\traceatlascard{Profile Card Grid: Highest-Value Profile}{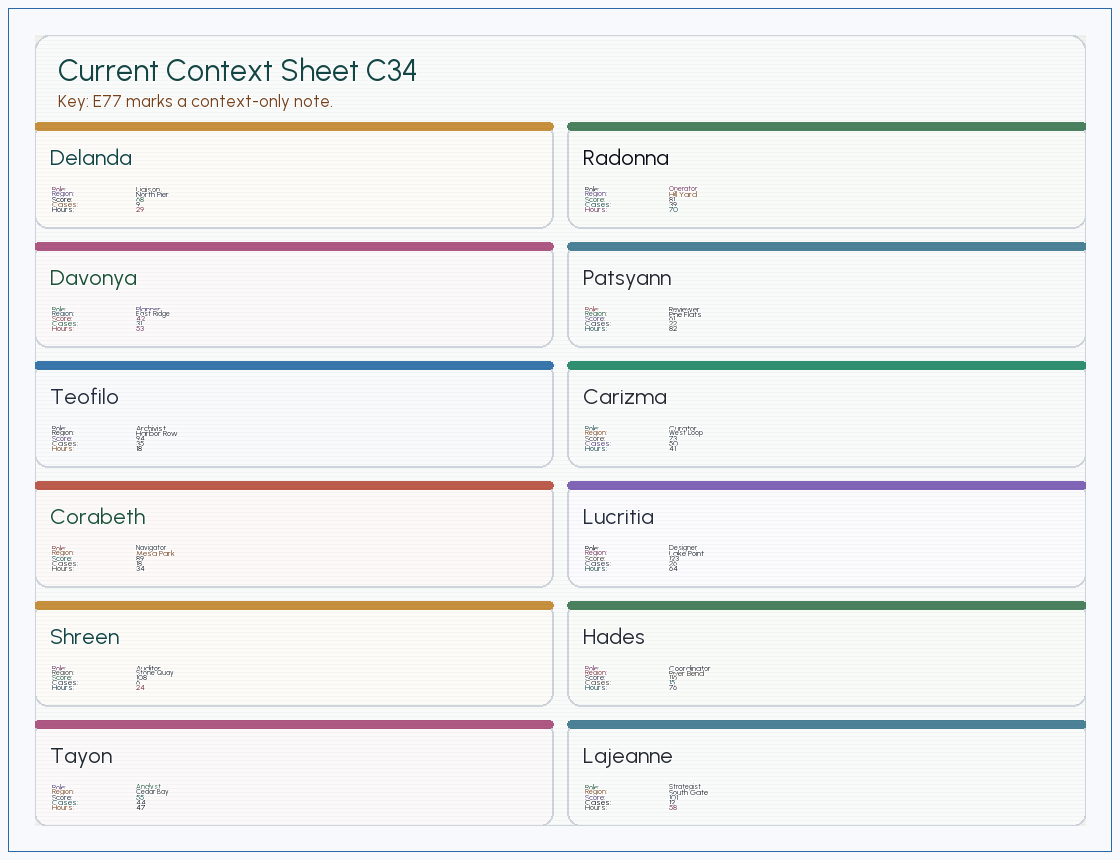}{This page shows a grid of profile cards, each with a visible profile name and labeled fields. Which visible profile ranks highest by Hours?}{Patsyann}{\traceatlaspromptnormal}
\caption{Representative page tasks.}
\label{fig:task-atlas-pages}
\end{figure}
\clearpage

\begin{figure}[p]
\centering
\traceatlascard{Ray Optics: Ray Bounce}{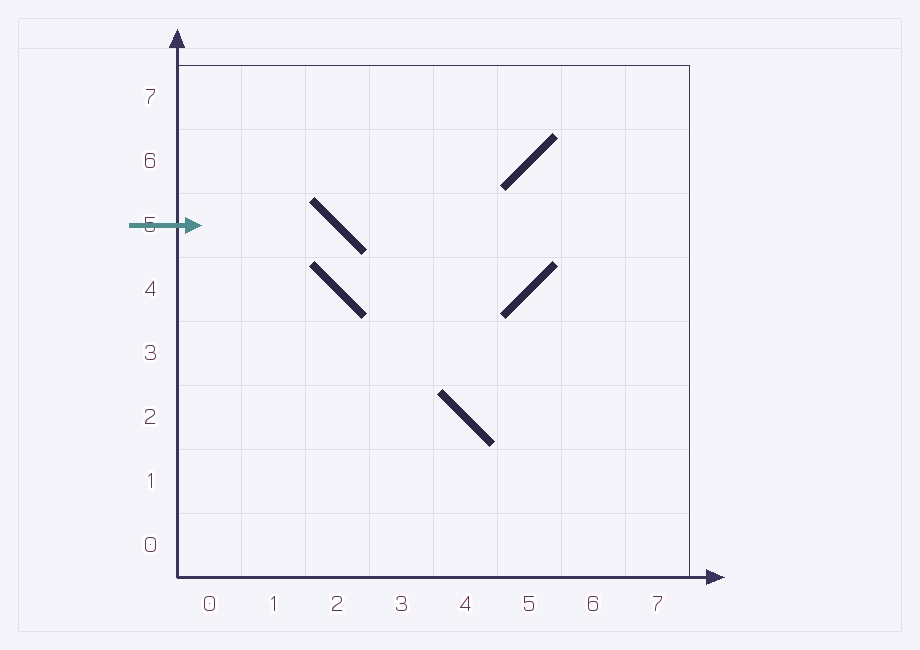}{This figure shows a reflected ray setup on a square grid. What is the total number of mirror bounces in the implied ray path?}{4}{\traceatlaspromptnormal}
\hfill
\traceatlascard{Electrostatic Field: Field Direction Choice}{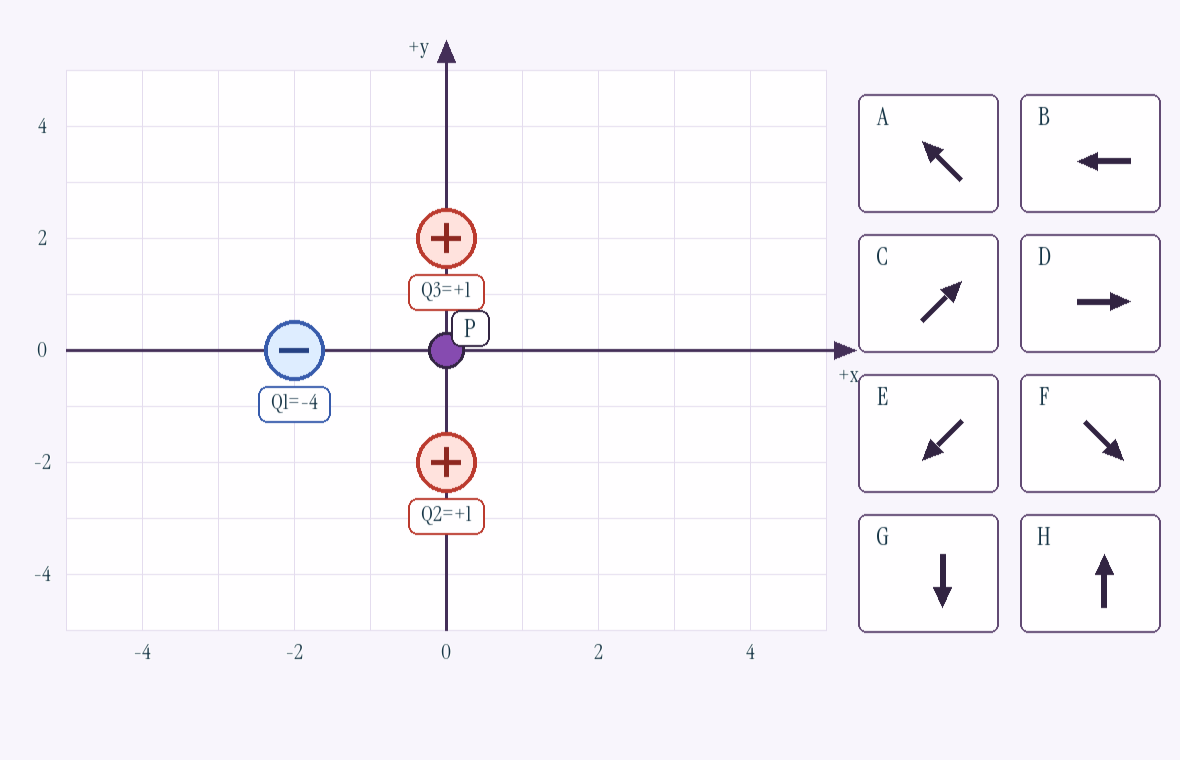}{The diagram shows an electrostatic coordinate grid with fixed point charges, a marked point or labeled candidate points, and visible option arrows or distance labels. Which option A-H points in the direction of the electric field at P?}{B}{\traceatlaspromptdense}
\hfill
\traceatlascard{Switch Circuit: Lit Bulb}{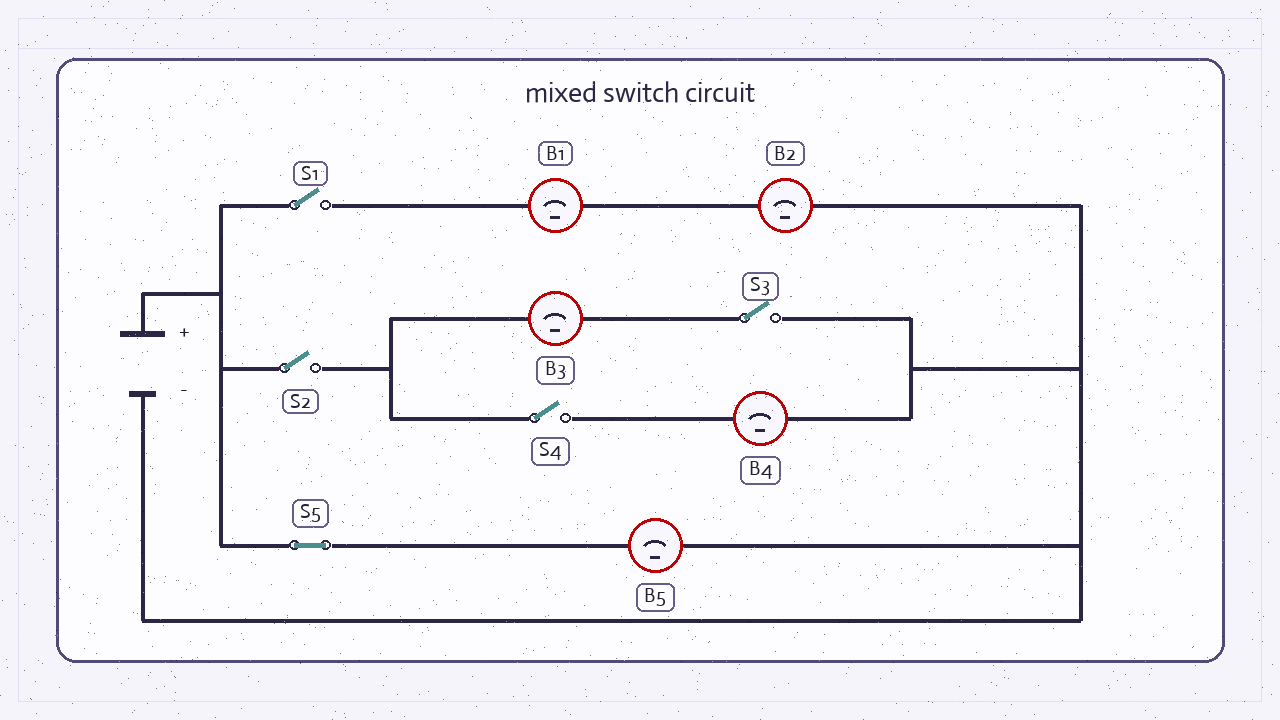}{The circuit diagram shows a mixed branch battery circuit with five labeled bulbs and visibly open or closed switches. Given the switch states, how many bulbs will be on?}{1}{\traceatlaspromptnormal}
\par\vspace{2pt}
\noindent
\traceatlascard{Circuit State Change: Bulb Brightness Change}{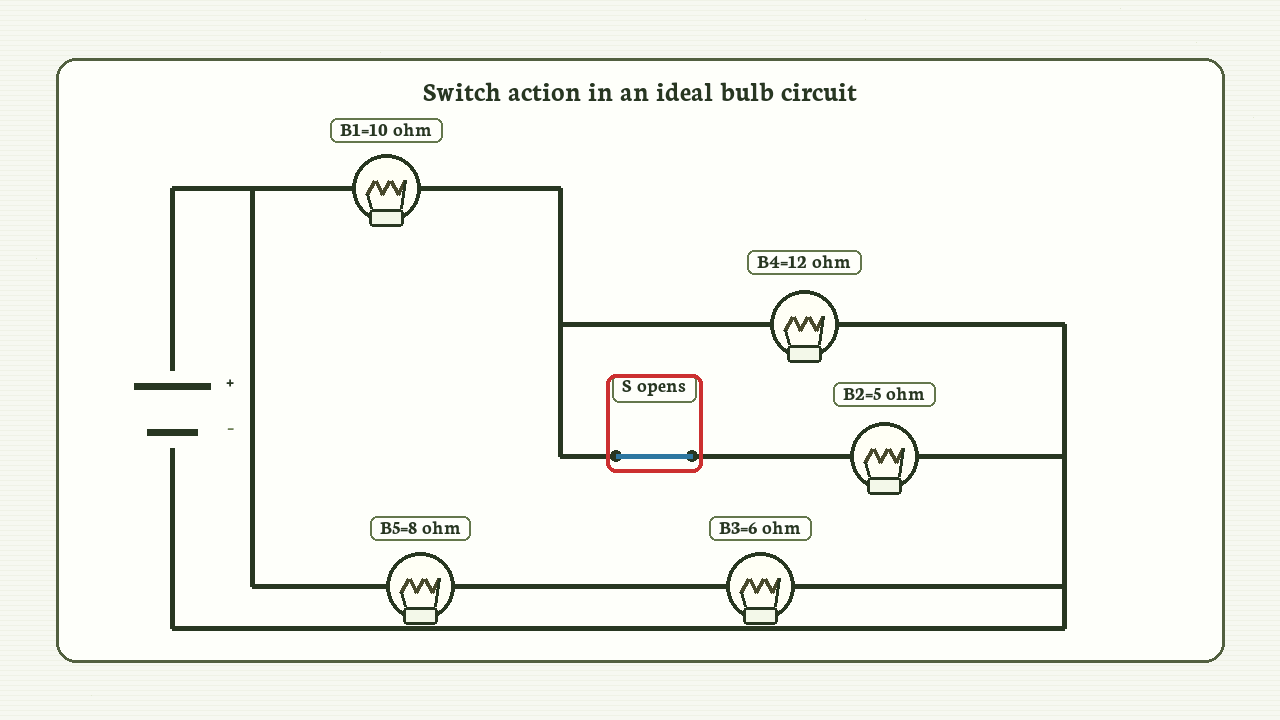}{The image shows a visible ideal-battery circuit with five labeled bulbs, resistance labels, and one red-boxed switch action cue. Using the bulb resistances and topology, which bulb becomes brighter after the switch action?}{B4}{\traceatlaspromptdense}
\hfill
\traceatlascard{Magnetic Force: Force Direction Choice}{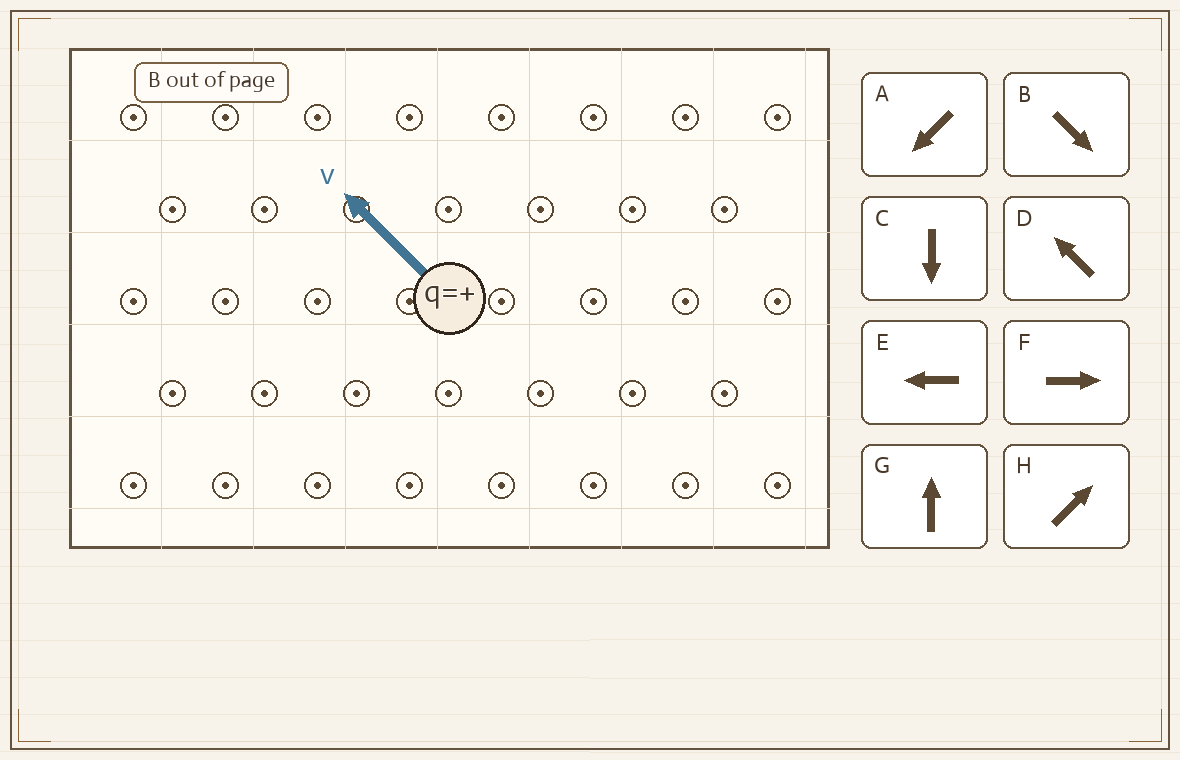}{The image shows a magnetic-field panel with field-direction symbols, a charged particle, a velocity arrow, and eight labeled candidate force arrows. Which option A-H points in the direction of the force on the charged particle?}{H}{\traceatlaspromptdense}
\hfill
\traceatlascard{Wave Interference: Path Difference}{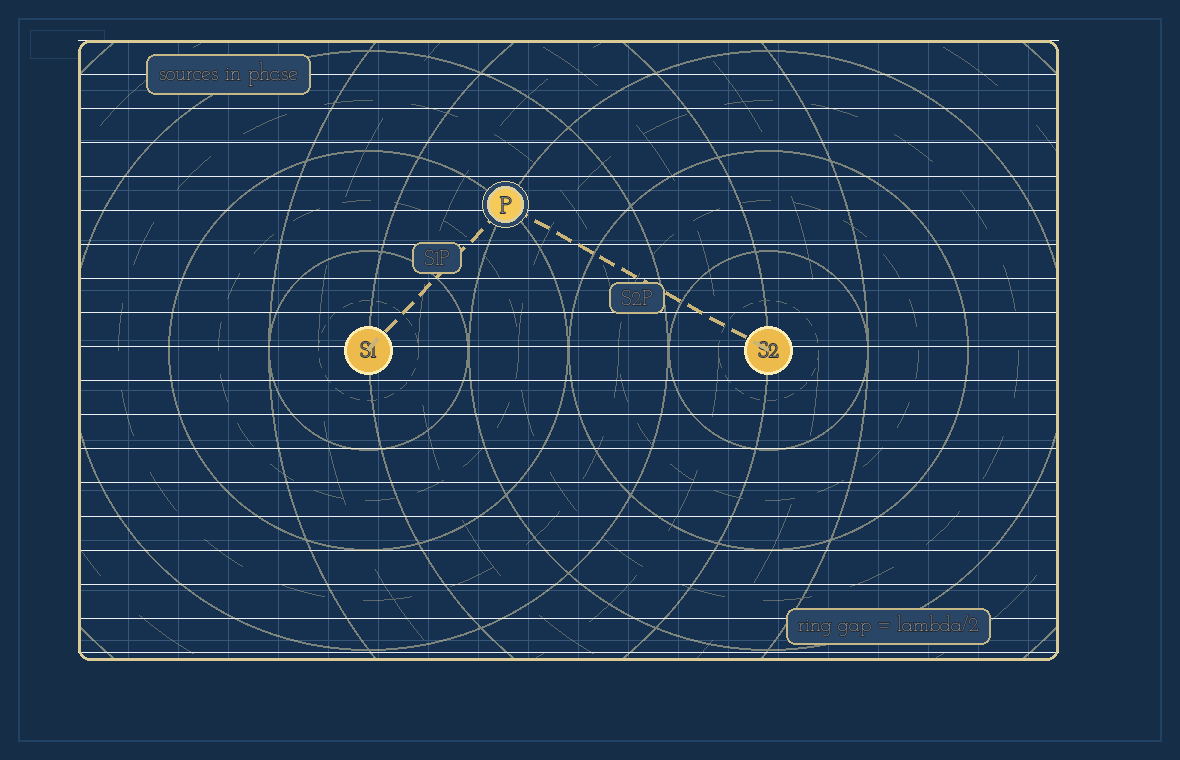}{The diagram shows two labeled wave sources and their circular interference wavefronts. Using the visible wavefront spacing and labels, determine the absolute path difference between S1P and S2P in half-wavelength steps.}{2}{\traceatlaspromptnormal}
\par\vspace{2pt}
\noindent
\traceatlascard{Spring: Spring Extension Difference}{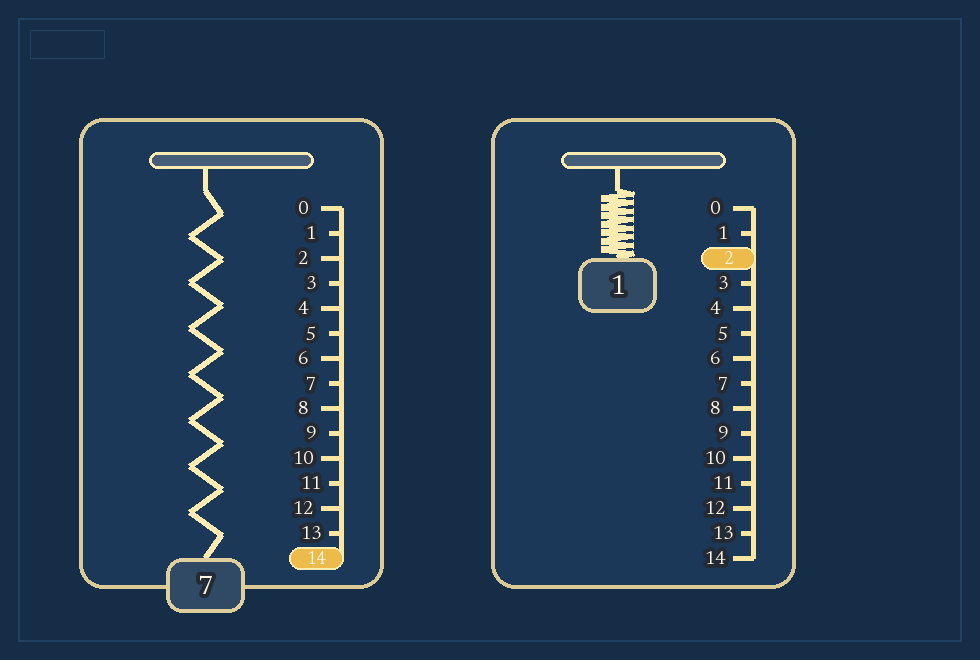}{The figure shows a pair of identical springs with weight blocks and ruler markers. Using the two ruler markers, what integer difference in extension do the springs show?}{12}{\traceatlaspromptnormal}
\hfill
\traceatlascard{Circuit Equivalent: Total Capacitance}{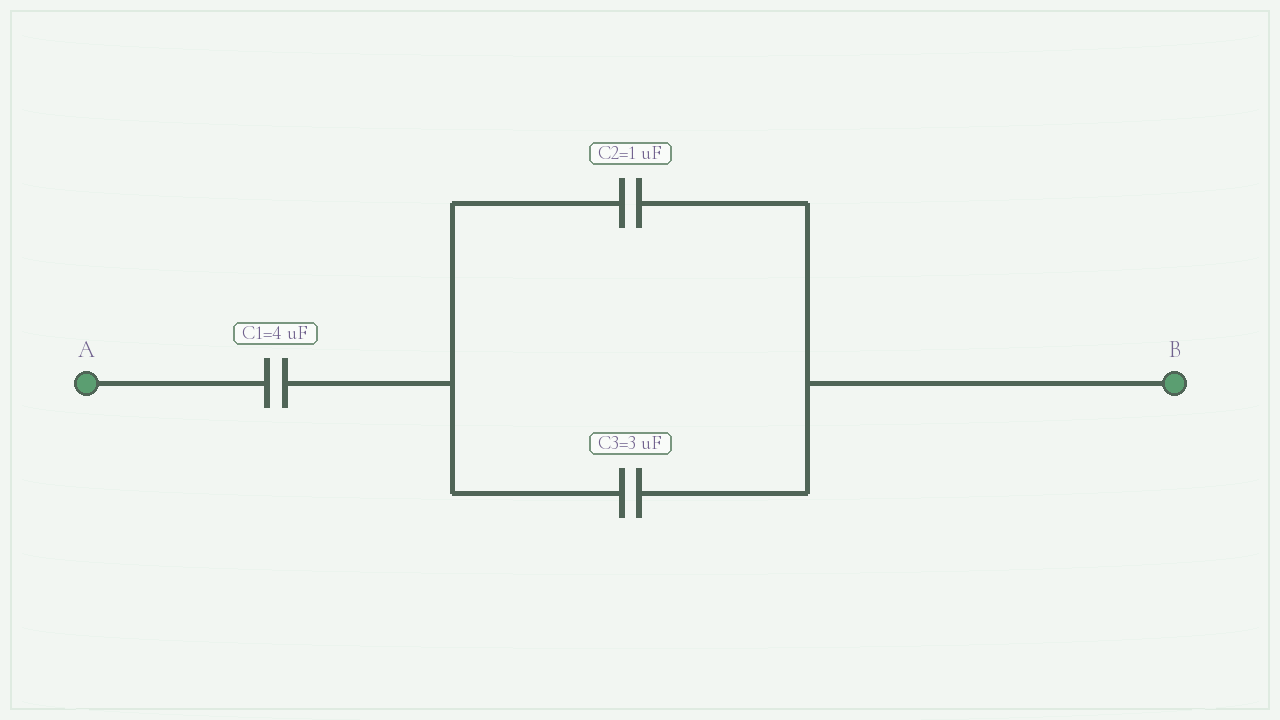}{The visual shows one capacitor circuit between terminals A and B with at least one labeled parallel-plate capacitor in series with one or two labeled parallel capacitor blocks. Using the shown capacitor values, what equivalent capacitance is seen between terminals A and B?}{2}{\traceatlaspromptdense}
\hfill
\traceatlascard{Electromagnetic Induction: Induced Current Direction}{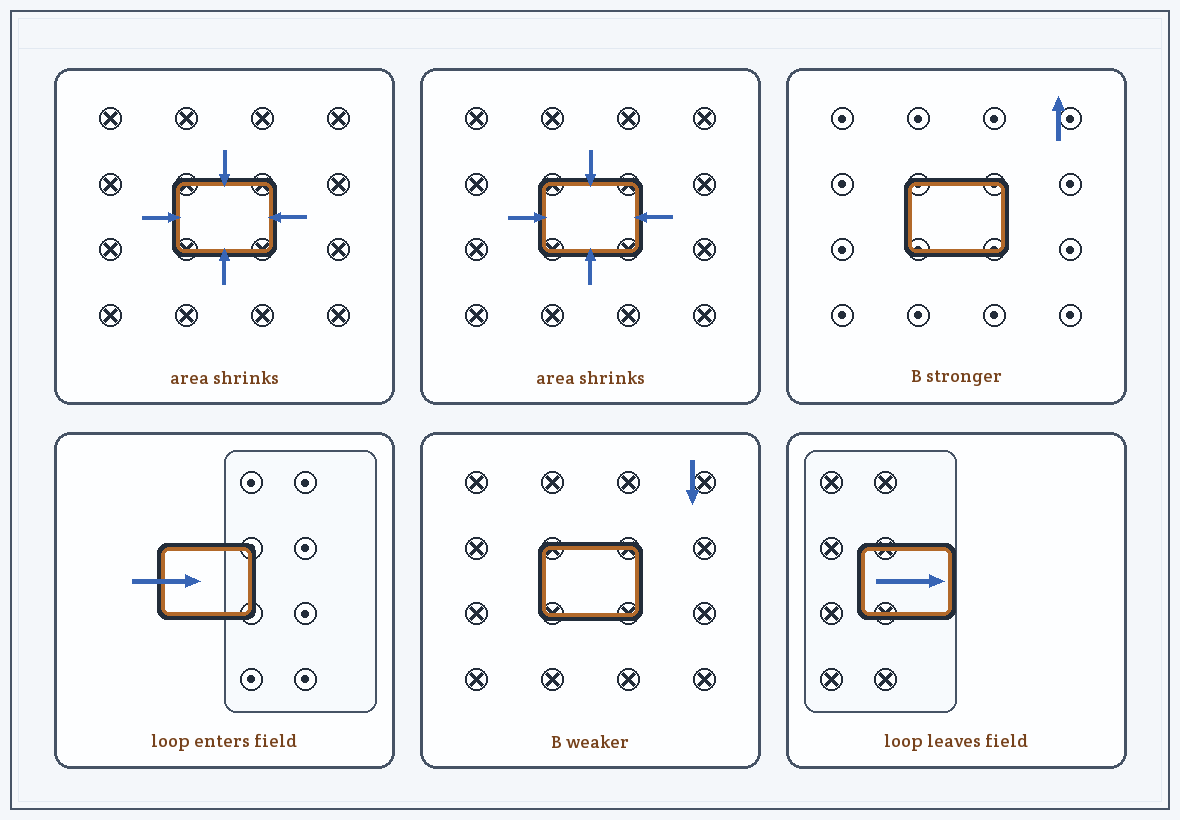}{The induction diagram shows six mini-panels showing conducting loops, into-page or out-of-page magnetic-field symbols, and visible cues for changing or unchanged magnetic flux. How many panels have a clockwise induced current in the loop?}{6}{\traceatlaspromptdense}
\par\vspace{2pt}
\noindent
\traceatlascard{Wire Magnetism: Wire Field Direction Choice}{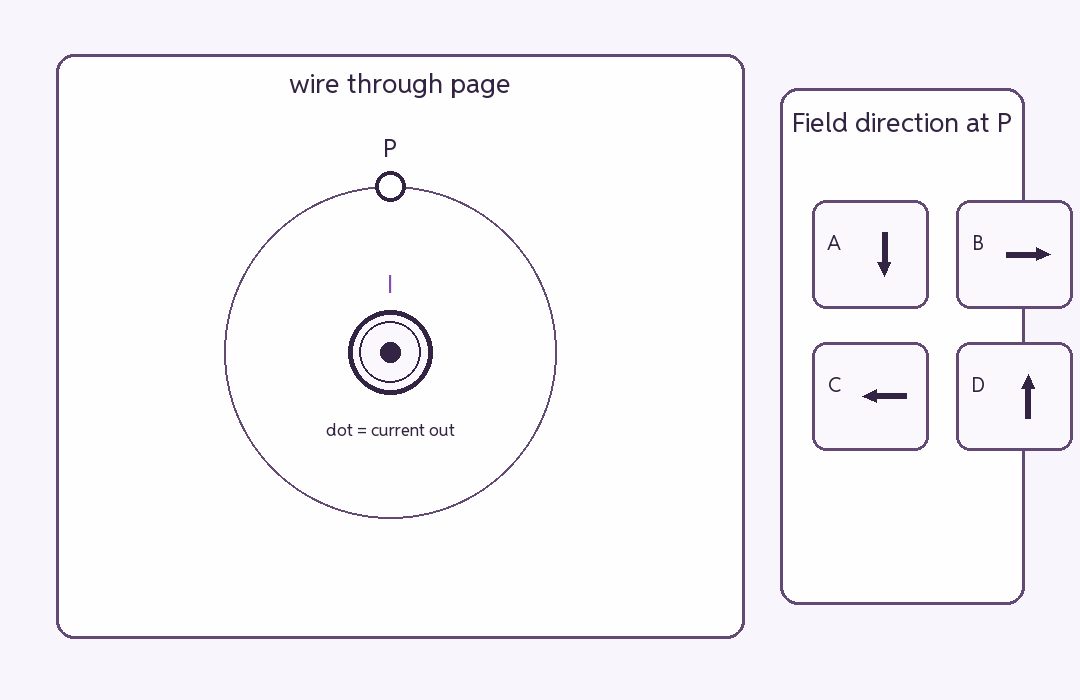}{The diagram shows a current-carrying wire through the page, point P, and labeled arrow options. Use the current cue to determine the circular magnetic-field direction at P. Which arrow option matches that direction?}{C}{\traceatlaspromptnormal}
\hfill
\traceatlascard{Stack Stability: Stability Status}{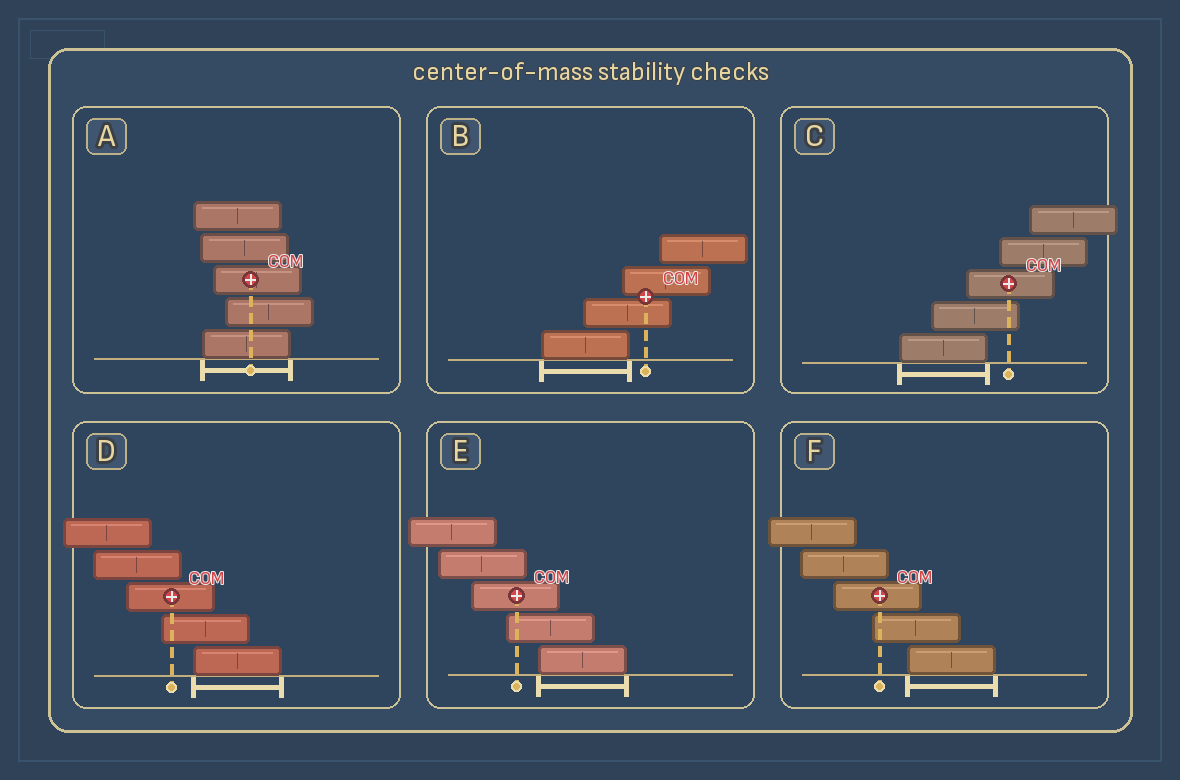}{The image shows six labeled stacks of equal-size brick blocks. Which labeled stack will stay upright because its center-of-mass projection falls within its support base?}{A}{\traceatlaspromptnormal}
\hfill
\traceatlascard{Lever: Missing Weight Balance}{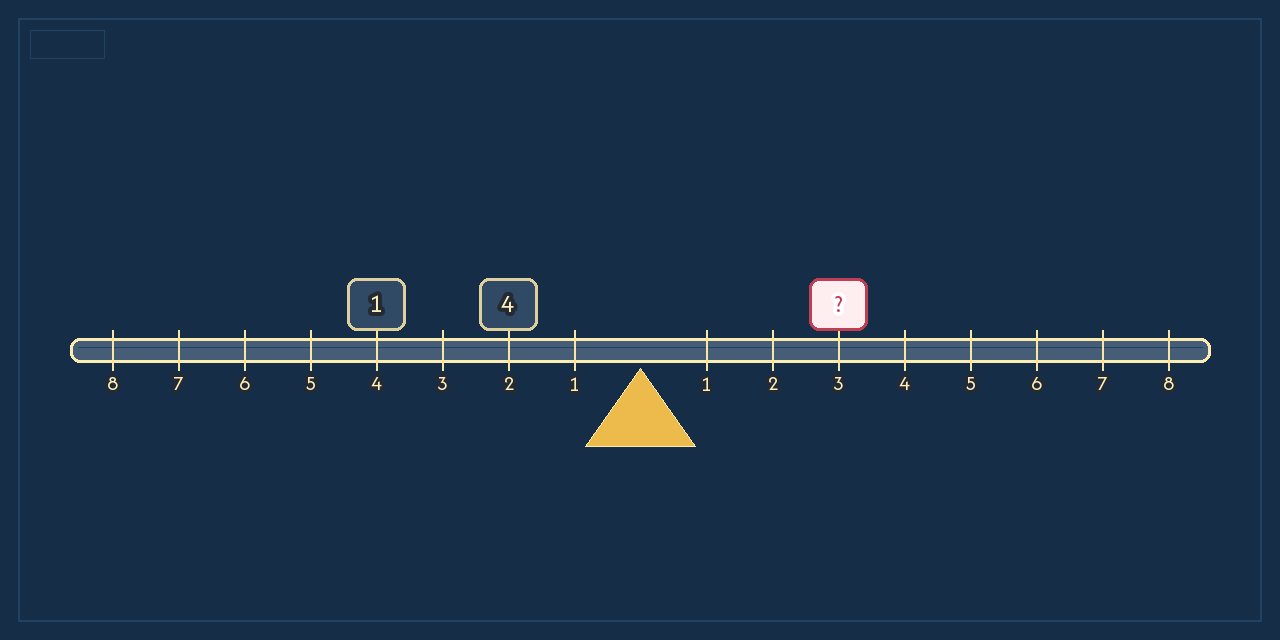}{The figure shows a lever with a fulcrum, distance marks, and weight blocks. What weight should replace the block marked with a question mark so the lever balances?}{4}{\traceatlaspromptnormal}
\caption{Representative physics tasks.}
\label{fig:task-atlas-physics}
\end{figure}
\clearpage

\begin{figure}[p]
\centering
\traceatlascard{Sheet Transform: Fold Projection Result}{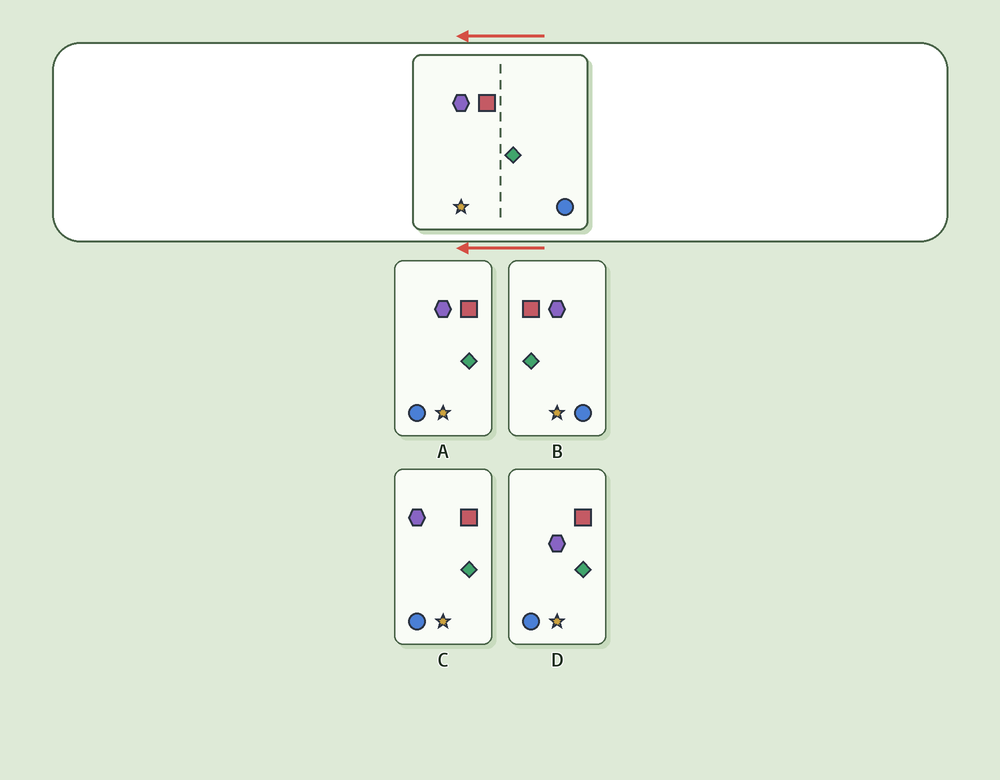}{The visual shows an outline-style paper-folding puzzle with a fold diagram above labeled result options. After the shown fold is made, which labeled option shows the visible mark pattern?}{A}{\traceatlaspromptnormal}
\hfill
\traceatlascard{Cube Net: Marked Edge Neighbor Face}{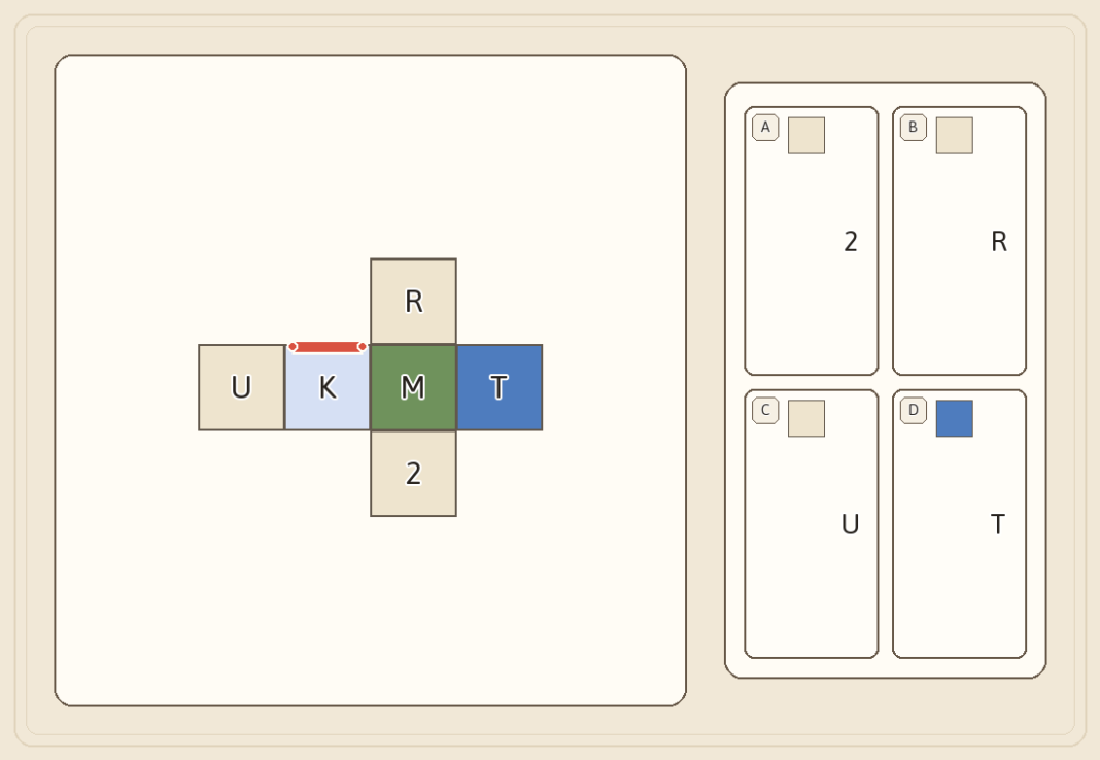}{The puzzle diagram shows a labeled cube net with one red marked edge and labeled face options. Which labeled option touches the red marked edge on the folded cube?}{B}{\traceatlaspromptnormal}
\hfill
\traceatlascard{Word Search: Search Location}{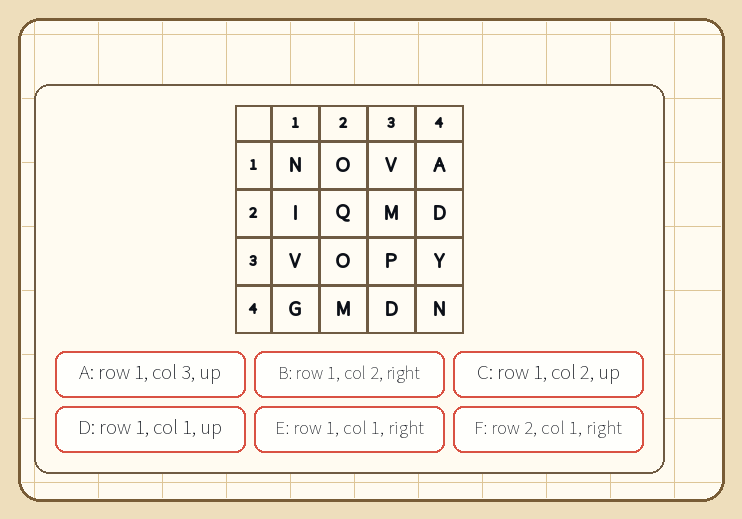}{The visual shows a row-and-column labeled word-search grid. Which labeled option below the grid matches the placement of "NOVA" in the word-search grid?}{E}{\traceatlaspromptnormal}
\par\vspace{2pt}
\noindent
\traceatlascard{Rubik's Cube Net: Post-Move Face Color Count}{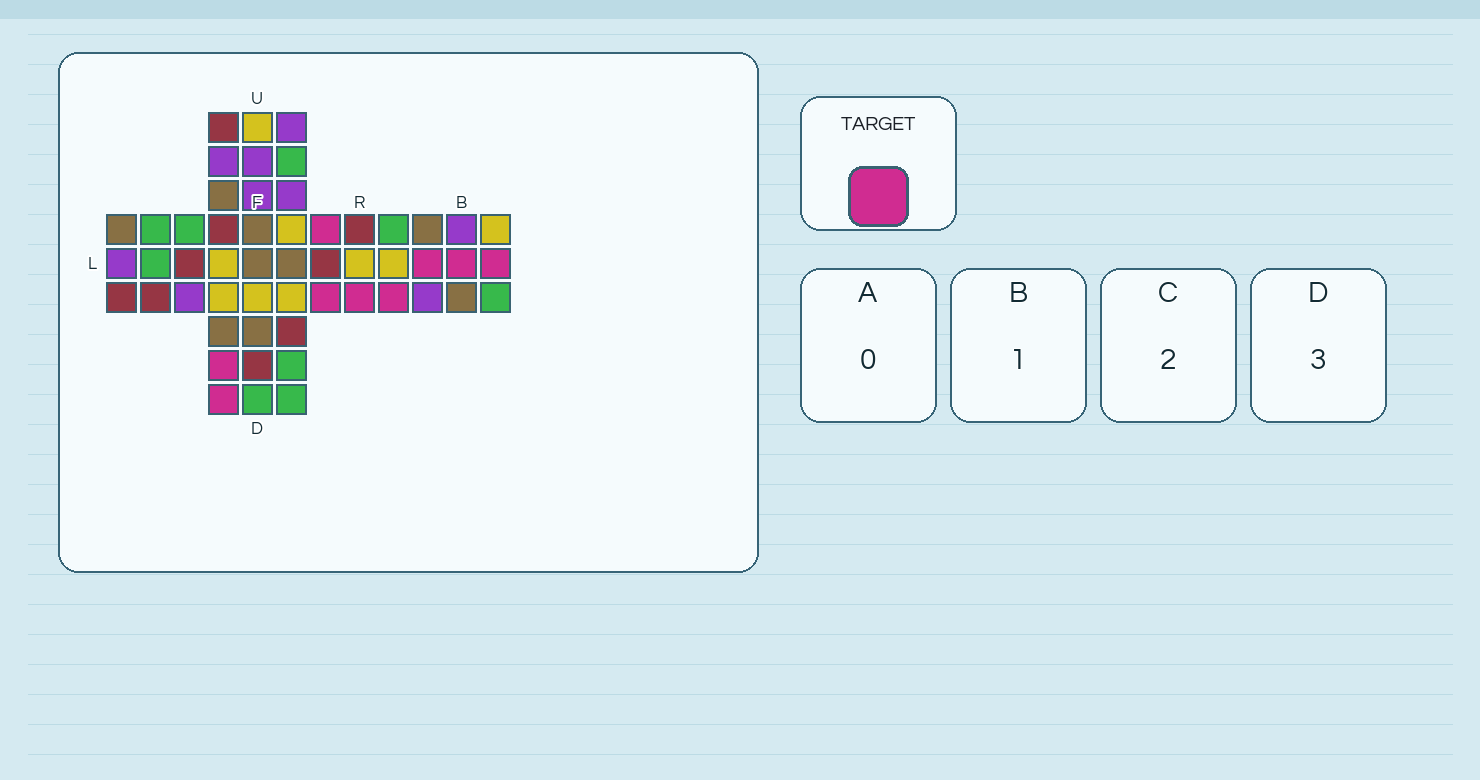}{The scene shows a Rubik's Cube net with face labels, a target-color swatch, and four numbered options. A prime mark denotes a counterclockwise turn as viewed from outside the affected face. After the sequence R' B' L, which option gives the number of target-color stickers on the upper face?}{B}{\traceatlaspromptdense}
\hfill
\traceatlascard{Toggle Grid: Toggle Result}{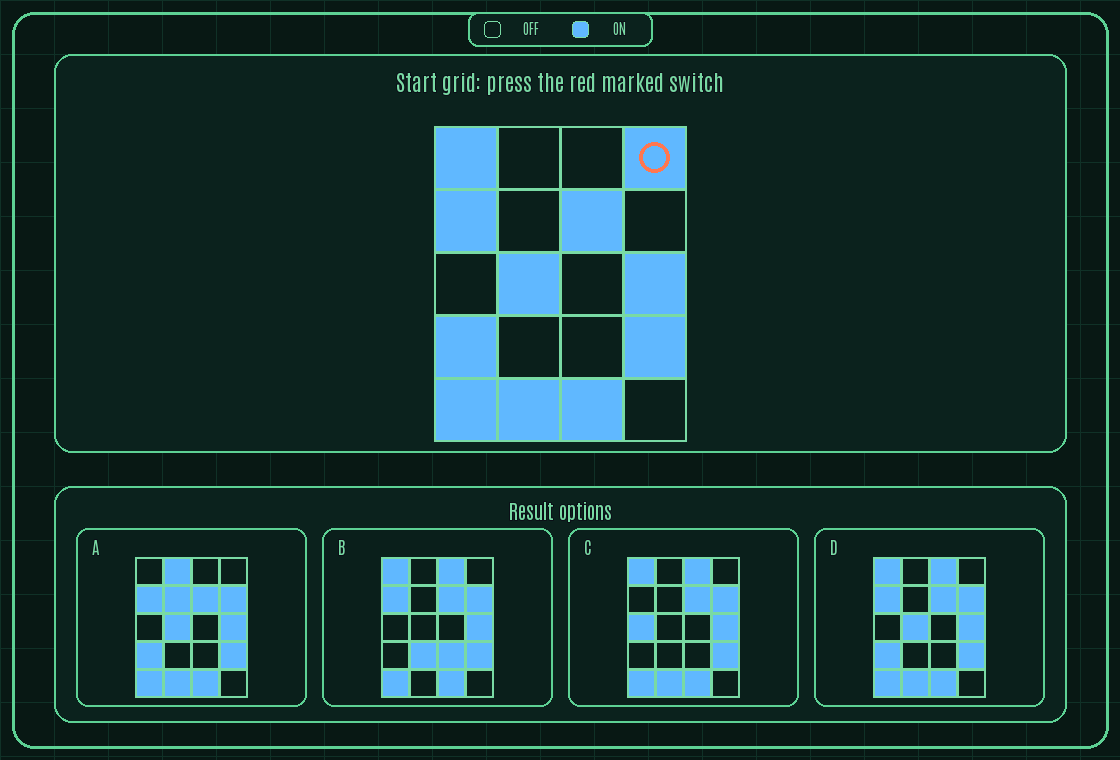}{The scene shows a toggle grid with switch cells, binary cell states, and visual answer options. Use the start grid and the red marked switch. After that one press toggles its cell and edge-neighbor cells, which labeled option matches the final state?}{D}{\traceatlaspromptdense}
\hfill
\traceatlascard{Maze: Nearest Exit}{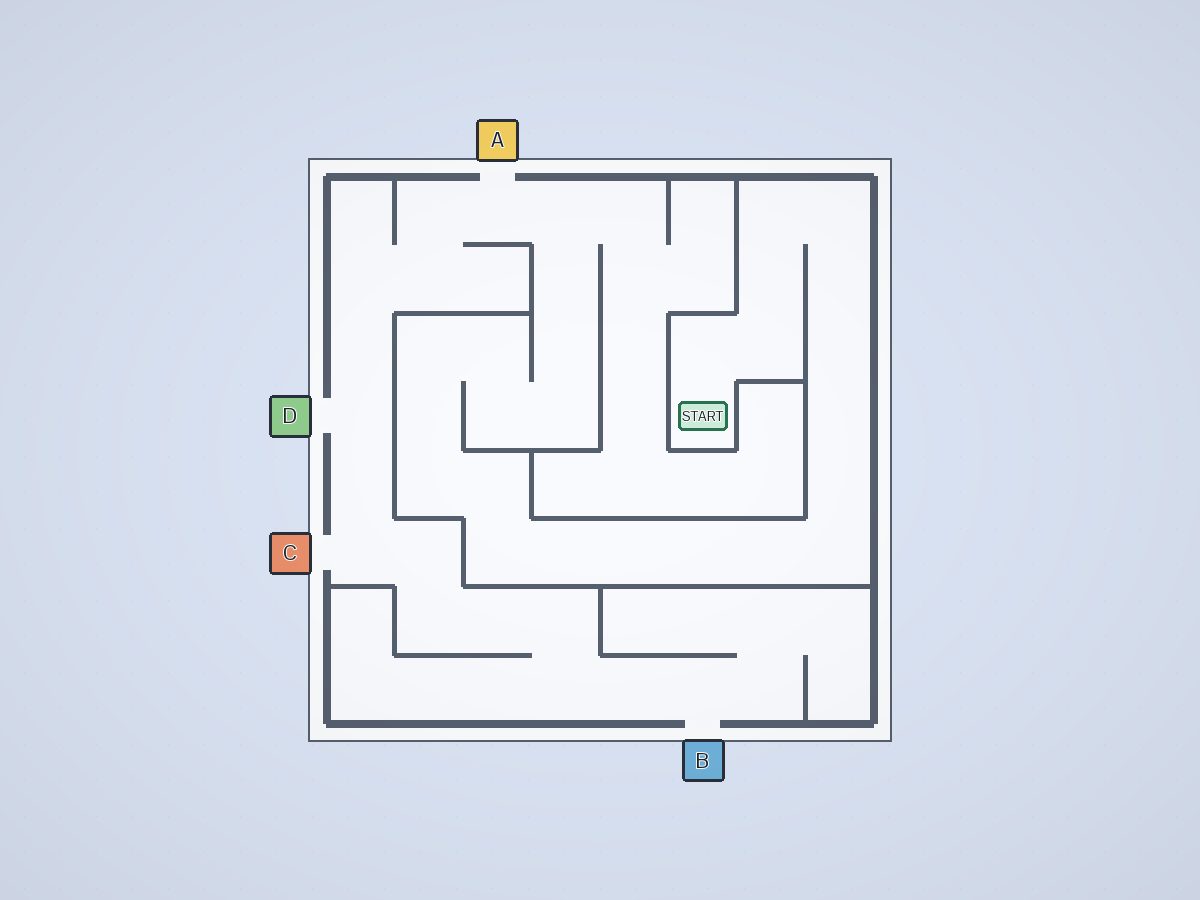}{The image shows an orthogonal wall maze with a START cell and labeled exits on the outer boundary. Which labeled exit is nearest to START by corridor path length?}{A}{\traceatlaspromptnormal}
\par\vspace{2pt}
\noindent
\traceatlascard{Pipe Flow: Repair Tile}{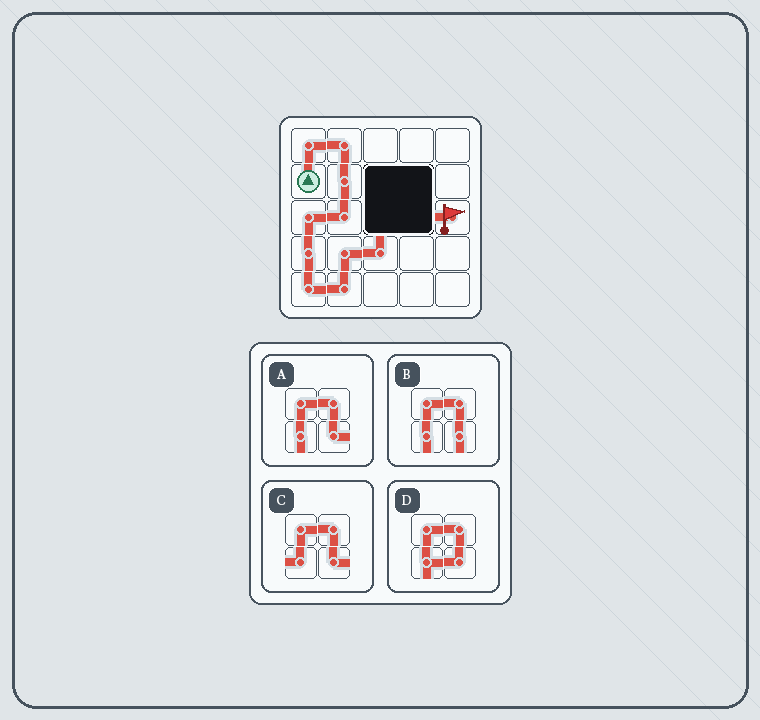}{The pipe puzzle shows an industrial conduit tile grid with a green start marker and red finish flag. Find the single option whose displayed openings make a continuous path from the green start marker to the red finish flag through the black 2x2 gap. Do not rotate or flip options.}{A}{\traceatlaspromptdense}
\hfill
\traceatlascard{Cyclic Order: Equivalent Arrangement}{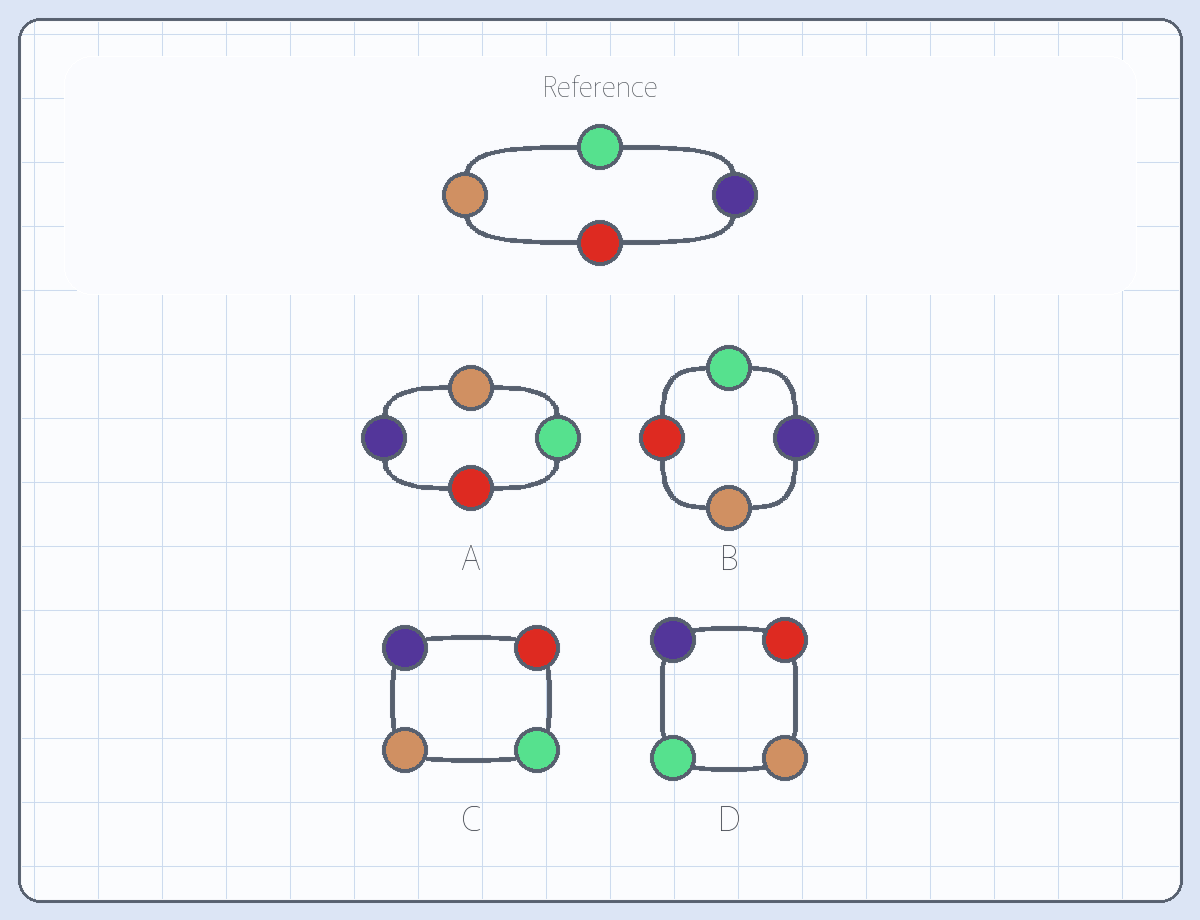}{The image shows a reference token loop above six labeled option loops. Use the token colors when comparing cyclic order. Exactly one option loop is equivalent to the reference loop. Rotations count as equivalent; reversed or reflected order does not.}{D}{\traceatlaspromptdense}
\hfill
\traceatlascard{Star Battle: Valid Cell Anywhere}{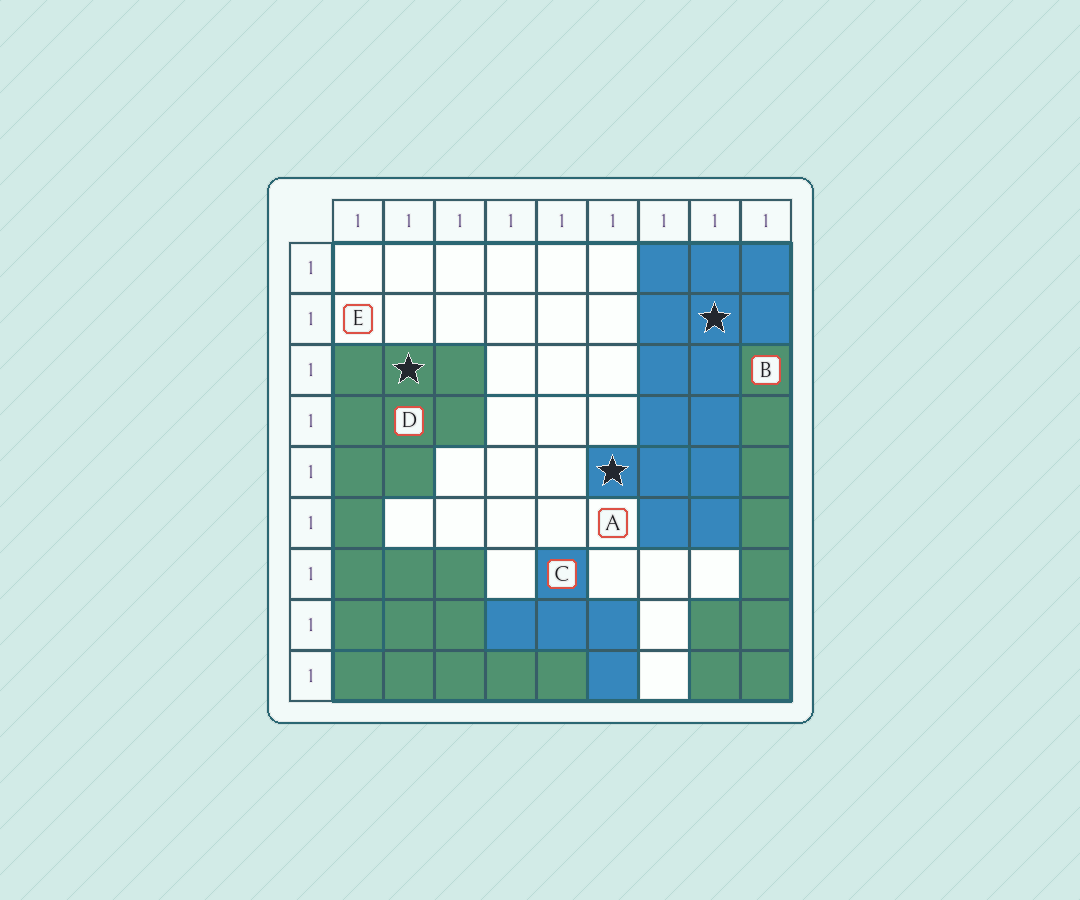}{The visual shows a partially filled Star Battle grid with colored regions, visible stars, and labeled candidate cells. Use these Star Battle rules: each row, column, and colored region must contain exactly one star; stars cannot touch, even diagonally. Choose the labeled cell where another star can be placed.}{C}{\traceatlaspromptdense}
\par\vspace{2pt}
\noindent
\traceatlascard{Tents: Violating Tent}{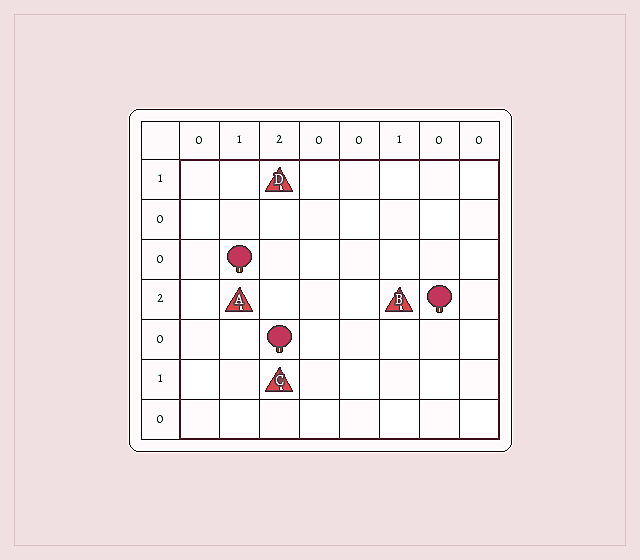}{The figure shows a Tents puzzle grid with row and column clues, trees, visible tents, and labeled cells or tents. Inspect the labeled tents. Which label marks the tent that has no orthogonally adjacent tree?}{D}{\traceatlaspromptnormal}
\hfill
\traceatlascard{Raven Matrix: Spatial Transform}{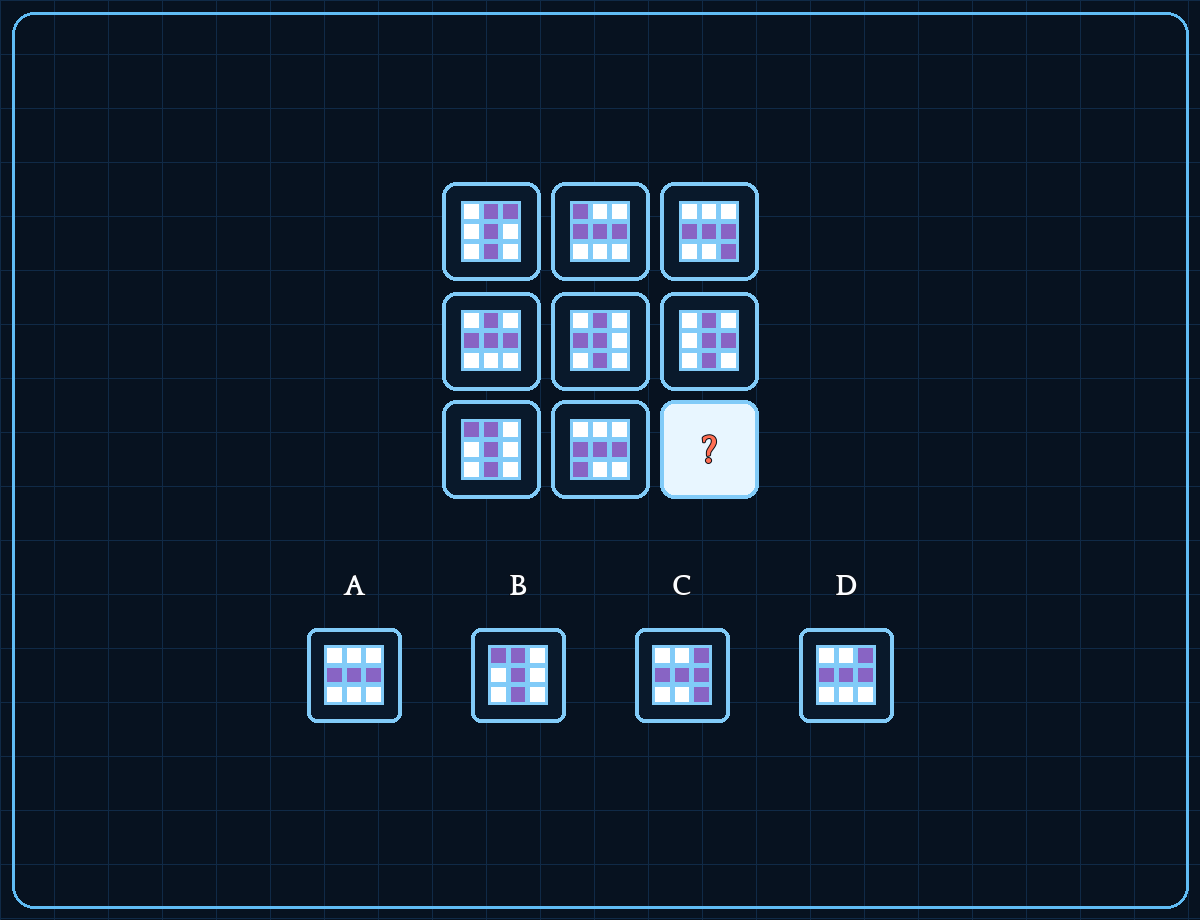}{The image shows a 3 by 3 Raven-style matrix puzzle with four labeled image options below it. Select the option whose mini-grid pattern fits the spatial arrangement rule.}{D}{\traceatlaspromptnormal}
\hfill
\traceatlascard{Nonogram: Candidate Solution}{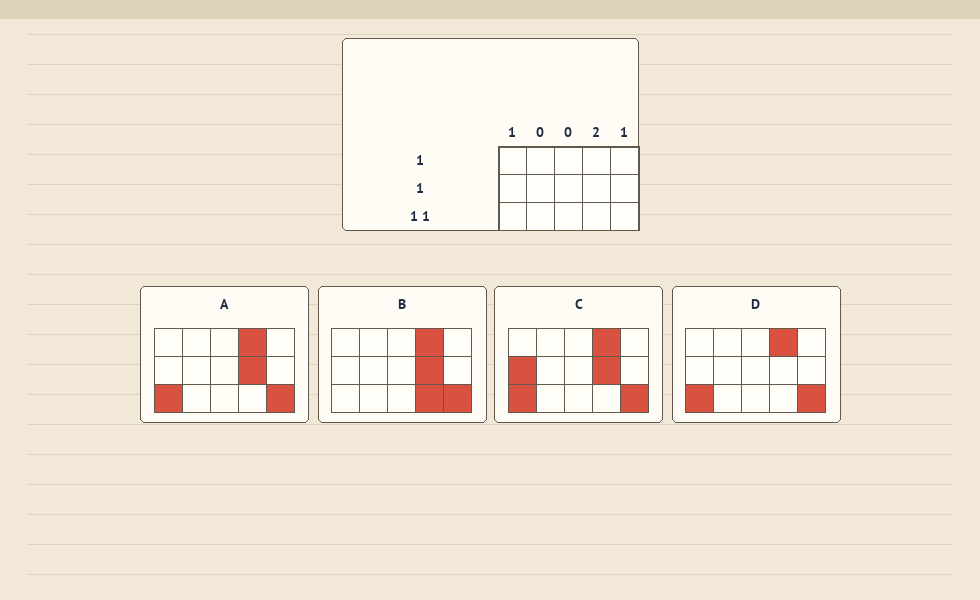}{This nonogram shows a clue grid with labeled filled-grid options. Use both clue rails to identify the correct candidate grid.}{A}{\traceatlaspromptnormal}
\caption{Representative puzzle tasks.}
\label{fig:task-atlas-puzzles}
\end{figure}
\clearpage

\begin{figure}[p]
\centering
\traceatlascard{Organic Structure: Bond Order}{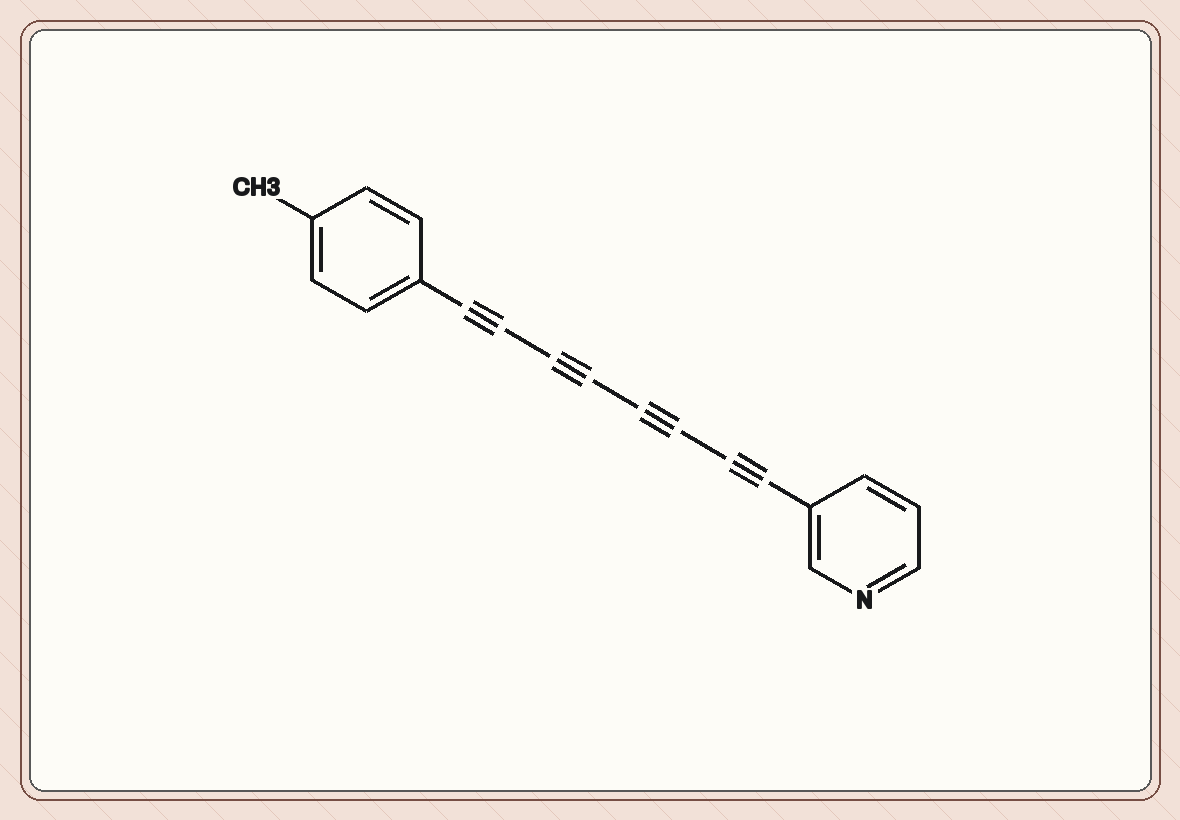}{The visual panel contains an organic skeletal structure with line-angle bonds, rings, and occasional atom or substituent labels. Count only the visible triple bonds.}{4}{\traceatlaspromptnormal}
\hfill
\traceatlascard{Life Automaton: One Step Cell State}{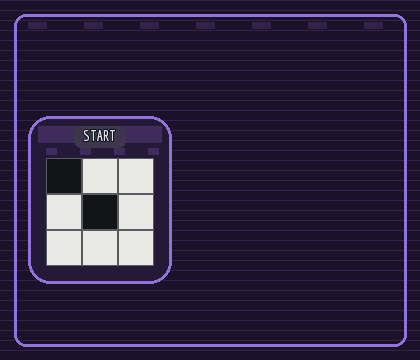}{In the START grid, dark cells are alive. An alive cell survives with two or three live neighbors; an empty cell becomes alive with exactly three; every other cell is empty next. After one update, how many cells are empty?}{9}{\traceatlaspromptdense}
\hfill
\traceatlascard{Dice: Joint Even-Value Probability}{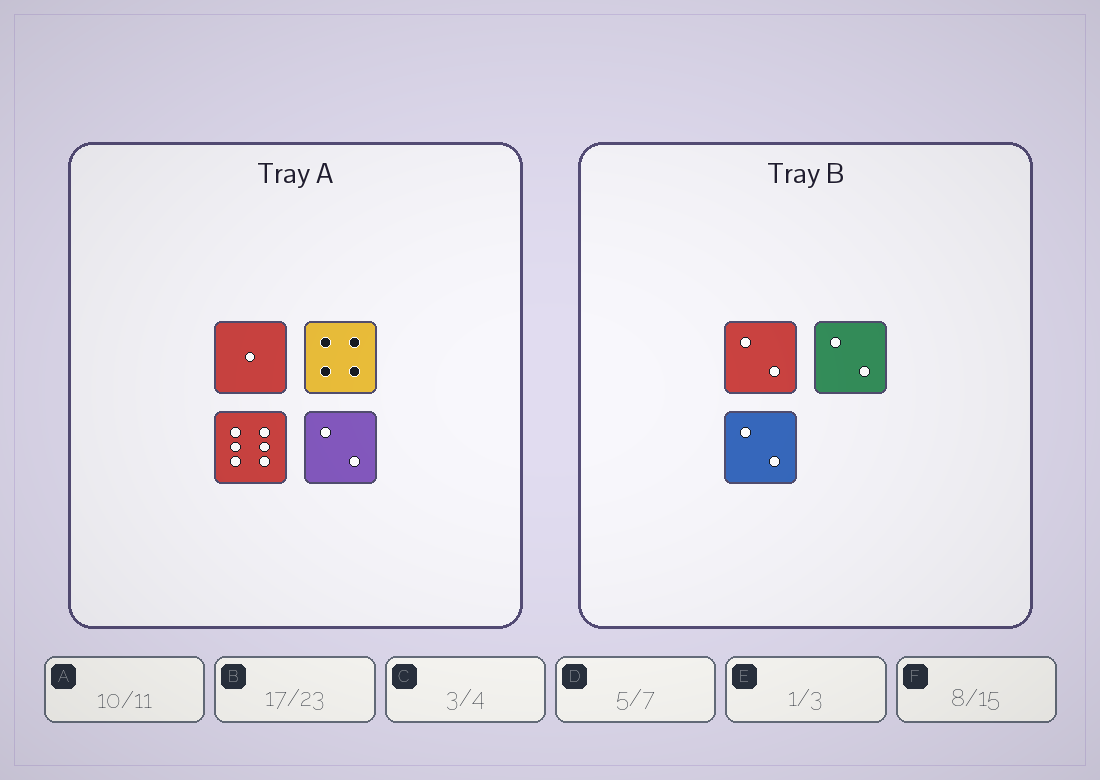}{The trays show each die's top value. One die is selected uniformly and independently from each tray. Which A-F option gives the probability that both selected dice show even values?}{C}{\traceatlaspromptdense}
\par\vspace{2pt}
\noindent
\traceatlascard{Spinner: Color-and-Shape Probability}{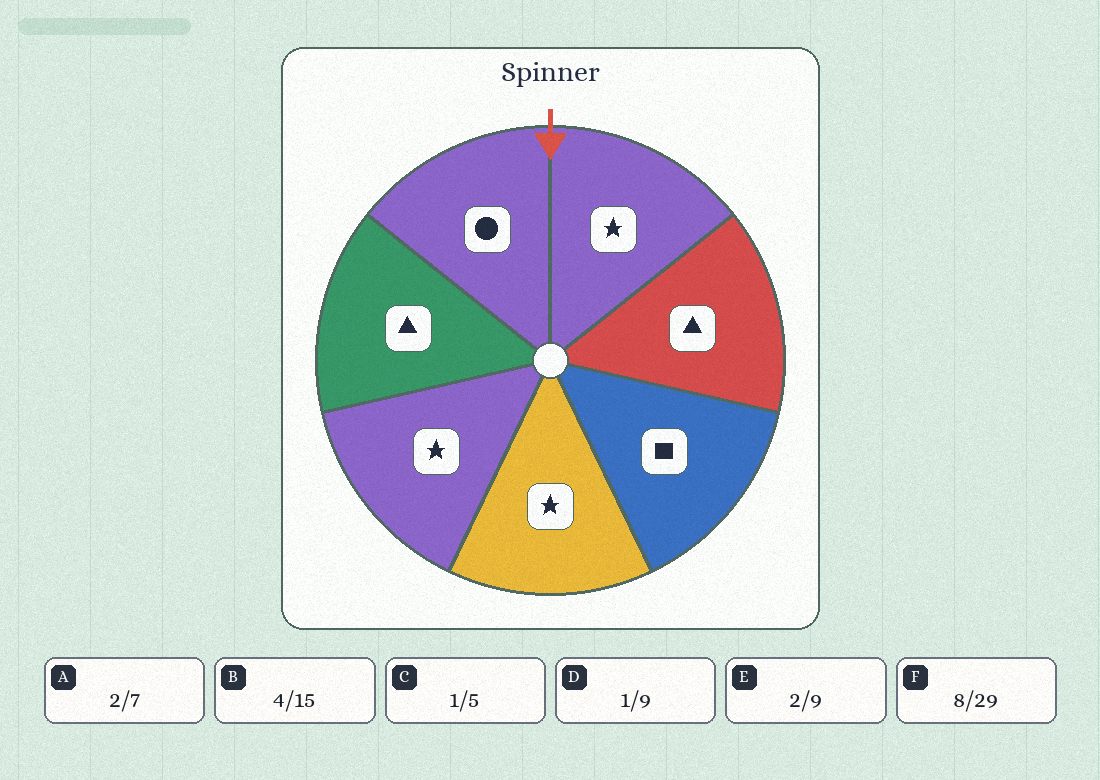}{Each spinner sector is equally likely and has a color and shape. Which A-F fraction option gives the probability of landing on a purple sector marked with a star?}{A}{\traceatlaspromptdense}
\hfill
\traceatlascard{Truth Table: Truth Pattern}{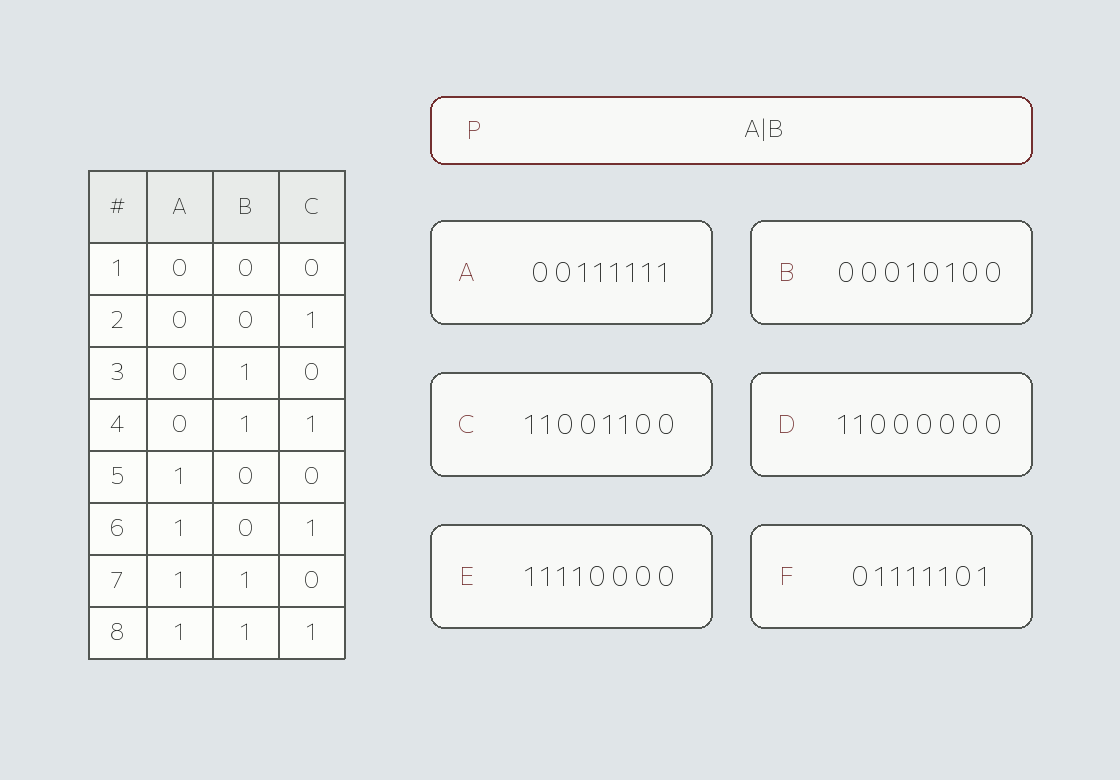}{A truth table shows an input table, a target expression P, and six labeled output-pattern options. Values use 1 for true and 0 for false. The operators are ! for NOT, \& for AND, | for OR, and \textasciicircum{} for XOR. Reading rows from top to bottom, which option matches the truth pattern for P?}{A}{\traceatlaspromptdense}
\hfill
\traceatlascard{Abacus: Displayed Value}{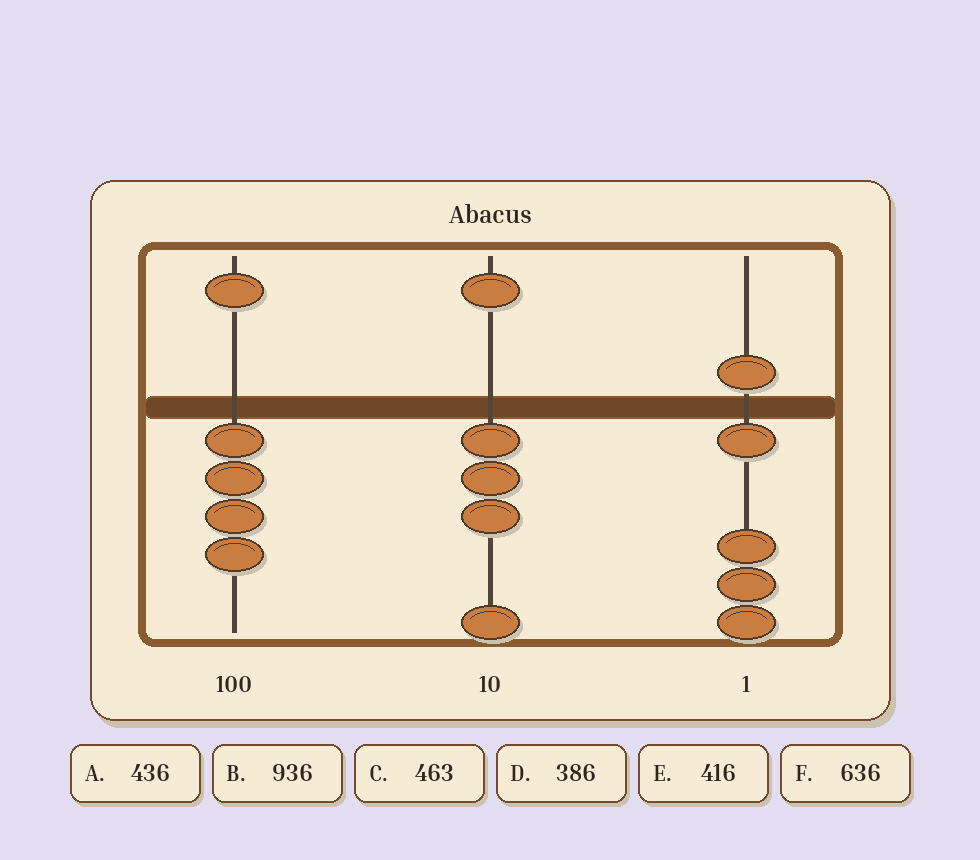}{On the three-column abacus, beads touching the center bar are active. An active upper bead is worth 5 and each active lower bead 1. Read the 100, 10, and 1 columns from left to right. Which option gives the displayed number?}{A}{\traceatlaspromptdense}
\par\vspace{2pt}
\noindent
\traceatlascard{Braille Cell: Matching Pattern}{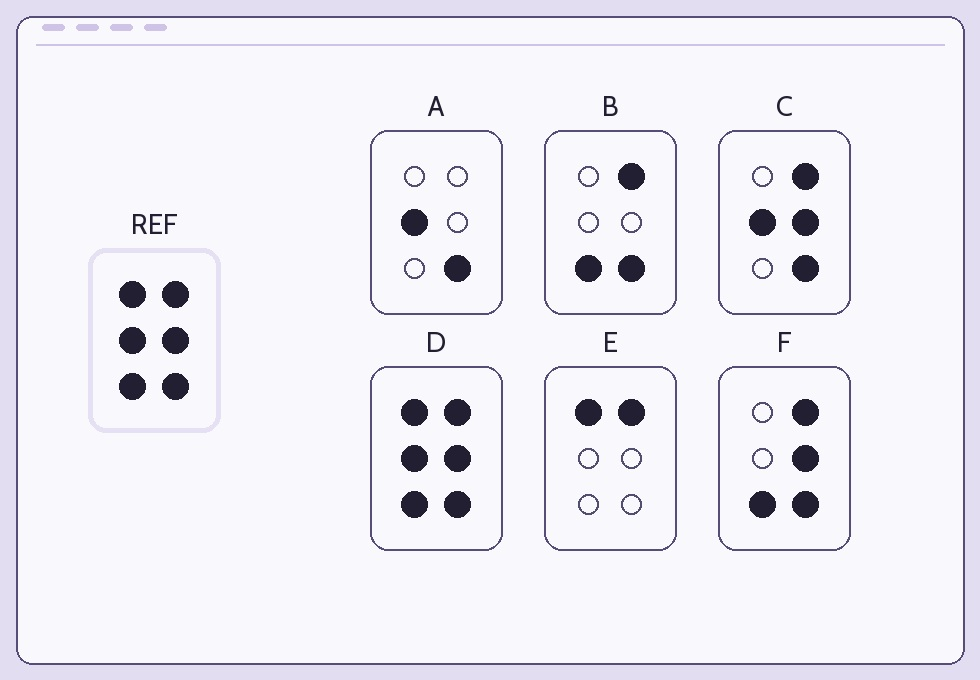}{This panel shows a Braille reference cell and six labeled Braille option cells. Which option letter matches the reference pattern?}{D}{\traceatlaspromptnormal}
\hfill
\traceatlascard{Morse Code: Morse Word Read}{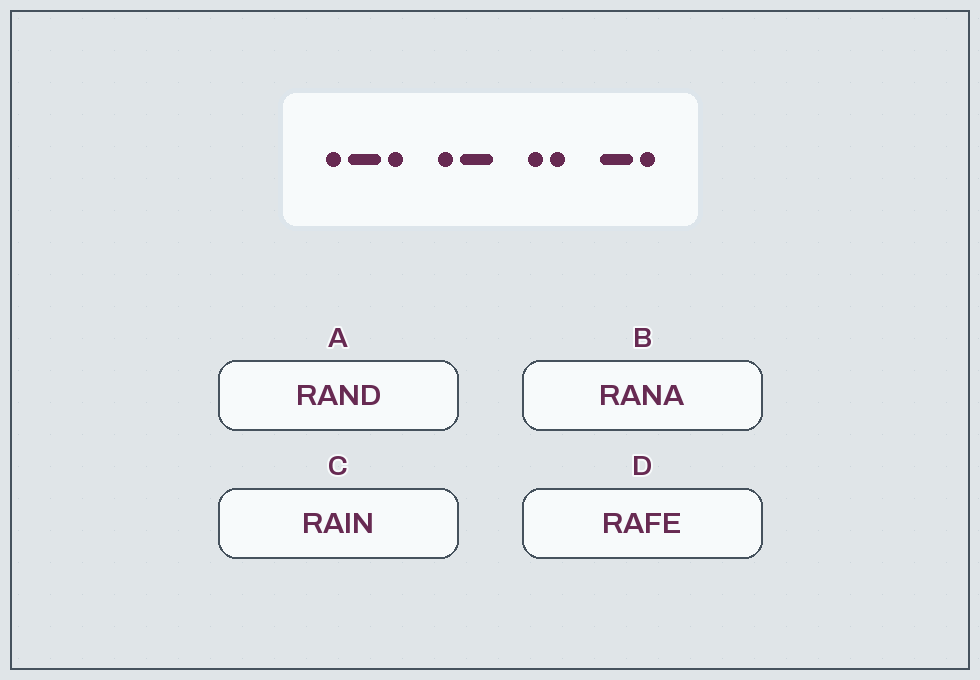}{A visual Morse-code panel presents a Morse-code word above four labeled word options. Which labeled option matches the Morse code?}{C}{\traceatlaspromptnormal}
\hfill
\traceatlascard{Music Staff: Scale Degree Function}{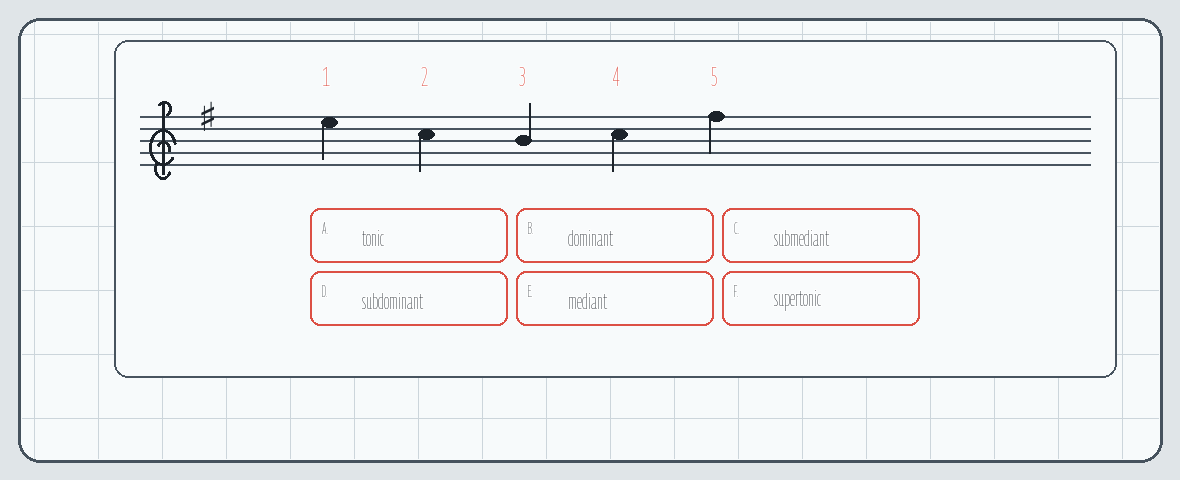}{The visual shows a sheet-music staff with marked notation items. Select the option naming the scale-degree function of note 3 in G major.}{E}{\traceatlaspromptnormal}
\par\vspace{2pt}
\noindent
\traceatlascard{Agent Automaton: Future Grid}{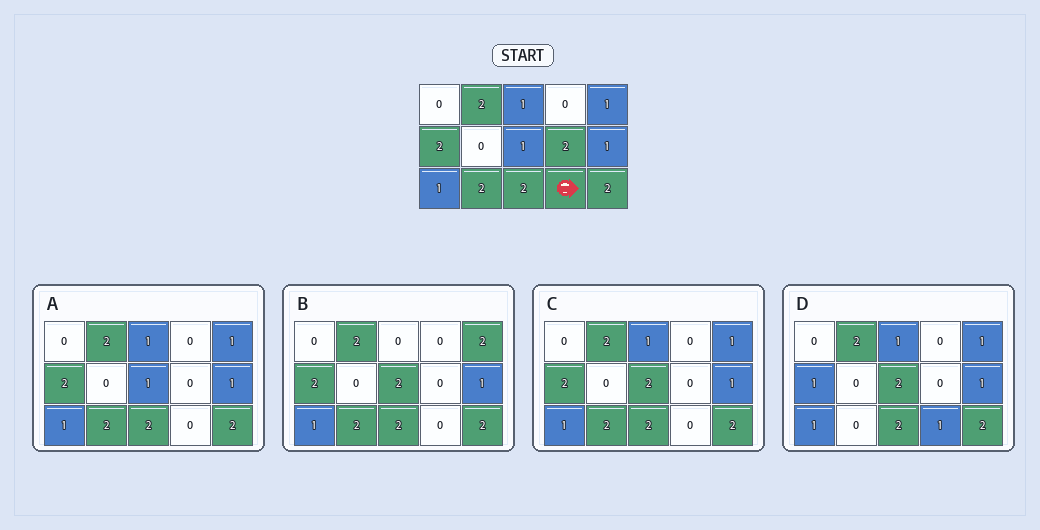}{Starting at the arrow, apply three updates. Before moving, state 0 turns right and becomes 1; state 1 goes straight and becomes 2; state 2 turns left and becomes 0. The agent then moves one cell forward, wrapping at the grid edge. Which option shows the resulting grid?}{C}{\traceatlaspromptdense}
\hfill
\traceatlascard{Turing Tape: Written-Symbol Count}{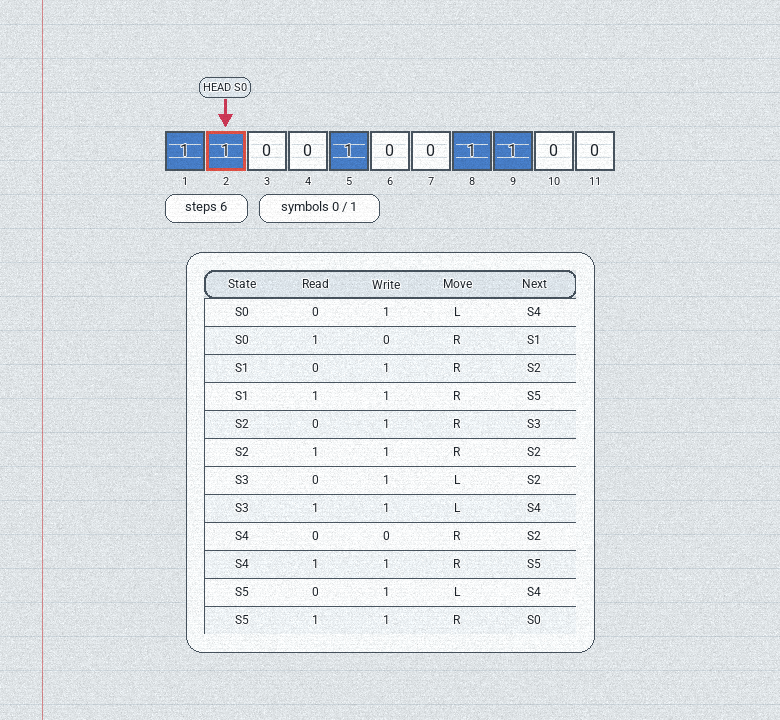}{Starting from the shown tape, head, and state, simulate six transitions. At each step, use the visible table to write a symbol, move left or right, and change state. How many tape cells contain symbol 0 afterward?}{5}{\traceatlaspromptdense}
\hfill
\traceatlascard{Logic Gate Circuit: Output Value}{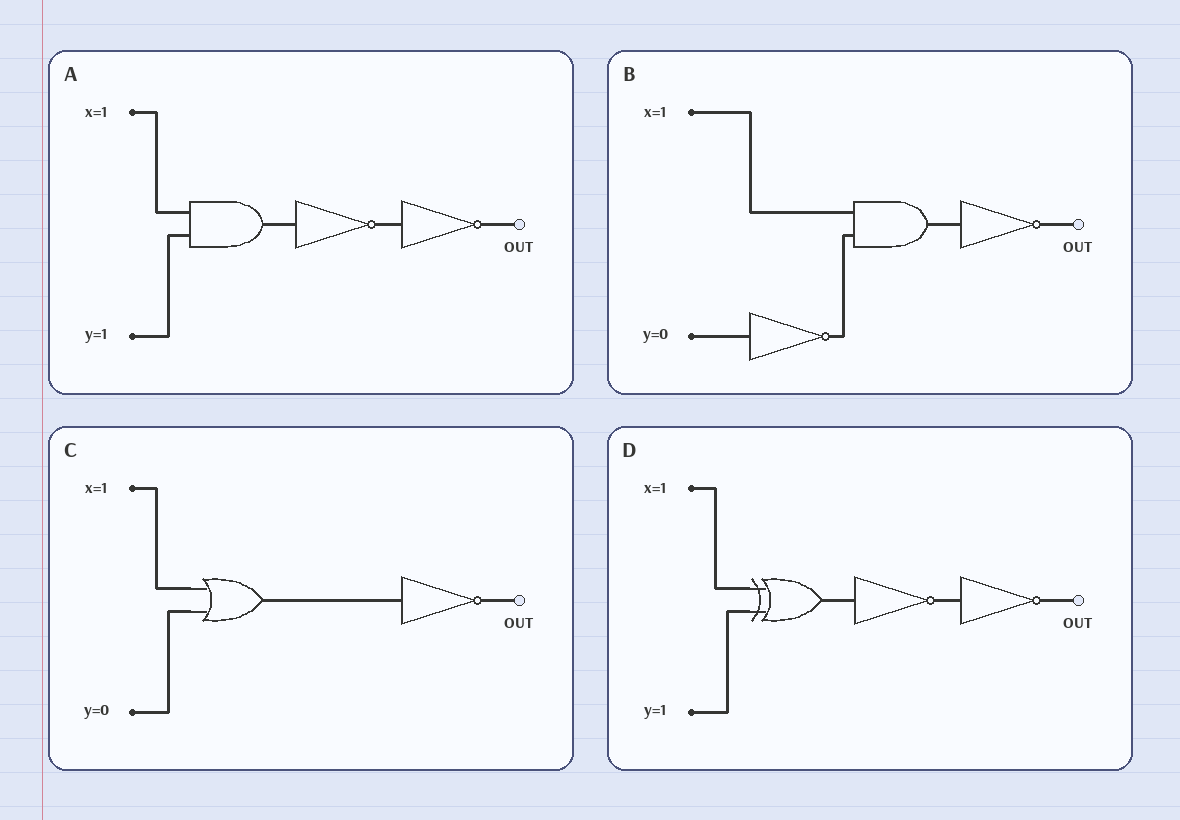}{A Boolean logic-gate panel shows a circuit with labeled inputs, standard gate symbols, wires, and final OUT nodes. Evaluate each circuit. Which option's final OUT value is 1?}{A}{\traceatlaspromptnormal}
\caption{Representative symbolic tasks.}
\label{fig:task-atlas-symbolic}
\end{figure}
\clearpage

\begin{figure}[p]
\centering
\traceatlascard{Object Cluster: Color-and-Type Exclusion}{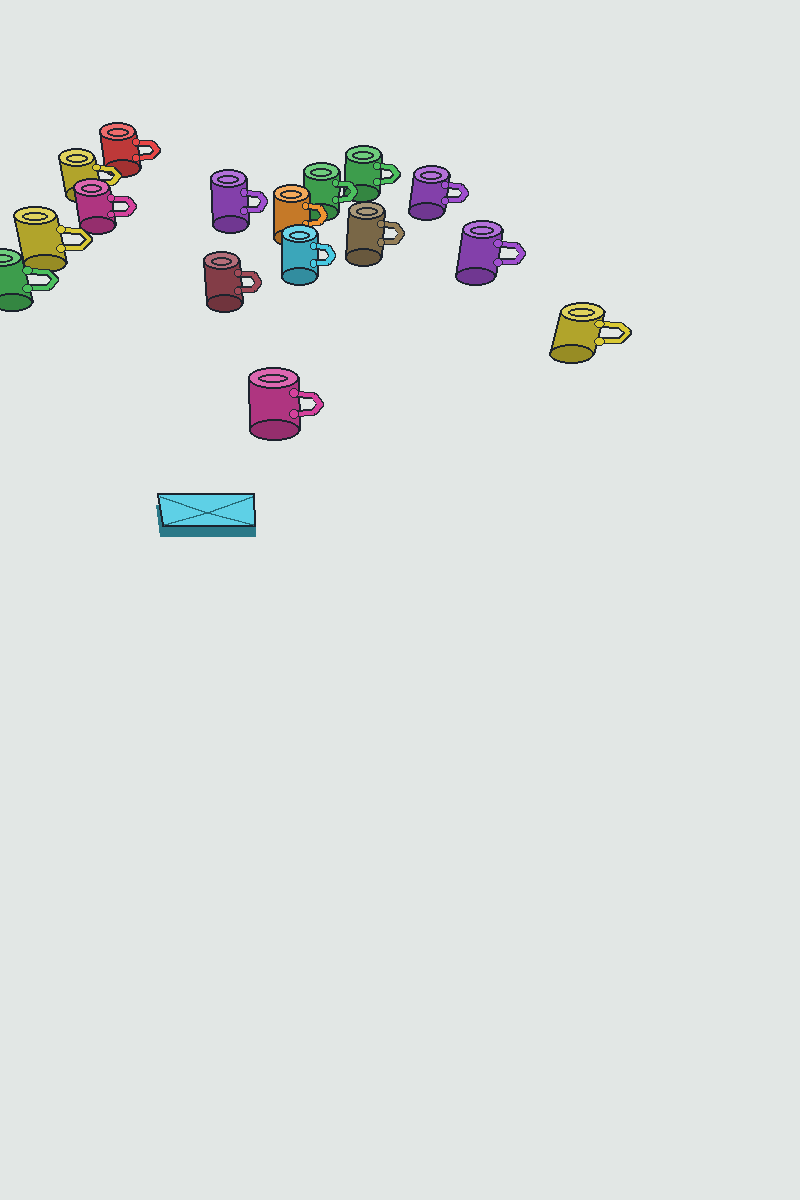}{The visible scene contains many small 3D objects arranged on a plain surface. How many cyan objects are not cups?}{1}{\traceatlaspromptnormal}
\hfill
\traceatlascard{Object Scene: Nearest to Reference}{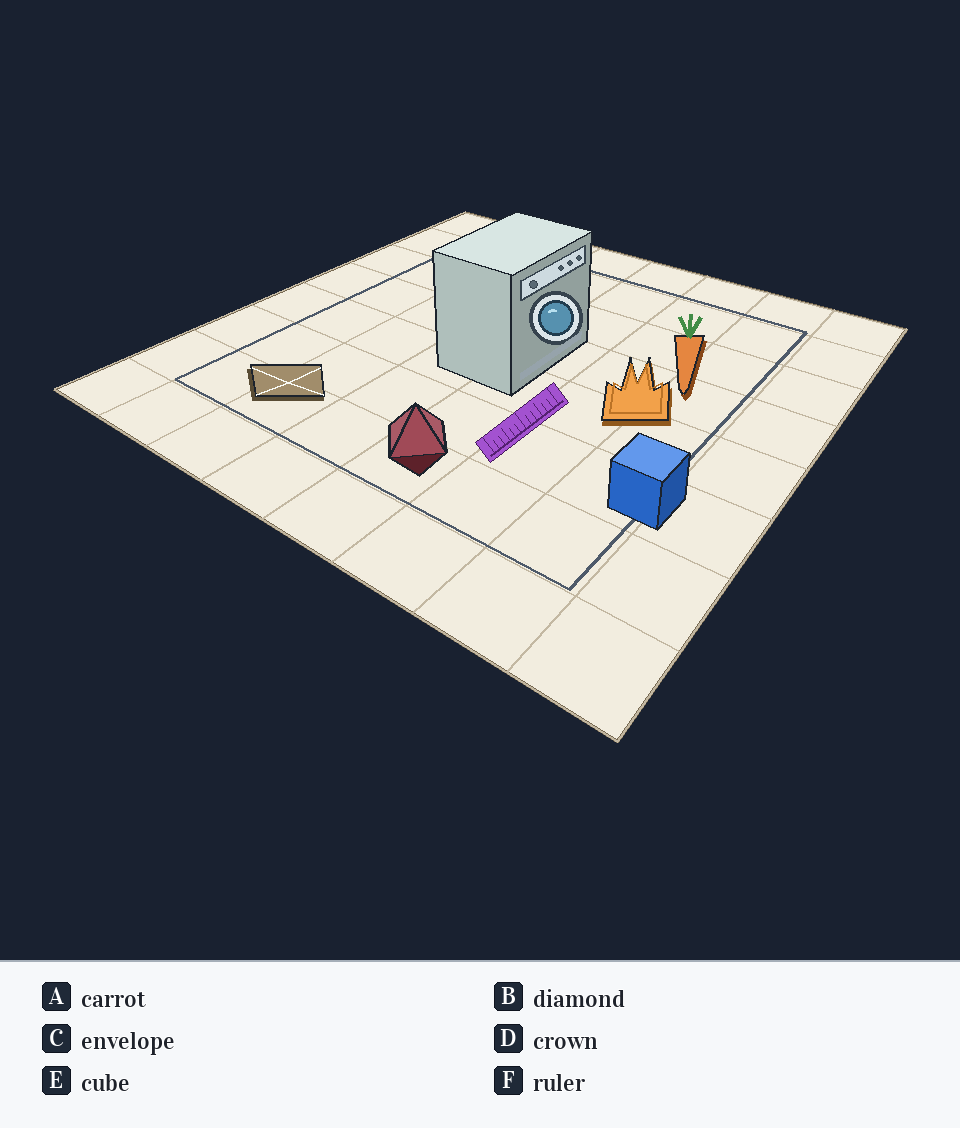}{The scene shows a washing machine as the reference and six smaller candidate objects. Which option identifies the candidate nearest the washing machine?}{F}{\traceatlaspromptnormal}
\hfill
\traceatlascard{Object Cluster: Type-or-Color Count}{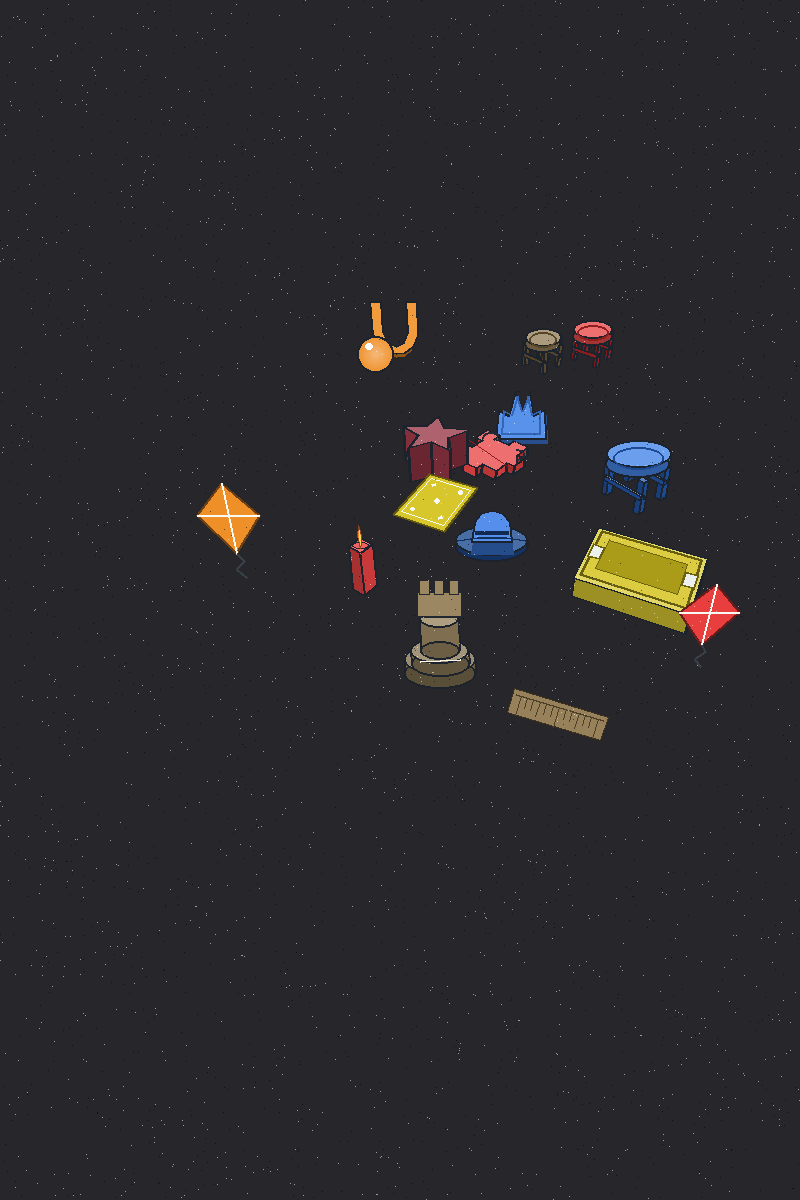}{The image shows many small 3D objects arranged on a plain surface. How many objects are either stools or blue?}{5}{\traceatlaspromptnormal}
\par\vspace{2pt}
\noindent
\traceatlascard{Surface Fixture: U-Bolt Count}{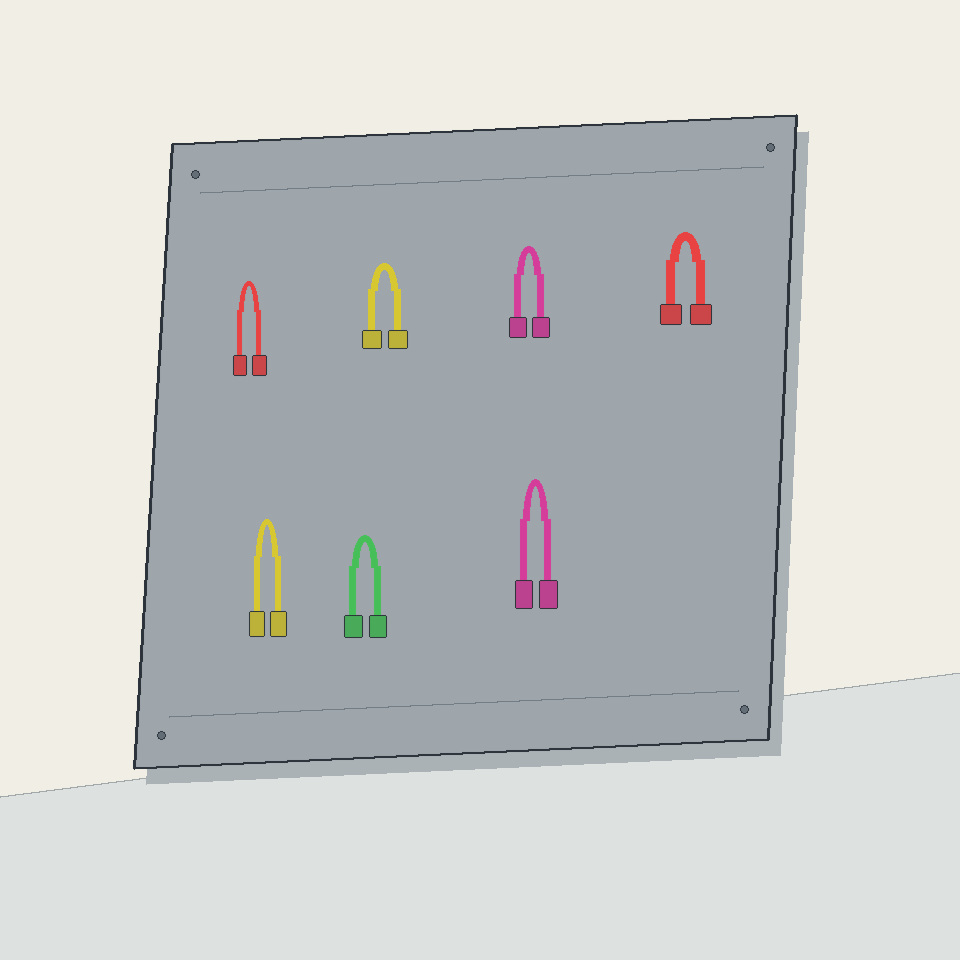}{The image shows a U-bolt plate with visible U-bolts. How many U-bolts are visible?}{7}{\traceatlaspromptnormal}
\hfill
\traceatlascard{Warehouse: Nearest Candidate to Reference}{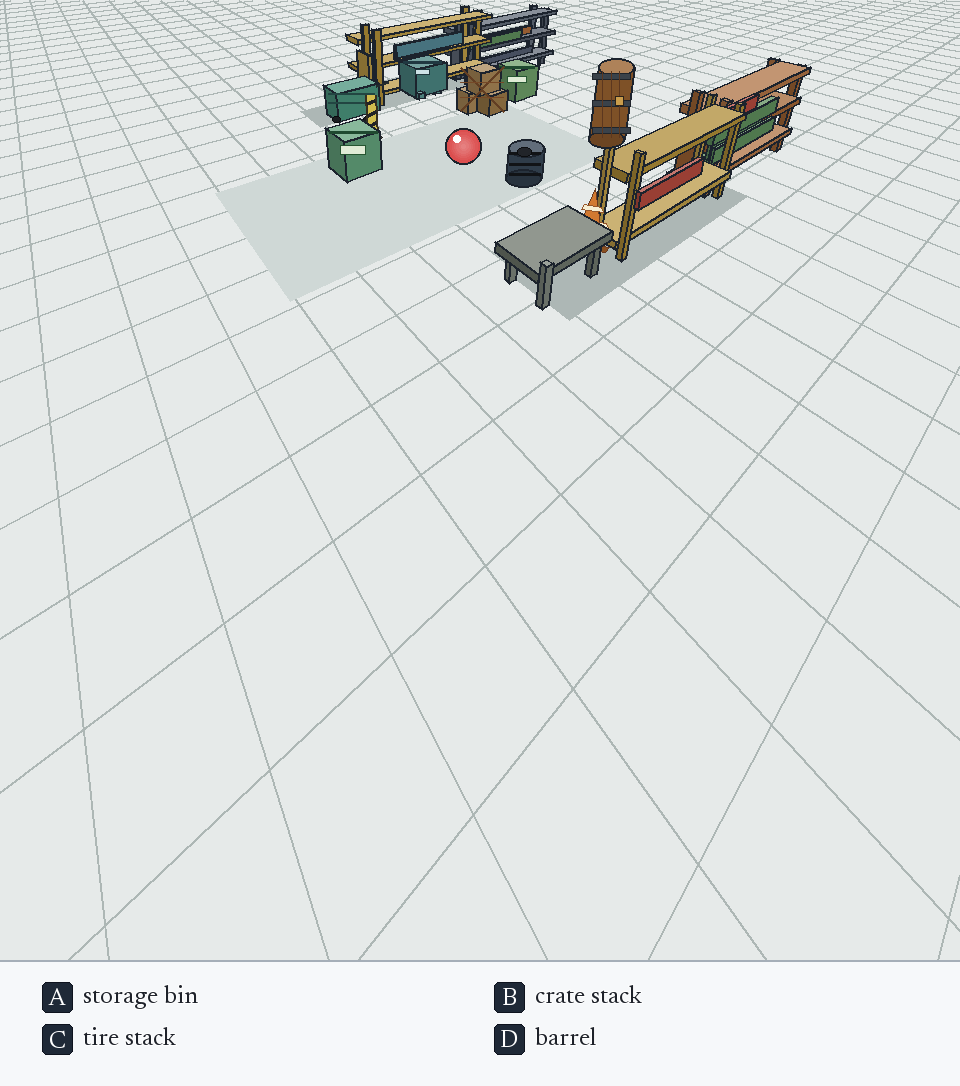}{The image shows a perspective warehouse aisle with shelf racks, warehouse equipment, one red sphere reference object, candidate warehouse objects, and a text option panel below the scene. Which option describes the warehouse object closest to the red sphere?}{C}{\traceatlaspromptdense}
\hfill
\traceatlascard{Room: Object on the Reference Wall}{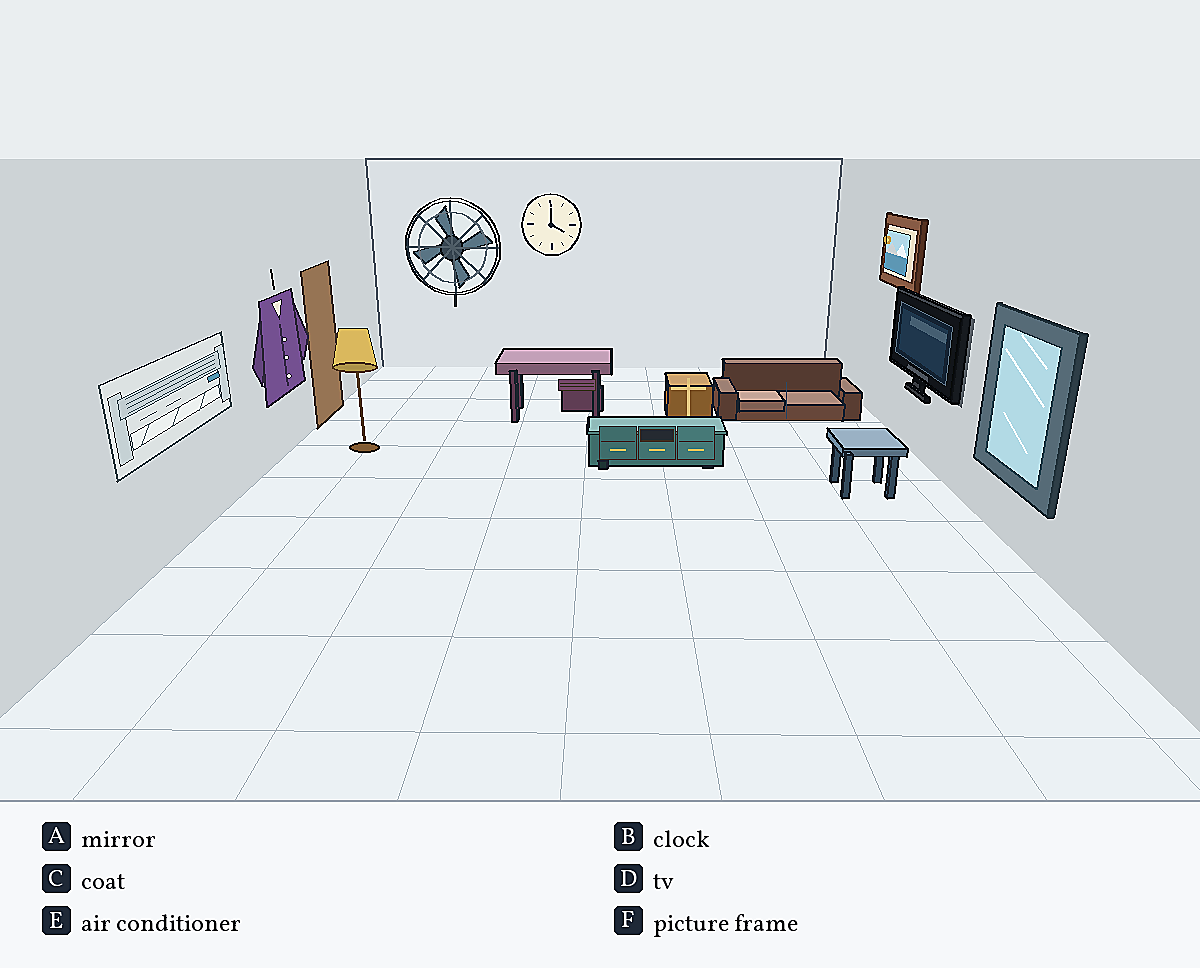}{The picture shows a perspective indoor room with floor, walls, furniture, room objects, and wall-mounted objects. Which option describes the object that shares a wall with the fan?}{B}{\traceatlaspromptnormal}
\par\vspace{2pt}
\noindent
\traceatlascard{Carousel: Ordered Adjacent Pair Count}{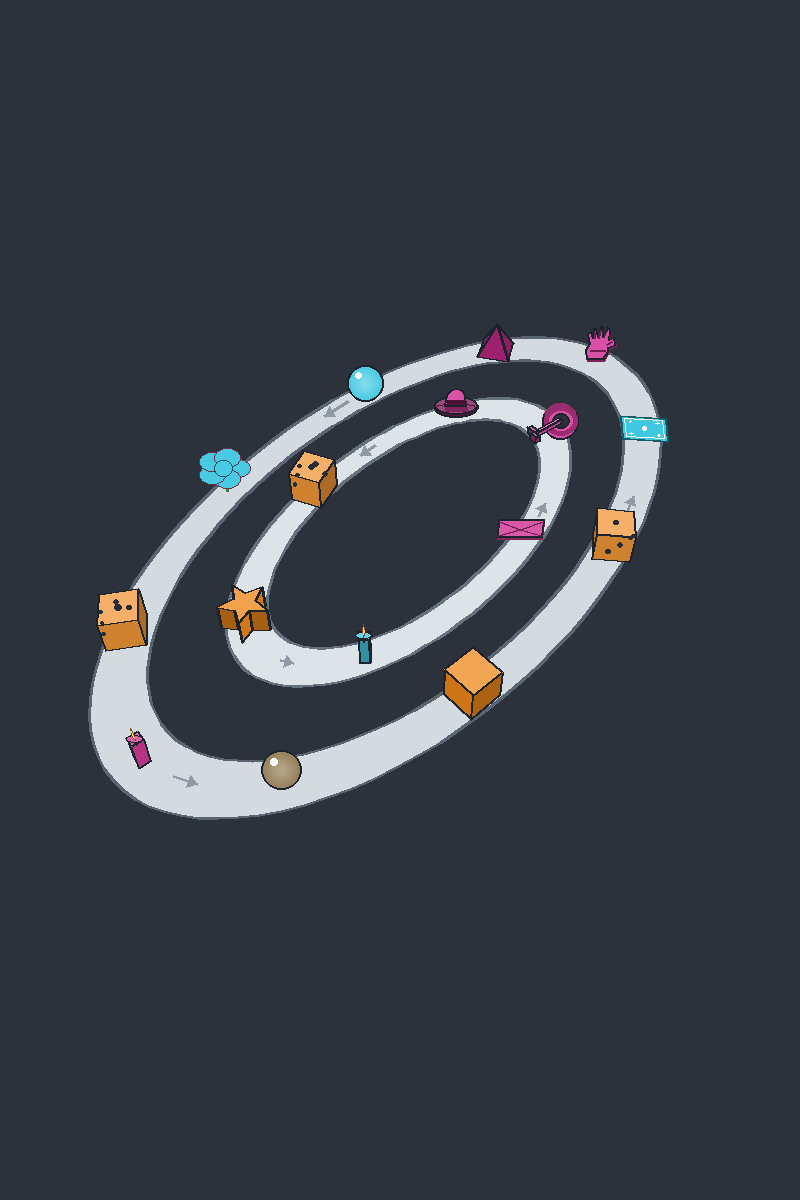}{The scene shows small 3D objects on two concentric elliptical belts. Following the order of the inner belt, count each occurrence of gloves immediately followed by pyramids. How many pairs are there?}{0}{\traceatlaspromptnormal}
\hfill
\traceatlascard{Street: Intersection Nearest}{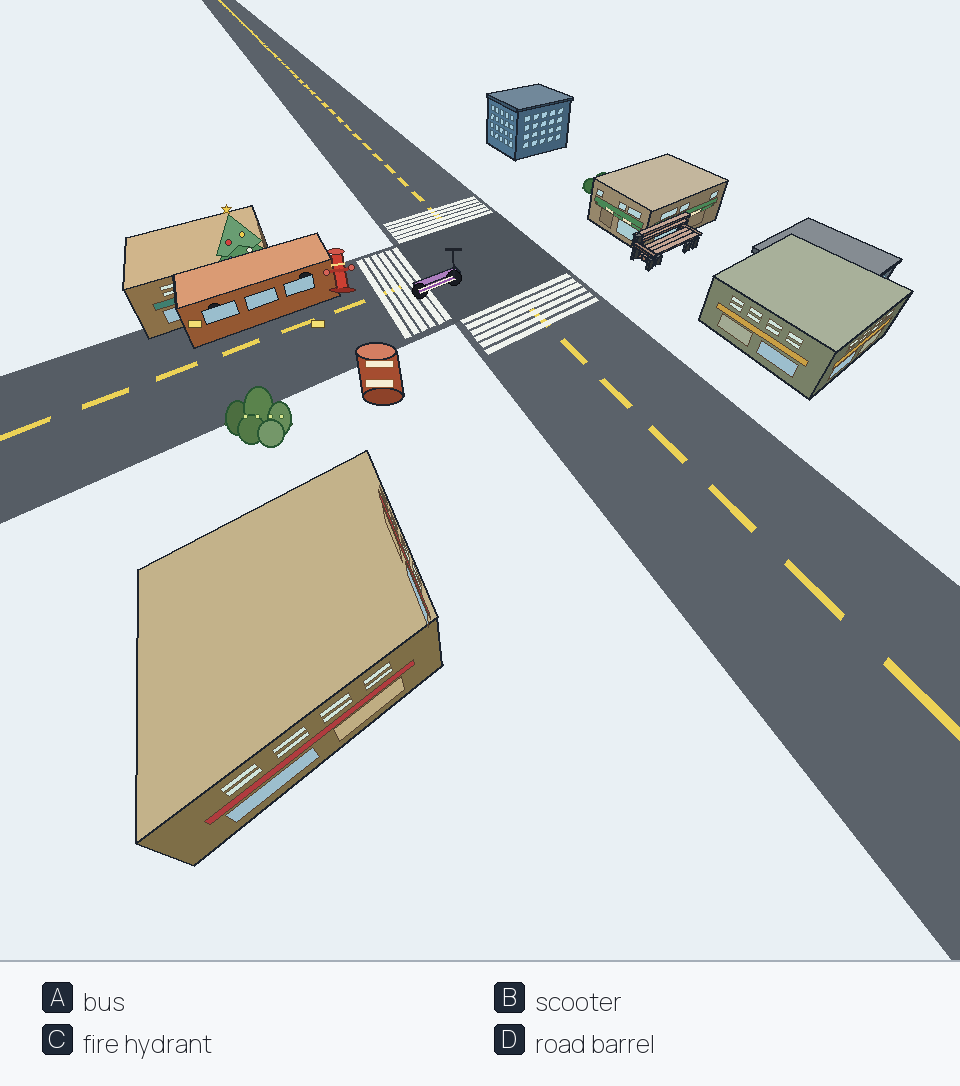}{The view shows a perspective street intersection or T intersection with roads, sidewalks, crosswalk markings, street context, candidate street objects, and a text option panel below the scene. Which option describes the object closest to where the roads meet?}{B}{\traceatlaspromptdense}
\hfill
\traceatlascard{Conveyor: Ordered Adjacent Pair Count}{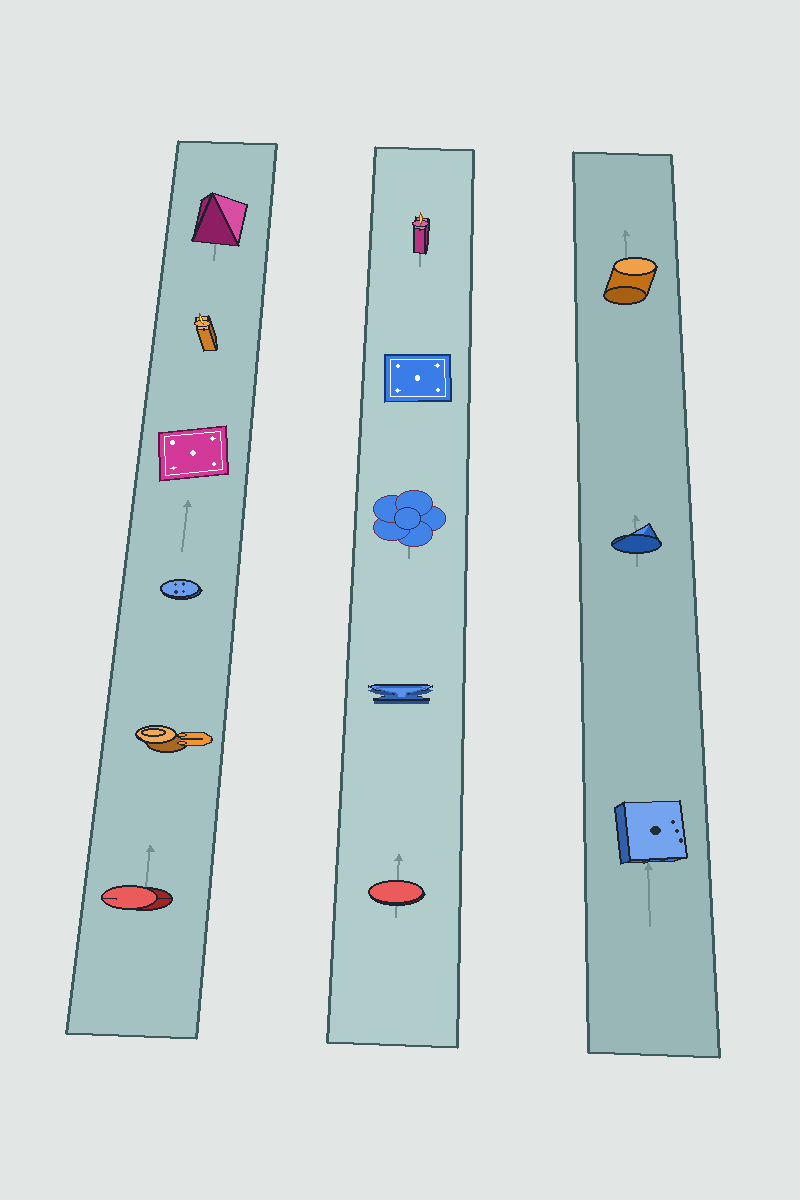}{The visual shows a three-lane conveyor with small objects on each lane. Following the order of the left belt, count each occurrence of flowers immediately followed by trophies. How many pairs are there?}{0}{\traceatlaspromptnormal}
\par\vspace{2pt}
\noindent
\traceatlascard{Conveyor: Objects on Selected Belt}{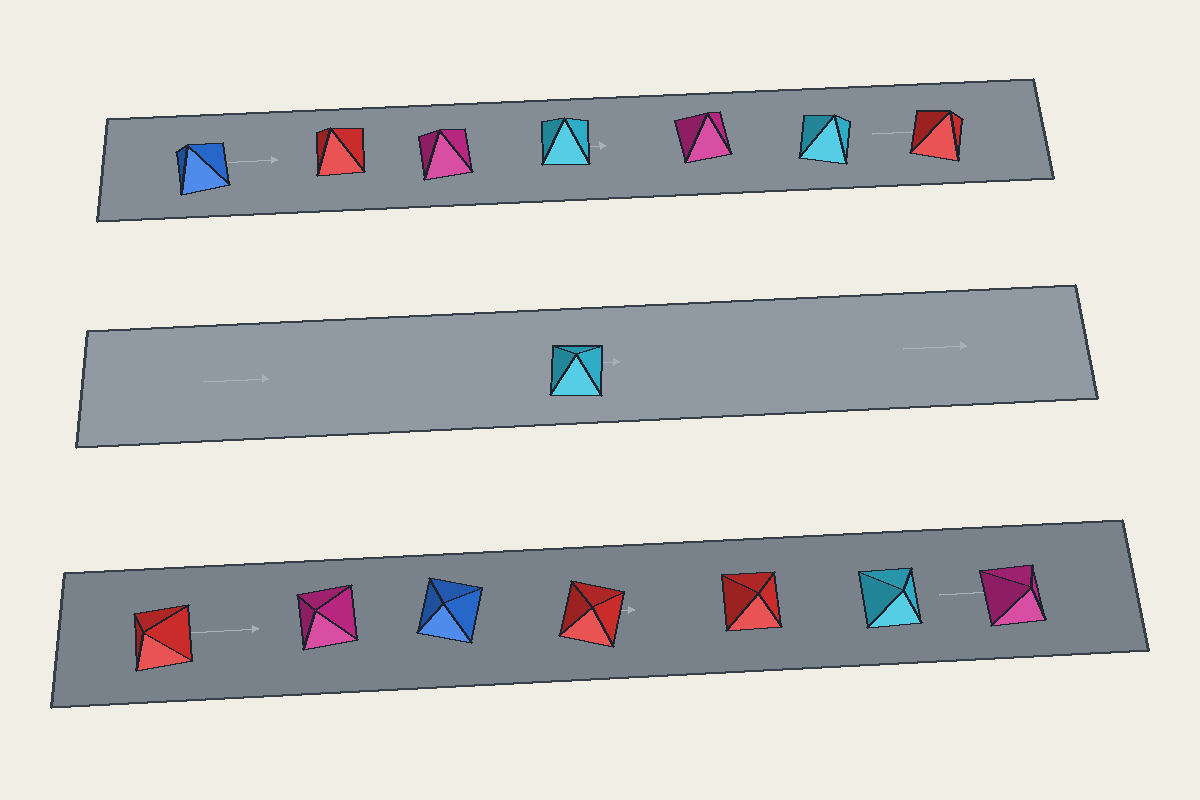}{The visual shows a three-lane conveyor with small objects on each lane. How many objects are on the MIDDLE belt?}{1}{\traceatlaspromptnormal}
\hfill
\traceatlascard{Carousel: Between Object Type Anchors}{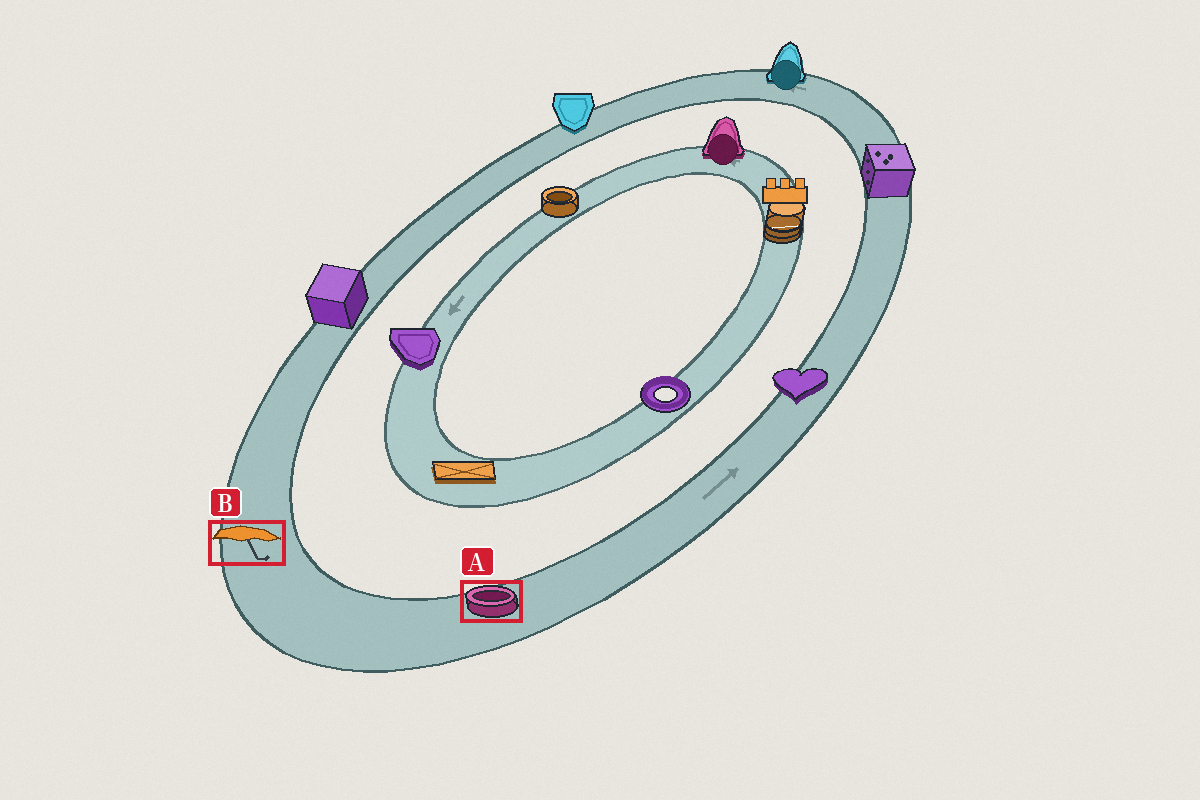}{A 3D conveyor carousel is shown with objects on the inner and outer belts. Follow the OUTER belt direction from the marked bowl A to the marked umbrella B. How many objects are strictly between them?}{5}{\traceatlaspromptnormal}
\hfill
\traceatlascard{Object Scene: Marked Point Depth Extremum}{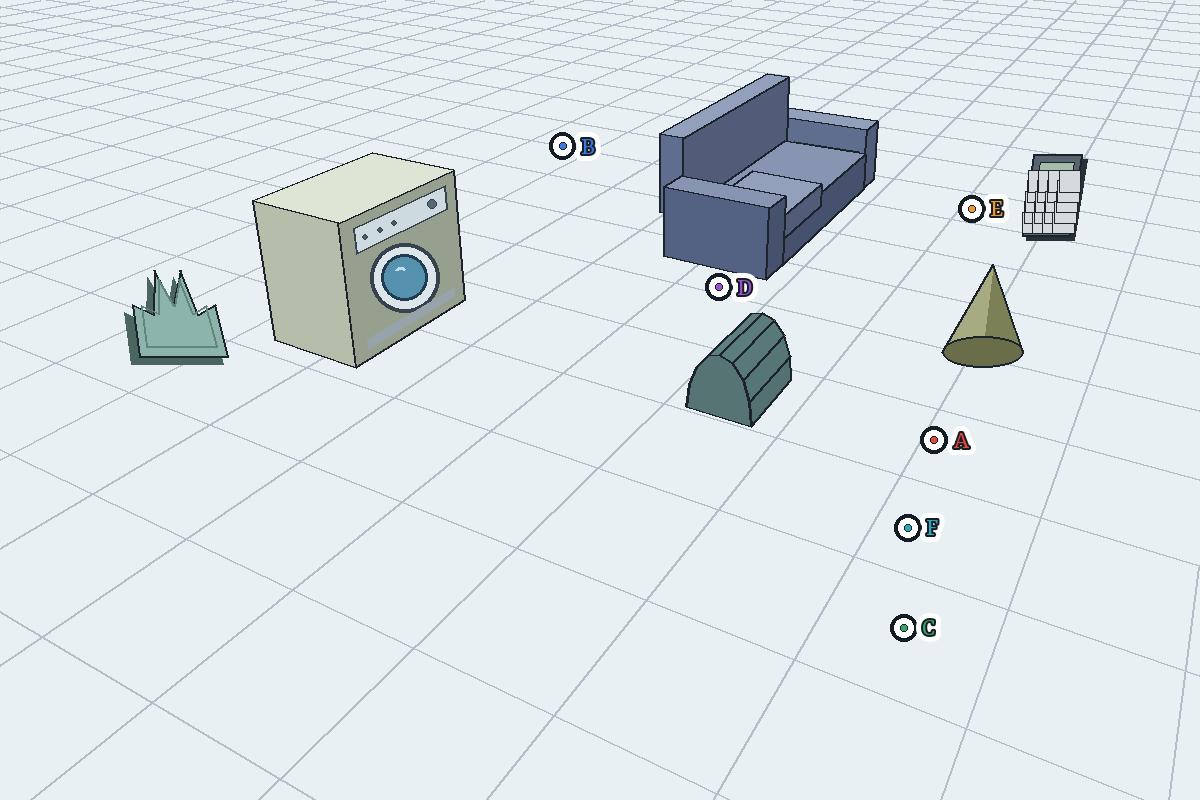}{The scene shows a perspective 3D platform scene with objects and lettered floor point markers. From this camera view, which lettered floor point is farthest away from the viewer?}{B}{\traceatlaspromptnormal}
\caption{Representative 3D tasks.}
\label{fig:task-atlas-three-d}
\end{figure}
\clearpage

\endgroup

\end{document}